\documentclass[10pt]{article} 
\usepackage[preprint]{tmlr}

\usepackage{amsmath,amsfonts,bm}

\def\eqref#1{equation~\ref{#1}}

\def\1{\bm{1}}

\DeclareMathAlphabet{\mathsfit}{\encodingdefault}{\sfdefault}{m}{sl}
\SetMathAlphabet{\mathsfit}{bold}{\encodingdefault}{\sfdefault}{bx}{n}

\usepackage[hypertexnames=false]{hyperref}
\usepackage{url}
\usepackage{graphicx}
\usepackage{algorithm}
\usepackage{algorithmic}

\usepackage{booktabs}
\usepackage{amsmath}
\usepackage{amssymb}
\usepackage{amsthm}
\theoremstyle{plain}
\newtheorem{theorem}{Theorem}
\newtheorem{lemma}{Lemma}
\newtheorem{remark}{Remark}
\newtheorem{assumption}{Assumption}
\newtheorem{proposition}{Proposition}
\newtheorem{corollary}{Corollary}
\newtheorem{definition}{Definition}
\usepackage{thmtools}
\usepackage{thm-restate}
\usepackage{subcaption}
\usepackage{etoc}

\ifdefined\XeTeXversion
\catcode"2013=\active
\def–{\textendash}
\catcode"2014=\active
\def—{\textemdash}
\fi

\title{Beyond Expectation: Tail-Sensitive RL through Exponential Utility}

\author{\name Chandradithya S Jonnalagadda\thanks{Work done while at Indian Institute of Technology, Tirupati. Now at Brown University.} \email cs21b059@iittp.ac.in \\
      \addr Indian Institute of Technology, Tirupati
      \AND
      \name Nandhavardhan Chowdhary Namboori\thanks{Work done while at Indian Institute of Technology, Tirupati. Now at Indian Institute of Science.} \email cs21b037@iittp.ac.in \\
      \addr Indian Institute of Technology, Tirupati
      \AND
      \name Deep Kumar Ganguly\thanks{Funded by the Deutsche Forschungsgemeinschaft (DFG) research training group ConVeY (GRK 2428).} \email deep.ganguly@tum.de\\
      \addr Technical University of Munich
      \AND
      \name Ajin George Joseph \email ajin@iittp.ac.in \\
      \addr Indian Institute of Technology, Tirupati
}

\def\month{MM}  
\def\year{YYYY} 
\def\openreview{\url{https://openreview.net/forum?id=XXXX}} 

\begin{document}
\etocdepthtag.toc{main}

\maketitle
\let\AND\relax 

\begin{abstract}
Reinforcement learning conventionally optimizes for expected cumulative reward, leaving policies vulnerable to rare, catastrophic tail events in safety-critical domains. While risk-sensitive methods utilizing risk measures or distributional RL address these vulnerabilities, they frequently suffer from time-inconsistency, non-convex optimization landscapes, and a lack of rigorous convergence guarantees under function approximation. To overcome these limitations, we present a tail-sensitive RL framework grounded in exponential utility, utilizing a single parameter for risk aversion and risk seeking, with an explicit risk-neutral branch at zero. Our central theoretical contribution is a proof that the resulting non-linear risk-sensitive Bellman operator for the nested entropic criterion is a $\gamma$-contraction under the sup-norm, guaranteeing existence and uniqueness of the optimal value function and motivating Risk-Sensitive Policy Iteration (RSPI). To enable scalable, model-free learning, we introduce the Relaxed-RSTD critic, which overcomes the numerical instability of the log-sum-exp Bellman target by deriving a tractable variational lower bound via the Donsker-Varadhan inequality, with an implicit Kullback-Leibler penalty replacing direct transition-distribution access. With sufficient ridge regularization, this formulation yields a strongly convex objective. Under the stated conditions, we obtain a finite-sample parameter convergence bound with linear function approximation. We couple this critic with a symmetric, simultaneous perturbation-based actor to obtain the fully on-policy algorithm: Relaxed Risk-Sensitive Actor-Critic (Relaxed-RS-AC). Empirical results across both discrete environments and high-dimensional continuous control tasks demonstrate that the proposed approach captures complex risk preferences and remains competitive with established baselines, while providing stable, interpretable control over the risk–return trade-off.
\end{abstract}

\section{Introduction}\label{sec:intro}
Reinforcement Learning (RL) achieves competitive performance across domains from game-playing \citep{doi:10.1126/science.aar6404} to robotics \citep{duan2016benchmarking} by optimizing expected cumulative reward. Real-world applications, however, involve high-stakes decisions with disproportionate downside, where optimizing the expectation alone yields policies vulnerable to rare but catastrophic events \citep{dulac2021challenges}: healthcare \citep{gottesman2019guidelines}, finance \citep{theate2021application}, robotics \citep{thananjeyan2021recovery}, and autonomous driving \citep{zhu2021survey} all demand explicit risk mitigation. Hence the interest in risk-sensitive RL (RSRL) \citep{howard1972risk, garcia2015comprehensive}, which accounts for variability and tail risk in the return distribution \citep{tamar2015optimizing, theate_risk-sensitive_2022}.

We use exponential utility to model preferences over cumulative rewards \citep{howard1972risk, heger1994consideration, Chapman2023, noorani_risk-sensitive_2023}. The static objective is the entropic certainty equivalent of the return. Negative $\beta$ is risk-averse and positive $\beta$ is risk-seeking. At $\beta=0$ we explicitly use the risk-neutral expected-return objective.

Our main contributions are:
\begin{itemize}
    \item \textbf{Theoretical foundation:} both the policy-specific and optimal risk-sensitive Bellman operators are $\gamma$-contractions in sup-norm and are exact for the nested entropic criterion, implying a unique optimal value function and motivating Risk-Sensitive Policy Iteration (RSPI) with monotonic improvement.
    \item \textbf{Tractable critic:} the Relaxed-RSTD critic uses a Donsker-Varadhan variational lower bound to convert the intractable log-sum-exp Bellman target into a KL-penalized objective that is provably strongly convex once the ridge parameter $\lambda$ exceeds an explicit curvature threshold. Under the stated gradient conditions, we obtain an $O(1/t)$ mean-square parameter error bound for the on-policy critic with linear value features in the discounted infinite-horizon setting.
    \item \textbf{Full algorithm:} coupling Relaxed-RSTD with a symmetric simultaneous perturbation-based actor yields Relaxed-RS-AC, a fully on-policy algorithm with almost-sure actor convergence.
    \item \textbf{Selecting $\beta$.} We use a normalized preference and empirical cumulant checks to guide selection. We also give theoretical certificates under explicit sufficient conditions and compare the resulting risk--return trade-offs with CVaR-based and distributional baselines.
\end{itemize}

\section{Related Work}\label{sec:related-work}
Risk-Sensitive RL \citep{howard1972risk, heger1994consideration} addresses the risk arising from environmental uncertainty by incorporating risk measures such as variance, Value-at-Risk (VaR), Conditional Value-at-Risk (CVaR) \citep{tamar2015optimizing, chow2015risk}, Entropic Value-at-Risk (EVaR) \citep{Ahmadi-Javid-EVaR, ganguly2025riskseeking}, and exponential utility \citep{borkar2002q, noorani_risk-sensitive_2021, noorani_risk-sensitive_2023}. These measures enable agent behavior to align with task-specific risk preferences \citep{petrik2012approximate, garcia2015comprehensive, achiam_constrained_2017}. However, traditional mean-variance formulations \citep{sobel1982variance, tamar2012policy} are inherently limited. By retaining only the first two statistical moments, they effectively represent the limit of zero risk aversion. At moderate levels of risk aversion, omitting higher-order corrections fails to yield qualitatively distinct, risk-aware policies. While CVaR-based approaches offer robust tail-risk sensitivity, optimizing them in sequential decision-making is notoriously difficult. CVaR often leads to time-inconsistent policies, non-convex optimization landscapes, and notably operates on the conditional left tail without a comprehensive cumulant representation \citep{heger1994consideration}. Direct entropic risk measures, which serve as a conceptual dual to EVaR \citep{Ahmadi-Javid-EVaR}, offer a way to implicitly weight all statistical moments without these optimization hurdles.

\citet{pmlr-v139-fei21a} obtain episodic regret bounds for entropic RL with function approximation by debiasing the finite-horizon regression targets. For the static discounted objective, time-indexed dynamic programs use a decaying risk level $\beta_t=\beta\gamma^t$ \citep{hau2022rasr, hau2023entropic}. \citet{hau2023dynamic} characterize when such decompositions are exact. Our stationary recursion keeps $\beta$ fixed and evaluates the nested entropic criterion, with an explicit bound on its gap to the static objective in Section~\ref{app:static-nested}. For fixed risk aversion, \citet{thoppe2026reinforcement} derive tabular Q-learning algorithms from this type of stationary recursion. \citet{naskar2026finitetime} establish finite-time bounds for both tabular algorithms under Markovian sampling. Our finite-sample result concerns the parameters of a variational TD critic with linear value features in a policy evaluation problem (Theorem~\ref{theorem:finitesamplerrorbound}).

Exponential-utility policy search includes Monte Carlo policy gradients \citep{noorani_risk-sensitive_2021-1} and actor-critic methods with multiplicative TD updates \citep{ExponentialTDLearning, noorani_risk-sensitive_2023}. \citet{enders_risk-sensitive_2024} adapt SAC to improve robustness to distribution shifts in discrete control. \citet{tasdighi2025improving} use log-exponential weighting to calibrate epistemic uncertainty in risk-neutral training. The rsEAC method uses normalization and clipping to stabilize log-domain critics \citep{granados2026rseac}. Relaxed-RS-AC uses a Donsker--Varadhan relaxation of the transition expectation and learns a policy through symmetric perturbations.

Distributional RL models the return distribution, allowing a risk preference to be applied to the learned distribution \citep{pmlr-v70-bellemare17a}. DSAC uses this approach in continuous control \citep{ma2020dsac}, and modified distributional Bellman operators support CVaR-aligned learning \citep{lim2022distributional}. The entropic criterion encodes the risk preference in a scalar value function.

\section{Risk-Sensitive MDP}\label{sec:risk-sensitive-mdp}
We model the agent–environment interaction as a Markov Decision Process (MDP) \citep{puterman2014markov, Sutton1998}  defined by the tuple $(\mathcal{S}, \mathcal{A}, P, r)$. Here, $\mathcal{S}$ is the state space, $\mathcal{A}$ is the action space, and $P: \mathcal{S} \times \mathcal{A} \rightarrow \Delta(\mathcal{S})$ denotes the transition probability distribution over next states $s' \in \mathcal{S}$. The function $r: \mathcal{S} \times \mathcal{A} \rightarrow \mathbb{R}$ determines the reward $r_t = r(s_t, a_t)$ received at time $t$ after taking action $a_t$ in state $s_t$. The agent behaves according to a deterministic policy $\pi \in \Pi$, where $\Pi$ denotes the set of tractable policies.
The operator and policy-iteration analysis considers all stationary deterministic Markov policies in the finite MDP. The actor in Section~\ref{sec:actor} searches a parameterized stochastic policy class $\Pi_\psi$. Relaxed-RSTD evaluates a fixed policy through its induced Markov chain.

Let $\gamma\in[0,1)$ be the discount factor, let $\tau=(s_0,a_0,r_0,s_1,a_1,r_1,\ldots)$ be a trajectory, and let $R(\tau)=\sum_{t=0}^\infty\gamma^t r_t$ be its return. For a policy $\pi$, the static entropic objective is
\begin{equation}
\label{eq:risk-rl-obj}
J_\beta(\pi)=
\begin{cases}
\displaystyle\frac{1}{\beta}\log\mathbb{E}_{\tau\sim\pi}[e^{\beta R(\tau)}], & \beta\ne0,\\[4pt]
\mathbb{E}_{\tau\sim\pi}[R(\tau)], & \beta=0.
\end{cases}
\end{equation}
We write $J_\beta(\psi)=J_\beta(\pi_\psi)$ for the parameterized policy. The nonzero-$\beta$ formula is undefined at zero. The continuous limit is the expected return, because the derivative at zero of $\log\mathbb{E}[e^{\beta R}]$ is $\mathbb{E}[R]$ for bounded returns. The objective and algorithms therefore use an explicit risk-neutral branch at $\beta=0$.

The Taylor expansion in Section~\ref{taylorseriesexp} gives the approximation
\begin{equation}
\label{eq:taylorexp}
J_\beta(\pi)\approx\mathbb{E}[R(\tau)]+\frac{\beta}{2}\operatorname{Var}[R(\tau)].
\end{equation}
The omitted remainder is $\mathcal{O}(\beta^2)$ as $\beta\to0$. Near zero, negative $\beta$ penalizes variance to first order and positive $\beta$ rewards it. For a bounded random variable $X$, we define the entropic certainty equivalent by
\[
\rho^{\mathrm{ent}}_\beta[X]=
\begin{cases}
\displaystyle\frac{1}{\beta}\log\mathbb{E}[e^{\beta X}], & \beta\ne0,\\[4pt]
\mathbb{E}[X], & \beta=0.
\end{cases}
\]
The conditional version uses a conditional expectation. Our recursion follows \citet{ruszczynski2010risk} for Markov risk evaluation and \citet{bauerle2018growth} for the risk-averse entropic case.

For a fixed policy $\pi$, its nested entropic value $U^\pi_\beta$ is the bounded solution of
\[
U^\pi_\beta(s)=r(s,\pi(s))+\gamma\rho^{\mathrm{ent}}_\beta[U^\pi_\beta(s')\mid s,\pi(s)].
\]
The criterion is nested because the conditional certainty equivalent acts on a continuation value defined by the same recursion.

For $\beta\ne0$, the critic uses the scaled value $\bar V^\pi_\beta=\beta U^\pi_\beta$. Multiplying the nested recursion by $\beta$ gives
\begin{equation}
\label{eq:bellman}
\bar V^\pi_\beta(s)=\beta r(s,\pi(s))+\gamma\log\mathbb{E}[e^{\bar V^\pi_\beta(s')}\mid s,\pi(s)].
\end{equation}
The stationary Bellman recursion in Equation~\ref{eq:bellman} evaluates the nested entropic criterion exactly. The static objective in Equation~\ref{eq:risk-rl-obj} applies a single certainty equivalent to the full discounted return. For a fixed policy and discount factor $\gamma$, the gap between the two values is $O(|\beta|)$ as $\beta$ approaches zero (Proposition~\ref{prop:static-nested} in Section~\ref{app:static-nested}).

We write $T^\pi_\beta$ for the fixed-policy operator on the right-hand side of Equation~\ref{eq:bellman}. Proposition~\ref{prop:nested-exact} gives the scaling identity. Policy improvement takes place in certainty-equivalent units. For any bounded function $U$, define the one-step score and both Bellman operators by
\begin{align*}
Q_{\beta,U}(s,a)&=r(s,a)+\gamma\rho^{\mathrm{ent}}_\beta[U(s')\mid s,a],\\
(\mathcal{T}^\pi_\beta U)(s)&=Q_{\beta,U}(s,\pi(s)),\\
(\mathcal{T}^*_\beta U)(s)&=\max_{a\in\mathcal{A}}Q_{\beta,U}(s,a).
\end{align*}
For $\beta\ne0$, the score is $r(s,a)+(\gamma/\beta)\log\mathbb{E}[e^{\beta U(s')}\mid s,a]$. For $\beta=0$, it is $r(s,a)+\gamma\mathbb{E}[U(s')\mid s,a]$. The greedy step maximizes the certainty equivalent for every choice of $\beta$. In scaled units, multiplying by a negative $\beta$ reverses the ordering. Maximizing the scaled score for $\beta<0$ would minimize the certainty equivalent. For example, when $\gamma=0$, maximizing $\beta r(s,a)$ with $\beta<0$ selects a smallest immediate reward, while maximizing $Q_{\beta,U}(s,a)$ selects a largest immediate reward.

\begin{lemma}
\label{lemma:bellman-operator-contraction-mapping}
For any fixed policy $\pi$ and any bounded $U_1,U_2:\mathcal{S}\to\mathbb{R}$, the operator $\mathcal{T}^\pi_\beta$ satisfies
\[
\|\mathcal{T}^\pi_\beta U_1-\mathcal{T}^\pi_\beta U_2\|_\infty\le\gamma\|U_1-U_2\|_\infty.
\]
\end{lemma}

\begin{lemma}
\label{lemma:fixed-point-optimal-bellman-operator}
The optimal certainty-equivalent value $U^*_\beta(s):=\sup_\pi U^\pi_\beta(s)$ is a fixed point of $\mathcal{T}^*_\beta$, so $\mathcal{T}^*_\beta U^*_\beta=U^*_\beta$.
\end{lemma}

\begin{lemma}
\label{lemma:optimal-bellman-contraction-mapping}
For any bounded $U_1,U_2:\mathcal{S}\to\mathbb{R}$, the operator $\mathcal{T}^*_\beta$ satisfies
\[
\|\mathcal{T}^*_\beta U_1-\mathcal{T}^*_\beta U_2\|_\infty\le\gamma\|U_1-U_2\|_\infty.
\]
\end{lemma}

Together, these results give a unique $U^\pi_\beta$ for each policy and identify $U^*_\beta$ as the unique fixed point of the optimal operator. They hold for both signs of $\beta$ and for the risk-neutral branch. Section~\ref{app:policy-iteration-proofs} gives the proofs.

\subsection{Risk-Sensitive Policy Iteration}
Risk-Sensitive Policy Iteration (RSPI) alternates between evaluating the fixed point $U^{\pi_t}_\beta=\mathcal{T}^{\pi_t}_\beta U^{\pi_t}_\beta$ and choosing a policy $\pi_{t+1}$ that is greedy for $Q_{\beta,U^{\pi_t}_\beta}$. Algorithm~\ref{alg:rspi} gives the procedure with access to the transition model.

\begin{theorem}
\label{thm:monotonic-improvement-rspi}
Let $\{\pi_t\}_{t\ge0}$ be the RSPI iterates obtained by exact policy evaluation and greedy improvement in certainty-equivalent units. For every state $s$ and iteration $t$, $U_\beta^{\pi_{t+1}}(s)\ge U_\beta^{\pi_t}(s)$. The sequence $U_\beta^{\pi_t}$ converges to $U_\beta^*$ for every $\beta\in\mathbb{R}$.
\end{theorem}
The guarantee applies to the nested entropic criterion in Section~\ref{sec:risk-sensitive-mdp}. Section~\ref{app:policy-iteration-proofs} proves both claims.

\begin{algorithm}[bp]
\footnotesize
\caption{Risk-Sensitive Policy Iteration (RSPI)}
\label{alg:rspi}
\begin{algorithmic}[1]
\REQUIRE Specify a finite MDP $(\mathcal{S},\mathcal{A},P,r)$, discount $\gamma\in[0,1)$, and risk parameter $\beta$.
\STATE Initialize a deterministic policy $\pi$.
\REPEAT
    \STATE Evaluate $U=\mathcal{T}^\pi_\beta U$ using the score $Q_{\beta,U}$.
    \STATE Set $\pi_{\mathrm{old}}\leftarrow\pi$.
    \FOR{each state $s\in\mathcal{S}$}
        \STATE Choose $\pi(s)\in\arg\max_{a\in\mathcal{A}}Q_{\beta,U}(s,a)$. Keep $\pi_{\mathrm{old}}(s)$ if it is greedy.
    \ENDFOR
\UNTIL{$\pi=\pi_{\mathrm{old}}$}
\RETURN the policy $\pi$ and its certainty-equivalent value $U$.
\end{algorithmic}
\end{algorithm}

\section{Relaxed Risk-Sensitive Actor-Critic}

In this section, we propose our algorithm, Relaxed Risk-Sensitive Actor-Critic (Relaxed RS-AC).

\subsection{Relaxed Risk-Sensitive TD-Learning}
\label{subsec:rstd}

Model-free estimation of \(\bar{V}^\pi_\beta\) sees only sampled transitions, and
because the next-state expectation sits inside a logarithm, naive sample averaging
is biased and the log-sum-exp is numerically unstable.
\citet{enders_risk-sensitive_2024} rewrite the RHS of (\ref{eq:bellman}) with a
smoothing trick \citep{nesterov2005smooth}, which fixes the instability but still
needs multiple trajectories per unbiased estimate. We convert the intractable log-sum-exp term into a tractable optimization problem by introducing a variational distribution $q(s_{t+1}\mid s_t; \omega)$, parameterized by $\omega$, that approximates the underlying transition distribution $p(s_{t+1}\mid s_t)$. The Donsker--Varadhan inequality \citep{donsker1976asymptotic} then lower-bounds the
log-sum-exp term,
$\log\mathbb{E}_p[e^{\bar V^\pi_\beta(s_{t+1})}]\ge
\mathbb{E}_q[\bar V^\pi_\beta(s_{t+1})]-\mathcal{D}_{\mathrm{KL}}(q\|p)$.

\paragraph{Continuous state spaces.}
For compact $\mathcal{S}\subset\mathbb{R}^n$ we take $q(\cdot\mid s;\omega)=\mathcal{N}(\mu_\omega(s),\Sigma_\omega(s))$ with spectrally normalized networks \citep{miyato2018spectral, khromov2024some} and $\sigma_{\min}^2I\preceq\Sigma_\omega\preceq\sigma_{\max}^2I$, so $\log q$ is closed-form and $\mathbb{E}_q[\phi^\top\theta]$ uses reparameterized sampling \citep{kingma2014vae}. Mixtures and normalizing flows with exact log-densities satisfy the same conditions \citep{papamakarios2021flows}. Section~\ref{app:critic-convergence} gives the assumption and all continuous-state constants.

Substituting into (\ref{eq:bellman}) lower-bounds $\bar V^\pi_\beta(s_t)$ by
$\beta r(s_t,a_t)+\gamma(\mathbb{E}_q[\bar V^\pi_\beta(s_{t+1})]-\mathcal{D}_{\mathrm{KL}}(q\|p))$.
Approximating the KL term by the negative log-likelihood of $q$ on the observed
data, which penalizes $q$ for deviating from $p$ using samples alone, gives the
tractable risk-sensitive Bellman equation
\begin{equation}
\label{eq:tractable-bellman}
\begin{split}
    \bar{V}^\pi_\beta&(s_t) \approx \beta r(s_t, a_t) \\
    &\quad + \gamma\left(\mathbb{E}_q\left[ \bar V^\pi_\beta(s_{t+1})\right] + \log q(s_{t+1}\mid s_t; \omega_t) \right).
\end{split}
\end{equation}

The tightness of the Donsker--Varadhan lower bound is governed by the
divergence between the learned variational distribution $q_t$ and the optimal
exponentially tilted distribution
$q^*_s(s_{t+1}) \propto e^{\bar V^\pi_\beta(s_{t+1})}\, p(s_{t+1}\mid s_t)$,
rather than between $q_t$ and $p$: the slack in the bound is exactly
$\mathcal{D}_{\mathrm{KL}}(q(\cdot\mid s_t)\|q^*_s)$, so it is tight iff $q=q^*_s$
(Section~\ref{app:ce-vs-kl}). However, the cross-entropy step maximizes the empirical log-likelihood of
$q_\omega$, hence minimizes the forward divergence
$\mathcal{D}_{\mathrm{KL}}(p\|q_\omega)$ up to the $\omega$-independent constant
$H(p(\cdot\mid s))$: $q_t$ converges to $p$, not to $q^*_s$, an inherent asymmetry
\citep{minka2005divergence}. At $q_t=p$ a residual Jensen gap persists
\begin{equation}
\label{eq:jensen-gap}
\begin{split}
    \delta_\pi(s) &:= \log\mathbb{E}_p\!\left[e^{\bar V^\pi_\beta(s')}\,\middle|\, s\right]
    - \mathbb{E}_p\!\left[\bar V^\pi_\beta(s')\,\middle|\, s\right] \\
    &\;= \mathcal{D}_{\mathrm{KL}}\big(p(\cdot\mid s)\,\big\|\,q^*_s\big)\;\ge\;0,
\end{split}
\end{equation}
with equality if and only if $\bar V^\pi_\beta(s')$ is constant
$p(\cdot\mid s)$-almost surely. In particular, $\delta_\pi \equiv 0$ under
deterministic dynamics. This irreducible offset is ultimately absorbed into
the critic, together with the entropy offset $-\gamma H(p(\cdot\mid s))$ that
the cross-entropy surrogate introduces: at $q_\omega = p$ the implemented
critic performs standard TD on the entropy-shaped reward
$\hat r(s) = \beta r(s,\pi(s)) - \gamma H(p(\cdot\mid s))$, and the induced
fixed-point bias is at most
$\tfrac{\gamma}{1-\gamma}\sup_s\left[\delta_\pi(s)+H(p(\cdot\mid s))\right]$
(see Section~\ref{app:ce-vs-kl} for the formal statements).

To approximate $\bar{V}^\pi_\beta$ we adopt \emph{linear function approximation}
(LFA), chosen for its finite-sample guarantees and scalability. We denote by
\(  \Phi \in \mathbb{R}^{\,\lvert S\rvert \times k},  k \ll \lvert S\rvert,
\)
the fixed feature matrix whose \(i^{\text{th}}\) row $\phi(s_i) \in \mathbb{R}^k$ embeds state \(s_i\).
With a weight vector \(\theta \in \mathbb{R}^k\) the approximate value is
\(
  \bar V_{\beta}^{\pi}(s)
  \;=\;
  \phi(s)^{\!\top}\theta
  \;=\;
  (\Phi\theta)_s,
\)
a compact differentiable surrogate that generalizes over large or continuous state
spaces without explicit enumeration.

\begin{assumption}
\label{assumption:standing}
$q$ is softmax-parameterized (conditionally Gaussian in continuous spaces), twice
continuously differentiable in $\omega$ with bounded derivatives of $\log q$. Every
induced chain is ergodic. The parameters $(\theta,\omega)$ lie in a convex compact
set $\mathcal{K}\times\Omega$ with bounded features. The step sizes satisfy $\sum_t\alpha_t=\infty$,
$\sum_t\alpha_t^2<\infty$.
\end{assumption}

Let $\phi(s)$ denote the feature vector, let $\theta$ be the linear value weights, and let $\omega$ parameterize the variational transition model $q$. Write $\mathcal{K}$ and $\Omega$ for their respective feasible sets. For a fixed policy and $\beta\ne0$, the risk-sensitive TD target is
\begin{equation}
y_t(\theta,\omega) := \beta r(s_t,a_t) + \gamma\Big(\mathbb{E}_{q}\big[\phi^\top(\bar{s}_{t+1})\theta\big] + \log q(s_{t+1}\mid s_t;\omega)\Big)\nonumber,
\end{equation}
Here $\bar{s}_{t+1}\sim q(\cdot\mid s_t;\omega)$. We define the TD residual as $\delta(\theta,\omega)=y_t(\theta,\omega)-\phi^\top(s_t)\theta$. The regularized objective averages the squared residual over the stationary state distribution of the fixed policy and its observed transitions:
\begin{equation}
\label{eq:Fthetaomega}
F(\theta,\omega) = \tfrac{1}{2}\,\mathbb{E}\big[(y_t(\theta,\omega) - \phi^\top(s_t)\theta)^2\big] + \tfrac{\lambda}{2}\big(\|\theta\|^2 + \|\omega\|^2\big),
\end{equation}
Here $\lambda>0$ is the ridge regularization weight. Parameter norms are Euclidean. To analyze SGD for this objective, write $x_t=(\theta_t,\omega_t)^\top$, let $\alpha_t$ be the step size, and let $g_t=\nabla\tilde F_t(x_t)$ be the stochastic gradient of the sample loss $\tilde F_t$ evaluated at the current parameters. We use the update
\begin{equation}
    \label{eq:sgdRelaxed}
    (\theta_{t+1}, \omega_{t+1})^{\top} = (\theta_{t}, \omega_{t})^{\top} - \alpha_t g_t.
\end{equation}

The ridge term makes $F$ strongly convex when $\lambda>L_H$, where $L_H$ bounds the negative curvature of the unregularized loss $\ell$. Section~\ref{app:critic-convergence} derives the threshold.

\begin{theorem}
\label{theorem:finitesamplerrorbound}
For the finite-state critic, let Assumptions~\ref{assumption:vardist1}--\ref{assumption:smoothness} hold and let $\lambda>L_H$. Then $F$ is $\mu_F$-strongly convex, where $\mu_F=\lambda-L_H>0$, and has a unique minimizer $x^*:=(\theta^*,\omega^*)^\top$ over $\mathcal{K}\times\Omega$.
Assume that the iterates $x_t$ of Equation~\ref{eq:sgdRelaxed} stay in $\mathcal{K}\times\Omega$ and that the stationary mean of $g_t$ at fixed parameters equals $\nabla F$.

\textbf{Almost-sure convergence.}
The iterates $x_t$ converge to $x^*$ almost surely.

\textbf{Finite-sample bound.}
Let $\mathcal F_t=\sigma\bigl(\{s_i,a_i,r_i,x_i\}_{i=0}^t\bigr)$ denote the history through iteration $t$. Suppose in addition that $g_t$ is conditionally unbiased, so $\mathbb{E}[g_t\mid\mathcal{F}_t]=\nabla F(x_t)$. Let $L=L_0+\lambda$ be the Lipschitz constant for $\nabla F$, where $L_0$ bounds $\|\nabla^2\ell\|$ on $\mathcal{K}\times\Omega$ (Lemma~\ref{lemma:FhasLlipcontgrad}). Let $\sigma^2$ uniformly bound the mean squared gradient noise $\mathbb{E}\|g_t-\nabla F(x_t)\|^2$ as in Lemma~\ref{lemma:boundedstochgradnoise}. For $\nabla F(x^*)=0$, these constants imply
\[
\mathbb{E}\|g_t\|^2\le L^2\mathbb{E}\|x_t-x^*\|^2+\sigma^2.
\]
If $\nabla F(x^*)\ne0$, choose $\sigma^2$ to also satisfy this second-moment bound. Lemma~\ref{lemma:boundedstochgradnoise} gives a uniform choice that works in either case.
With step size $\alpha_t=1/(\mu_F t)$ for $t\ge1$, the iterates satisfy
\[
\mathbb{E}\|x_t-x^*\|^2\le\frac{4\sigma^2}{\mu_F^2t}+\mathcal{O}(t^{-2})
\]
for sufficiently large $t$.
\end{theorem}
Section~\ref{app:critic-convergence} gives the proofs and the explicit constant $L_H$. It also gives the continuous-state regularity bounds under Assumption~\ref{assumption:contq}. 

This controls the distance to the minimizer of $F$, not the mismatch between that
fixed point and $\bar V^\pi_\beta$. Section~\ref{app:ce-vs-kl} gives a value-space decomposition
$\|\Phi\theta_t-\bar V^\pi_\beta\|_\infty\le\varepsilon_{\mathrm{opt}}(t)+
\varepsilon_{\mathrm{approx}}+\varepsilon_{\mathrm{model}}+\varepsilon_{\mathrm{relax}}$
separating the $\mathcal{O}(1/\sqrt t)$ optimization error from the LFA/ridge,
variational-model, and irreducible relaxation biases.

Algorithm~\ref{alg:rrstd} gives the practical critic update. For $\beta=0$, it uses the ordinary reward TD target and omits the variational update. Theorem~\ref{theorem:finitesamplerrorbound} applies when the stochastic updates satisfy its stated gradient conditions.

\begin{algorithm}[tbp]
\footnotesize
\caption{Relaxed Risk-Sensitive Temporal Difference (Relaxed-RSTD)}
\label{alg:rrstd}
\begin{algorithmic}[1]
\REQUIRE Use a transition $(s,a,r,s')$, discount $\gamma\in[0,1)$, and risk parameter $\beta$.
\STATE Use value weights $\theta$, dynamics weights $\omega$, learning rates $\alpha_\theta,\alpha_\omega$, and regularization weights $\lambda_\theta,\lambda_\omega$.
\STATE \textit{(1. Compute the TD target, taking $V_\theta(s') = 0$ if $s'$ is terminal)}
\IF{$\beta \neq 0$}
    \STATE $y \leftarrow \beta r + \gamma \left( V_\theta(s') + \log q_\omega(s'|s) \right)$
\ELSE
    \STATE $y \leftarrow r + \gamma V_\theta(s')$
\ENDIF
\STATE \textit{(2. Value function update)}
\STATE $\delta \leftarrow y - V_\theta(s)$
\STATE $\theta \leftarrow \theta + \alpha_\theta \left( \delta \nabla_\theta V_\theta(s) - \lambda_\theta \theta \right)$
\STATE \textit{(3. Dynamics model update)}
\IF{$\beta \neq 0$}
    \STATE $\omega \leftarrow \omega + \alpha_\omega \left( \nabla_\omega \log q_\omega(s'|s) - \lambda_\omega \omega \right)$
\ENDIF
\RETURN the updated critic weights $\theta$ and $\omega$.
\end{algorithmic}
\end{algorithm}

\subsection{Simultaneous Perturbation-Based Actor}
\label{sec:actor}

Gradient-free policy optimization in high dimensions can perturb all coordinates at
once and infer the ascent direction from the resulting performance change. We use a
symmetric two-sided estimate: for a policy $\psi_t\in\mathbb{R}^d$, draw
$\Delta_t\in\{-1,1\}^d$ and a vanishing $c_t>0$, set
$\psi_t^{\pm}:=\psi_t\pm c_t\Delta_t$, run one rollout under each and feed their
initial states $s_0^{\pm}$ to the Relaxed-RSTD critic.
For $\beta\ne0$, the critic output $\phi(s)^\top\theta$ estimates the scaled nested value $\bar V^\pi_\beta(s)=\beta U^\pi_\beta(s)$. We convert each perturbed critic output to reward units before comparing the two estimates:
\begin{equation}
  \widehat{J}_{\psi_t^{\pm}}
  \;:=\;
  \begin{cases}
    \dfrac{1}{\beta}\,\phi(s_0^{\pm})^{\top}\theta, & \beta \neq 0,\\[6pt]
    \phi(s_0^{\pm})^{\top}\theta, & \beta = 0,
  \end{cases}.
  \label{eq:actor-ce-estimate}
\end{equation}
For $\beta>0$, division by $\beta$ preserves the ordering of the scaled critic estimates. For $\beta<0$, it reverses that ordering so the ascent step favors the larger certainty-equivalent estimate. The $\beta=0$ branch reads the risk-neutral critic, whose target is already in reward units. The symmetric estimate and the actor update with step size $a_t$ are
\begin{equation}
  \widehat{g}_t
  \;=\;
  \frac{\widehat{J}_{\psi_t^{+}}-\widehat{J}_{\psi_t^{-}}}{2c_t}
  \,\Delta_t^{-1},
  \qquad
  \psi_{t+1} = \psi_t + a_t\,\widehat{g}_t.
  \label{eq:twosided-update}
\end{equation}

Let $J$ be the objective evaluated at the perturbed policies and let $\mathcal F_t$ be the actor history before drawing $\Delta_t$. Write $\widehat g_t^{\,\circ}$ for the estimator in Equation~\ref{eq:twosided-update} with the exact values $J(\psi_t^\pm)$ in place of $\widehat J_{\psi_t^\pm}$.

\begin{theorem}[Convergence of the two-sided actor]
\label{thm:twosided_convergence}
Let $J\in C^3(\mathbb R^d)$. Assume the iterates of Equation~\ref{eq:twosided-update} are almost surely bounded, and let the step sizes and independent Rademacher perturbations satisfy Assumptions~\ref{ass:ts-step} and~\ref{ass:ts-rad}. Suppose the two estimates each satisfy $\mathbb E|\widehat J_{\psi_t^\pm}-J(\psi_t^\pm)|^2=O(N_t^{-1})$ for $N_t$ inner critic steps, with $\sum_t a_t/(c_t\sqrt{N_t})<\infty$. Then the exact-value estimator has conditional bias
\[
\bigl\|\mathbb E[\widehat g_t^{\,\circ}\mid\mathcal F_t]-\nabla J(\psi_t)\bigr\|=O(c_t^2),
\]
with a uniform constant on bounded parameter sets. The iterates satisfy $\|\nabla J(\psi_t)\|\to0$ almost surely, and all their limit points are critical points of $J$.

Let $\psi^\star$ be a critical point with $\nabla^2J(\psi^\star)\prec0$. If from some deterministic time onward the iterates stay almost surely in a sufficiently small closed ball about $\psi^\star$, then $\psi_t\to\psi^\star$ almost surely. In this case, for $a_t=t^{-\alpha}$ with $\tfrac23<\alpha<1$, the rate is $\mathbb E\|\psi_t-\psi^\star\|^2=O(t^{-\alpha})$ if $c_t^4+(c_t^2N_t)^{-1}=O(a_t)$. One schedule meeting these requirements is $c_t=t^{-\alpha/4}$ and $N_t\ge t^{3\alpha/2}$.
\end{theorem}

Section~\ref{app:proof-twosided} proves Theorem~\ref{thm:twosided_convergence}. For a fixed initial state $s_0$, the critic-based actor optimizes the surrogate $J_c(\psi)=\phi(s_0)^\top\theta^*(\psi)/\beta$ for $\beta\ne0$ and $J_c(\psi)=\phi(s_0)^\top\theta^*(\psi)$ for $\beta=0$. Here $\theta^*(\psi)$ is the value-weight component of the minimizer of $F$ under $\pi_\psi$ (Corollary~\ref{cor:actor-surrogate-spsa}). Theorem~\ref{theorem:finitesamplerrorbound} provides the inner critic error bound when its finite-sample assumptions hold uniformly.

If $\sup_{\psi}|J_c(\psi)-J_\beta(\psi)|\le\varepsilon_{\mathrm{tot}}$, every maximizer of $J_c$ is $2\varepsilon_{\mathrm{tot}}$-optimal for $J_\beta$ (Proposition~\ref{prop:actor-surrogate-gap}). Section~\ref{app:ce-vs-kl} bounds the optimization, approximation, model, and relaxation errors. Section~\ref{app:static-nested} bounds the gap between the nested and static criteria. Section~\ref{app:proof-twosided} also treats the signed $O(1/m)$ bias of the Monte Carlo estimator that takes the logarithm of an empirical mean of $m$ samples. This bias is optimistic when $\beta<0$.

Algorithm~\ref{alg:rrsac} combines Relaxed-RSTD with the symmetric actor update. The estimates in Equation~\ref{eq:actor-ce-estimate} divide the scaled critic values by $\beta$ for both nonzero signs. When $\beta<0$, that division puts the two perturbed policy estimates in reward units before the ascent step in Equation~\ref{eq:twosided-update}. At $\beta=0$, the algorithm reads the risk-neutral critic directly.

\begin{algorithm}[!hbp]
\footnotesize
\caption{Relaxed Risk-Sensitive Actor-Critic (Relaxed-RS-AC)}
\label{alg:rrsac}
\begin{algorithmic}[1]
\REQUIRE Specify initial policy and critic parameters $\psi,\theta$, and a risk parameter $\beta$.
\STATE Use actor step size $a$, perturbation magnitude $c$, and discount $\gamma\in[0,1)$.
\WHILE{not converged}
    \STATE \textit{(1. Update the base critic and maintain the central value estimate)}
    \STATE Collect trajectory $\tau \sim \pi(\cdot|s;\psi)$
    \STATE Train central critic $\theta$ on $\tau$ using Relaxed-RSTD (Algorithm~\ref{alg:rrstd})
    \STATE \textit{(2. Symmetric SPSA perturbations)}
    \STATE Sample perturbation vector $\Delta \sim \text{Uniform}(\{-1,1\}^d)$
    \STATE $\psi^+ \leftarrow \psi + c\Delta$
    \STATE $\psi^- \leftarrow \psi - c\Delta$
    \STATE \textit{(3. Evaluate positive perturbation)}
    \STATE $\theta^+ \leftarrow \theta$
    \STATE Collect trajectory $\tau^+ \sim \pi(\cdot|s;\psi^+)$
    \STATE Train critic $\theta^+$ on $\tau^+$ using Relaxed-RSTD
    \STATE Compute $\widehat J^+$ at the initial state $s_0^+$ of $\tau^+$ using Equation~\ref{eq:actor-ce-estimate}.
    \STATE \textit{(4. Evaluate negative perturbation)}
    \STATE $\theta^- \leftarrow \theta$
    \STATE Collect trajectory $\tau^- \sim \pi(\cdot|s;\psi^-)$
    \STATE Train critic $\theta^-$ on $\tau^-$ using Relaxed-RSTD
    \STATE Compute $\widehat J^-$ at the initial state $s_0^-$ of $\tau^-$ using Equation~\ref{eq:actor-ce-estimate}.
    \STATE \textit{(5. Actor update)}
    \STATE $\hat{g} \leftarrow \frac{\widehat J^+ - \widehat J^-}{2c}\Delta^{-1}$
    \STATE $\psi \leftarrow \psi + a\hat{g}$
\ENDWHILE
\RETURN the learned policy and critic parameters $\psi$ and $\theta$.
\end{algorithmic}
\end{algorithm}

\section{Experimental Results}
We empirically validate Relaxed-RS-AC across a diverse suite of discrete and continuous control benchmarks. In discrete settings, we benchmark against established CVaR baselines, while in continuous domains, we compare against risk-neutral SAC \citep{haarnoja_soft_2018} and risk-averse DSAC \citep{ma2020dsac}. Evaluating all three risk profiles (averse, neutral, seeking), our analysis focuses on asymptotic mean reward, tail-risk mitigation, and learning dynamics. \S\ref{app:choosing-beta} details our methodology to choose the risk-sensitivity parameter $\beta$. The main text presents results for CliffWalking (\S\ref{sec:gridworlds}), glycemic control (\S\ref{sec:simglucose}), and dynamic portfolio optimization (\S\ref{sec:portfolio_optimization_main}). The appendix reports results for Windy Gridworld (\S\ref{sec:gridworlds-windy}) and active network management (\S\ref{appendix:anm}). Furthermore, comprehensive ablation studies isolate the impact of the risk parameter in finance (\S\ref{sec:portfolio_optimization_main} \& \S\ref{sec:portfolio-experiments}) and feature dimensionality on stability in locomotion tasks (\S\ref{subsec:feature_dim_ablation} \& \S\ref{appendix:feature_dim_ablation}). We report results across five random seeds. The learning experiments in the main text and appendix use the methods described for each comparison. The model-based RSPI procedure in Algorithm~\ref{alg:rspi} is a theoretical result and was not used in these experiments. Section~\ref{appendix:implementation-details} gives the implementation details.

\subsection{A Principled Procedure for Selecting $\beta$}
\label{app:choosing-beta}

Near $\beta=0$, the entropic objective of a return $R$ has the cumulant expansion
\begin{equation}
J_\beta(\pi)=\mathbb{E}[R]+\frac{\beta}{2}\operatorname{Var}[R]
+\sum_{n\ge3}\frac{\beta^{n-1}}{n!}\kappa_n(R),
\end{equation}
where $\kappa_n(R)$ is the $n$th cumulant of the return under $\pi$.

In practice, we first evaluate the risk-neutral baseline policy to estimate the return mean $\hat\mu$ and variance $\hat\sigma^2$. For a positive estimated variance, we use a normalized preference $\rho\in[-1,1]$ to set $\beta=\rho|\hat\mu|/\hat\sigma^2$. We then check the estimated third- and fourth-cumulant contributions to see how large they are relative to the variance term. These checks are empirical diagnostics. The restriction $|\rho|\le1$ limits the nominal scale of $\beta$, but by itself gives no guarantee that the variance term dominates the full remainder. The reported sweeps explore the response to $\beta$ without restricting every configuration to a theoretically certified range.

Section~\ref{app:beta-guardrails} gives complementary theoretical bounds. Proposition~\ref{prop:beta-guardrail}(c) controls the full remainder, and Corollary~\ref{cor:beta-certified}(iii) gives a distribution-dependent sufficient condition for variance dominance. The worst-case return-span bound can make the certified range particularly restrictive at long horizons, as discussed in Remark~\ref{rem:beta-guardrail-practice}. The normalized rule and low-order cumulant checks guide the practical choice of $\beta$. An all-orders certificate requires the hypotheses of the appendix bounds.

\subsection{CliffWalking}
\label{sec:gridworlds}
We evaluate Relaxed-RS-AC in CliffWalking \citep{Sutton1998} at $\beta\in\{-0.01,0,0.01\}$. The CVaR baselines represent complementary approaches to tail-risk optimization. CVaR-PG directly optimizes tail returns \citep{chow2015risk}, while SDPG-CVaR and our simplified D4PG-CVaR use return-distribution estimates \citep{pmlr-v120-singh20a, barth-maron2018distributional}. Together, they provide reference points for evaluating tail-risk sensitivity and the risk--return trade-off under matched feature representations. All agents use linear function approximation with a feature dimension of $20$, and the CVaR baselines use risk level $0.1$. We also compare with entropic Q-learning \citep{borkar2002q} at the matched nonzero $\beta$ values.

\begin{figure*}[t]
    \centering
    \begin{subfigure}[b]{0.30\textwidth}
        \centering
        \includegraphics[width=\linewidth]{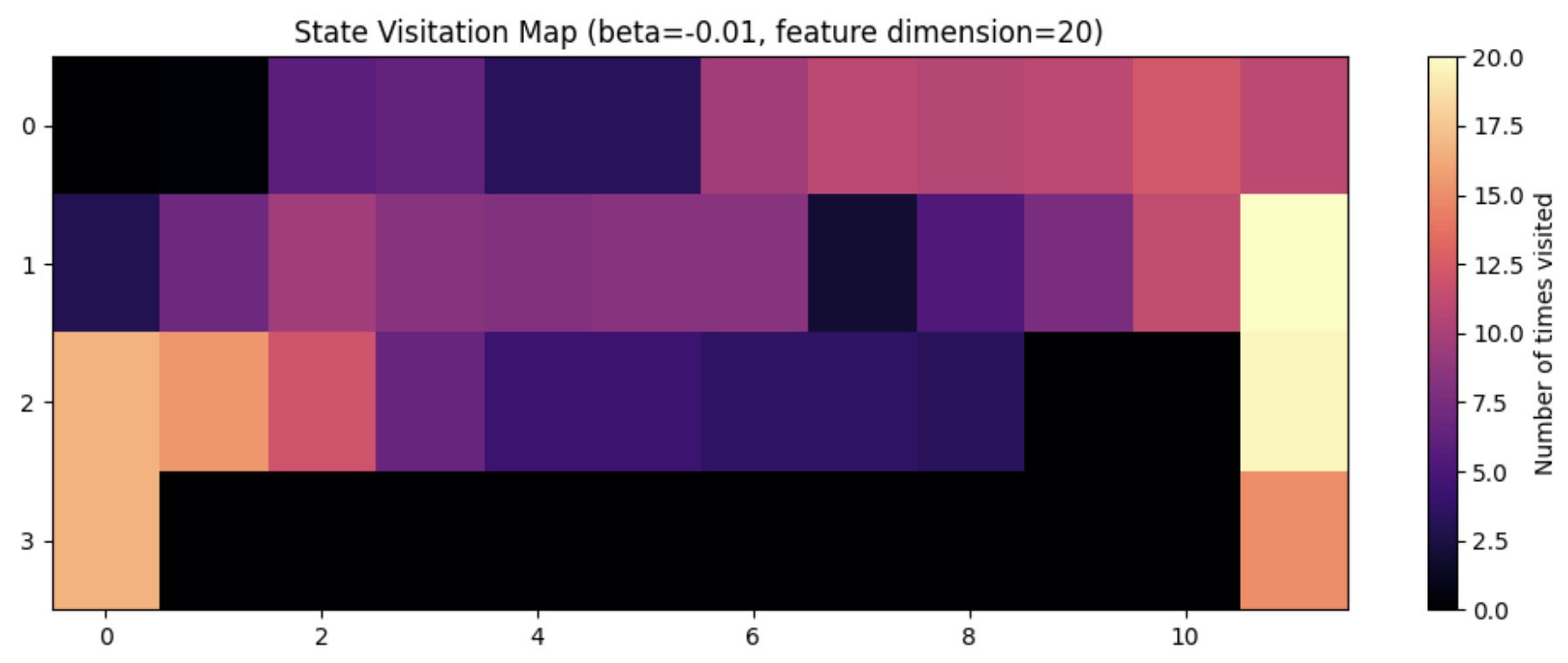}
        \caption{Relaxed-RS-AC $(\beta=-0.01)$}
        \label{fig:cliff_averse}
    \end{subfigure}
    \hfill 
    \begin{subfigure}[b]{0.30\textwidth}
        \centering
        \includegraphics[width=\linewidth]{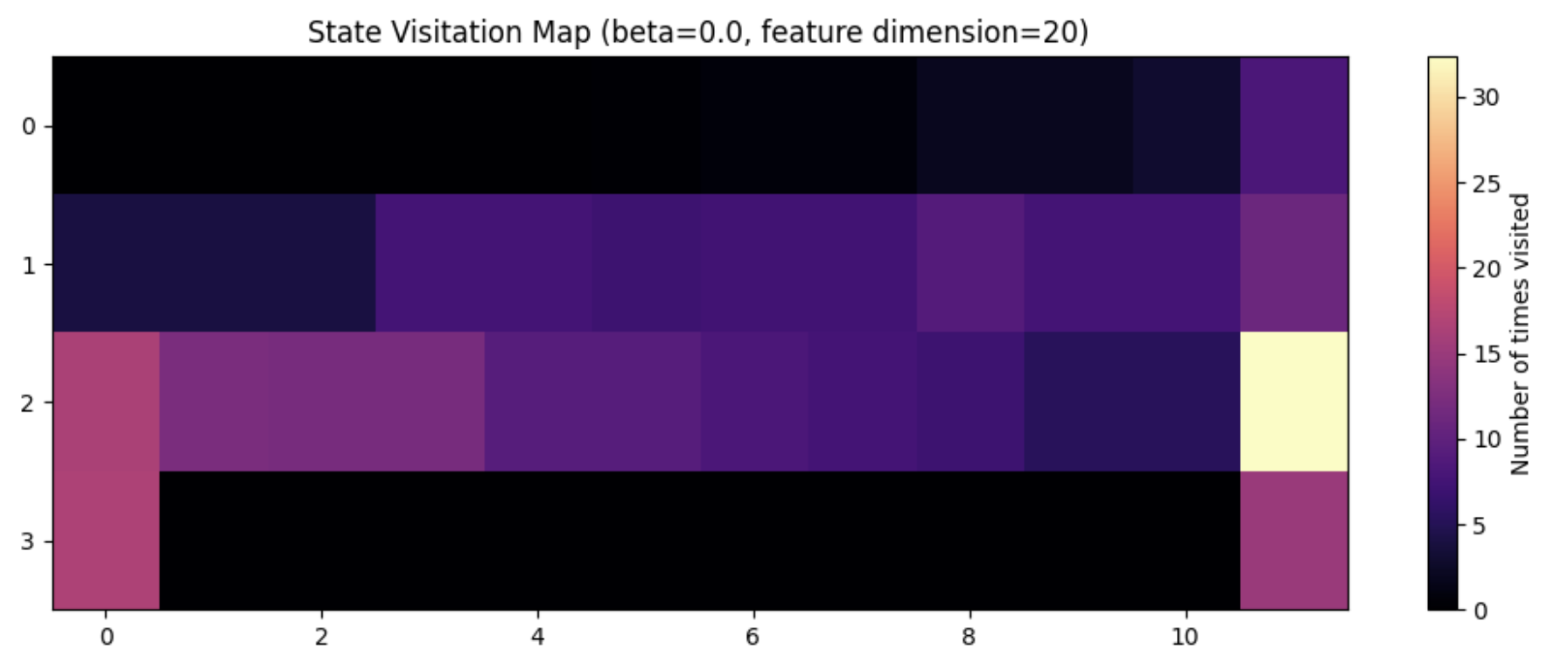}
        \caption{Relaxed-RS-AC $(\beta=0)$}
        \label{fig:cliff_neutral}
    \end{subfigure}
    \hfill
    \begin{subfigure}[b]{0.30\textwidth}
        \centering
        \includegraphics[width=\linewidth]{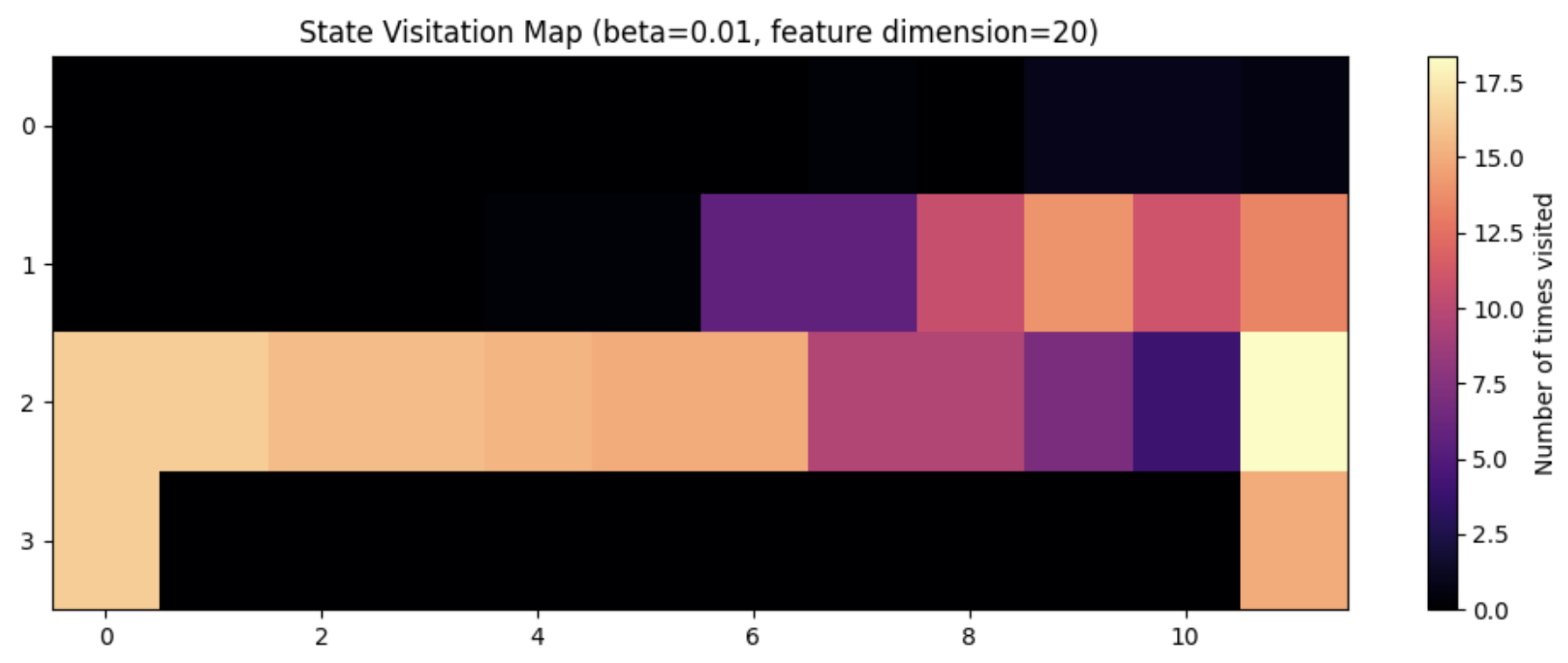}
        \caption{Relaxed-RS-AC $(\beta=0.01)$}
        \label{fig:cliff_seeking}
    \end{subfigure}
    
    
    \begin{subfigure}[b]{0.30\textwidth}
        \centering
        \includegraphics[width=\linewidth]{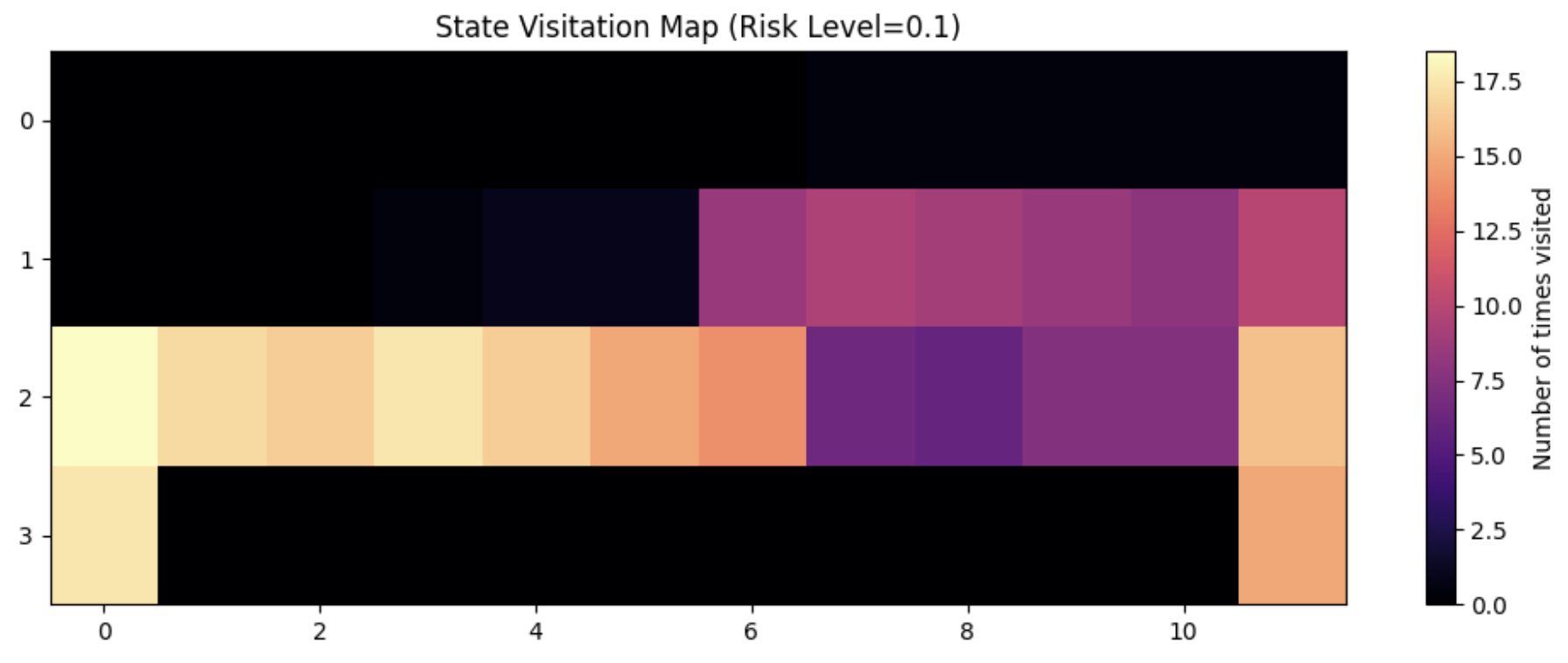}
        \caption{CVaR-PG}
        \label{fig:cliff_cvarpg}
    \end{subfigure}
    \hfill
    \begin{subfigure}[b]{0.30\textwidth}
        \centering
        \includegraphics[width=\linewidth]{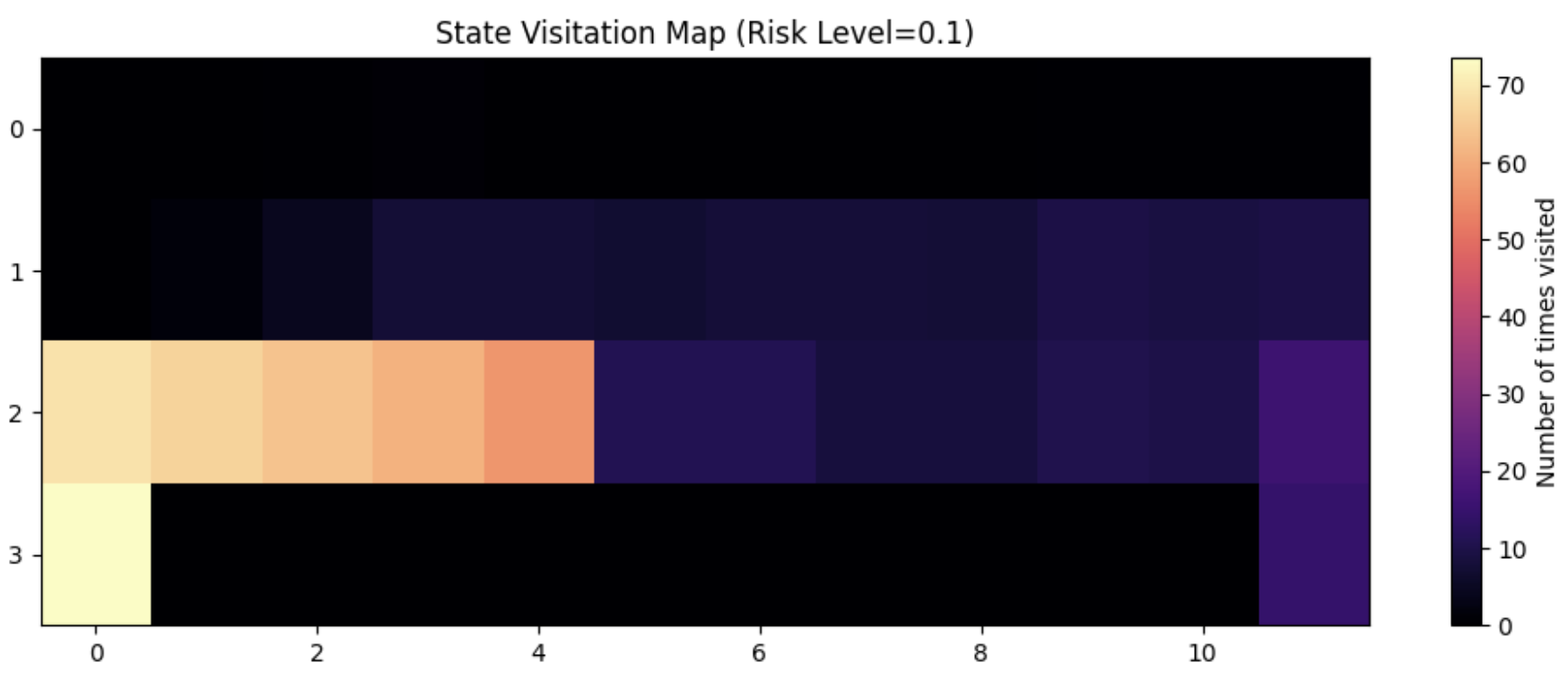}
        \caption{D4PG-CVaR}
        \label{fig:cliff_d4pgcvar}
    \end{subfigure}
    \hfill
    \begin{subfigure}[b]{0.30\textwidth}
        \centering
        \includegraphics[width=\linewidth]{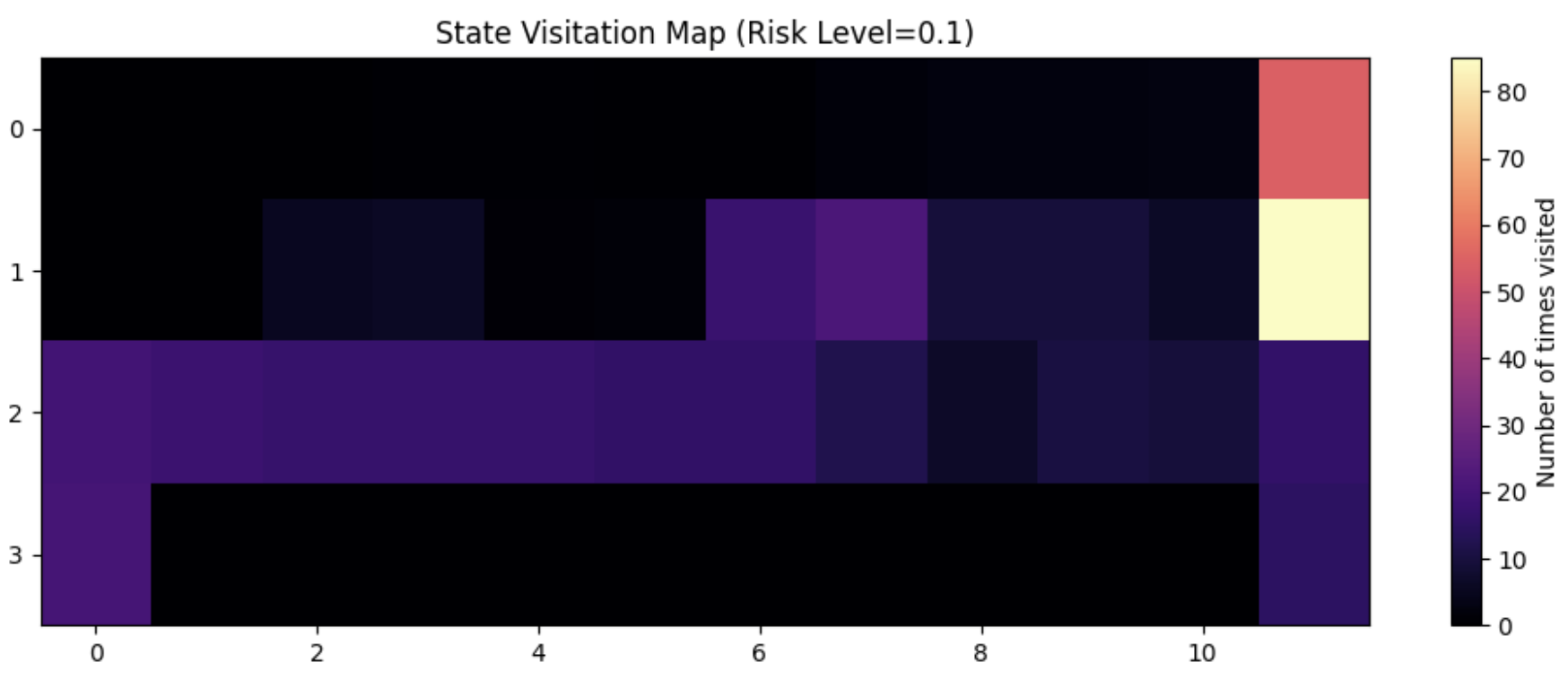}
        \caption{SDPG-CVaR}
        \label{fig:cliff_sdpgcvar}
    \end{subfigure}
    \caption{\textbf{State visitation in CliffWalking} over $15$ evaluation episodes. Lighter areas indicate more frequent visitation. \textbf{(a)--(c)} Relaxed-RS-AC spans a clear behavioral spectrum, from the risk-averse agent hugging the upper boundary to the risk-seeking agent following a cliff-adjacent path. \textbf{(d)--(f)} The CVaR baselines use risk level $0.1$ and keep no consistent safety margin.}
    \label{fig:cliffwalking_heatmaps}
\end{figure*}

In CliffWalking, $|\beta|=0.01$ corresponds to $|\rho|\approx0.6$. Figure~\ref{fig:cliffwalking_heatmaps} shows state visitation across $15$ evaluation episodes. Table~\ref{tab:cliffwalking_final_rewards} reports final trajectory lengths across five seeds.

The risk-averse agent stays farthest from the cliff, the risk-neutral agent takes an intermediate route, and the risk-seeking agent follows a route adjacent to the cliff. CVaR-PG and D4PG-CVaR have shorter mean final trajectories than the risk-seeking Relaxed-RS-AC agent, whose reported length is $15.0\pm0.0$.

Risk seeking does not require a high-variance final policy. CliffWalking has deterministic transitions, so a policy that concentrates on particular actions can repeatedly follow the same route with little variability. Positive $\beta$ favors favorable returns without guaranteeing that training finds the shortest route. Entropic Q-learning shows the same qualitative pattern. Its trajectory-length standard deviation decreases from $3.79$ at negative $\beta$ to $1.07$ at positive $\beta$. These results alone do not establish that risk-seeking policies always have lower final-policy variability.

\begin{table}[tbp]
    \centering
    \footnotesize
    \setlength{\tabcolsep}{4pt}
    \begin{tabular}{lc}
        \toprule
        \textbf{Algorithm} & \textbf{Final Lengths $(\downarrow)$} \\
        \midrule
        Relaxed-RS-AC ($\beta = -0.01$) & $17.0 \pm 1.63$ \\
        Relaxed-RS-AC ($\beta = 0$)     & $15.45 \pm 0.74$ \\
        Relaxed-RS-AC ($\beta = 0.01$)  & $15.0 \pm 0.0$ \\
        Entropic Q-Learning ($\beta = -0.01$) & $20.50 \pm 3.79$ \\
        Entropic Q-Learning ($\beta = 0.01$) & $15.50 \pm 1.07$ \\
        CVaR-PG                          & $14.5 \pm 1.02$ \\
        D4PG-CVaR                        & $14.2 \pm 1.6$ \\
        SDPG-CVaR                        & $16.72 \pm 6.64$ \\
        \bottomrule
    \end{tabular}
    \caption{\textbf{CliffWalking} final trajectory lengths (mean $\pm$ s.d., 5 seeds).}
    \label{tab:cliffwalking_final_rewards}
\end{table}

\subsection{Glycemic Control}
\label{sec:simglucose}
\begin{figure}[tb]
    \centering
    \begin{subfigure}[b]{0.48\linewidth}
        \includegraphics[width=\linewidth]{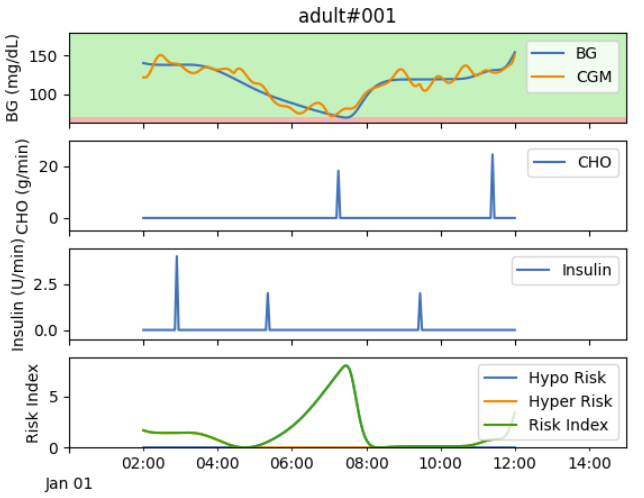}
        \caption{Relaxed-RS-AC}
        \label{fig:rsac-simglucose}
    \end{subfigure}
    \hfill 
    \begin{subfigure}[b]{0.48\linewidth}
        \includegraphics[width=\linewidth]{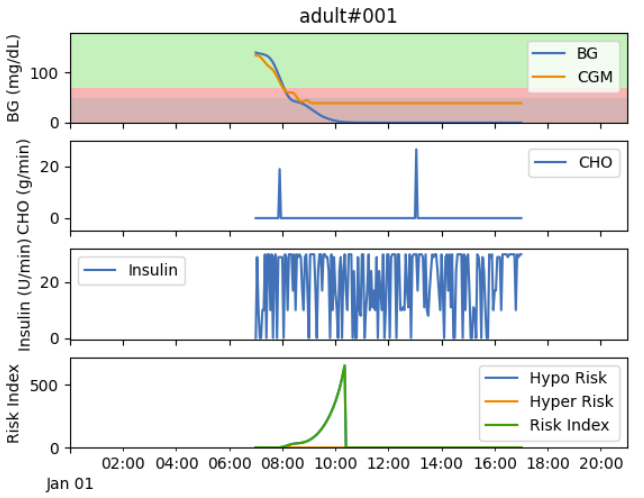}
        \caption{SAC}
        \label{fig:sac-simglucose}
    \end{subfigure}
    \caption{\textbf{Safety validation in the T1DM Simglucose task.} While the risk-neutral SAC (right) induces control instability leading to severe hypoglycemia, the proposed risk-averse Relaxed-RS-AC (left) maintains homeostasis.}
    \label{fig:simglucose}
\end{figure}

To validate our algorithm in a high-stakes medical control setting, we utilize the Simglucose environment \citep{xie2018simglucose}, a high-fidelity proxy for Type-1 Diabetes Mellitus (T1DM) physiology. The control objective is to maintain blood glucose (BG) within a normoglycemic range, a task governed by a heavy-tailed reward structure where deviations into hypoglycemic or hyperglycemic states incur disproportionately severe penalties. A critical challenge in this domain is the limited observation space, which consists solely of the patient's current blood glucose level. The agent must operate without access to exogenous variables such as carbohydrate intake (meals), age, or underlying metabolic rates. Under these conditions of partial observability and aleatoric uncertainty, standard risk-neutral and risk-seeking baselines fail catastrophically. As shown in Figure \ref{fig:simglucose}, for an adult profile, the risk-neutral LFA SAC exhibits high-variance control, oscillating between extremes and ultimately driving the patient into fatal hypoglycemia (BG $\to$ 0). In contrast, the risk-averse Relaxed-RS-AC ($\beta=-0.01$) demonstrates robust stabilization despite the informational bottleneck. By explicitly penalizing the variance of the return distribution, the policy adopts a conservative control strategy that naturally buffers against unobserved disturbances (e.g., meal onsets). Conversely, risk-seeking and risk-neutral variants prioritize aggressive expected-value corrections, systematically over-administering insulin during transient spikes and irreversibly destabilizing the system. This stark divergence underscores the critical necessity of exponential tail-risk penalization in life-critical medical applications.

\subsection{Portfolio Optimization}
\label{sec:portfolio_optimization_main}

We evaluate Relaxed-RS-AC on a real-world equity portfolio optimization task using ten large-cap S\&P~500 equities together with a rest-of-market proxy, spanning January 2013--December 2023.
Models are trained on January 2013--December 2019, validated on the 2020 COVID-19 crash year, and evaluated out-of-sample on the held-out test window January 2021--December 2023, whose dominant
stress event is the 2022 Federal Reserve tightening cycle. The policy outputs a vector $a_t\in\mathbb{R}^{d}$ of portfolio weights with $\sum_i a_{ti}=1$. Negative weights denote short positions, and the policy is not constrained to long-only. All algorithms use a common feature matrix $\Phi$ with feature dimension 20 for linear
function approximation. We compare against two DSAC baselines, DSAC-Wang~\citep{bellemare2023distributional} and DSAC-CVaR, which represent the dominant risk-sensitive distributional RL approaches. We report historical compounded cumulative returns, uncompounded maximum drawdown, and the annualized Sharpe ratio. A complete set of additional risk metrics (Ulcer Index, average drawdown, per-asset risk contribution) is reported in \S\ref{sec:portfolio-experiments}.

We swept $\beta\in\{-0.1,-0.05,-0.01,0,0.01,0.05,0.1\}$ for Relaxed-RS-AC and evaluated DSAC-Wang at $\kappa\in\{-0.75,+0.75\}$ and DSAC-CVaR at $\alpha=0.9$, selecting, for each method, the configuration with the highest Sharpe ratio (zero risk-free rate) on the validation year. Table~\ref{tab:beta-summary-main} reports out-of-sample (2021--2023) performance for every configuration in the sweep, with validation-selected models in bold.

\begin{table}[tb]
\centering
\footnotesize
\setlength{\tabcolsep}{3pt}
\begin{tabular}{lccc}
\toprule
\textbf{Config.} & \textbf{Cum. Ret.} & \textbf{Max DD} & \textbf{SR (V/T)} \\
\midrule
\multicolumn{4}{l}{\textit{Relaxed-RS-AC}} \\
$\beta=-0.10$ & $+46\%$ & $-24.9\%$ & $0.35\,/\,\text{-}0.86$ \\
$\beta=-0.05$ & $+24\%$ & $-13.4\%$ & $0.02\,/\,\text{-}0.40$ \\
$\boldsymbol{\beta=-0.01}$ & $\mathbf{+22\%}$ & $\mathbf{-23.1\%}$ & $\mathbf{0.85\,/\,0.06}$ \\
$\beta=0$     & $-90\%$ & $-129.8\%$ & $0.42\,/\,\text{-}0.68$ \\
$\beta=+0.01$ & $-85\%$ & $-213.1\%$ & $\text{-}3.48\,/\,\text{-}1.48$ \\
$\beta=+0.05$ & $-75\%$ & $-167.4\%$ & $\text{-}3.79\,/\,\text{-}1.20$ \\
$\beta=+0.10$ & $-50\%$ & $-182.2\%$ & $\text{-}1.42\,/\,\text{-}0.64$ \\
\midrule
\multicolumn{4}{l}{\textit{Baselines}} \\
DSAC-Wang ($-0.75$) & $+44\%$ & $-41.8\%$ & $1.77\,/\,0.72$ \\
DSAC-Wang ($+0.75$) & $+52\%$ & $-30.6\%$ & $1.31\,/\,\text{-}0.10$ \\
\textbf{DSAC-CVaR ($0.9$)} & $\mathbf{+47\%}$ & $\mathbf{-19.1\%}$ & $\mathbf{1.83\,/\,0.66}$ \\
\bottomrule
\end{tabular}
\caption{\textbf{Portfolio Optimization.} Out-of-sample (2021--2023) cumulative return, max.\ drawdown, and Sharpe ratio (validation/test) across the full $\beta$ sweep and all baselines. Bold rows are selected by validation-year (2020) Sharpe.}
\label{tab:beta-summary-main}
\end{table}

Performance across the $\beta$ sweep is non-monotonic and separates cleanly by sign: every risk-averse configuration remains profitable out-of-sample with bounded leverage, while every $\beta\ge0$ configuration collapses out-of-sample, as unconstrained exponential amplification of the actor's gradient drives the policy toward highly levered, variance-seeking allocations that do not transfer beyond the training regime. Selecting by validation-year Sharpe ratio identifies $\beta=-0.01$ as the best-performing Relaxed-RS-AC configuration, and this choice remains the best of the sweep on the test window as well. We defer a full analysis of this ablation, including learned allocations and a stress-event breakdown, to \S\ref{sec:portfolio-experiments}.

Under the validation-selected configuration ($\beta=-0.01$), Relaxed-RS-AC constructs a moderately levered, quality-tilted long
book that remains profitable out-of-sample, despite the 2022 tightening cycle falling entirely outside both the training and validation windows. This indicates that the exponential-utility objective, tuned using only data through the 2020 validation year, transfers a stable risk-averse preference into an unseen macroeconomic regime rather than overfitting to the training-period return premia.

The strongest DSAC baseline, DSAC-CVaR ($\alpha=0.9$), attains a higher cumulative return, a shallower maximum
drawdown, and a higher Sharpe ratio over the same test window. We emphasize that the goal of this comparison is not to outperform state-of-the-art distributional RL methods, which model the full return distribution via a quantile-regression critic, but to establish that a substantially simpler exponential-utility objective under LFA achieves broadly comparable, risk-sensitive out-of-sample performance with a single continuously tunable scalar $\beta$. Relaxed-RS-AC's principal advantage, discussed further in \S\ref{sec:portfolio-experiments}, is controllability: unlike the distributional baselines, whose learned risk preference is fixed by a discrete distortion function, $\beta$ smoothly interpolates the trained policy along the risk-return frontier, from defensive, drawdown-minimizing books at large $|\beta|$ to return-maximizing books as $\beta\to0$.

\subsection{Impact of Feature Dimension}
\label{subsec:feature_dim_ablation}

We further investigate the sensitivity of Relaxed-RS-AC to the linear critic's feature dimension ($k \in \{32, 64, 128\}$) across $2$ MuJoCo locomotion environments~\citep{MuJoCoTassa2012} (\texttt{Inverted-Double-Pendulum/v5}, \texttt{Hopper/v5}) and \texttt{Mountain-Car-Continuous/v0} \citep{Moore-1990-13223}. Our findings reveal that representational requirements are strictly coupled to the agent's risk profile ($\beta$). Under risk-averse objectives ($\beta < 0$), higher capacity ($k = 128$) is essential to accurately map complex failure boundaries and isolate safe trajectories, resulting in minimal variance and enhanced stability. Conversely, under risk-seeking objectives ($\beta > 0$), high feature capacity aggressively over-exploits rare, high-variance tail trajectories, inducing policy collapse. In this setting, lower capacity ($k = 32$) acts as a necessary structural regularizer against dangerous exploration. A comprehensive empirical breakdown across distinct environment geometries and detailed theoretical implications are provided in \S\ref{appendix:feature_dim_ablation}.

\section{Conclusion}
We presented a risk-sensitive RL framework via exponential utility whose stationary
Bellman operators are $\gamma$-contractions and exact for the nested entropic
criterion. Its Relaxed-RSTD critic relaxes the log-sum-exp target by a
Donsker--Varadhan bound, giving an $O(1/t)$ finite-sample parameter-convergence
guarantee for on-policy discounted exponential-utility TD under linear function
approximation when the ridge weight exceeds the curvature threshold and the
stated gradient conditions hold. Paired with a simultaneous-perturbation actor,
it yields Relaxed-RS-AC, with almost-sure actor convergence under the stated
assumptions and tail control via a single risk parameter $\beta$, selected
using the practical procedure in Section~\ref{app:choosing-beta}.

Across a broad set of experiments, varying $\beta$ produced interpretable
changes in the trade-off between mean performance and downside exposure.
The feature ablations also showed that critic capacity and risk preference
jointly affect learning stability. Future work includes scaling Relaxed-RS-AC
to larger state and action spaces and extending it to more expressive policy
classes.

\subsubsection*{Broader Impact Statement}
Our work advances methods for Risk-Sensitive Reinforcement Learning with exponential utility. The experiments in this paper were conducted in simulated environments, with no involvement of human subjects or privacy concerns, and we do not foresee any ethical or societal risks from this work. 

\subsubsection*{Acknowledgments}
Part of this research was conducted using computational resources and services at the Department of Computer Science and Engineering, Indian Institute of Technology, Tirupati.

\bibliography{main}
\bibliographystyle{tmlr}

\clearpage

\appendix
\etocdepthtag.toc{appendix}

\begingroup
\renewcommand{\contentsname}{Appendix Contents}
\etocsetnexttocdepth{subsection}
\etocsettagdepth{main}{none}
\etocsettagdepth{appendix}{subsection}
\hypersetup{hidelinks}
\pdfbookmark[1]{Appendix Contents}{appendix-contents}
\tableofcontents
\endgroup
\clearpage

\section{Static Entropic Objective and Risk-Parameter Bounds}

We derive the local expansion of the static objective, then bound the difference from the risk-neutral value and the higher-order remainder.

\subsection{Taylor Series Expansion}
\label{taylorseriesexp}
We derive Equation~\ref{eq:taylorexp} by expanding the nonzero-$\beta$ branch of Equation~\ref{eq:risk-rl-obj}.
\begin{equation}
J_{\beta}(\pi) = \frac{1}{\beta} \log \mathbb{E}_{\tau \sim \pi} \left[ e^{\beta R(\tau)} \right], \nonumber
\end{equation}
where $\tau = (s_0, a_0, r_0, \dots)$ denotes a sample trajectory and $R(\tau)$ is the cumulative return. Expanding the exponential and logarithm gives the approximation in Equation~\ref{eq:taylorexp} after discarding terms of order $\beta^2$ and higher in the objective.

\begin{align}
    \frac{1}{\beta}\log \mathbb{E}\left[e^{\beta R(\tau)}\right] 
    &\approx \frac{1}{\beta}\log \mathbb{E}\left[1 + \beta R(\tau) + \frac{1}{2}\beta^2 R(\tau)^2 + \mathcal{O}(\beta^3)\right] \nonumber \\
    &= \frac{1}{\beta}\log \left(1 + \beta \mathbb{E}[R(\tau)] + \frac{1}{2}\beta^2 \mathbb{E}[R(\tau)^2] + \mathcal{O}(\beta^3)\right) \nonumber \\
    &= \frac{1}{\beta}\left( \beta \mathbb{E}[R(\tau)] + \frac{1}{2}\beta^2 \mathbb{E}[R(\tau)^2] - \frac{1}{2}\big(\beta \mathbb{E}[R(\tau)]\big)^2 + \mathcal{O}(\beta^3) \right) \nonumber \\
    &= \mathbb{E}[R(\tau)] + \frac{\beta}{2}\left(\mathbb{E}[R(\tau)^2] - (\mathbb{E}[R(\tau)])^2\right) + \mathcal{O}(\beta^2) \nonumber \\
    &\approx \mathbb{E}\left[\sum_{t=0}^{\infty} \gamma^t r_t\right] + \frac{\beta}{2}\text{Var}\left[\sum_{t=0}^{\infty} \gamma^t r_t\right]. \nonumber
\end{align}

\subsection{Risk-Parameter Bounds}
\label{app:beta-guardrails}

Proposition~\ref{prop:beta-guardrail} gives the bounds behind the $\beta$-selection procedure in Section~\ref{app:choosing-beta}. It establishes monotonicity of the entropic objective, bounds the deviation from the risk-neutral value, and identifies when the variance term dominates the higher-order cumulants. Corollary~\ref{cor:beta-certified} turns these bounds into certified ranges for $\beta$. All bounds are distribution-level statements about the return
$R$. The only assumption about the MDP is boundedness of rewards.

\paragraph{Setup and notation.}
Let $R$ be an essentially bounded real random variable (the discounted
return under the evaluated policy $\pi$) with law $P_{R}$ on $\mathbb{R}$
(the subscript distinguishes the law of the return from the transition
kernel $P$ of the MDP). Write
$R_{\mathrm{lo}}:=\operatorname{ess\,inf}R$, $R_{\mathrm{hi}}:=\operatorname{ess\,sup}R$,
and $\mathrm{sp}(R):=R_{\mathrm{hi}}-R_{\mathrm{lo}}$ for its span. The
cumulant generating function (CGF) is
\[
    K(u):=\log\mathbb{E}\bigl[e^{uR}\bigr],\qquad u\in\mathbb{R},
\]
which is finite for every $u$ because
$e^{uR}\le e^{|u|\max(|R_{\mathrm{lo}}|,|R_{\mathrm{hi}}|)}$ almost surely. For each
$u\in\mathbb{R}$, the \emph{exponentially tilted} law $Q_u$ is defined by the
density
\[
    \frac{\mathrm{d}Q_u}{\mathrm{d}P_{R}}(x)=e^{u x-K(u)} ,
\]
so that $Q_0=P_{R}$ and $Q_u$ is equivalent to $P_{R}$ (mutually absolutely
continuous) for every $u$. In particular, $Q_u$ is supported in
$[R_{\mathrm{lo}},R_{\mathrm{hi}}]$. The entropic objective is
$J_\beta(\pi)=K(\beta)/\beta$ for $\beta\neq0$ and
$J_0(\pi):=\mathbb{E}[R]$. When rewards are bounded,
$R=\sum_{t\ge0}\gamma^t r_t\in
[\inf r/(1-\gamma),\,\sup r/(1-\gamma)]$ almost surely, whence
\begin{equation}
    \mathrm{sp}(R)\;\le\;\frac{\mathrm{sp}(r)}{1-\gamma},
    \qquad \mathrm{sp}(r):=\sup r-\inf r.
    \label{eq:apriori-span}
\end{equation}
If $\mathrm{sp}(R)=0$ (degenerate return), then $J_\beta\equiv\mathbb{E}[R]$
and every inequality below holds with both sides equal to zero. We therefore
carry the degenerate case along without further comment.

\subsubsection{Cumulants and Exponential Tilting}

The following identities also apply to any bounded random variable. In Appendix~\ref{appendix:rspi}, we use them for conditional distributions of continuation values.

\begin{lemma}[CGF identities]
\label{lem:cgf-tilt}
Let $R$, $P_{R}$, $K$, $Q_u$ be as above. Then:
\begin{enumerate}
    \item[(i)] $K$ is infinitely differentiable on $\mathbb{R}$, $K(0)=0$, and
    for every $u\in\mathbb{R}$,
    \[
        K'(u)=\mathbb{E}_{Q_u}[R]=:m_u,\qquad
        K''(u)=\mathrm{Var}_{Q_u}[R]\;\ge\;0,\qquad
        K'''(u)=\mathbb{E}_{Q_u}\bigl[(R-m_u)^{3}\bigr].
    \]
    \item[(ii)] For every $u\in\mathbb{R}$,
    \[
        u\,K'(u)-K(u)=\mathcal{D}_{\mathrm{KL}}\bigl(Q_u\,\|\,P_{R}\bigr)\;\ge\;0,
    \]
    with equality if and only if $Q_u=P_{R}$. For $u\neq0$, $Q_u=P_{R}$ if and only
    if $R$ is $P_{R}$-almost surely constant.
    \item[(iii)] For every $\beta\in\mathbb{R}$ (including $\beta=0$),
    \[
        J_\beta=\int_0^1 K'(t\beta)\,\mathrm{d}t ,
    \]
    and $J$ is continuously differentiable on $\mathbb{R}$ with
    \[
        J'(\beta)=\int_0^1 t\,K''(t\beta)\,\mathrm{d}t\;\ge\;0,
        \qquad
        J'(\beta)=\frac{\mathcal{D}_{\mathrm{KL}}(Q_\beta\,\|\,P_{R})}{\beta^{2}}
        \ \ (\beta\neq0),
        \qquad
        J'(0)=\frac{\mathrm{Var}[R]}{2}.
    \]
    \item[(iv)] $\lim_{\beta\to+\infty}J_\beta=R_{\mathrm{hi}}$ and
    $\lim_{\beta\to-\infty}J_\beta=R_{\mathrm{lo}}$.
\end{enumerate}
\end{lemma}

\begin{proof}
\emph{(i).} Fix $U>0$ and let $C:=\max(|R_{\mathrm{lo}}|,|R_{\mathrm{hi}}|)$. For
$|u|\le U$ and every $k\in\mathbb{N}$, $|R^{k}e^{uR}|\le C^{k}e^{UC}$ almost
surely, so differentiation under the expectation is justified by dominated
convergence on every compact interval \citep[Lemma~2.2.5]{dembo2010large}.
Hence $u\mapsto\mathbb{E}[e^{uR}]$ is smooth with
$\frac{\mathrm{d}^k}{\mathrm{d}u^k}\mathbb{E}[e^{uR}]=\mathbb{E}[R^{k}e^{uR}]$,
and $K$ is smooth because $\mathbb{E}[e^{uR}]\ge e^{-UC}>0$. Directly,
\[
    K'(u)=\frac{\mathbb{E}[R\,e^{uR}]}{\mathbb{E}[e^{uR}]}
    =\mathbb{E}_{Q_u}[R]=m_u .
\]
More generally, for any bounded measurable $f$, writing
$m_f(u):=\mathbb{E}_{Q_u}[f(R)]=\mathbb{E}[f(R)e^{uR}]/\mathbb{E}[e^{uR}]$,
the quotient rule gives
\begin{equation}
    \frac{\mathrm{d}}{\mathrm{d}u}\,m_f(u)
    =\mathbb{E}_{Q_u}[f(R)\,R]-\mathbb{E}_{Q_u}[f(R)]\,\mathbb{E}_{Q_u}[R]
    =\mathrm{Cov}_{Q_u}\bigl(f(R),R\bigr).
    \label{eq:tilt-derivative}
\end{equation}
Applying \eqref{eq:tilt-derivative} with $f(x)=x$ yields
$K''(u)=m_u'=\mathrm{Var}_{Q_u}[R]\ge0$. Writing
$m_k(u):=\mathbb{E}_{Q_u}[R^{k}]$ and applying \eqref{eq:tilt-derivative}
with $f(x)=x^{2}$ and $f(x)=x$,
\[
    K'''(u)=\frac{\mathrm{d}}{\mathrm{d}u}\bigl(m_2(u)-m_1(u)^{2}\bigr)
    =\bigl(m_3-m_1m_2\bigr)-2m_1\bigl(m_2-m_1^{2}\bigr)
    =m_3-3m_1m_2+2m_1^{3},
\]
and expanding $\mathbb{E}_{Q_u}[(R-m_1)^{3}]
=m_3-3m_1m_2+3m_1^{2}m_1-m_1^{3}=m_3-3m_1m_2+2m_1^{3}$ shows the two agree.

\emph{(ii).} By definition of $Q_u$,
\[
    \mathcal{D}_{\mathrm{KL}}(Q_u\|P_{R})
    =\mathbb{E}_{Q_u}\Bigl[\log\tfrac{\mathrm{d}Q_u}{\mathrm{d}P_{R}}\Bigr]
    =\mathbb{E}_{Q_u}\bigl[uR-K(u)\bigr]
    =u\,K'(u)-K(u),
\]
using (i). Nonnegativity and the equality case are the information
inequality: $\mathcal{D}_{\mathrm{KL}}(Q_u\|P_{R})\ge0$ with equality iff
$Q_u=P_{R}$ \citep[Theorem~2.6.3]{cover2006elements}. Since
$\mathrm{d}Q_u/\mathrm{d}P_{R}=e^{uR-K(u)}$, the condition $Q_u=P_{R}$ means
$e^{uR-K(u)}=1$ $P_{R}$-a.s., i.e., $uR=K(u)$ $P_{R}$-a.s. For $u\neq0$, this holds
iff $R$ is a.s.\ constant.

\emph{(iii).} $K$ is $C^{1}$ with $K(0)=0$, so by the fundamental theorem of
calculus and the substitution $u=t\beta$,
\[
    K(\beta)=\int_0^{\beta}K'(u)\,\mathrm{d}u
    =\beta\int_0^{1}K'(t\beta)\,\mathrm{d}t .
\]
Dividing by $\beta\neq0$ gives the representation. At $\beta=0$, the integral
equals $K'(0)=\mathbb{E}[R]=J_0$, so the representation extends continuously
through $0$ and shows that the convention $J_0=\mathbb{E}[R]$ is the
continuous extension. The integrand $(t,\beta)\mapsto K'(t\beta)$ is $C^{1}$
with $\partial_\beta K'(t\beta)=t\,K''(t\beta)$, jointly continuous and
bounded on $[0,1]\times[-U,U]$ for every $U$ (by (i) and boundedness of $R$),
so Leibniz's rule gives
$J'(\beta)=\int_0^1 t\,K''(t\beta)\,\mathrm{d}t\ge0$, continuous in $\beta$.
For $\beta\neq0$ the quotient rule and (ii) give
\[
    J'(\beta)=\frac{\beta K'(\beta)-K(\beta)}{\beta^{2}}
    =\frac{\mathcal{D}_{\mathrm{KL}}(Q_\beta\|P_{R})}{\beta^{2}} ,
\]
and at $\beta=0$, $J'(0)=K''(0)\int_0^1 t\,\mathrm{d}t=\mathrm{Var}[R]/2$.

\emph{(iv).} For $\beta>0$, $e^{\beta R}\le e^{\beta R_{\mathrm{hi}}}$ a.s.\ gives
$J_\beta\le R_{\mathrm{hi}}$. Conversely, fix $\varepsilon>0$ and let
$p_\varepsilon:=P_{R}(R>R_{\mathrm{hi}}-\varepsilon)>0$ (positive by definition of the
essential supremum). Then
$\mathbb{E}[e^{\beta R}]\ge p_\varepsilon e^{\beta(R_{\mathrm{hi}}-\varepsilon)}$,
so $J_\beta\ge R_{\mathrm{hi}}-\varepsilon+\beta^{-1}\log p_\varepsilon
\to R_{\mathrm{hi}}-\varepsilon$ as $\beta\to\infty$. Since $\varepsilon>0$ was
arbitrary, $J_\beta\to R_{\mathrm{hi}}$. The limit $\beta\to-\infty$ follows by
applying the same argument to $-R$, using
$J_\beta(R)=-J_{-\beta}(-R)$ and
$\operatorname{ess\,sup}(-R)=-R_{\mathrm{lo}}$.
\end{proof}

\begin{lemma}[The certainty equivalent is $\mathrm{sp}^{2}/8$-Lipschitz in
its parameter]
\label{lemma:ce-lipschitz}
Let $X$ be an essentially bounded random variable on $(\Omega,\mathcal{F},P)$
and write $\mathrm{sp}(X):=\operatorname*{ess\,sup}X-\operatorname*{ess\,inf}X$,
let $K(u):=\log\mathbb{E}[e^{uX}]$ be its cumulant
generating function, and let $C(u):=K(u)/u$ for $u\neq0$, $C(0):=\mathbb{E}[X]$,
so that $C(u)=\rho^{\mathrm{ent}}_{u}[X]$. Then $C$ is continuously
differentiable and nondecreasing on $\mathbb{R}$, and for every $u\neq0$
\begin{equation}
  0 \;\le\; C'(u)
  \;=\; \frac{uK'(u)-K(u)}{u^{2}}
  \;=\; \frac{\mathcal{D}_{\mathrm{KL}}\!\left(Q_{u}\,\|\,P\right)}{u^{2}}
  \;=\; \frac{1}{u^{2}}\int_{0}^{u}v\,K''(v)\,dv
  \;\le\; \frac{\mathrm{sp}(X)^{2}}{8},
  \label{eq:ce-lipschitz}
\end{equation}
where $dQ_{u}\propto e^{uX}dP$. Also,
$C'(0)=\tfrac12\mathrm{Var}[X]\le\mathrm{sp}(X)^2/8$. Consequently
\[
  \big|\rho^{\mathrm{ent}}_{u_{1}}[X]-\rho^{\mathrm{ent}}_{u_{2}}[X]\big|
  \;\le\;\frac{\mathrm{sp}(X)^{2}}{8}\,|u_{1}-u_{2}|
  \qquad\text{for all } u_{1},u_{2}\in\mathbb{R}.
\]
\end{lemma}

\begin{proof}
Apply Lemma~\ref{lem:cgf-tilt} with $X$ in place of $R$. Part (ii) gives the KL identity and part (iii) gives continuous differentiability, the derivative at zero, and the representation $C'(u)=\int_0^1 tK''(tu)\,dt$. For $u\neq0$, the change of variable $v=tu$ gives the integral in Equation~\ref{eq:ce-lipschitz}. Part (i) identifies $K''(v)$ with the variance under $Q_v$. The tilted law stays within the essential range of $X$, so Popoviciu's inequality gives $0\le K''(v)\le\mathrm{sp}(X)^2/4$. Integrating $t$ over $[0,1]$ gives $0\le C'(u)\le\mathrm{sp}(X)^2/8$. The mean value theorem gives the Lipschitz bound.
\end{proof}

\subsubsection{A Sharp Span Bound on the Third Tilted Moment}

\begin{lemma}[Third-moment span bound]
\label{lem:third-moment-span}
Let $Q$ be any probability law supported in an interval of length
$L\ge0$, let $X\sim Q$ and $m:=\mathbb{E}_Q[X]$. Then
\[
    \bigl|\mathbb{E}_Q\bigl[(X-m)^{3}\bigr]\bigr|
    \;\le\;\mathbb{E}_Q\bigl[|X-m|^{3}\bigr]
    \;\le\;\frac{L^{3}}{8},
\]
and the constant $1/8$ is attained (for the absolute moment) by the
symmetric two-point law on the interval endpoints. Consequently, for the
tilted laws of Lemma~\ref{lem:cgf-tilt},
\[
    \sup_{u\in\mathbb{R}}\bigl|K'''(u)\bigr|
    \;\le\;\frac{\mathrm{sp}(R)^{3}}{8},
\]
since every $Q_u$ is supported in $[R_{\mathrm{lo}},R_{\mathrm{hi}}]$.
\end{lemma}

\begin{proof}
If $L=0$, both sides vanish. Assume $L>0$. The first inequality is the
triangle inequality for expectations. For the second, central moments are
translation invariant, so we may assume the supporting interval is $[0,L]$.
then $m\in[0,L]$. The map $h(x):=|x-m|^{3}$ is convex on $\mathbb{R}$: it is
the composition of the convex map $x\mapsto|x-m|$ with the nondecreasing
convex map $t\mapsto t^{3}$ on $t\ge0$. Every $x\in[0,L]$ is the convex
combination $x=(1-\tfrac{x}{L})\cdot0+\tfrac{x}{L}\cdot L$, so convexity
gives the pointwise bound
\[
    h(x)\;\le\;\Bigl(1-\frac{x}{L}\Bigr)h(0)+\frac{x}{L}\,h(L),
    \qquad x\in[0,L].
\]
Taking $\mathbb{E}_Q$ and using $\mathbb{E}_Q[X]=m$,
\[
    \mathbb{E}_Q\bigl[|X-m|^{3}\bigr]
    \;\le\;\Bigl(1-\frac{m}{L}\Bigr)m^{3}+\frac{m}{L}\,(L-m)^{3}
    \;=\;L^{3}\,t(1-t)\bigl(t^{2}+(1-t)^{2}\bigr),
    \qquad t:=\frac{m}{L}\in[0,1].
\]
With $q:=t(1-t)\in[0,\tfrac14]$ we have
$t^{2}+(1-t)^{2}=1-2q$, and the map $q\mapsto q(1-2q)$ has derivative
$1-4q\ge0$ on $[0,\tfrac14]$, hence is maximized at $q=\tfrac14$ (i.e.\
$t=\tfrac12$) with value $\tfrac18$. This proves
$\mathbb{E}_Q[|X-m|^{3}]\le L^{3}/8$. Sharpness: for $Q$ uniform on the two
endpoints $\{0,L\}$, $m=L/2$ and
$\mathbb{E}_Q[|X-m|^{3}]=(L/2)^{3}=L^{3}/8$. The final claim follows because
$Q_u\ll P_{R}$ implies that $Q_u$ is supported in $[R_{\mathrm{lo}},R_{\mathrm{hi}}]$, an
interval of length $\mathrm{sp}(R)$, and
$K'''(u)=\mathbb{E}_{Q_u}[(R-m_u)^{3}]$ by Lemma~\ref{lem:cgf-tilt}(i).
\end{proof}

\begin{remark}[Sharp constant for the signed cumulant]
\label{rem:third-moment-sharp}
For the \emph{signed} third central moment the sharp constant is smaller:
$\sup\bigl|\mathbb{E}_Q[(X-m)^{3}]\bigr|=\tfrac{\sqrt{3}}{18}L^{3}
\approx0.0962\,L^{3}$ over laws $Q$ supported in an interval of length $L$.
Indeed, with $m$ fixed, $Q\mapsto\mathbb{E}_Q[(X-m)^{3}]$ is linear, and the
extreme points of the set of laws on $[0,L]$ with mean $m$ are supported on
at most two atoms \citep{winkler1988extreme}. For a two-point law with atoms
$x_1<m<x_2$ and $a:=m-x_1$, $b:=x_2-m$, a direct computation gives
$\mathbb{E}_Q[(X-m)^{3}]=ab(b-a)$, which is homogeneous of degree $3$ in
$(a,b)$ and hence maximized on $a+b=L$, where
$\max_{a+b=L}ab(b-a)=\tfrac{\sqrt{3}}{18}L^{3}$, attained at
$a=\tfrac{3-\sqrt3}{6}L$. We only use the elementary constant $1/8$ of
Lemma~\ref{lem:third-moment-span}, which requires no extreme-point
machinery. The gap ($0.125$ vs.\ $0.0962$) is a factor $\approx1.3$.
\end{remark}

\subsubsection{Bounds and Certified Ranges}

\begin{proposition}[Guardrails for the risk parameter $\beta$]
\label{prop:beta-guardrail}
Let the per-step reward be bounded, so that $R=\sum_{t\ge0}\gamma^t r_t$
satisfies $\mathrm{sp}(R)\le\mathrm{sp}(r)/(1-\gamma)<\infty$. Then:
\begin{enumerate}
    \item[(a)] \emph{(Monotonicity.)} The map $\beta\mapsto J_\beta(\pi)$ is
    continuously differentiable and nondecreasing on all of $\mathbb{R}$, with
    \[
        \frac{\mathrm{d}J_\beta}{\mathrm{d}\beta}
        =\frac{\mathcal{D}_{\mathrm{KL}}\!\left(Q_\beta\,\|\,P_{R}\right)}{\beta^{2}}\ \ (\beta\neq0),
        \qquad
        \frac{\mathrm{d}J_\beta}{\mathrm{d}\beta}\Big|_{\beta=0}
        =\frac{\mathrm{Var}[R]}{2}.
    \]
    It is strictly increasing unless $R$ is almost surely constant (in which
    case $J_\beta\equiv\mathbb{E}[R]$), and
    $\lim_{\beta\to+\infty}J_\beta=R_{\mathrm{hi}}$,
    $\lim_{\beta\to-\infty}J_\beta=R_{\mathrm{lo}}$.
    \item[(b)] \emph{(Hoeffding certificate.)} For every $\beta\in\mathbb{R}$,
    \[
        0\;\le\;\operatorname{sgn}(\beta)\bigl(J_\beta(\pi)-\mathbb{E}[R]\bigr)
        \;\le\;\frac{|\beta|\,\mathrm{sp}(R)^{2}}{8},
    \]
    and the constant $1/8$ cannot be improved: for the symmetric two-point
    return the ratio of the two sides tends to $1$ as $\beta\to0$.
    \item[(c)] \emph{(Cumulant-remainder control.)} With
    $M_3(\beta):=\sup_{|u|\le|\beta|}|K'''(u)|$,
    \[
        \Bigl|J_\beta(\pi)-\mathbb{E}[R]-\tfrac{\beta}{2}\mathrm{Var}[R]\Bigr|
        \;\le\;\frac{\beta^{2}}{6}\,M_3(\beta)
        \;\le\;\frac{\beta^{2}\,\mathrm{sp}(R)^{3}}{48},
    \]
    and the higher-order remainder is at most
    $\tfrac12\cdot\tfrac{|\beta|}{2}\mathrm{Var}[R]$ whenever
    $|\beta|\le\tfrac{3}{2}\mathrm{Var}[R]/\overline{M}_3$ with
    $\overline{M}_3:=\sup_{u\in\mathbb{R}}|K'''(u)|\le\mathrm{sp}(R)^{3}/8$,
    and a fortiori whenever
    $|\beta|\le12\,\mathrm{Var}[R]/\mathrm{sp}(R)^{3}$.
\end{enumerate}
\end{proposition}

\begin{proof}
\emph{(a) Monotonicity.} By Lemma~\ref{lem:cgf-tilt}(iii), $J$ is
continuously differentiable with
$J'(\beta)=\int_0^1 t\,K''(t\beta)\,\mathrm{d}t\ge0$, since
$K''=\mathrm{Var}_{Q_u}[R]\ge0$ pointwise. Hence $J$ is nondecreasing on all
of $\mathbb{R}$, and the two derivative identities are
Lemma~\ref{lem:cgf-tilt}(iii). For strictness, suppose $R$ is not almost
surely constant. Then $\mathrm{Var}_{Q_u}[R]>0$ for every $u$: $Q_u$ is
equivalent to $P_{R}$, so $R$ is $Q_u$-a.s.\ constant iff it is $P_{R}$-a.s.\
constant. Hence $J'(\beta)=\int_0^1 t\,K''(t\beta)\,\mathrm{d}t>0$ for every
$\beta$ (the integrand is continuous and strictly positive for $t\in(0,1]$),
so $J$ is strictly increasing. Equivalently, for $\beta\neq0$,
$\mathcal{D}_{\mathrm{KL}}(Q_\beta\|P_{R})>0$ by Lemma~\ref{lem:cgf-tilt}(ii).
If $R$ is a.s.\ constant, $K(u)=u\,\mathbb{E}[R]$ and
$J_\beta\equiv\mathbb{E}[R]$. The limits are Lemma~\ref{lem:cgf-tilt}(iv).
The monotone interpretation of $\beta$ as a risk knob is classical for
entropic utilities \citep{follmer2016stochastic}. The proof above keeps the
statement self-contained and supplies the derivative formula used in (b) and
in the certificates.

\emph{(b) Hoeffding certificate.} Hoeffding's lemma
\citep{hoeffding1963probability}, in the form of
\citet[\S 2.6]{boucheron2013concentration}, states that a random variable
$X$ with $a\le X\le b$ almost surely satisfies
$\log\mathbb{E}\bigl[e^{u(X-\mathbb{E}X)}\bigr]\le u^{2}(b-a)^{2}/8$ for all
$u\in\mathbb{R}$. Applying this to $X=R$ with $a=R_{\mathrm{lo}}$, $b=R_{\mathrm{hi}}$, and
combining with Jensen's inequality
$\log\mathbb{E}[e^{\beta R}]\ge\beta\,\mathbb{E}[R]$, we obtain, for every
$\beta\in\mathbb{R}$,
\begin{equation}
    0\;\le\;K(\beta)-\beta\,\mathbb{E}[R]\;\le\;\frac{\beta^{2}\,\mathrm{sp}(R)^{2}}{8}.
    \label{eq:hoeffding-K}
\end{equation}
For $\beta>0$, dividing \eqref{eq:hoeffding-K} by $\beta$ gives
$0\le J_\beta-\mathbb{E}[R]\le\beta\,\mathrm{sp}(R)^{2}/8$. For $\beta<0$,
dividing by $\beta$ reverses both inequalities:
$0\ge J_\beta-\mathbb{E}[R]\ge\beta\,\mathrm{sp}(R)^{2}/8
=-|\beta|\,\mathrm{sp}(R)^{2}/8$. Both cases are summarized by
$0\le\operatorname{sgn}(\beta)(J_\beta-\mathbb{E}[R])
\le|\beta|\,\mathrm{sp}(R)^{2}/8$ (the case $\beta=0$ is trivial). The
one-sidedness is also a direct consequence of (a) via
$J_0=\mathbb{E}[R]$. For $\beta\neq0$, equality holds throughout iff $R$ is
degenerate (at $\beta=0$ all three members vanish for every $R$). For
sharpness of the constant, take $R$ uniform on $\{\pm L/2\}$ with
$L=\mathrm{sp}(R)$: then $\mathbb{E}[R]=0$,
$J_\beta=\beta^{-1}\log\cosh(\beta L/2)$, and since
$\log\cosh(x)=x^{2}/2+O(x^{4})$,
\[
    \frac{J_\beta-\mathbb{E}[R]}{|\beta|\,L^{2}/8}
    =\frac{\log\cosh(\beta L/2)}{\beta^{2}L^{2}/8}
    \;\longrightarrow\;1
    \qquad(\beta\to0),
\]
so no constant smaller than $1/8$ can hold uniformly over bounded returns.

\emph{(c) Cumulant-remainder control.} The case $\beta=0$ is trivial, so let
$\beta\neq0$. $K$ is $C^{3}$ by Lemma~\ref{lem:cgf-tilt}(i), so Taylor's
theorem with integral remainder gives
\[
    K(\beta)=\underbrace{K(0)}_{=0}
    +K'(0)\,\beta+\frac{K''(0)}{2}\,\beta^{2}
    +\int_0^{\beta}\frac{(\beta-u)^{2}}{2}\,K'''(u)\,\mathrm{d}u ,
\]
with $K'(0)=\mathbb{E}[R]$ and $K''(0)=\mathrm{Var}[R]$
(Lemma~\ref{lem:cgf-tilt}(i) at $u=0$). Dividing by $\beta$ and substituting
$u=s\beta$, $s\in[0,1]$ (valid for either sign of $\beta$, with
$\mathrm{d}u=\beta\,\mathrm{d}s$),
\[
    J_\beta-\mathbb{E}[R]-\frac{\beta}{2}\mathrm{Var}[R]
    =\frac{1}{\beta}\int_0^{\beta}\frac{(\beta-u)^{2}}{2}\,K'''(u)\,\mathrm{d}u
    =\beta^{2}\int_0^{1}\frac{(1-s)^{2}}{2}\,K'''(s\beta)\,\mathrm{d}s .
\]
Since $|s\beta|\le|\beta|$ for $s\in[0,1]$, taking absolute values yields
\[
    \Bigl|J_\beta-\mathbb{E}[R]-\tfrac{\beta}{2}\mathrm{Var}[R]\Bigr|
    \;\le\;\beta^{2}\,M_3(\beta)\int_0^{1}\frac{(1-s)^{2}}{2}\,\mathrm{d}s
    \;=\;\frac{\beta^{2}}{6}\,M_3(\beta),
\]
with $M_3(\beta)=\sup_{|u|\le|\beta|}|K'''(u)|$. For a return law with $K'''(0)\ne0$, the ratio of the absolute remainder to $\beta^2 M_3(\beta)/6$ tends to $1$ as $\beta\to0$ by continuity of $K'''$. This proves sharpness of the factor $1/6$. Lemma~\ref{lem:third-moment-span}
gives $M_3(\beta)\le\overline{M}_3\le\mathrm{sp}(R)^{3}/8$, whence the
uniform bound $\beta^{2}\mathrm{sp}(R)^{3}/48$. Finally, if
$|\beta|\le\tfrac{3}{2}\mathrm{Var}[R]/\overline{M}_3$, then
\[
    \frac{\beta^{2}}{6}\,M_3(\beta)
    \;\le\;\frac{|\beta|}{6}\cdot|\beta|\,\overline{M}_3
    \;\le\;\frac{|\beta|}{6}\cdot\frac{3}{2}\mathrm{Var}[R]
    \;=\;\frac{1}{2}\cdot\frac{|\beta|}{2}\,\mathrm{Var}[R],
\]
i.e., the entire cubic-and-higher remainder is at most half the magnitude of
the variance term $\tfrac{\beta}{2}\mathrm{Var}[R]$, which is the announced
dominance. Substituting $\overline{M}_3\le\mathrm{sp}(R)^{3}/8$ gives the
a-fortiori threshold $|\beta|\le12\,\mathrm{Var}[R]/\mathrm{sp}(R)^{3}$.
\end{proof}

\begin{corollary}[Certified operating range for $\beta$]
\label{cor:beta-certified}
Let $\mathrm{sp}(R)>0$ and $\mathrm{Var}[R]>0$, and fix a tolerance
$\varepsilon>0$.
\begin{enumerate}
    \item[(i)] Every $\beta$ with $|\beta|\le 8\varepsilon/\mathrm{sp}(R)^{2}$
    satisfies $|J_\beta-\mathbb{E}[R]|\le\varepsilon$. Conversely, any chosen
    $\beta$ certifies
    $|J_\beta-\mathbb{E}[R]|\le|\beta|\,\mathrm{sp}(R)^{2}/8$.
    \item[(ii)] Under $\beta=\rho\,|\mathbb{E}[R]|/\mathrm{Var}[R]$,
    $\bigl|J_\beta-\mathbb{E}[R]\bigr|
    \le|\rho|\,|\mathbb{E}[R]|\,\mathrm{sp}(R)^{2}/(8\,\mathrm{Var}[R])$.
    \item[(iii)] Under the same rule, the variance term dominates the
    higher-order remainder (in the sense of
    Proposition~\ref{prop:beta-guardrail}(c)) whenever
    $|\rho|\le\tfrac{3}{2}\mathrm{Var}[R]^{2}/(|\mathbb{E}[R]|\,\overline{M}_3)$,
    and a fortiori whenever
    $|\rho|\le12\,\mathrm{Var}[R]^{2}/(|\mathbb{E}[R]|\,\mathrm{sp}(R)^{3})$.
    The bounds apply for $\mathbb{E}[R]\neq0$. If $\mathbb{E}[R]=0$, the rule gives $\beta=0$
    and dominance holds trivially.
    \item[(iv)] Since $\mathrm{sp}(R)\le\mathrm{sp}(r)/(1-\gamma)$, the
    a-priori range $|\beta|\le8\varepsilon(1-\gamma)^{2}/\mathrm{sp}(r)^{2}$
    certifies the tolerance $\varepsilon$ before any rollout data is
    collected.
\end{enumerate}
\end{corollary}

\begin{proof}
(i) is Proposition~\ref{prop:beta-guardrail}(b) read in the two directions:
if $|\beta|\le8\varepsilon/\mathrm{sp}(R)^{2}$ then
$|J_\beta-\mathbb{E}[R]|\le|\beta|\mathrm{sp}(R)^{2}/8\le\varepsilon$, and
any fixed $\beta$ certifies the stated worst case. (ii) substitute
$|\beta|=|\rho|\,|\mathbb{E}[R]|/\mathrm{Var}[R]$ into (b). (iii) the rule
gives $|\beta|\le\tfrac{3}{2}\mathrm{Var}[R]/\overline{M}_3$ iff
$|\rho|\le\tfrac{3}{2}\mathrm{Var}[R]^{2}/(|\mathbb{E}[R]|\,\overline{M}_3)$. Apply Proposition~\ref{prop:beta-guardrail}(c), and use
$\overline{M}_3\le\mathrm{sp}(R)^{3}/8$ for the a-fortiori form. (iv) combine
(i) with the a-priori span bound \eqref{eq:apriori-span}: if
$|\beta|\le8\varepsilon(1-\gamma)^{2}/\mathrm{sp}(r)^{2}$ then
$|\beta|\,\mathrm{sp}(R)^{2}/8\le
|\beta|\,\mathrm{sp}(r)^{2}/(8(1-\gamma)^{2})\le\varepsilon$.
\end{proof}

\begin{remark}[Deployment guardrails]
\label{rem:beta-deployment}
For risk-averse deployment ($\beta<0$),
Proposition~\ref{prop:beta-guardrail}(b) places the reported objective in the
one-sided band
$\bigl[\mathbb{E}[R]-|\beta|\mathrm{sp}(R)^{2}/8,\;\mathbb{E}[R]\bigr]$. This certificate is for the static entropic objective of the return distribution. Sections~\ref{app:discount-placement} and~\ref{app:static-nested} give the one-sided fixed-policy comparisons for discount placement and the nested criterion, respectively. The learned critic also has the optimization, approximation, and model errors described in Corollary~\ref{cor:error-decomposition}.
For
risk-seeking deployment ($\beta>0$), the same certificate caps overconfidence:
no admissible $\beta$ in the certified range of
Corollary~\ref{cor:beta-certified}(i) can inflate the objective by more than
$\varepsilon$ above the risk-neutral value. These are concrete guardrails for
the broader-impact concern that entropic risk-seeking might silently chase
heavy right tails.
\end{remark}

\begin{remark}[Interpreting the bounds on $\beta$]
\label{rem:beta-guardrail-practice}
Three points deserve emphasis. First, the certificate in
Corollary~\ref{cor:beta-certified} is one-sided in exactly the direction that
matters for deployment: for $\beta<0$ the entropic value can only sit
\emph{below} the risk-neutral value, by at most the certified amount, so the
certified band lies on the conservative side of the risk-neutral value, as described in Remark~\ref{rem:beta-deployment}. Second, the a-priori range in (iv) uses the worst-case span
$\mathrm{sp}(r)/(1-\gamma)$ and is therefore conservative for $\gamma$ close
to $1$. Whenever rollout data is available, a tighter valid upper bound on $\mathrm{sp}(R)$ can improve the certificate. An observed rollout span is an empirical diagnostic unless it is independently justified as an upper bound. Third, the
dominance thresholds in Proposition~\ref{prop:beta-guardrail}(c) and
Corollary~\ref{cor:beta-certified}(iii) are \emph{sufficient} conditions. The checks in Section~\ref{app:choosing-beta} assess estimated low-order contributions and do not certify the full remainder.
\end{remark}

\section{Bellman Operators and Risk-Sensitive Policy Iteration}
\label{appendix:rspi}
We prove the contraction and policy iteration results for Algorithm~\ref{alg:rspi}, then identify the objective evaluated by the stationary Bellman recursion. Sections~\ref{app:discount-placement} and~\ref{app:static-nested} quantify the effect of discount placement and compare the nested criterion with the static objective, respectively.

\subsection{Bellman Contraction and Policy Improvement}
\label{app:policy-iteration-proofs}

The policy iteration proofs use certainty-equivalent values and action scores.

\begin{proof}[Proof of Lemma~\ref{lemma:bellman-operator-contraction-mapping}]
For $\beta>0$, the map $\rho^{\mathrm{ent}}_\beta$ is monotone increasing because exponentiation, expectation, and the scaled logarithm preserve order. Monotonicity also holds for $\beta<0$ because multiplication by $\beta$ and division by $\beta$ each reverse the inequality. At $\beta=0$ it follows from monotonicity of expectation. For all $\beta$, adding a constant $d$ to the input adds $d$ to the certainty equivalent. With $d=\|U_1-U_2\|_\infty$, the bounds $U_2-d\le U_1\le U_2+d$ therefore give
\[
|\rho^{\mathrm{ent}}_\beta[U_1(s')\mid s,a]-\rho^{\mathrm{ent}}_\beta[U_2(s')\mid s,a]|\le d.
\]
The reward cancels in the difference between the fixed-policy updates. Multiplying by $\gamma$ and taking the supremum over states gives the contraction. Banach's fixed-point theorem gives the unique bounded fixed point $U^\pi_\beta$.
\end{proof}

\begin{proof}[Proof of Lemma~\ref{lemma:fixed-point-optimal-bellman-operator}]
The preceding monotonicity property applies to $\mathcal{T}^\pi_\beta$ and $\mathcal{T}^*_\beta$ for every $\beta$. For any $\pi$, we have
\[
\mathcal{T}^*_\beta U^*_\beta\ge \mathcal{T}^\pi_\beta U^*_\beta
\ge \mathcal{T}^\pi_\beta U^\pi_\beta=U^\pi_\beta.
\]
Taking the supremum over policies shows $\mathcal{T}^*_\beta U^*_\beta\ge U^*_\beta$. Choose a policy $\check\pi$ greedy for $U^*_\beta$. By construction,
\begin{equation}
\label{eq:greedy_equiv}
\mathcal{T}^{\check\pi}_\beta U^*_\beta=\mathcal{T}^*_\beta U^*_\beta.
\end{equation}
Iteration of $\mathcal{T}^{\check\pi}_\beta$ from $U^*_\beta$ is nondecreasing and converges to $U^{\check\pi}_\beta$ by Lemma~\ref{lemma:bellman-operator-contraction-mapping}. Thus $U^{\check\pi}_\beta\ge U^*_\beta$. The definition of $U^*_\beta$ gives the reverse inequality. Substituting their equality into Equation~\ref{eq:greedy_equiv} proves the fixed-point identity.
\end{proof}

\begin{proof}[Proof of Lemma~\ref{lemma:optimal-bellman-contraction-mapping}]
The same one-step bound used in Lemma~\ref{lemma:bellman-operator-contraction-mapping} bounds the change in each action score by $\gamma\|U_1-U_2\|_\infty$. The maximum satisfies $|\max_a f(a)-\max_a g(a)|\le\max_a|f(a)-g(a)|$. Taking the supremum over states proves the contraction for every $\beta$.
\end{proof}

\begin{proof}[Proof of Theorem~\ref{thm:monotonic-improvement-rspi}]
\textbf{Monotone improvement.}
Write $W_t:=U^{\pi_t}_\beta$. The fixed-point identity for the current policy and greediness of $\pi_{t+1}$ give
\begin{equation}
\label{eq:improvement_step}
W_t=\mathcal{T}^{\pi_t}_\beta W_t\le \mathcal{T}^*_\beta W_t=\mathcal{T}^{\pi_{t+1}}_\beta W_t.
\end{equation}
Monotone iteration of $\mathcal{T}^{\pi_{t+1}}_\beta$ from $W_t$ converges to $W_{t+1}$, so $W_t\le\mathcal{T}^*_\beta W_t\le W_{t+1}$.

\textbf{Convergence to the optimal value.}
Since $W_{t+1}\le U^*_\beta$ and $U^*_\beta=\mathcal{T}^*_\beta U^*_\beta$, the improvement bound gives
\[
0\le U^*_\beta-W_{t+1}\le \mathcal{T}^*_\beta U^*_\beta-\mathcal{T}^*_\beta W_t.
\]
Lemma~\ref{lemma:optimal-bellman-contraction-mapping} gives
$\|U^*_\beta-W_{t+1}\|_\infty\le\gamma\|U^*_\beta-W_t\|_\infty$.
Hence the values converge to $U^*_\beta$ for every $\beta$. With the tie rule in Algorithm~\ref{alg:rspi}, a policy change requires strict improvement of the action score in at least one state. The value then improves strictly in that state, so the algorithm cannot revisit a policy. In a finite MDP it therefore terminates at a greedy policy for its own value, which is optimal by uniqueness of the fixed point of $\mathcal{T}^*_\beta$.
\end{proof}

\subsection{Stationary Entropic Operators}
\label{app:exact-surrogate-operators}

For the fixed-policy results in Sections~\ref{app:exact-surrogate-operators}--\ref{app:static-nested}, fix a deterministic Markov policy $\pi$ and write
$p(\cdot\mid s)$ for $P(\cdot\mid s,\pi(s))$ and $r_\pi(s)$ for
$r(s,\pi(s))$. The state space $\mathcal{S}$ is finite or standard Borel and
the action set $\mathcal{A}$ is finite, as in the main text. All
functions $W:\mathcal{S}\to\mathbb{R}$ below are bounded and measurable, so
that every expectation written is finite. Rewards are bounded,
$|r(s,a)|\le r_{\max}<\infty$, and we write
$\mathrm{sp}(W):=\sup_{s}W(s)-\inf_{s}W(s)$ for the span seminorm and
$\mathrm{sp}(r):=\sup_{s,a}r(s,a)-\inf_{s,a}r(s,a)$. The discount is
$\gamma\in[0,1)$ and the risk parameter is $\beta\neq 0$ unless stated
otherwise. Two one-step operators act on bounded $W$:
\begin{align}
  (T^{\pi}_{\beta}W)(s)
  &:= \beta r_\pi(s) + \log\mathbb{E}_{p}\!\left[e^{\gamma W(s')}\,\middle|\, s\right]
  &&\text{(discount \emph{inside} in the multiplicative recursion),}
  \label{eq:op-exact}\\
  (\widetilde{T}^{\pi}_{\beta}W)(s)
  &:= \beta r_\pi(s) + \gamma\log\mathbb{E}_{p}\!\left[e^{W(s')}\,\middle|\, s\right]
  &&\text{(discount \emph{outside} in Equation~\ref{eq:bellman}).}
  \label{eq:op-surrogate}
\end{align}

The main text denotes the discount-outside operator by $T^\pi_\beta$. Here we write it as $\widetilde{T}^\pi_\beta$ and use $T^\pi_\beta$ for the discount-inside operator. The latter moves the discount power inside the conditional expectation of the multiplicative recursion. This interchange generally approximates the static criterion in Equation~\ref{eq:static-ce}. The operators have unique fixed points $\bar V_T$ and $\bar V_{\widetilde T}$ by Lemma~\ref{lemma:exact-contraction} and Proposition~\ref{prop:nested-exact}.

For $u\neq 0$ the \emph{entropic certainty equivalent} of a
bounded random variable $X$ is
\begin{equation}
  \rho^{\mathrm{ent}}_{u}[X] := \tfrac{1}{u}\log\mathbb{E}\!\left[e^{uX}\right],
  \qquad \rho^{\mathrm{ent}}_{0}[X] := \mathbb{E}[X],
  \label{eq:entropic-ce}
\end{equation}
and $\rho^{\mathrm{ent}}_{u}[X\mid s]$ denotes the conditional version with
$X=f(s')$, $s'\sim p(\cdot\mid s)$. Three elementary properties are used
repeatedly and follow directly from \eqref{eq:entropic-ce}: (P1)
\emph{monotonicity}: $X\le Y$ a.s.\ implies
$\rho^{\mathrm{ent}}_u[X]\le\rho^{\mathrm{ent}}_u[Y]$. (P2)
\emph{translation equivariance}: $\rho^{\mathrm{ent}}_u[X+c]=
\rho^{\mathrm{ent}}_u[X]+c$ for every constant $c\in\mathbb{R}$. (P3)
\emph{range}: $\operatorname*{ess\,inf}X\le\rho^{\mathrm{ent}}_u[X]\le
\operatorname*{ess\,sup}X$, since $e^{u\operatorname*{ess\,inf}X}\le
\mathbb{E}[e^{uX}]\le e^{u\operatorname*{ess\,sup}X}$ for $u>0$ (with the
two outer inequalities swapped for $u<0$, in which case dividing by $u<0$
restores the stated order).

\subsubsection{Discount-Inside Contraction}

The discount-inside recursion also contracts. For a fixed policy we state the result in scaled units, as used in the fixed-point comparison below. For control we state it in certainty-equivalent units.

\begin{lemma}[Discount-inside contraction]
\label{lemma:exact-contraction}
For all bounded $W_1,W_2:\mathcal{S}\to\mathbb{R}$, the fixed-policy operator satisfies
\[
\|T^\pi_\beta W_1-T^\pi_\beta W_2\|_\infty\le\gamma\|W_1-W_2\|_\infty.
\]
The optimal discount-inside operator in certainty-equivalent units is
\[
(\mathcal{T}^{*,\mathrm{in}}_\beta U)(s)
=\max_{a\in\mathcal A}\left\{r(s,a)+\rho^{\mathrm{ent}}_\beta[\gamma U(s')\mid s,a]\right\}.
\]
It too is a $\gamma$-contraction. Each operator has a unique bounded fixed point.
\end{lemma}

\begin{proof}
For $T^\pi_\beta$, use the sup-norm non-expansion of the log-moment map, which is $\rho^{\mathrm{ent}}_1$, on $\gamma W_1(s')$ and $\gamma W_2(s')$. The reward term cancels and the pointwise difference is bounded by $\gamma\|W_1-W_2\|_\infty$. For the optimal operator, the same argument uses $\rho^{\mathrm{ent}}_\beta$ on $\gamma U_1(s')$ and $\gamma U_2(s')$ at each fixed state and action. Lemma~\ref{lemma:bellman-operator-contraction-mapping} establishes the non-expansion for both signs of $\beta$. Taking the maximum over actions and then the supremum over states gives the bound, as in Lemma~\ref{lemma:optimal-bellman-contraction-mapping}. Banach's fixed-point theorem gives uniqueness.
\end{proof}

\subsubsection{Exactness for the Nested Criterion}

The discount-outside operator is exact for the nested entropic criterion. This criterion applies a conditional certainty equivalent at each step in the Markov risk formulation of \citet{ruszczynski2010risk}. The discount-inside operator belongs to the same family with per-step risk parameter $\gamma\beta$.

\begin{proposition}[Nested-criterion exactness]
\label{prop:nested-exact}
Let $\beta\neq0$ and $\gamma\in[0,1)$.
\begin{enumerate}
  \item[(a)] Define the nested entropic Bellman operator
  $(\mathcal{T}^{\pi}_{\beta}U)(s):=r_\pi(s)+
  \gamma\,\rho^{\mathrm{ent}}_{\beta}\!\left[U(s')\,\middle|\,s\right]$
  on bounded $U:\mathcal{S}\to\mathbb{R}$. Then for every bounded $U$,
  \begin{equation}
    \widetilde{T}^{\pi}_{\beta}(\beta U) \;=\; \beta\,\mathcal{T}^{\pi}_{\beta}U .
    \label{eq:conjugation}
  \end{equation}
  Consequently $W^{*}$ is the fixed point of $\widetilde{T}^\pi_\beta$ if and
  only if $U^{*}:=W^{*}/\beta$ is the unique fixed point of
  $\mathcal{T}^\pi_\beta$, i.e.\ the unique bounded solution of the nested
  recursion
  \begin{equation}
    U^{\pi}_{\beta}(s) \;=\; r(s,\pi(s)) \;+\;
    \gamma\,\rho^{\mathrm{ent}}_{\beta}\!\left[U^{\pi}_{\beta}(s')\,\middle|\,s\right],
    \qquad s\in\mathcal{S}.
    \label{eq:nested-fp}
  \end{equation}
  \item[(b)] For $\gamma\in(0,1)$ the same substitution applied to the discount-inside
  operator gives
  $T^{\pi}_{\beta}(\beta U)=\beta\big(r_\pi+
  \gamma\,\rho^{\mathrm{ent}}_{\gamma\beta}[U(s')\mid\cdot]\big)$, so
  $\bar{V}_{T}/\beta$ is the unique bounded solution of the nested recursion
  \eqref{eq:nested-fp} with per-step risk parameter $\gamma\beta$ in place of
  $\beta$.
\end{enumerate}
\end{proposition}

\begin{proof}
(a) Fix $s$ and a bounded $U$. By definition
\eqref{eq:op-surrogate},
\[
  \big(\widetilde{T}^{\pi}_{\beta}(\beta U)\big)(s)
  = \beta r_\pi(s) + \gamma\log\mathbb{E}_p\!\left[e^{\beta U(s')}\,\middle|\,s\right]
  = \beta\Big( r_\pi(s) + \gamma\cdot\tfrac{1}{\beta}\log
    \mathbb{E}_p\!\left[e^{\beta U(s')}\,\middle|\,s\right]\Big)
  = \beta\,(\mathcal{T}^{\pi}_{\beta}U)(s),
\]
where the middle equality multiplies and divides the second summand by
$\beta\neq0$ (an identity, valid for either sign of $\beta$), and the last
equality is definition \eqref{eq:entropic-ce}. Since $U\mapsto\beta U$ is a
bijection of the space of bounded functions onto itself,
\eqref{eq:conjugation} maps fixed points of $\mathcal{T}^\pi_\beta$ bijectively onto fixed points of $\widetilde{T}^\pi_\beta$. Lemma~\ref{lemma:bellman-operator-contraction-mapping} and Banach's fixed-point theorem give the unique $U^\pi_\beta$. The conjugacy gives the unique scaled fixed point and transfers the sup-norm contraction bound to $\widetilde{T}^\pi_\beta$, since the factors $|\beta|$ and $1/|\beta|$ cancel.

(b) Identically, using \eqref{eq:op-exact} and $\gamma>0$:
\begin{align*}
  \big(T^{\pi}_{\beta}(\beta U)\big)(s)
  &= \beta r_\pi(s) + \log\mathbb{E}_p\!\left[e^{(\gamma\beta) U(s')}\,\middle|\,s\right] \\
  &= \beta\Big(r_\pi(s) + \gamma\cdot\tfrac{1}{\gamma\beta}
    \log\mathbb{E}_p\!\left[e^{(\gamma\beta)U(s')}\,\middle|\,s\right]\Big) \\
  &= \beta\Big(r_\pi(s)+\gamma\,\rho^{\mathrm{ent}}_{\gamma\beta}
    \!\left[U(s')\,\middle|\,s\right]\Big),
\end{align*}
and the fixed-point correspondence follows as in (a), with uniqueness from
Lemma~\ref{lemma:exact-contraction}.
\end{proof}

\begin{remark}[The nested entropic recursion]
\label{remark:what-is-exact}
Recursion \eqref{eq:nested-fp} has precisely the form of the discounted
risk-averse dynamic-programming equation of \citet{ruszczynski2010risk},
with the one-step conditional risk mapping instantiated by the entropic
certainty equivalent $\rho^{\mathrm{ent}}_\beta$. (That reference develops
the theory for \emph{coherent} Markov risk mappings. The mapping $\rho^{\mathrm{ent}}_\beta$
is not positively homogeneous, hence not coherent. As a function of its
argument it is convex for $\beta>0$ and concave for $\beta<0$. In the
risk-averse regime $\beta<0$ it is the reward-convention mirror image
$X\mapsto-\rho^{\mathrm{ent}}_{|\beta|}[-X]$ of the convex conditional risk
mappings studied by \citet{ruszczynski2006conditional}. In either regime it
is monotone and translation-equivariant --- properties (P1)--(P2) above,
from which (P3) also follows by applying (P1)--(P2) to the constants
$\operatorname*{ess\,inf}X$ and $\operatorname*{ess\,sup}X$ together with
$\rho^{\mathrm{ent}}_u[0]=0$. The contraction proofs use (P1)--(P2). The quantitative gap bounds also use the smoothness of the cumulant generating function from Lemma~\ref{lem:cgf-tilt}.)
Recursion \eqref{eq:nested-fp} is thus the standard
\emph{time-consistent} formulation in which the certainty equivalent of the
next-state value is taken one step at a time.
Proposition~\ref{prop:nested-exact}(a) relates the scaled critic operator to the nested policy-evaluation operator. Lemmas~\ref{lemma:bellman-operator-contraction-mapping}--\ref{lemma:optimal-bellman-contraction-mapping} and Theorem~\ref{thm:monotonic-improvement-rspi} compare and optimize the nested values in certainty-equivalent units for either sign of $\beta$. The Relaxed-RSTD critic uses a variational approximation to the scaled target. By part~(b), the discount-inside operator belongs to the same nested family with per-step parameter $\gamma\beta$. Thus the
$T$-versus-$\widetilde{T}$ discrepancy quantified in
Lemma~\ref{lemma:chord-gap} and Proposition~\ref{prop:fixed-point-bias}
below is exactly a reparameterization $\beta\leftrightarrow\gamma\beta$
within one family of well-posed criteria. Neither fixed point coincides in
general with the \emph{static} objective of Equation~\ref{eq:risk-rl-obj} of the main text,
whose exact dynamic programming requires a time-varying risk schedule. The
static-versus-nested discrepancy is bounded, with explicit constants, in
Proposition~\ref{prop:static-nested}.
\end{remark}

\subsection{Discount Placement and Fixed-Point Bias}
\label{app:discount-placement}

We first bound the effect of moving the discount inside the one-step operator. We then propagate this bound through the fixed-point equations and express its sign in certainty-equivalent units.

\subsubsection{Per-Step Chord Gap}

\begin{lemma}[Chord gap]
\label{lemma:chord-gap}
Fix a bounded $W:\mathcal{S}\to\mathbb{R}$ and $s\in\mathcal{S}$, and define
the conditional cumulant generating function
\[
  g_s(u) := \log\mathbb{E}_{p(\cdot\mid s)}\!\left[e^{uW(s')}\right],
  \qquad u\in\mathbb{R}.
\]
Then $g_s$ is convex and infinitely differentiable with $g_s(0)=0$,
$g_s'(u)=\mathbb{E}_{Q^s_u}[W(s')]$ and
$g_s''(u)=\mathrm{Var}_{Q^s_u}[W(s')]\ge0$, where $Q^s_u$ is the
exponentially tilted kernel $dQ^s_u\propto e^{uW}\,dp(\cdot\mid s)$.
Moreover, for every $\gamma\in[0,1]$,
\begin{equation}
  0 \;\le\; \big(\widetilde{T}^{\pi}_{\beta}W - T^{\pi}_{\beta}W\big)(s)
  \;=\; \gamma\,g_s(1)-g_s(\gamma)
  \;=\; (1-\gamma)\!\int_{0}^{\gamma}\! u\,g_s''(u)\,du
        \;+\;\gamma\!\int_{\gamma}^{1}\!(1-u)\,g_s''(u)\,du
  \label{eq:chord-identity}
\end{equation}
and consequently
\begin{equation}
  \big(\widetilde{T}^{\pi}_{\beta}W - T^{\pi}_{\beta}W\big)(s)
  \;\le\; \frac{\gamma(1-\gamma)}{2}\,
     \sup_{u\in[0,1]}\mathrm{Var}_{Q^s_u}\!\left[W(s')\right]
  \;\le\; \frac{\gamma(1-\gamma)}{8}\,\mathrm{sp}_s(W)^2
  \;\le\; \frac{\gamma(1-\gamma)}{8}\,\mathrm{sp}(W)^2,
  \label{eq:chord-bound}
\end{equation}
where $\mathrm{sp}_s(W)$ is the span of $W$ restricted to the support of
$p(\cdot\mid s)$. The gap vanishes identically when $\gamma\in\{0,1\}$, when
$p(\cdot\mid s)$ is a point mass (deterministic transition at $s$), and, more
generally, when $W$ is constant on the support of $p(\cdot\mid s)$.
\end{lemma}

\begin{proof}
\emph{Regularity.} Apply Lemma~\ref{lem:cgf-tilt}(i) to $W(s')$ under $p(\cdot\mid s)$. The function $g_s$ is smooth with $g_s(0)=0$ and $g_s''(u)=\mathrm{Var}_{Q^s_u}[W(s')]\ge0$, so it is convex.

\emph{First equality in \eqref{eq:chord-identity}.} By definitions
\eqref{eq:op-exact}--\eqref{eq:op-surrogate}, the reward terms
$\beta r_\pi(s)$ cancel and
$(\widetilde{T}^{\pi}_{\beta}W-T^{\pi}_{\beta}W)(s)
=\gamma\log\mathbb{E}_p[e^{W(s')}\mid s]-\log\mathbb{E}_p[e^{\gamma
W(s')}\mid s]=\gamma g_s(1)-g_s(\gamma)$. The difference does not
depend on $\beta$.

\emph{One-sidedness.} Since $g_s$ is convex, $g_s(0)=0$, and
$\gamma=\gamma\cdot1+(1-\gamma)\cdot0$ with $\gamma\in[0,1]$,
\[
  g_s(\gamma)\;\le\;\gamma\,g_s(1)+(1-\gamma)\,g_s(0)\;=\;\gamma\,g_s(1):
\]
the chord of a convex function lies above the function, whence
$\gamma g_s(1)-g_s(\gamma)\ge0$.

\emph{Integral identity.} Taylor's theorem with integral remainder for the
$C^2$ function $g_s$, expanded at $\gamma$, gives
\[
  g_s(1)=g_s(\gamma)+(1-\gamma)g_s'(\gamma)+\int_{\gamma}^{1}(1-u)\,g_s''(u)\,du,
  \qquad
  0=g_s(0)=g_s(\gamma)-\gamma g_s'(\gamma)+\int_{0}^{\gamma}u\,g_s''(u)\,du,
\]
where the second identity uses
$\int_{\gamma}^{0}(0-u)g_s''(u)\,du=\int_{0}^{\gamma}u\,g_s''(u)\,du$.
The second identity rearranges to
$\gamma g_s'(\gamma)-g_s(\gamma)=\int_{0}^{\gamma}u\,g_s''(u)\,du$.
Therefore
\begin{align*}
  \gamma g_s(1)-g_s(\gamma)
  &= \gamma\Big(g_s(\gamma)+(1-\gamma)g_s'(\gamma)
     +\int_{\gamma}^{1}(1-u)g_s''(u)\,du\Big)-g_s(\gamma)\\
  &= (1-\gamma)\big(\gamma g_s'(\gamma)-g_s(\gamma)\big)
     +\gamma\int_{\gamma}^{1}(1-u)\,g_s''(u)\,du\\
  &= (1-\gamma)\int_{0}^{\gamma}u\,g_s''(u)\,du
     +\gamma\int_{\gamma}^{1}(1-u)\,g_s''(u)\,du,
\end{align*}
which is \eqref{eq:chord-identity}.

\emph{Variance bound.} Bounding $g_s''(u)\le
\Sigma_s:=\sup_{u\in[0,1]}\mathrm{Var}_{Q^s_u}[W(s')]$ on $[0,1]$ inside
both integrals of \eqref{eq:chord-identity} and evaluating
\[
  (1-\gamma)\int_{0}^{\gamma}u\,du+\gamma\int_{\gamma}^{1}(1-u)\,du
  = (1-\gamma)\frac{\gamma^{2}}{2}+\gamma\frac{(1-\gamma)^{2}}{2}
  = \frac{\gamma(1-\gamma)}{2}\big(\gamma+(1-\gamma)\big)
  = \frac{\gamma(1-\gamma)}{2}
\]
yields the first inequality in \eqref{eq:chord-bound}.

\emph{Span bound.} The tilted law $Q^s_u$ is supported on the support of
$p(\cdot\mid s)$, on which $W$ takes values in an interval of length
$\mathrm{sp}_s(W)$. For any random variable $Y$ with $a\le Y\le b$ a.s.,
\[
  \mathrm{Var}[Y]\;\le\;\mathbb{E}\Big[\big(Y-\tfrac{a+b}{2}\big)^{2}\Big]
  \;\le\;\Big(\tfrac{b-a}{2}\Big)^{2},
\]
the first step because $\mathbb{E}[(Y-c)^2]$ is minimized over $c$ at
$c=\mathbb{E}[Y]$, the second because $|Y-\tfrac{a+b}{2}|\le\tfrac{b-a}{2}$
almost surely. The combination is Popoviciu's variance inequality. Hence
$\Sigma_s\le\mathrm{sp}_s(W)^2/4\le\mathrm{sp}(W)^2/4$, giving the last two
inequalities in \eqref{eq:chord-bound}.

\emph{Exactness cases.} If $\gamma\in\{0,1\}$ then
$\gamma g_s(1)-g_s(\gamma)$ equals $0-g_s(0)=0$ or $g_s(1)-g_s(1)=0$. If
$p(\cdot\mid s)$ is a point mass at $s'_{0}$, or more generally $W\equiv c$
on the support of $p(\cdot\mid s)$, then $g_s(u)=uc$ is linear, so
$g_s''\equiv0$ and \eqref{eq:chord-identity} vanishes. Conversely, the
variance-form constant $\gamma(1-\gamma)/2$ in \eqref{eq:chord-bound} is
approached when $g_s''$ is (nearly) constant on $[0,1]$, and Popoviciu's
inequality is tight for a two-point conditional law with equal masses at the
two extremes of $W$.
\end{proof}

\subsubsection{Fixed-Point Bias and Horizon Dependence}

\begin{proposition}[One-sided, non-compounding fixed-point bias]
\label{prop:fixed-point-bias}
Let $\bar{V}_{T}$ and $\bar{V}_{\widetilde{T}}$ be the fixed points of
$T^\pi_\beta$ and $\widetilde{T}^\pi_\beta$, and define
\[
  \bar\sigma^{2} \;:=\; \sup_{s\in\mathcal{S}}\;\sup_{u\in[0,1]}\;
  \mathrm{Var}_{Q^{s}_{u}}\!\left[\bar{V}_{T}(s')\right],
  \qquad dQ^{s}_{u}\propto e^{u\bar{V}_{T}}\,dp(\cdot\mid s).
\]
Then:
\begin{enumerate}
  \item[(i)] (One-sidedness) $\bar{V}_{\widetilde{T}}(s)\ge\bar{V}_{T}(s)$
  for every $s\in\mathcal{S}$.
  \item[(ii)] (Horizon cancellation)
  \begin{equation}
    \big\|\bar{V}_{\widetilde{T}}-\bar{V}_{T}\big\|_{\infty}
    \;\le\; \frac{1}{1-\gamma}\,
      \sup_{s}\big(\widetilde{T}^{\pi}_{\beta}\bar{V}_{T}
      -T^{\pi}_{\beta}\bar{V}_{T}\big)(s)
    \;\le\; \frac{1}{1-\gamma}\cdot\frac{\gamma(1-\gamma)}{2}\,\bar\sigma^{2}
    \;=\; \frac{\gamma}{2}\,\bar\sigma^{2}:
    \label{eq:horizon-cancellation}
  \end{equation}
  the per-step factor $(1-\gamma)$ from Lemma~\ref{lemma:chord-gap} cancels
  the $1/(1-\gamma)$ horizon factor of the contraction--perturbation
  argument exactly, so the bias does not compound over the horizon.
  \item[(iii)] (Fully explicit worst case)
  $\mathrm{sp}(\bar{V}_{T})\le\dfrac{|\beta|\,\mathrm{sp}(r)}{1-\gamma}$, and
  therefore
  \[
    \big\|\bar{V}_{\widetilde{T}}-\bar{V}_{T}\big\|_{\infty}
    \;\le\;\frac{\gamma}{2}\cdot\frac{\mathrm{sp}(\bar{V}_{T})^{2}}{4}
    \;\le\;\frac{\gamma}{8}\left(\frac{|\beta|\,\mathrm{sp}(r)}{1-\gamma}\right)^{\!2}.
  \]
\end{enumerate}
The bias vanishes when the dynamics are deterministic and when
$\gamma=0$ (in both cases the per-step gap of
Lemma~\ref{lemma:chord-gap} is identically zero), and it is
$O(\beta^{2})$ as $\beta\to0$ by (iii). Note that the \emph{per-step} gap
also vanishes as $\gamma\uparrow1$, but the fixed-point bound
\eqref{eq:horizon-cancellation} does not: the vanishing per-step factor is
exactly consumed by the growing horizon factor.
\end{proposition}

\begin{proof}
(i) By Lemma~\ref{lemma:chord-gap} applied with $W=\bar{V}_{T}$,
$\widetilde{T}^\pi_\beta\bar{V}_{T}\ge T^\pi_\beta\bar{V}_{T}
=\bar{V}_{T}$ pointwise. The operator $\widetilde{T}^\pi_\beta$ is
monotone: if $W_1\le W_2$ pointwise then
$\log\mathbb{E}_p[e^{W_1}\mid s]\le\log\mathbb{E}_p[e^{W_2}\mid s]$ and
multiplying by $\gamma\ge0$ and adding $\beta r_\pi(s)$ preserves the
order. By induction, from
$\widetilde{T}^\pi_\beta\bar{V}_{T}\ge\bar{V}_{T}$ and monotonicity,
$(\widetilde{T}^\pi_\beta)^{n+1}\bar{V}_{T}\ge
(\widetilde{T}^\pi_\beta)^{n}\bar{V}_{T}\ge\bar{V}_{T}$ for all $n\ge0$.
Since $\widetilde{T}^\pi_\beta$ is a $\gamma$-contraction
(Proposition~\ref{prop:nested-exact}),
$(\widetilde{T}^\pi_\beta)^{n}\bar{V}_{T}\to\bar{V}_{\widetilde{T}}$ in the
sup-norm, and pointwise inequalities pass to the limit:
$\bar{V}_{\widetilde{T}}\ge\bar{V}_{T}$.

(ii) Using the triangle inequality, the fixed-point property of both
operators, and the $\gamma$-contraction of $\widetilde{T}^\pi_\beta$,
\begin{align*}
  \|\bar{V}_{\widetilde{T}}-\bar{V}_{T}\|_\infty
  &= \|\widetilde{T}^\pi_\beta\bar{V}_{\widetilde{T}}
      -T^\pi_\beta\bar{V}_{T}\|_\infty \\
  &\le \|\widetilde{T}^\pi_\beta\bar{V}_{\widetilde{T}}
      -\widetilde{T}^\pi_\beta\bar{V}_{T}\|_\infty
    + \|\widetilde{T}^\pi_\beta\bar{V}_{T}-T^\pi_\beta\bar{V}_{T}\|_\infty \\
  &\le \gamma\,\|\bar{V}_{\widetilde{T}}-\bar{V}_{T}\|_\infty
    + \sup_s\big(\widetilde{T}^\pi_\beta\bar{V}_{T}
      -T^\pi_\beta\bar{V}_{T}\big)(s),
\end{align*}
where the last sup equals the sup-norm of the difference because the
difference is nonnegative by Lemma~\ref{lemma:chord-gap}. Rearranging (the
left side is finite since both fixed points are bounded) gives the first
inequality of \eqref{eq:horizon-cancellation}. The second inequality is
\eqref{eq:chord-bound} with $W=\bar{V}_{T}$, whose tilted variances are by
definition bounded by $\bar\sigma^{2}$. The final equality is the explicit
cancellation $\frac{1}{1-\gamma}\cdot\frac{\gamma(1-\gamma)}{2}
=\frac{\gamma}{2}$.

(iii) First, the span bound for $\bar{V}_{T}$. For any bounded $W$ and any
$s$, property (P3)-style bounds give
$\gamma\inf W\le\log\mathbb{E}_p[e^{\gamma W(s')}\mid s]\le\gamma\sup W$,
so the function $s\mapsto\log\mathbb{E}_p[e^{\gamma W(s')}\mid s]$ has span
at most $\gamma\,\mathrm{sp}(W)$. Since the span seminorm is subadditive,
\[
  \mathrm{sp}(T^\pi_\beta W)
  \;\le\;\mathrm{sp}\big(\beta r_\pi\big)+\gamma\,\mathrm{sp}(W)
  \;\le\;|\beta|\,\mathrm{sp}(r)+\gamma\,\mathrm{sp}(W),
\]
using $\mathrm{sp}(\beta r_\pi)=|\beta|\,\mathrm{sp}(r_\pi)\le
|\beta|\,\mathrm{sp}(r)$. Applying this at the fixed point
$W=\bar{V}_{T}=T^\pi_\beta\bar{V}_{T}$ yields
$\mathrm{sp}(\bar{V}_{T})\le|\beta|\,\mathrm{sp}(r)+
\gamma\,\mathrm{sp}(\bar{V}_{T})$, i.e.\
$\mathrm{sp}(\bar{V}_{T})\le|\beta|\,\mathrm{sp}(r)/(1-\gamma)$. Second, by
Popoviciu's inequality as in the proof of Lemma~\ref{lemma:chord-gap},
$\bar\sigma^{2}\le\mathrm{sp}(\bar{V}_{T})^{2}/4$. Combining with (ii)
proves (iii).
\end{proof}

\begin{remark}[Horizon dependence of the fixed-point bound]
\label{remark:gamma-independence-honest}
In Equation~\ref{eq:horizon-cancellation}, the factors $(1-\gamma)$ and
$1/(1-\gamma)$ cancel, leaving $\frac{\gamma}{2}\bar\sigma^{2}$.
The tilted next-state variance $\bar\sigma^{2}$ still depends on $\gamma$
through $\bar{V}_{T}$. It is controlled by the span of $\bar{V}_{T}$ across
successor states. The explicit worst-case bound in part~(iii) scales as
$(1-\gamma)^{-2}$.
\end{remark}

\begin{corollary}[The surrogate is conservative in the risk-averse regime]
\label{cor:conservative}
Write the fixed points in certainty-equivalent units,
$U_{T}:=\bar{V}_{T}/\beta$ and
$U_{\widetilde{T}}:=\bar{V}_{\widetilde{T}}/\beta$
(cf.\ Proposition~\ref{prop:nested-exact}), and let
$\bar\sigma_{U}^{2}:=\bar\sigma^{2}/\beta^{2}$, which equals
$\sup_{s}\sup_{u\in[0,1]}\mathrm{Var}_{Q^{s}_{u}}[U_{T}(s')]$ with the same
tilted kernels $dQ^s_u\propto e^{u\beta U_T}dp$. If $\beta<0$ then
\[
  0 \;\le\; U_{T}(s)-U_{\widetilde{T}}(s)
  \;\le\; \frac{\gamma}{2|\beta|}\,\bar\sigma^{2}
  \;=\; \frac{\gamma\,|\beta|}{2}\,\bar\sigma_{U}^{2}
  \qquad\text{for all } s:
\]
the surrogate \emph{under}-estimates the exact certainty equivalent, i.e.\
the approximation errs on the conservative side exactly in the risk-averse
regime, and the certainty-equivalent bias is $O(|\beta|)$ at fixed
$\bar\sigma_U^2$. If $\beta>0$ the inequalities reverse
($0\le U_{\widetilde{T}}-U_{T}\le\frac{\gamma|\beta|}{2}\bar\sigma_U^2$):
the surrogate over-estimates in the risk-seeking regime.
\end{corollary}

\begin{proof}
By Proposition~\ref{prop:fixed-point-bias}(i)--(ii),
$0\le\bar{V}_{\widetilde{T}}-\bar{V}_{T}\le\frac{\gamma}{2}\bar\sigma^{2}$
pointwise. Dividing by $\beta<0$ reverses both inequalities and gives
$0\le U_{T}-U_{\widetilde{T}}\le\frac{\gamma}{2|\beta|}\bar\sigma^{2}$. Dividing by $\beta>0$ preserves them. Finally
$\mathrm{Var}[\bar{V}_{T}(s')]=\beta^{2}\mathrm{Var}[U_{T}(s')]$ under any
fixed probability measure, so $\bar\sigma^{2}=\beta^{2}\bar\sigma_{U}^{2}$.
\end{proof}

\subsection{Static versus Nested Criterion}
\label{app:static-nested}

The comparison in Section~\ref{app:discount-placement} concerns two stationary operators. We now compare the nested value $U^\pi_\beta$ in Equation~\ref{eq:nested-fp} with the static objective in Equation~\ref{eq:risk-rl-obj}, conditioned on the initial state.
\begin{equation}
  C^{\pi}_{\beta}(s) \;:=\; \frac{1}{\beta}\log\mathbb{E}\!\left[
    e^{\beta\sum_{t=0}^{\infty}\gamma^{t}r_{t}}\,\middle|\,s_{0}=s\right]
  \;=\; \rho^{\mathrm{ent}}_{\beta}\!\left[\textstyle\sum_{t\ge0}\gamma^{t}r_{t}
    \,\middle|\,s_{0}=s\right],
  \label{eq:static-ce}
\end{equation}
the expectation being over trajectories generated by $\pi$ and $P$. Lemma~\ref{lemma:ce-lipschitz} gives the Lipschitz estimate in the risk parameter used below.

\begin{proposition}[Static versus nested entropic criterion]
\label{prop:static-nested}
Let $U^{\pi}_{\beta}$ be the nested value \eqref{eq:nested-fp} and
$C^{\pi}_{\beta}$ the static certainty equivalent \eqref{eq:static-ce},
with $\beta\neq0$ and $\gamma\in(0,1)$. Define, for $t\ge0$,
\[
  h_{t}(s) \;:=\; \rho^{\mathrm{ent}}_{\beta\gamma^{t}}\!\left[
    \textstyle\sum_{l\ge t}\gamma^{\,l-t}\,r_{l}\,\middle|\,s_{t}=s\right],
  \qquad\text{so that}\quad h_{0}=C^{\pi}_{\beta}.
\]
Then:
\begin{enumerate}
  \item[(i)] (Exact time-varying decomposition and the RASR observation
  \citep{hau2022rasr}) For every $t\ge0$ and $s\in\mathcal{S}$,
  \begin{equation}
    h_{t}(s) \;=\; r(s,\pi(s)) \;+\;
    \gamma\,\rho^{\mathrm{ent}}_{\beta\gamma^{t+1}}\!\left[h_{t+1}(s')\,\middle|\,s\right],
    \label{eq:rasr-recursion}
  \end{equation}
  with no approximation. The static criterion thus satisfies exact dynamic
  programming with the nonstationary risk schedule
  $\beta_{t}=\beta\gamma^{t}$, while the nested criterion
  \eqref{eq:nested-fp} freezes the schedule at $\beta$.
  \item[(ii)] (Sign) If $\beta<0$ then $C^{\pi}_{\beta}(s)\ge
  U^{\pi}_{\beta}(s)$ for all $s$. If $\beta>0$ then
  $C^{\pi}_{\beta}(s)\le U^{\pi}_{\beta}(s)$ for all $s$. In particular, in
  the risk-averse regime the nested value --- the quantity our operators
  compute exactly --- \emph{under}-estimates the static certainty
  equivalent: the discrepancy, like that of
  Corollary~\ref{cor:conservative}, is conservative.
  \item[(iii)] (Explicit bound)
  \begin{equation}
    \big\|C^{\pi}_{\beta}-U^{\pi}_{\beta}\big\|_{\infty}
    \;\le\; \frac{|\beta|\,\mathrm{sp}(U^{\pi}_{\beta})^{2}}{8}
      \sum_{t=0}^{\infty}\gamma^{t+1}\big(1-\gamma^{t+1}\big)
    \;=\; \frac{|\beta|\,\gamma}{8\,(1-\gamma^{2})}\,
      \mathrm{sp}(U^{\pi}_{\beta})^{2}
    \;\le\; \frac{|\beta|\,\gamma\,\mathrm{sp}(r)^{2}}
      {8\,(1-\gamma^{2})(1-\gamma)^{2}}\,.
    \label{eq:static-nested-bound}
  \end{equation}
\end{enumerate}
The two criteria coincide ($C^{\pi}_{\beta}=U^{\pi}_{\beta}$) when the
transition kernel is deterministic, and both converge to the risk-neutral
value $V^{\pi}$ with the discrepancy vanishing linearly as $\beta\to0$. As
$\gamma\to0^+$, both sides of \eqref{eq:static-nested-bound} tend to $0$. As
$\gamma\uparrow1$ the fully explicit form on the right of
\eqref{eq:static-nested-bound} grows as
$\mathrm{sp}(r)^{2}|\beta|/\big(16(1-\gamma)^{3}\big)$. The static-versus-nested discrepancy therefore compounds with the horizon in this worst-case bound. The $T$-versus-$\widetilde{T}$ bias of Proposition~\ref{prop:fixed-point-bias} does not. This horizon dependence is
the price paid here for replacing the exact time-varying schedule
$\beta_{t}=\beta\gamma^{t}$ by a stationary single-parameter recursion.
\end{proposition}

\begin{proof}
Throughout, $\|h_{t}\|_\infty\le r_{\max}/(1-\gamma)$ and
$\|U^{\pi}_{\beta}\|_\infty\le r_{\max}/(1-\gamma)$: the first by property
(P3), since $|\sum_{l\ge t}\gamma^{l-t}r_{l}|\le r_{\max}/(1-\gamma)$
almost surely. The second bound follows from \eqref{eq:nested-fp} and (P3), which give
$\|U^{\pi}_{\beta}\|_\infty\le r_{\max}+\gamma\|U^{\pi}_{\beta}\|_\infty$.

(i) Write $B_{t}:=\beta\gamma^{t}\neq0$ and
$G_{t+1}:=\sum_{l\ge t+1}\gamma^{\,l-(t+1)}r_{l}$, so that
$\sum_{l\ge t}\gamma^{l-t}r_{l}=r_{t}+\gamma G_{t+1}$ and $r_t=r(s,\pi(s))$
is a deterministic function of $s_t=s$. Then
\[
  B_{t}\,h_{t}(s)
  = \log\mathbb{E}\!\left[e^{B_{t}(r_{t}+\gamma G_{t+1})}\,\middle|\,s_{t}=s\right]
  = B_{t}\,r(s,\pi(s))
    + \log\mathbb{E}\!\left[e^{B_{t+1}G_{t+1}}\,\middle|\,s_{t}=s\right],
\]
using $B_{t}\gamma=B_{t+1}$. By the Markov property (the conditional law of
$G_{t+1}$ given $(s_{t},s_{t+1})$ depends only on $s_{t+1}$, and, the
policy and kernel being time-homogeneous, coincides with the law of the
same functional started at $s_{t+1}$) and the tower property,
\[
  \mathbb{E}\!\left[e^{B_{t+1}G_{t+1}}\,\middle|\,s_{t}=s\right]
  = \mathbb{E}_{s'\sim p(\cdot\mid s)}\!\left[
      \mathbb{E}\!\left[e^{B_{t+1}G_{t+1}}\,\middle|\,s_{t+1}=s'\right]\right]
  = \mathbb{E}_{s'\sim p(\cdot\mid s)}\!\left[e^{B_{t+1}h_{t+1}(s')}\right],
\]
the last step by the definition of $h_{t+1}$ (note the tilt parameter
matches: $h_{t+1}$ is defined with parameter $\beta\gamma^{t+1}=B_{t+1}$).
Dividing $B_{t}h_{t}(s)=B_{t}r(s,\pi(s))+
\log\mathbb{E}_{s'}[e^{B_{t+1}h_{t+1}(s')}]$ by $B_{t}$ and inserting
$1=\gamma/ B_{t+1}\cdot B_t$, i.e.\ $\tfrac{1}{B_t}=\tfrac{\gamma}{B_{t+1}}$,
into the second term yields exactly \eqref{eq:rasr-recursion}. Finally
$h_{0}=C^{\pi}_{\beta}$ by definition \eqref{eq:static-ce} with
$B_{0}=\beta$.

For (ii) and (iii), introduce the (nonstationary) operators
\[
  (\mathcal{T}_{t}f)(s):=r_\pi(s)+\gamma\,
    \rho^{\mathrm{ent}}_{\beta\gamma^{t+1}}\!\left[f(s')\,\middle|\,s\right],
  \qquad
  (\mathcal{T}f)(s):=r_\pi(s)+\gamma\,
    \rho^{\mathrm{ent}}_{\beta}\!\left[f(s')\,\middle|\,s\right],
\]
so that $h_{t}=\mathcal{T}_{t}h_{t+1}$ by (i) and
$U^{\pi}_{\beta}=\mathcal{T}U^{\pi}_{\beta}$ by \eqref{eq:nested-fp}. Each
$\mathcal{T}_{t}$ (and $\mathcal{T}$) is \emph{monotone} by (P1). It is also
\emph{translation-affine}, $\mathcal{T}_{t}(f+c)=\mathcal{T}_{t}f+\gamma c$
for constants $c$ by (P2), and a \emph{$\gamma$-contraction} in the
sup-norm: for any fixed parameter $u\in\mathbb{R}$ and bounded $f_{1},f_{2}$
with $d:=\|f_{1}-f_{2}\|_{\infty}$, the pointwise bound $f_{1}\le f_{2}+d$
together with (P1) and (P2) gives
$\rho^{\mathrm{ent}}_{u}[f_{1}(s')\mid s]\le
\rho^{\mathrm{ent}}_{u}[f_{2}(s')\mid s]+d$ at every $s$. Exchanging
$f_{1},f_{2}$ and multiplying by $\gamma$ (the $r_\pi$ terms cancel) yields
$\|\mathcal{T}_{t}f_{1}-\mathcal{T}_{t}f_{2}\|_{\infty}\le\gamma d$. This is
the argument of Lemma~\ref{lemma:bellman-operator-contraction-mapping} with
the exponential-monotonicity step routed through (P1)--(P2), which cover
both signs of $u$ (at $u=0$, $\rho^{\mathrm{ent}}_{0}=\mathbb{E}$ is
trivially nonexpansive).

(ii) Let $\beta<0$. Then $\beta\gamma^{t+1}\in[\beta,0)$, so
$\beta\gamma^{t+1}\ge\beta$, and by the monotonicity of the certainty
equivalent in its parameter (Lemma~\ref{lemma:ce-lipschitz},
$C'\ge0$), $\rho^{\mathrm{ent}}_{\beta\gamma^{t+1}}[X]\ge
\rho^{\mathrm{ent}}_{\beta}[X]$ for every bounded $X$. Hence
$\mathcal{T}_{t}f\ge\mathcal{T}f$ pointwise for every bounded $f$ and every
$t$. Define $m_{t}:=\inf_{s}\big(h_{t}(s)-U^{\pi}_{\beta}(s)\big)$, finite
since both functions are bounded (by at most $2r_{\max}/(1-\gamma)$ in
absolute difference). From $h_{t+1}\ge U^{\pi}_{\beta}+m_{t+1}$,
monotonicity and translation-affinity give
\[
  h_{t}=\mathcal{T}_{t}h_{t+1}
  \;\ge\;\mathcal{T}h_{t+1}
  \;\ge\;\mathcal{T}\big(U^{\pi}_{\beta}+m_{t+1}\big)
  \;=\;\mathcal{T}U^{\pi}_{\beta}+\gamma m_{t+1}
  \;=\;U^{\pi}_{\beta}+\gamma m_{t+1},
\]
so $m_{t}\ge\gamma m_{t+1}$ for every $t\ge0$. Iterating,
$m_{0}\ge\gamma^{T}m_{T}\ge-\gamma^{T}\cdot 2r_{\max}/(1-\gamma)$ for every
$T\ge1$, and letting $T\to\infty$ gives $m_{0}\ge0$, i.e.\
$C^{\pi}_{\beta}=h_{0}\ge U^{\pi}_{\beta}$. For $\beta>0$ we have
$\beta\gamma^{t+1}\le\beta$, hence $\mathcal{T}_{t}f\le\mathcal{T}f$, and
the symmetric argument with $M_{t}:=\sup_{s}(h_{t}-U^{\pi}_{\beta})(s)$
gives $M_{t}\le\gamma M_{t+1}$ and $C^{\pi}_{\beta}\le U^{\pi}_{\beta}$.

(iii) For each $t$, applying Lemma~\ref{lemma:ce-lipschitz} conditionally
at each $s$ to $X=U^{\pi}_{\beta}(s')$, $s'\sim p(\cdot\mid s)$ --- whose
span is at most $\mathrm{sp}(U^{\pi}_{\beta})$ --- with parameters
$u_{1}=\beta\gamma^{t+1}$ and $u_{2}=\beta$:
\[
  \varepsilon_{t}:=\big\|\mathcal{T}_{t}U^{\pi}_{\beta}
    -\mathcal{T}U^{\pi}_{\beta}\big\|_{\infty}
  = \gamma\,\sup_{s}\Big|\rho^{\mathrm{ent}}_{\beta\gamma^{t+1}}
      \!\left[U^{\pi}_{\beta}(s')\mid s\right]
    -\rho^{\mathrm{ent}}_{\beta}\!\left[U^{\pi}_{\beta}(s')\mid s\right]\Big|
  \;\le\;\gamma\,\frac{\mathrm{sp}(U^{\pi}_{\beta})^{2}}{8}\,
    |\beta|\big(1-\gamma^{t+1}\big).
\]
Next, the perturbation recursion: for every $t$,
\[
  \|h_{t}-U^{\pi}_{\beta}\|_{\infty}
  = \|\mathcal{T}_{t}h_{t+1}-\mathcal{T}U^{\pi}_{\beta}\|_{\infty}
  \le \|\mathcal{T}_{t}h_{t+1}-\mathcal{T}_{t}U^{\pi}_{\beta}\|_{\infty}
    + \varepsilon_{t}
  \le \gamma\,\|h_{t+1}-U^{\pi}_{\beta}\|_{\infty}+\varepsilon_{t},
\]
using the $\gamma$-contraction of $\mathcal{T}_{t}$. Unrolling this
inequality $T$ times,
\[
  \|h_{0}-U^{\pi}_{\beta}\|_{\infty}
  \;\le\;\sum_{t=0}^{T-1}\gamma^{t}\varepsilon_{t}
    +\gamma^{T}\,\|h_{T}-U^{\pi}_{\beta}\|_{\infty}
  \;\le\;\sum_{t=0}^{\infty}\gamma^{t}\varepsilon_{t}
    +\gamma^{T}\cdot\frac{2r_{\max}}{1-\gamma},
\]
and letting $T\to\infty$ leaves
$\|h_{0}-U^{\pi}_{\beta}\|_{\infty}\le\sum_{t\ge0}\gamma^{t}\varepsilon_{t}$.
Substituting the bound on $\varepsilon_{t}$ and evaluating the sum with the
index shift $k=t+1$,
\[
  \sum_{t=0}^{\infty}\gamma^{t}\cdot\gamma\big(1-\gamma^{t+1}\big)
  = \sum_{k=1}^{\infty}\big(\gamma^{k}-\gamma^{2k}\big)
  = \frac{\gamma}{1-\gamma}-\frac{\gamma^{2}}{1-\gamma^{2}}
  = \frac{\gamma(1+\gamma)-\gamma^{2}}{1-\gamma^{2}}
  = \frac{\gamma}{1-\gamma^{2}},
\]
which gives the equality in \eqref{eq:static-nested-bound}. For the final
inequality, the span bound
$\mathrm{sp}(U^{\pi}_{\beta})\le\mathrm{sp}(r)/(1-\gamma)$ follows exactly
as in Proposition~\ref{prop:fixed-point-bias}(iii): by (P3),
$s\mapsto\rho^{\mathrm{ent}}_{\beta}[U^{\pi}_{\beta}(s')\mid s]$ takes
values in $[\inf U^{\pi}_{\beta},\sup U^{\pi}_{\beta}]$, hence has span at
most $\mathrm{sp}(U^{\pi}_{\beta})$, so \eqref{eq:nested-fp} gives
$\mathrm{sp}(U^{\pi}_{\beta})\le\mathrm{sp}(r_\pi)+
\gamma\,\mathrm{sp}(U^{\pi}_{\beta})\le\mathrm{sp}(r)+
\gamma\,\mathrm{sp}(U^{\pi}_{\beta})$.

\emph{Degenerate cases.} If the kernel is deterministic, then conditionally
on each $s$ the variable $U^{\pi}_{\beta}(s')$ is a constant and, in
recursion \eqref{eq:rasr-recursion}, so is $h_{t+1}(s')$. Every certainty
equivalent of a constant equals that constant regardless of its parameter,
so $\mathcal{T}_{t}=\mathcal{T}$ on the relevant functions and
$h_{t}=U^{\pi}_{\beta}$ for all $t$ by uniqueness of the bounded solution
(both solve the same deterministic recursion, as follows formally from
$\varepsilon_{t}=0$ in the display above, using the per-state span
$\mathrm{sp}_{s}$ in Lemma~\ref{lemma:ce-lipschitz}, which vanishes). As
$\beta\to0$ with $\gamma$ fixed, the right side of
\eqref{eq:static-nested-bound} tends to $0$ linearly in $|\beta|$, and both
$C^{\pi}_{\beta}$ and $U^{\pi}_{\beta}$ converge uniformly to the
risk-neutral value $V^{\pi}$ (by Lemma~\ref{lemma:ce-lipschitz},
$|\rho^{\mathrm{ent}}_{\beta}[X]-\mathbb{E}[X]|\le|\beta|\,\mathrm{sp}(X)^2/8$
applied inside either recursion). As $\gamma\to0^{+}$ both
$h_{0}$ and $U^{\pi}_{\beta}$ converge uniformly to $r_\pi$ and the bound
\eqref{eq:static-nested-bound} tends to $0$.
\end{proof}

\begin{remark}[Implications for policy evaluation]
\label{remark:beta-operating-regime}
Proposition~\ref{prop:static-nested} gives a one-sided $O(|\beta|)$ bound with constant $\gamma\,\mathrm{sp}(U^\pi_\beta)^2/(8(1-\gamma^2))$. Both evaluation and greedy improvement in Theorem~\ref{thm:monotonic-improvement-rspi} use the nested certainty-equivalent value $U^\pi_\beta$. Proposition~\ref{prop:nested-exact} relates it to the scaled critic value $\beta U^\pi_\beta$ for $\beta\ne0$. Section~\ref{sec:related-work} discusses prior work on time-indexed dynamic programming for the static objective.
\end{remark}

\section{Theoretical Guarantees of Relaxed Risk-Sensitive Actor-Critic}
\label{appendix:relaxed-rs-ac}

We establish convergence of the critic and quantify its value error before analyzing the actor. We then state how the actor's surrogate objective relates to the static objective.

\subsection{Relaxed-RSTD Convergence}
\label{app:critic-convergence}

This section gives the detailed critic assumptions and proves both parts of Theorem~\ref{theorem:finitesamplerrorbound}. The first part establishes almost-sure convergence to the minimizer of $F$. The second part bounds the expected squared distance to that minimizer under the stated gradient conditions. Algorithm~\ref{alg:rrstd} describes the Relaxed-RSTD procedure.

\begin{assumption}
    \label{assumption:vardist1}
    For finite state-action spaces, $q$ is softmax-parameterizable. Given a state $s$, $q$ outputs a vector of probabilities over the next states. $q(s_{t+1}\mid s_t;\omega_t)$ denotes the softmax probability assigned to $s_{t+1}$ given $s_t$. We assume that $\omega$ is bounded. Following \citet{miyato2018spectral} and \citet{khromov2024some}, we assume that $\nabla_\omega q(s;\omega)$ is bounded.
\end{assumption}

\begin{assumption}
    \label{assumption:mdpboundedergodic}
    For each $\pi\in\Pi$, the induced Markov chain is ergodic. The mixing argument uses the geometric total-variation bound in Equation~\ref{ineq:levinemarkovmix}.
\end{assumption}

\begin{assumption}
\label{assumption:relaxedboundedweights}
    Parameters $\theta$ are bounded,  $\theta \in \mathcal{K}$, where $\mathcal{K}$ is a compact set with smooth boundary.
\end{assumption}

\begin{assumption}
\label{assumption:relaxedlinfuncapp}
    Feature map $\Phi$ is bounded $\|\phi^\top(s)\| \le \|B^\top_\phi\| \quad \forall s \in S$.
\end{assumption}

\begin{assumption}
\label{eq:robbins_monro}
The learning rate $\alpha_t$ satisfies 
$\sum_{t=0}^\infty \alpha_t = \infty,  \quad \sum_{t=0}^\infty \alpha_t^2 < \infty$.
\end{assumption}

\begin{assumption}[Smooth variational class]
\label{assumption:smoothness}
The parameter domain is $\mathcal{K}\times\Omega$, where
$\mathcal{K}\subseteq\{\theta:\|\theta\|\le B_\theta\}$ (Assumption~\ref{assumption:relaxedboundedweights}) and
$\Omega\subseteq\{\omega:\|\omega\|\le B_\omega\}$ are convex and compact
(e.g., Euclidean balls).
For every $s,s'\in\mathcal{S}$, the map $\omega\mapsto q(s'\mid s;\omega)$ is twice
continuously differentiable on an open neighborhood of $\Omega$, and there
exist finite constants $C_\omega, C'_\omega$ such that, for all
$s,s'\in\mathcal{S}$ and $\omega\in\Omega$,
\begin{equation}
\big\|\nabla_\omega \log q(s'\mid s;\omega)\big\| \le C_\omega,
\qquad
\big\|\nabla^2_\omega \log q(s'\mid s;\omega)\big\| \le C'_\omega .
\label{eq:logq-deriv-bounds}
\end{equation}
This extends Assumption~\ref{assumption:vardist1}, which bounds only first
derivatives.
\end{assumption}

\begin{remark}[Softmax constants]
\label{remark:softmaxconstants}
Under the softmax parameterization of Assumption~\ref{assumption:vardist1},
$\log q(s'\mid s;\omega) = z_\omega(s,s') -
\log\sum_{\tilde s} e^{z_\omega(s,\tilde s)}$. Suppose the logits satisfy
$\|\nabla_\omega z_\omega(s,\tilde s)\|\le L_z$ and
$\|\nabla^2_\omega z_\omega(s,\tilde s)\|\le L'_z$ for all $s,\tilde s$ and
$\omega\in\Omega$. Both bounds are enforceable by spectral normalization of
the logit network \citep{miyato2018spectral, khromov2024some}. Write
$p_{\tilde s} := q(\tilde s\mid s;\omega)$ and
$g := \mathrm{LSE}(z_\omega(s,\cdot))$. Then
$\nabla_\omega \log q = \nabla_\omega z_\omega(s,s') -
\sum_{\tilde s} p_{\tilde s}\nabla_\omega z_\omega(s,\tilde s)$, so
$\|\nabla_\omega\log q\| \le L_z + L_z = 2L_z$. Thus
Equation~\ref{eq:logq-deriv-bounds} holds with $C_\omega = 2L_z$, consistent
with the Lipschitz constant $L_q = 2L_z$ derived in
Equation~\ref{ineq:logqbounds}. For the second derivative, the chain rule
gives
$\nabla^2_\omega g = \sum_{\tilde s} p_{\tilde s}\,
\nabla^2_\omega z_\omega(s,\tilde s) + J^\top\big(\mathrm{diag}(p) -
pp^\top\big)J$, where $J$ stacks the rows
$\nabla_\omega z_\omega(s,\tilde s)^\top$. Since
$0 \preceq J^\top(\mathrm{diag}(p)-pp^\top)J
= \sum_{\tilde s} p_{\tilde s}\,\nabla z_{\tilde s}\nabla z_{\tilde s}^\top
- \big(\textstyle\sum_{\tilde s} p_{\tilde s}\nabla z_{\tilde s}\big)
  \big(\textstyle\sum_{\tilde s} p_{\tilde s}\nabla z_{\tilde s}\big)^\top
\preceq L_z^2 I$,
we obtain $\|\nabla^2_\omega g\| \le L'_z + L_z^2$ and therefore
$\|\nabla^2_\omega \log q\| \le \|\nabla^2_\omega z_\omega(s,s')\| +
\|\nabla^2_\omega g\| \le 2L'_z + L_z^2 =: C'_\omega$.
\end{remark}

Recall the TD target
$y_t(\theta,\omega) = \beta r(s_t,a_t) +
\gamma\big(\mathbb{E}_{q}\big[\phi^\top(\bar{s}_{t+1})\theta\big] +
\log q(s_{t+1}\mid s_t;\omega)\big)$ with
$\bar{s}_{t+1}\sim q(\cdot\mid s_t;\omega)$, and define the residual
\begin{equation}
\delta(\theta,\omega) := y_t(\theta,\omega) - \phi^\top(s_t)\theta
= v(\omega)^\top\theta + c(\omega),
\label{eq:residual-affine}
\end{equation}
where, for the sampled transition $(s_t, a_t, s_{t+1})$,
\begin{equation*}
v(\omega) := \gamma\,\mathbb{E}_{q_\omega}\big[\phi(\bar s_{t+1})\big] - \phi(s_t)
\in\mathbb{R}^k,
\qquad
c(\omega) := \beta r(s_t,a_t) + \gamma \log q(s_{t+1}\mid s_t;\omega)
\in\mathbb{R}.
\end{equation*}
That is, $\delta$ is \emph{affine in $\theta$} with $\omega$-dependent
coefficient $v(\omega)$, and depends on $\omega$ nonlinearly through both
$\mathbb{E}_{q_\omega}[\phi^\top\theta]$ and $\log q$. It is precisely this
nonlinear $\omega$-dependence that invalidates a direct joint-convexity
argument for the squared loss. The following lemma replaces it with a
Hessian perturbation bound. With
$R_{\max}:=\sup_{s,a}|r(s,a)|$ and, with slight abuse of notation, the
scalar $B_\phi := \|B_\phi^\top\|$
(Assumption~\ref{assumption:relaxedlinfuncapp}), define
\begin{align}
D_{\max} &:= |\beta| R_{\max} + (1+\gamma) B_\phi B_\theta
             + \gamma\big(C_q + L_q B_\omega\big), \nonumber\\
H_{\max} &:= \gamma\Big[\,C'_\omega + B_\phi C_\omega
             + B_\phi B_\theta \big(C_\omega^2 + C'_\omega\big)\Big],
\qquad
L_H := D_{\max} H_{\max},
\label{eq:LHdef-supp}
\end{align}
where $C_q, L_q$ are the $\log q$ constants of
Equation~\ref{ineq:logqbounds}.

\begin{lemma}[Strong convexity of $F$ for $\lambda > L_H$]
\label{lemma:Fstrconvex-revised}
Let Assumptions~\ref{assumption:vardist1}--\ref{assumption:relaxedlinfuncapp}
and \ref{assumption:smoothness} hold, and let
$\ell(\theta,\omega):=\tfrac12\,\mathbb{E}\big[\delta(\theta,\omega)^2\big]$
be the unregularized loss, the expectation taken over $s_t\sim\mu$ (the
stationary distribution of the chain induced by $\pi$,
Assumption~\ref{assumption:mdpboundedergodic}), $a_t=\pi(s_t)$,
$s_{t+1}\sim P(\cdot\mid s_t,a_t)$. Then:
\begin{enumerate}
    \item\label{claim:strong-convexity-hessian} $\ell\in C^2$ on a neighborhood of $\mathcal{K}\times\Omega$ and
    $\nabla^2\ell(\theta,\omega) \succeq -L_H I$ for all
    $(\theta,\omega)\in\mathcal{K}\times\Omega$.
    \item\label{claim:strong-convexity-inequality} If $\lambda > L_H$, then
    $F = \ell + \tfrac{\lambda}{2}(\|\theta\|^2+\|\omega\|^2)$ is
    $\mu_F$-strongly convex on $\mathcal{K}\times\Omega$ with modulus
    $\mu_F := \lambda - L_H > 0$. In particular, for all
    $x, y \in \mathcal{K}\times\Omega$,
    $F(y) \ge F(x) + \langle\nabla F(x),\, y - x\rangle
    + \tfrac{\mu_F}{2}\|y-x\|^2$.
    \item\label{claim:strong-convexity-minimizer} $F$ attains a unique minimizer
    $(\theta^*,\omega^*)\in\mathcal{K}\times\Omega$, and for all
    $x\in\mathcal{K}\times\Omega$,
    $\langle\nabla F(x),\, x - x^*\rangle \ge \mu_F\|x-x^*\|^2$ with
    $x^* = (\theta^*,\omega^*)^\top$.
    \item\label{claim:strong-convexity-theta} For every fixed $\omega\in\Omega$, the $\theta$-block Hessian
    satisfies
    $\nabla^2_{\theta\theta} F(\theta,\omega)
    = \mathbb{E}\big[v(\omega)v(\omega)^\top\big] + \lambda I
    \succeq \lambda I$, with no condition on $\lambda$.
\end{enumerate}
\end{lemma}

\begin{proof}
\textbf{Step 0 (Regularity and differentiation under the expectation).}
For finite $\mathcal{S}$, the expectation defining $\ell$ is a finite
weighted sum over transitions $(s_t, a_t, s_{t+1})$ with weights
$\mu(s_t)P(s_{t+1}\mid s_t, a_t)$, and
$\mathbb{E}_{q_\omega}[\phi^\top(\bar s_{t+1})\theta] =
\sum_{\bar s} q(\bar s\mid s_t;\omega)\,\phi^\top(\bar s)\theta$ is likewise
a finite sum, so no stochastic estimator enters the population objective.
By Assumption~\ref{assumption:smoothness}, each summand is $C^2$ in
$(\theta,\omega)$ on a neighborhood of $\mathcal{K}\times\Omega$. Hence
$\ell\in C^2$ and derivatives pass through the (finite) expectation exactly.
(For general state spaces, the interchange follows from the dominated
convergence theorem, since all first and second derivatives appearing below
are uniformly bounded by the constants of
Assumptions~\ref{assumption:relaxedlinfuncapp} and
\ref{assumption:smoothness}.)

\textbf{Step 1 (Hessian decomposition).}
Writing $\nabla, \nabla^2$ for derivatives in the joint variable
$x=(\theta,\omega)^\top\in\mathbb{R}^{k+d_\omega}$, the product rule gives
\begin{equation}
\nabla^2 \ell(\theta,\omega)
= \mathbb{E}\big[\nabla\delta\,\nabla\delta^\top\big]
+ \mathbb{E}\big[\delta\,\nabla^2\delta\big].
\label{eq:hess-decomp}
\end{equation}
The first term is positive semidefinite: for any
$a\in\mathbb{R}^{k+d_\omega}$,
$a^\top\mathbb{E}[\nabla\delta\nabla\delta^\top]a
= \mathbb{E}[(\nabla\delta^\top a)^2]\ge 0$. Only the second term can
contribute negative curvature. We bound its operator norm below.

\textbf{Step 2 (Uniform residual bound $D_{\max}$).}
For all $(\theta,\omega)\in\mathcal{K}\times\Omega$ and all transitions,
each term of $\delta$ in Equation~\ref{eq:residual-affine} is bounded:
$|\beta r(s_t,a_t)|\le|\beta|R_{\max}$. Also,
$\gamma\,|\mathbb{E}_{q_\omega}[\phi^\top\theta]| \le
\gamma\,\mathbb{E}_{q_\omega}[\|\phi\|\|\theta\|] \le \gamma B_\phi B_\theta$
(Cauchy--Schwarz, then Assumption~\ref{assumption:relaxedlinfuncapp} and
$\|\theta\|\le B_\theta$). By Equation~\ref{ineq:logqbounds},
$\gamma\,|\log q(s_{t+1}\mid s_t;\omega)| \le \gamma(C_q + L_q\|\omega\|)
\le \gamma(C_q + L_q B_\omega)$. Finally,
$|\phi^\top(s_t)\theta|\le B_\phi B_\theta$. By the triangle inequality,
\begin{equation}
\sup_{(\theta,\omega)\in\mathcal{K}\times\Omega}\ \sup_{(s_t,a_t,s_{t+1})}
|\delta(\theta,\omega)|
\;\le\; |\beta|R_{\max} + (1+\gamma)B_\phi B_\theta
+ \gamma(C_q + L_q B_\omega) \;=\; D_{\max}.
\label{eq:Dmax-bound}
\end{equation}

\textbf{Step 3 (Uniform curvature bound $H_{\max}$ on $\nabla^2\delta$).}
Fix a transition and write $\nabla^2\delta$ in block form with respect to
$(\theta,\omega)$:
\begin{equation*}
\nabla^2\delta =
\begin{pmatrix}
0 & A \\ A^\top & B
\end{pmatrix},
\qquad
A := \nabla^2_{\theta\omega}\delta \in \mathbb{R}^{k\times d_\omega},
\quad
B := \nabla^2_{\omega\omega}\delta \in \mathbb{R}^{d_\omega\times d_\omega}.
\end{equation*}
\emph{(a) $\theta\theta$-block.} By
Equation~\ref{eq:residual-affine}, $\delta$ is affine in $\theta$, so
$\nabla^2_{\theta\theta}\delta = 0$.

\emph{(b) Cross block.} Since $\nabla_\theta\delta = v(\omega)$ and
$\nabla_\omega q(\bar s\mid s_t;\omega) = q(\bar s\mid s_t;\omega)
\nabla_\omega \log q(\bar s\mid s_t;\omega)$,
\begin{equation*}
A = \nabla_\omega v(\omega)
= \gamma \sum_{\bar s} \phi(\bar s)\,
\nabla_\omega q(\bar s\mid s_t;\omega)^\top
= \gamma\,\mathbb{E}_{q_\omega}\big[\phi(\bar s)\,
\nabla_\omega\log q(\bar s\mid s_t;\omega)^\top\big],
\end{equation*}
whence $\|A\| \le \gamma\,\mathbb{E}_{q_\omega}\big[\|\phi(\bar s)\|
\|\nabla_\omega\log q(\bar s\mid s_t;\omega)\|\big] \le \gamma B_\phi
C_\omega$ (Jensen's inequality applied to the operator norm, then
Assumptions~\ref{assumption:relaxedlinfuncapp} and
\ref{assumption:smoothness}).

\emph{(c) $\omega\omega$-block.} From
$\nabla^2_\omega q = q\,(\nabla_\omega\log q\,\nabla_\omega\log q^\top +
\nabla^2_\omega\log q)$ (chain rule applied to $q = e^{\log q}$),
\begin{equation*}
B = \gamma\Big(\underbrace{\sum_{\bar s} \phi^\top(\bar s)\theta\,
\nabla^2_\omega q(\bar s\mid s_t;\omega)}_{=\;\nabla^2_\omega
\mathbb{E}_{q_\omega}[\phi^\top\theta]}
\;+\; \nabla^2_\omega \log q(s_{t+1}\mid s_t;\omega)\Big),
\end{equation*}
so, using $|\phi^\top(\bar s)\theta|\le B_\phi B_\theta$ and
Equation~\ref{eq:logq-deriv-bounds},
\begin{equation*}
\|B\| \le \gamma\Big(\mathbb{E}_{q_\omega}\big[|\phi^\top\theta|\,
\big(\|\nabla_\omega\log q\|^2 + \|\nabla^2_\omega\log q\|\big)\big]
+ C'_\omega\Big)
\le \gamma\Big(B_\phi B_\theta\big(C_\omega^2 + C'_\omega\big)
+ C'_\omega\Big).
\end{equation*}
\emph{(d) Assembly.} Decompose
$\nabla^2\delta =
\left(\begin{smallmatrix} 0 & A \\ A^\top & 0\end{smallmatrix}\right)
+ \left(\begin{smallmatrix} 0 & 0 \\ 0 & B\end{smallmatrix}\right)$. The first
(anti-diagonal) matrix is symmetric with eigenvalues
$\{\pm\sigma_i(A)\}$, the signed singular values of $A$, so its operator
norm equals $\|A\|$. The second has operator norm $\|B\|$. By the triangle
inequality for the operator norm,
\begin{equation}
\big\|\nabla^2\delta\big\| \le \|A\| + \|B\|
\le \gamma\Big[\,B_\phi C_\omega + B_\phi B_\theta(C_\omega^2 + C'_\omega)
+ C'_\omega\Big] = H_{\max}.
\label{eq:Hmax-bound}
\end{equation}

\textbf{Step 4 (Operator-norm bound on the curvature term).}
Let $M := \mathbb{E}[\delta\,\nabla^2\delta]$. The matrix $M$ is symmetric as an
expectation of scalar multiples of symmetric matrices. By Jensen's
inequality (convexity of the operator norm) and
Equations~\ref{eq:Dmax-bound}--\ref{eq:Hmax-bound},
\begin{equation*}
\|M\| \le \mathbb{E}\big[\,|\delta|\;\big\|\nabla^2\delta\big\|\,\big]
\le D_{\max} H_{\max} = L_H .
\end{equation*}
A symmetric matrix with $\|M\|\le L_H$ satisfies $M \succeq -L_H I$.
Combining with Equation~\ref{eq:hess-decomp} and the positive
semidefiniteness of the first term proves claim~(\ref{claim:strong-convexity-hessian}):
$\nabla^2\ell \succeq 0 - L_H I = -L_H I$ on $\mathcal{K}\times\Omega$.

\textbf{Step 5 (Strong convexity of $F$).}
The regularizer contributes Hessian $\lambda I$ exactly, so for
$\lambda > L_H$,
\begin{equation*}
\nabla^2 F(\theta,\omega) = \nabla^2\ell(\theta,\omega) + \lambda I
\succeq (\lambda - L_H) I = \mu_F I
\qquad \text{on } \mathcal{K}\times\Omega .
\end{equation*}
Since $\mathcal{K}\times\Omega$ is convex
(Assumption~\ref{assumption:smoothness}), the pointwise Hessian bound
integrates to strong convexity: for $x, y\in\mathcal{K}\times\Omega$ let
$g(t):=F(x + t(y-x))$ for $t\in[0,1]$. Then
$g''(t) = (y-x)^\top \nabla^2F\big(x+t(y-x)\big)(y-x) \ge \mu_F\|y-x\|^2$
(the segment stays in $\mathcal{K}\times\Omega$ by convexity), and the
second-order Taylor identity
$g(1) = g(0) + g'(0) + \int_0^1 (1-t)\,g''(t)\,dt$ yields
$F(y) \ge F(x) + \langle\nabla F(x), y-x\rangle
+ \tfrac{\mu_F}{2}\|y-x\|^2$. This proves claim~(\ref{claim:strong-convexity-inequality}).

\textbf{Step 6 (Existence, uniqueness, and the gradient inequality).}
$F$ is continuous on the nonempty compact set $\mathcal{K}\times\Omega$, so
a minimizer $x^*$ exists (Weierstrass). If $x_1^*\neq x_2^*$ were two
minimizers with common value $F^*$, set
$m := \tfrac12(x_1^*+x_2^*) \in \mathcal{K}\times\Omega$ (convexity).
Applying claim~(\ref{claim:strong-convexity-inequality}) with base point $x = m$, once with $y = x_1^*$ and once
with $y = x_2^*$, and averaging the two inequalities --- the gradient terms
cancel since $x_1^* - m = -(x_2^* - m)$, and
$\|x_i^* - m\| = \tfrac12\|x_1^*-x_2^*\|$ --- yields
$F^* \ge F(m) + \tfrac{\mu_F}{8}\|x_1^*-x_2^*\|^2 > F(m)$, contradicting
the minimality of $F^*$. The minimizer is therefore unique. For the gradient inequality, first-order
optimality of $x^*$ over the convex set $\mathcal{K}\times\Omega$ gives
$\langle\nabla F(x^*),\, x - x^*\rangle \ge 0$ for all
$x\in\mathcal{K}\times\Omega$ (including when the minimizer lies on the boundary). If $x^*$ is
interior, $\nabla F(x^*)=0$. Applying claim~(\ref{claim:strong-convexity-inequality}) twice,
\begin{equation*}
\langle \nabla F(x),\, x - x^*\rangle
\;\ge\; F(x)-F(x^*) + \tfrac{\mu_F}{2}\|x-x^*\|^2
\;\ge\; \langle\nabla F(x^*), x-x^*\rangle + \mu_F\|x-x^*\|^2
\;\ge\; \mu_F\|x-x^*\|^2,
\end{equation*}
proving claim~(\ref{claim:strong-convexity-minimizer}).

\textbf{Step 7 ($\theta$-block).}
By Equation~\ref{eq:residual-affine}, for fixed $\omega$ the map
$\theta\mapsto\delta$ is affine with gradient $v(\omega)$, so
$\nabla^2_{\theta\theta}\ell = \mathbb{E}[v(\omega)v(\omega)^\top]
\succeq 0$ and
$\nabla^2_{\theta\theta}F = \mathbb{E}[v(\omega)v(\omega)^\top] + \lambda I
\succeq \lambda I$, proving claim~(\ref{claim:strong-convexity-theta}).

\textbf{Degenerate and exact cases.}
If $q$ is held fixed (not learned), then $C_\omega = C'_\omega = 0$, hence
$H_{\max}=0$, $L_H = 0$ and $\mu_F = \lambda$: the lemma recovers exactly
the classical $\lambda$-strong convexity of ridge-regularized least squares,
with no slack. The sign of $\beta$ is immaterial ($\beta$ enters only
through $|\beta|$). The $\beta=0$ algorithm uses the separate risk-neutral update.
The bound $L_H$ uses worst-case suprema and a triangle inequality over Hessian blocks, which can make the sufficient ridge threshold conservative.
\end{proof}

\begin{remark}[Blockwise structure, and why $\lambda > L_H$ is conservative]
\label{remark:blockwise}
The joint condition $\lambda > L_H$ suffices for the analysis. The
optimization problem has considerably more benign blockwise structure.
\emph{(i) $\theta$-subproblem.} For fixed $\omega$, $\delta$ is affine in
$\theta$ with coefficient vector
$v(\omega) = \gamma\,\mathbb{E}_{q_\omega}[\phi(\bar s_{t+1})] - \phi(s_t)$
(Equation~\ref{eq:residual-affine}), so $F(\cdot\,,\omega)$ is a
ridge-regularized convex quadratic with Hessian
$\mathbb{E}[v(\omega)v(\omega)^\top]+\lambda I \succeq \lambda I$:
$\lambda$-strongly convex for \emph{every} $\lambda>0$, with no condition
relating $\lambda$ to $L_H$.
\emph{(ii) $\omega$-subproblem.} If the logits are linear in $\omega$
(log-linear, i.e., exponential-family $q$), then
$-\log q(s'\mid s;\omega)$ is convex in $\omega$ (an affine function plus a
log-sum-exp of affine functions), so the regularized cross-entropy
objective $\mathbb{E}[-\log q(s_{t+1}\mid s_t;\omega)] +
\tfrac{\lambda}{2}\|\omega\|^2$ --- which is precisely what the
$\omega$-update of Algorithm~\ref{alg:rrstd} descends, with the
algorithm's $\lambda_\omega$ in the role of $\lambda$ --- is
$\lambda$-strongly convex, again with no condition on $\lambda$. In the
\emph{joint} squared-TD objective, by contrast, $\omega$ enters the residual
$\delta$ both through the tilting term
$\mathbb{E}_{q_\omega}[\phi^\top\theta]$ and through $\gamma\log q$, and
squaring destroys convexity: the curvature term
$\mathbb{E}[\delta\,\nabla^2\delta]$ in Equation~\ref{eq:hess-decomp}
carries no sign in general. Indeed, since
$\nabla^2_{\theta\theta}\delta = 0$, the matrix $\nabla^2\delta$ is itself
indefinite whenever its cross block $A = \nabla^2_{\theta\omega}\delta$ is
nonzero (evaluate the quadratic form at $(tAb,\, b)$ with $Ab\neq 0$: it
equals $2t\|Ab\|^2 + b^\top B b$, which takes both signs as
$t\to\pm\infty$). The averaged matrix
$\mathbb{E}[\delta\,\nabla^2\delta]$ inherits the zero $\theta\theta$-block,
and a symmetric matrix with a vanishing diagonal block is positive
semidefinite only if the corresponding off-diagonal block vanishes as well
(with cross block $C\neq 0$, evaluate the quadratic form at $(-tCv,\,v)$
for $Cv\neq 0$ and let $t\to+\infty$). Hence
$\mathbb{E}[\delta\,\nabla^2\delta]$ is indefinite whenever its averaged
cross block $\mathbb{E}[\delta\,\nabla^2_{\theta\omega}\delta]$ is nonzero,
whatever the sign of $\delta$. The nonconvex coupling is thus confined to the residual's
$\omega$-dependence --- dominantly the term
$\mathbb{E}_{q_\omega}[\phi^\top\theta]$, which couples $\theta$ and
$\omega$ --- and vanishes when $q$ is not learned
($C_\omega=C'_\omega=0$, so $L_H=0$).
\emph{(iii) Two-timescale alternative.} Running the $\omega$-update on a
fast timescale (regularized maximum likelihood, as in
Algorithm~\ref{alg:rrstd}) and the $\theta$-update on a slow timescale
yields, by (i)--(ii), two $\lambda$-strongly convex subproblems without any
lower bound on $\lambda$. A two-timescale stochastic-approximation analysis
in the sense of \citet{borkar2008stochastic} could then replace the joint
condition $\lambda > L_H$. We state this as an alternative and do not claim
it here. The single-timescale analysis in Theorem~\ref{theorem:finitesamplerrorbound} uses the conservative joint condition.
\end{remark}

\begin{assumption}[Continuous-state variational family]
\label{assumption:contq}
Let the state space $\mathcal{S}\subset\mathbb{R}^n$ be compact with diameter $D_S:=\sup_{s,s'\in\mathcal{S}}\|s-s'\|_2<\infty$, and let the variational parameters range over a compact convex set $\Omega\subset\mathbb{R}^{d_\omega}$. The variational family is conditionally Gaussian, $q(\cdot\mid s;\omega)=\mathcal{N}\big(\mu_\omega(s),\Sigma_\omega(s)\big)$, where the subscripted map $\mu_\omega:\mathcal{S}\to\mathbb{R}^n$ denotes the Gaussian mean (not to be confused with the stationary distribution $\mu$), and:
\begin{enumerate}
    \item[(i)] \emph{(Noise floor and ceiling.)} There exist $0<\sigma_{\min}\le\sigma_{\max}<\infty$ with $\sigma_{\min}^2 I_n\preceq\Sigma_\omega(s)\preceq\sigma_{\max}^2 I_n$ for all $(s,\omega)\in\mathcal{S}\times\Omega$.
    \item[(ii)] \emph{(Range of the mean.)} A smooth squashing of the network output onto $\mathcal{S}$ enforces $\mu_\omega(s)\in\mathcal{S}$ for all $(s,\omega)$. If instead only $\sup_{s,\omega}\operatorname{dist}(\mu_\omega(s),\mathcal{S})\le d_0<\infty$, every statement below holds with $D_S$ replaced by $D_S+d_0$.
    \item[(iii)] \emph{(Smoothness in $\omega$.)} $\omega\mapsto\mu_\omega(s)$ and $\omega\mapsto\Sigma_\omega(s)$ are twice continuously differentiable, uniformly in $s$: for every unit vector $v\in\mathbb{R}^{d_\omega}$ and all $(s,\omega)\in\mathcal{S}\times\Omega$,
    \[
        \|D_v\mu_\omega(s)\|_2\le L_\mu,\quad
        \|D_v\Sigma_\omega(s)\|_{\mathrm{op}}\le L_\Sigma,\quad
        \|D^2_v\mu_\omega(s)\|_2\le L'_\mu,\quad
        \|D^2_v\Sigma_\omega(s)\|_{\mathrm{op}}\le L'_\Sigma,
    \]
    where $D_v$ ($D^2_v$) is the first (second) directional derivative in $\omega$ along $v$. In practice, these bounds are enforced by spectral normalization of every layer \citep{miyato2018spectral, khromov2024some}, exactly as in the discrete case of Assumption~\ref{assumption:vardist1}.
    \item[(iv)] \emph{(Features off $\mathcal{S}$.)} The feature map $\phi$ extends to a continuously differentiable map on $\mathbb{R}^n$ with $\|\phi(x)\|\le\|B_\phi^\top\|$ and $\|\nabla_x\phi(x)\|_{\mathrm{op}}\le L_\phi$ for all $x\in\mathbb{R}^n$ (satisfied, e.g., by radial-basis or random Fourier features). This strengthens Assumption~\ref{assumption:relaxedlinfuncapp} to all of $\mathbb{R}^n$ and is needed because a Gaussian sample $\bar s_{t+1}$ may fall outside $\mathcal{S}$.
    \item[(v)] \emph{(Dominating dynamics.)} The transition kernel admits a density with a Gaussian-tail lower bound: there exist $c_0>0$, $c_1\ge0$, and a measurable anchor $a:\mathcal{S}\to\mathcal{S}$ with $p(x\mid s)\ge c_0\,e^{-c_1\|x-a(s)\|_2^2}$ for all $x\in\mathbb{R}^n$. Otherwise, $p(\cdot\mid s)$ is replaced throughout the analysis by its mollification $p_{\sigma_0}(\cdot\mid s)=p(\cdot\mid s)*\mathcal{N}(0,\sigma_0^2 I_n)$ with $0<\sigma_0\le\sigma_{\min}$, which satisfies the tail bound with $c_0=(2\pi\sigma_0^2)^{-n/2}e^{-D_S^2/\sigma_0^2}$ and $c_1=\sigma_0^{-2}$ (Remark~\ref{remark:contq-entropy}). In either case
    \[
        \mathcal{D}_{\mathrm{KL}}\big(q(\cdot\mid s;\omega)\,\|\,p(\cdot\mid s)\big)
        \;\le\;
        -\tfrac{n}{2}\log\!\big(2\pi e\,\sigma_{\min}^2\big)-\log c_0+c_1\big(\sqrt{n}\,\sigma_{\max}+D_S\big)^2
        \;<\;\infty ,
    \]
    so the Donsker--Varadhan lower bound \citep{donsker1976asymptotic} is non-vacuous for every member of the family. By (i), $\mathbb{E}_q[\log q]=-\tfrac12\log\det(2\pi e\,\Sigma_\omega(s))\le-\tfrac{n}{2}\log(2\pi e\,\sigma_{\min}^2)$. The tail bound gives $-\mathbb{E}_q[\log p]\le-\log c_0+c_1\mathbb{E}_q\|x-a(s)\|^2$. In addition, $\mathbb{E}_q\|x-a(s)\|_2^2\le\big(\sqrt{\operatorname{tr}\Sigma_\omega(s)}+\|\mu_\omega(s)-a(s)\|_2\big)^2\le(\sqrt{n}\,\sigma_{\max}+D_S)^2$ by (i), (ii), and Jensen's inequality.
\end{enumerate}
The Gaussian choice is for concreteness: any conditional family with an exact closed-form log-density and a reparameterizable sampler --- finite Gaussian mixtures whose mixture weights are bounded away from zero and whose component covariances obey (i), or normalizing flows with spectrally normalized, invertible couplings \citep{papamakarios2021flows} --- satisfies Remark~\ref{remark:contq-mechanics} verbatim, with $(\sigma_{\min},\sigma_{\max},L_\mu,L_\Sigma,L'_\mu,L'_\Sigma)$ replaced by the corresponding density and log-det-Jacobian bounds.
\end{assumption}

\begin{remark}[Mechanics of the continuous-state critic]
\label{remark:contq-mechanics}
Under Assumption~\ref{assumption:contq} every quantity appearing in the Relaxed-RSTD objective $F(\theta,\omega)$ of Equation~\ref{eq:Fthetaomega} of the main text is computable from a single transition, and every constant used in Lemmas~\ref{lemma:relcasemartingale}, \ref{lemma:FhasLlipcontgrad}, and \ref{lemma:boundedstochgradnoise} has an explicit continuous-state value. We record the four ingredients.

\emph{(a) Reparameterized sampling and the pathwise gradient.}
Writing $\bar s_{t+1}=\mu_\omega(s_t)+\Sigma_\omega(s_t)^{1/2}\varepsilon$ with $\varepsilon\sim\mathcal{N}(0,I_n)$ gives $\bar s_{t+1}\sim q(\cdot\mid s_t;\omega)$ exactly, so $\phi(\bar s_{t+1})^\top\theta$ is an unbiased single-sample estimate of $\mathbb{E}_q[\phi^\top\theta]$. Because the map $\omega\mapsto\phi\big(\mu_\omega(s)+\Sigma_\omega(s)^{1/2}\varepsilon\big)^\top\theta$ is continuously differentiable with the $\varepsilon$-integrable envelope $\big(L_\mu+\tfrac{L_\Sigma}{2\sigma_{\min}}\|\varepsilon\|_2\big)L_\phi B_\theta$, where $B_\theta:=\max_{\theta\in\mathcal{K}}\|\theta\|$ (Assumption~\ref{assumption:relaxedboundedweights}), dominated convergence permits differentiation under the integral sign and yields the pathwise (reparameterization) gradient \citep{kingma2014vae}
\[
    \nabla_\omega\,\mathbb{E}_q\big[\phi(\bar s)^\top\theta\big]
    =\mathbb{E}_{\varepsilon}\Big[\big(D_\omega\mu_\omega(s)+D_\omega\Sigma_\omega(s)^{1/2}[\varepsilon]\big)^\top\nabla_x\big(\phi(x)^\top\theta\big)\big|_{x=\bar s}\Big],
\]
with the explicit bound
\[
    \big\|\nabla_\omega\,\mathbb{E}_q[\phi(\bar s)^\top\theta]\big\|
    \;\le\;
    \Big(L_\mu+\frac{\sqrt{n}\,L_\Sigma}{2\sigma_{\min}}\Big)L_\phi B_\theta ,
\]
using $\mathbb{E}\|\varepsilon\|_2\le\sqrt{n}$ and the matrix-square-root perturbation bound $\|D_v\Sigma^{1/2}\|_{\mathrm{op}}\le L_\Sigma/(2\sigma_{\min})$. The perturbation bound follows because $X:=D_v\Sigma^{1/2}$ solves the Sylvester equation $\Sigma^{1/2}X+X\Sigma^{1/2}=D_v\Sigma$ (differentiate $(\Sigma^{1/2})^2=\Sigma$), whence $X=\int_0^\infty e^{-t\Sigma^{1/2}}(D_v\Sigma)\,e^{-t\Sigma^{1/2}}\,dt$ and $\|X\|_{\mathrm{op}}\le L_\Sigma\int_0^\infty e^{-2t\sigma_{\min}}dt=L_\Sigma/(2\sigma_{\min})$, since $\lambda_{\min}(\Sigma^{1/2})\ge\sigma_{\min}$ by Assumption~\ref{assumption:contq}(i). The single-sample pathwise estimator therefore has second moment at most $2\big(L_\mu^2+\tfrac{n L_\Sigma^2}{4\sigma_{\min}^2}\big)L_\phi^2 B_\theta^2$. In the discrete case (Equation~\eqref{ineq:derivsamplestate}) the score-function (log-derivative) estimator was used because $\bar s_{t+1}$ is a discrete draw and no differentiable sampling path exists. That estimator remains valid in continuous spaces, but its variance scales with $\|\nabla_\omega\log q\|^2\le (C^{\mathrm{c}}_\omega)^2$ from \eqref{eq:contq-scorebound} below --- which grows as $\sigma_{\min}^{-4}$ --- and does not vanish for near-constant integrands, whereas the pathwise estimator's variance is controlled by the integrand's slope $L_\phi$. The pathwise form is therefore the recommended default in continuous spaces \citep{mohamed2020monte}.

\emph{(b) Closed-form log-density.}
For every observed transition $(s_t,s_{t+1})$,
\begin{equation}
\label{eq:contq-logq}
    \log q(s_{t+1}\mid s_t;\omega)
    = -\frac{n}{2}\log(2\pi)-\frac12\log\det\Sigma_\omega(s_t)-\frac12\,\xi^\top\Sigma_\omega(s_t)^{-1}\xi ,
    \qquad \xi:=s_{t+1}-\mu_\omega(s_t),
\end{equation}
which requires one Cholesky factorization per transition. The cost is $O(n^3)$, or $O(n)$ for diagonal $\Sigma_\omega$.

\emph{(c) Two-sided bound on $\log q$ (the $C_q$ analogue).}
For all $s_t,s_{t+1}\in\mathcal{S}$ and $\omega\in\Omega$,
\begin{equation}
\label{eq:contq-logq-bounds}
    -\frac{n}{2}\log\!\big(2\pi\sigma_{\max}^2\big)-\frac{D_S^2}{2\sigma_{\min}^2}
    \;\le\;
    \log q(s_{t+1}\mid s_t;\omega)
    \;\le\;
    -\frac{n}{2}\log\!\big(2\pi\sigma_{\min}^2\big).
\end{equation}
\emph{Upper bound:} the quadratic form in \eqref{eq:contq-logq} is nonnegative, and $\log\det\Sigma_\omega(s_t)\ge n\log\sigma_{\min}^2$ because every eigenvalue of $\Sigma_\omega(s_t)$ is at least $\sigma_{\min}^2$ by Assumption~\ref{assumption:contq}(i). The bound is thus governed by the noise \emph{floor} $\sigma_{\min}$ (a small floor lets the density peak at $(2\pi\sigma_{\min}^2)^{-n/2}$). It is attained exactly iff $\Sigma_\omega(s_t)=\sigma_{\min}^2 I_n$ and $s_{t+1}=\mu_\omega(s_t)$, and it is positive iff $\sigma_{\min}^2<1/(2\pi)$.
\emph{Lower bound:} $\log\det\Sigma_\omega(s_t)\le n\log\sigma_{\max}^2$ (each eigenvalue is at most $\sigma_{\max}^2$), while $\xi^\top\Sigma_\omega(s_t)^{-1}\xi\le\|\xi\|_2^2/\sigma_{\min}^2\le D_S^2/\sigma_{\min}^2$, since $\lambda_{\max}(\Sigma_\omega^{-1})\le\sigma_{\min}^{-2}$ and $\|\xi\|_2\le D_S$ by Assumption~\ref{assumption:contq}(ii) ($s_{t+1}$ and $\mu_\omega(s_t)$ both lie in $\mathcal{S}$). Note that \emph{both} extremes enter the lower bound --- $\sigma_{\max}$ through the normalizer and $\sigma_{\min}$ through $\lambda_{\max}(\Sigma^{-1})$. Both extremes are simultaneously realized only when $\sigma_{\min}=\sigma_{\max}$, in which case the bound is attained at $\|\xi\|_2=D_S$. Otherwise the bound is strict: taking $\Sigma$ with one eigenvalue $\sigma_{\min}^2$ along $\xi$ and $\sigma_{\max}^2$ elsewhere and $\|\xi\|_2=D_S$ leaves a slack of exactly $\tfrac12\log(\sigma_{\max}^2/\sigma_{\min}^2)$. Consequently the discrete-case constant $C_q$ of Equation~\eqref{ineq:logqbounds} is replaced by
\[
    \sup_{s_t,s_{t+1}\in\mathcal{S},\,\omega\in\Omega}\big|\log q(s_{t+1}\mid s_t;\omega)\big|
    \;\le\;
    C^{\mathrm{c}}_q
    :=\max\Big\{\tfrac{n}{2}\big|\log(2\pi\sigma_{\min}^2)\big|,\;
    \tfrac{n}{2}\big|\log(2\pi\sigma_{\max}^2)\big|+\tfrac{D_S^2}{2\sigma_{\min}^2}\Big\}.
\]

\emph{(d) Score and Lipschitz constants (the $C_\omega$ and $L_q$ analogues).}
Differentiating \eqref{eq:contq-logq} along a unit direction $v$ in $\omega$-space gives
\begin{equation}
\label{eq:contq-score}
    D_v\log q(s_{t+1}\mid s_t;\omega)
    = -\tfrac12\operatorname{tr}\!\big(\Sigma^{-1}D_v\Sigma\big)
    +(D_v\mu)^\top\Sigma^{-1}\xi
    +\tfrac12\,\xi^\top\Sigma^{-1}(D_v\Sigma)\Sigma^{-1}\xi ,
\end{equation}
(arguments $s_t,\omega$ suppressed), and bounding the three terms by $\tfrac{n L_\Sigma}{2\sigma_{\min}^2}$, $\tfrac{L_\mu D_S}{\sigma_{\min}^2}$, and $\tfrac{D_S^2 L_\Sigma}{2\sigma_{\min}^4}$ respectively (using $|\operatorname{tr}(AB)|\le n\|A\|_{\mathrm{op}}\|B\|_{\mathrm{op}}$, $\|\Sigma^{-1}\|_{\mathrm{op}}\le\sigma_{\min}^{-2}$, $\|\xi\|_2\le D_S$, and Assumption~\ref{assumption:contq}(iii)) yields, since $\|\nabla_\omega\log q\|=\sup_{\|v\|=1}D_v\log q$,
\begin{equation}
\label{eq:contq-scorebound}
    \big\|\nabla_\omega\log q(s_{t+1}\mid s_t;\omega)\big\|
    \;\le\;
    C^{\mathrm{c}}_\omega
    :=\frac{n L_\Sigma}{2\sigma_{\min}^2}+\frac{L_\mu D_S}{\sigma_{\min}^2}+\frac{D_S^2 L_\Sigma}{2\sigma_{\min}^4}.
\end{equation}
Because $\Omega$ is convex and compact, the mean value theorem gives that $\omega\mapsto\log q(s_{t+1}\mid s_t;\omega)$ is Lipschitz on $\Omega$ with constant $L^{\mathrm{c}}_q:=C^{\mathrm{c}}_\omega$. Hence Equation~\eqref{ineq:logqbounds} and Remark~\ref{remark:vardistf} hold verbatim with $(C_q,L_q,C_\omega)\to(C^{\mathrm{c}}_q,L^{\mathrm{c}}_q,C^{\mathrm{c}}_\omega)$, the bound of part~(a) replaces Equation~\eqref{ineq:derivsamplestate}, and the algebra of Lemma~\ref{lemma:relcasemartingale} goes through unchanged.

\emph{(e) Curvature (for Lemmas~\ref{lemma:FhasLlipcontgrad} and \ref{lemma:boundedstochgradnoise}).}
Differentiating \eqref{eq:contq-score} once more along $v$ and bounding every term with Assumption~\ref{assumption:contq}(i)--(iii) (each occurrence of $\Sigma^{-1}$ contributes $\sigma_{\min}^{-2}$, each of $D_v\Sigma$, $D^2_v\Sigma$, $D_v\mu$, $D^2_v\mu$ contributes $L_\Sigma$, $L'_\Sigma$, $L_\mu$, $L'_\mu$, and $\|\xi\|_2\le D_S$) gives $\sup_{\|v\|=1}|D^2_v\log q|\le M^{\mathrm{c}}_q$ with
\begin{equation}
\label{eq:contq-smooth}
    M^{\mathrm{c}}_q
    :=\frac{n}{2}\Big(\frac{L_\Sigma^2}{\sigma_{\min}^4}+\frac{L'_\Sigma}{\sigma_{\min}^2}\Big)
    +\frac{L'_\mu D_S+L_\mu^2}{\sigma_{\min}^2}
    +\frac{2L_\mu L_\Sigma D_S}{\sigma_{\min}^4}
    +\frac{D_S^2 L_\Sigma^2}{\sigma_{\min}^6}
    +\frac{D_S^2 L'_\Sigma}{2\sigma_{\min}^4},
\end{equation}
so $\nabla_\omega\log q$ is $M^{\mathrm{c}}_q$-Lipschitz on $\Omega$ (the Hessian of $\log q$ in $\omega$ is symmetric, so its operator norm equals $\sup_{\|v\|=1}|D^2_v\log q|$). Together with the boundedness of $\phi$, $\nabla_x\phi$, $\theta$, and $\omega$, this supplies the finite smoothness constant $L$ of Lemma~\ref{lemma:FhasLlipcontgrad} and the finite noise level $\sigma^2$ of Lemma~\ref{lemma:boundedstochgradnoise} in the continuous case.
\end{remark}

\begin{remark}[Differential entropy and the shaped reward in continuous spaces]
\label{remark:contq-entropy}
In the shaped-reward interpretation (Proposition~\ref{prop:shaped-reward}), averaging the surrogate target over $s_{t+1}\sim p(\cdot\mid s_t)$ replaces $\log q(s_{t+1}\mid s_t;\omega)$ by $\mathbb{E}_p[\log q]=-H\big(p(\cdot\mid s_t)\big)-\mathcal{D}_{\mathrm{KL}}\big(p(\cdot\mid s_t)\,\|\,q(\cdot\mid s_t;\omega)\big)$, i.e., the $\beta$-scaled reward $\beta r(s_t,a_t)$ acquires the per-step shaping $-\gamma\big(H(p(\cdot\mid s_t))+\mathcal{D}_{\mathrm{KL}}(p\|q_\omega)\big)$. In finite spaces $H\ge0$, so the shaping is always a penalty on transition stochasticity. In continuous spaces, $H(p(\cdot\mid s))$ is the \emph{differential} entropy, which can be negative --- e.g.\ $H(\mathcal{N}(0,\sigma^2 I_n))=\tfrac{n}{2}\log(2\pi e\sigma^2)<0$ whenever $\sigma^2<1/(2\pi e)$ --- so at the level of the log-value the shaping becomes a \emph{bonus} in near-deterministic regions of the state space once $q_\omega\approx p$ (there $\mathcal{D}_{\mathrm{KL}}(p\|q_\omega)\approx0$ and the shaping is $\approx-\gamma H>0$). After rescaling by $1/\beta$ to value units this rewards predictable transitions when $\beta>0$ and penalizes them when $\beta<0$. The realized shaping is nevertheless bounded on the assumed family irrespective of its sign: by Equation~\eqref{eq:contq-logq-bounds}, $\gamma\,\mathbb{E}_p[\log q]\in\big[-\gamma\big(\tfrac{n}{2}\log(2\pi\sigma_{\max}^2)+\tfrac{D_S^2}{2\sigma_{\min}^2}\big),\,\gamma\,\tfrac{n}{2}\log\tfrac{1}{2\pi\sigma_{\min}^2}\big]$, so the noise floor caps the maximal bonus at $\gamma\tfrac{n}{2}\log\tfrac{1}{2\pi\sigma_{\min}^2}$ (and if $\sigma_{\min}^2\ge1/(2\pi)$ there is no bonus at all). The deterministic limit requires care and is where the mollified view of Assumption~\ref{assumption:contq}(v) is not optional: if $p(\cdot\mid s)=\delta_{f(s)}$ then $H(p(\cdot\mid s))=-\infty$ (no Lebesgue density) and $\mathcal{D}_{\mathrm{KL}}(q\|p)=+\infty$ for every absolutely continuous $q$, so the Donsker--Varadhan bound, while formally valid, is vacuous. Replacing $p$ by $p_{\sigma_0}=p*\mathcal{N}(0,\sigma_0^2 I_n)$ with $0<\sigma_0\le\sigma_{\min}$ restores finiteness on both sides: $H\big(p_{\sigma_0}(\cdot\mid s)\big)\ge H\big(\mathcal{N}(0,\sigma_0^2 I_n)\big)=\tfrac{n}{2}\log(2\pi e\sigma_0^2)>-\infty$, because conditioning reduces entropy, $H(X+Z)\ge H(X+Z\mid X)=H(Z)$ for independent $X\sim p$, $Z\sim\mathcal{N}(0,\sigma_0^2I_n)$ \citep[Ch.~8]{cover2006elements}, and
\[
    \mathcal{D}_{\mathrm{KL}}\big(q(\cdot\mid s;\omega)\,\|\,p_{\sigma_0}(\cdot\mid s)\big)
    \;\le\;
    \frac{n}{2}\log\frac{\sigma_0^2}{e\,\sigma_{\min}^2}
    +\frac{\big(\sqrt{n}\,\sigma_{\max}+D_S\big)^2}{2\sigma_0^2}\;<\;\infty ,
\]
which follows from $\mathbb{E}_q[\log q]\le-\tfrac{n}{2}\log(2\pi e\sigma_{\min}^2)$ and the mixture lower bound $p_{\sigma_0}(x\mid s)\ge(2\pi\sigma_0^2)^{-n/2}e^{-(\|x-\mu_\omega(s)\|_2+D_S)^2/(2\sigma_0^2)}$ (every atom of $p(\cdot\mid s)$ lies in $\mathcal{S}$, within distance $D_S$ of $\mu_\omega(s)$), combined with $\mathbb{E}_q\big(\|x-\mu_\omega(s)\|_2+D_S\big)^2\le\big(\sqrt{n}\,\sigma_{\max}+D_S\big)^2$. As $\sigma_0\downarrow0$ the mollified entropy bonus grows like $\tfrac{n}{2}\log(1/\sigma_0^2)$, but the shaping the algorithm can actually realize stays capped by the $\sigma_{\min}$-bound above: the noise floor $\sigma_{\min}>0$ is exactly what keeps the surrogate objective, its gradients, and the entropy bonus finite. We stress that the mollification is an interpretive device for $H$ and $\mathcal{D}_{\mathrm{KL}}$: the implemented updates evaluate $\log q$ only at observed transitions via \eqref{eq:contq-logq}, which is finite either way, so the algorithm itself is unchanged.
\end{remark}

The parameter update rule is
\begin{equation}
    (\theta_{t+1}, \omega_{t+1})^{\top} = (\theta_{t}, \omega_{t})^{\top} - \alpha_t\nabla \tilde{F}(\theta_t, \omega_t), \nonumber
\end{equation}
where,
\begin{equation}
    \tilde{F}(\theta, \omega) = \frac{1}{2}\bigg(\beta r(s_t, a_t) + \gamma\Big(\phi^\top(\bar{s}_{t+1})\theta
     + \log q(s_{t+1}\mid s_t; \omega) \Big) - \phi^\top(s_t)\theta \bigg)^2
     + \frac{\lambda}{2}\big(\|\theta\|^2 + \|\omega\|^2\big).
    \end{equation}

This expression uses one draw $\bar s_{t+1}$ from $q_\omega$ inside the squared residual. For a fixed observed transition, averaging this square adds the variance of $\gamma\,\phi^\top(\bar s_{t+1})\theta$ to the square of the residual based on the inner expectation. The mean-gradient condition in Theorem~\ref{theorem:finitesamplerrorbound} therefore needs to be checked for the estimator used.

The gradient of $\tilde F$ is $\nabla\tilde F = \left(\nabla_\theta\tilde F, \nabla_\omega\tilde F\right)^\top$, with components
\begin{align}
\nabla_\theta\tilde F &= \Big(\beta r(s_t, a_t) + \gamma\big(\phi^\top(\bar{s}_{t+1})\theta + \log q(s_{t+1}\mid s_t;\omega)\big) - \phi^\top(s_t)\theta\Big) \notag\\
&\quad \times \big(\gamma\phi(\bar{s}_{t+1}) - \phi\big) + \lambda\theta, \label{eq:derivFtheta}\\
\nabla_\omega \tilde F &= \Big(\beta r(s_t, a_t) + \gamma\big(\phi^\top(\bar{s}_{t+1})\theta + \log q(s_{t+1}\mid s_t;\omega)\big) - \phi^\top(s_t)\theta\Big) \notag\\
&\quad \times \gamma\big(\nabla_\omega \phi^\top(\bar{s}_{t+1})\theta + \nabla_\omega\log q(s_{t+1}\mid s_t; \omega)\big) + \lambda\omega.
\label{eq:derivFomega}
\end{align}

Let $g_t = \left(\nabla_\theta\tilde{F}(\theta_t, \omega_t), \nabla_\omega\tilde{F}(\theta_t, \omega_t)\right)^\top$. The stochastic approximation update can be rewritten as
\begin{align}
\begin{bmatrix}
\theta_{t+1} \\
\omega_{t+1}
\end{bmatrix}
= \begin{bmatrix}
\theta_{t} \\
\omega_{t}
\end{bmatrix} + \alpha_t\big(h(\theta_t, \omega_t) + \mathbb{M}_{t+1} + b_{t+1} \big),
\label{eq:SGD-update}
\end{align}
where $\mathcal{F}_t = \sigma ( \{s_i, a_i, r_i, \theta_i, \omega_i\}_{i=0}^t)$ is the history of the stochastic process and
\begin{align}
    h(\theta_t, \omega_t) &= -\mathbb{E}_{s_t \sim \mu}[g_t], \nonumber \\
    \mathbb{M}_{t+1} &= -g_t + \mathbb{E}[g_t \mid \mathcal{F}_t], \nonumber \\
    b_{t+1} &= -\mathbb{E}[g_t \mid \mathcal{F}_t] + \mathbb{E}_{s_t \sim \mu}[g_t] \nonumber.
\end{align}
For asymptotic convergence of stochastic update in Equation \ref{eq:SGD-update}, we follow the framework proposed in \cite{borkar2008stochastic}. It requires the following lemma to hold.

\begin{lemma}
\label{lemma:relcasemartingale}
$\mathbb{E}[\|\mathbb{M}_{t+1}\|^2|\mathcal{F}_t] \le C(1+\|\left(\theta_t, \omega_t\right)^\top\|^2)$, where C > 0 is a constant.
\end{lemma}
\begin{proof}
\label{proof:lemma:relcasemartingale}
The left hand side can be represented as the conditional variance of $g_t$ given the history $\mathcal{F}_t$
\begin{align} \label{ineq:mub1}
    \mathbb{E}[\|\mathbb{M}_{t+1}\|^2|\mathcal{F}_t] &= \mathbb{E}[\|-g_t + \mathbb{E}[g_t|\mathcal{F}_t]\|^2|\mathcal{F}_t] \nonumber\\
&= \text{Var}[g_t|\mathcal{F}_t] = \mathbb{E}[\|g_t\|^2|\mathcal{F}_t] - \|\mathbb{E}[g_t|\mathcal{F}_t]\|^2 \nonumber\\
&\le \mathbb{E}[\|g_t\|^2|\mathcal{F}_t]
\end{align}
\\
Let's compute an upper bound for $\|g_t\|^2 = \left\|\nabla_\theta\tilde{F}(\theta_t, \omega_t)\right\|^2 + \left\|\nabla_\omega\tilde{F}(\theta_t, \omega_t)\right\|^2$.
\begin{align} \label{ineq:relgtub1}
\left\|\nabla_\theta\tilde{F}(\theta_t, \omega_t)\right\| \le & \left|\beta r(s_t, a_t) + \gamma\left(\phi^\top(\bar{s}_{t+1})\theta_t + \log q(s_{t+1}\mid s_t;\omega_t)\right) - \phi^\top(s_t)\theta_t\right| \nonumber\\
& \times\left\|\gamma\phi(\bar{s}_{t+1}) - \phi(s_t)\right\| + \lambda\|\theta_t\| \quad\text{(Triangle ineq. on Equation } \ref{eq:derivFtheta} \text{)}
\end{align}

From Assumption \ref{assumption:vardist1}, $\log q$ is bounded due to the finite state space. Since the pre-softmax logits $z_\omega(s)$ are $L_z$-Lipschitz in parameters, and the log-sum-exp function is non-expansive (Lipschitz with constant 1 in the $\ell_\infty$ norm), we have
\begin{align*}
    |\log q(s'|s;\omega_1) &- \log q(s'|s;\omega_2)| \\
    &\leq |z_{\omega_1}(s,s') - z_{\omega_2}(s,s')| + \left|\log \sum_{\tilde{s}} e^{z_{\omega_1}(s,\tilde{s})} - \log \sum_{\tilde{s}} e^{z_{\omega_2}(s,\tilde{s})}\right| \\
    &\leq L_z \|\omega_1 - \omega_2\| + \max_{\tilde{s}} |z_{\omega_1}(s,\tilde{s}) - z_{\omega_2}(s,\tilde{s})| \\
    &\leq 2L_z \|\omega_1 - \omega_2\|.
\end{align*}
Thus, $\log q$ is Lipschitz continuous with constant $L_q = 2L_z$.
Hence, we can bound $|\log q(s_{t+1}\mid s_t; \omega_t)|$ as follows
\begin{equation}
\label{ineq:logqbounds}
   |\log q(s_{t+1} \mid s_t; \omega_t)| \leq |\log q(s_{t+1} \mid s_t; 0)| + L_q \|\omega_t\| \leq C_q + L_q \|\omega_t\|,
\end{equation}
where $C_q = \max_{s_{t+1}, s_t} |\log q(s_{t+1} \mid s_t; 0)|$ is a constant and is positive. Now, we continue (\ref{ineq:relgtub1}).
\begin{align}
\left\|\nabla_\theta\tilde{F}(\theta_t, \omega_t)\right\| &\le \left(|\beta R| + \gamma\left(2\|B^\top_\phi\|\|\theta_t| + C_q + L_q \|\omega_t\|\right)\right) \quad\text{(Assumption \ref{assumption:relaxedlinfuncapp} and (\ref{ineq:logqbounds}))}\nonumber\\
&\qquad \times (\gamma+1)\|B^\top_\phi\|\|\theta_t\| + \lambda\|\theta_t\| \nonumber\\
&= C_1\|\theta_t\| + C_2 \|\theta_t\|\|\omega_t\| + C_3 \|\omega_t\|^2 \label{ineq:relgtub2}
\end{align}
where $C_1 = \left(|\beta R| + \gamma C_q\right)(\gamma+1)\|B_\phi^\top\| + \lambda$ and $C_2 = \gamma(\gamma +1) L_q\|B_\phi^\top\|$ and $C_3 = 2\gamma(\gamma+1) \|B_\phi^\top\|^2$ are all positive constants. Next, we apply Cauchy-Schwarz on  (\ref{eq:derivFomega})
\begin{align} \label{ineq:relgtub3}
\left\|\nabla_\omega\tilde{F}(\theta_t, \omega_t)\right\| \le & \left|\beta r(s_t, a_t) + \gamma\left(\phi^\top(\bar{s}_{t+1})\theta_t + \log q(s_{t+1}\mid s_t;\omega_t)\right) - \phi^\top(s_t)\theta_t\right| \nonumber\\
& \times \gamma\left\|\nabla_\omega \phi^\top(\bar{s}_{t+1})\theta_t\right\| + \gamma\left\|\nabla_\omega \log q(s_{t+1}\mid s_t;\omega-t)\right\| + \lambda\|\omega_t\|.
\end{align}

\begin{remark}
    \label{remark:vardistf}
    Let $q$ be implemented by a neural network. Following \cite{miyato2018spectral} and \cite{khromov2024some}, we ensure that $\nabla_\omega q(s;\omega)$ is bounded.
\end{remark}

From Remark \ref{remark:vardistf}, we state that $\|\nabla_\omega\log q(s_{t+1}\mid s_t;\omega_t)\|$ is upper bounded by a positive constant $C_\omega$ for all states $s \in S$. Now we wish to find a bound for $\left\|\nabla_\omega \phi^\top(\bar{s}_{t+1})\theta_t\right\|$.
\begin{equation}
    \left\|\nabla_\omega \phi^\top(\bar{s}_{t+1})\theta_t\right\| = \left\|\nabla_\omega \bar{s}_{t+1} \nabla_{\bar{s}_{t+1}}\phi^\top(\bar{s}_{t+1})\theta_t\right\| \nonumber
\end{equation}
Since $\bar{s}_{t+1}$ is sampled from a discrete distribution, the gradient does not flow through $\bar{s}_{t+1}$ directly. Instead, we treat it as a random variable and use the log-derivative trick \citep{sutton1999policy}.
\begin{align}
\label{ineq:derivsamplestate}
    \left\|\nabla_\omega \phi^\top(\bar{s}_{t+1})\theta_t\right\| &= |\phi^\top(\bar{s}_{t+1})\theta_t| |\log q(\bar s_{t+1} \mid s_t; \omega_t)| \nonumber\\
    &\le  \|B^\top_\phi\|\|\theta_t\|\left(C_q + L_q\|\omega_t\|\right) \quad \text{(From  (\ref{ineq:logqbounds}))} \nonumber \\
\end{align}
Continuing (\ref{ineq:relgtub3})
\begin{align}
\left\|\nabla_\omega\tilde{F}(\theta_t, \omega_t)\right\| &\le \left(|\beta R| + \gamma\left(2\|B^\top_\phi\|\|\theta_t\| + C_q + L_q\|\omega_t\|\right)\right) \nonumber\\
&\qquad \times \ \gamma \|B^\top_\phi\|\|\theta_t\|\left(C_q + L_q\|\omega_t\|\right) + C_\omega + \lambda\|\omega_t\| \nonumber\\
&= C_4\|\theta_t\| + C_5 \|\omega_t\| + C_6 \|\theta_t\|\|\omega_t\| + C_7\|\theta_t\|^2 + C_8\|\theta_t\|\|\omega_t\|^2 + C_\omega, \label{ineq:relgtub4}
\end{align}
where $C_4 = \gamma \|B_\phi^\top\| \left( |\beta R| C_q + \gamma C_q^2 \right)$, $C_5 = \lambda$, $C_6 = \gamma \|B_\phi^\top\| \left( |\beta R| L_q + \gamma C_q L_q + \gamma L_q C_q + 2\gamma \|B_\phi^\top\| C_q \right)$, $C_7 = 2 \gamma^2 \|B_\phi^\top\|^2 C_q$, $C_8 = \gamma^2 \|B_\phi^\top\| L_q^2$, and $C_\omega$ are all positive constants. Combining (\ref{ineq:relgtub2}) and (\ref{ineq:relgtub4}), we can upper bound $\|g_t\|^2$ as follows
\begin{align}
    \left\|g_t\right\|^2 &\le \left(C_1 + C_2\|\theta_t\| + C_3 \|\omega_t\|\right)^2 + \left(C_4 + C_5\|\theta_t\| + C_6 \|\omega_t\|+ C_7 \|\theta_t\|\|\omega_t\|\right)^2. \nonumber
\end{align}
Expanding the above and leveraging the following inequalities, $\|\theta_t\| \le 1 + \|\theta_t\|^2$, $\|\omega_t\| \le 1 + \|\omega_t\|^2$ and $\|\theta_t\|\|\omega_t\| \le \frac{1}{2}\left(\|\theta_t\|^2 + \|\omega_t\|^2\right)$ and bounding the higher order terms using the boundedness of $\|\theta_t\|$ and $\|\omega_t\|$ (Assumptions \ref{assumption:vardist1} and \ref{assumption:relaxedboundedweights}), we arrive at
\begin{equation} \label{ineq:relgtub5}
    \|g_t\|^2 \le C(1 + \|\theta_t\|^2 + \|\omega_t\|^2) \le C(1+\|\left(\theta_t, \omega_t\right)^\top\|^2)
\end{equation}
where $C > 0$ is a constant. Taking an expectation conditioned on $\mathcal{F}_t$,
\begin{align}
    \mathbb{E}[\|g_t\|^2\mid\mathcal{F}_t] &\le \mathbb{E}\left[C(1+\|\left(\theta_t, \omega_t\right)^\top\|^2)\mid\mathcal{F}_t\right] \le C(1+\|\left(\theta_t, \omega_t\right)^\top\|^2)) \nonumber\\
\therefore \mathbb{E}[\|\mathbb{M}_{t+1}\|^2\mid\mathcal{F}_t]         &\le C(1+\|\left(\theta_t, \omega_t\right)^\top\|^2)).
\end{align}
\end{proof}

Finally, \cite{borkar2008stochastic} requires the following to hold.
\begin{restatable}{lemma}{relcasemartingalediesrelaxedbiasdies}\label{lemma:relcasemartingaledies-relaxedbiasdies}
The Martingale difference $\mathbb{M}_t$ and the conditional bias $b_t$ satisfy
\begin{flalign}
&\mathbb{P}\left(\lim_{m \rightarrow \infty}\sum_{t>m} \alpha_t \mathbb{M}_{t+1} = 0\right) = 1, \quad
\mathbb{P}\left(\lim_{m\rightarrow\infty} \sum_{t>m}\alpha_t b_{t+1} = 0 \right) = 1 \nonumber
\end{flalign}
\end{restatable}
\begin{proof}
\label{proof:lemma:relcasemartingaledies-relcasebiasdies}
Taking the expectation of $\mathbb{M}_{t+1}$ over the entire history
\begin{align} \label{ineq:relmTailSum}
\mathbb{E}[-g_t + \mathbb{E}[g_t|\mathcal{F}_t]|\mathcal{F}_t] &= \mathbb{E}[-g_t|\mathcal{F}_t] + \mathbb{E}[\mathbb{E}[g_t|\mathcal{F}_t]|\mathcal{F}_t] \nonumber\\
&= \mathbb{E}[-g_t|\mathcal{F}_t] + \mathbb{E}[g_t|\mathcal{F}_t] \quad \text{(Tower Property)} \nonumber\\
&= 0 \nonumber
\end{align}
This implies that $\{\mathbb{M}_t\}$ is a Martingale Difference Sequence and thereby, $\{g_t\}$ is a Martingale Sequence. Let us now define a Martingale Sequence $\{\zeta_t\}$ as
\begin{equation}
    \zeta_t = \sum_{m=0}^{t-1}\alpha_m\mathbb{M}_{m+1}.\nonumber
\end{equation}
We have
\begin{align}
    \sum_{m\ge 0}\mathbb{E}\left[\|\zeta_{m+1} - \zeta_m\|^2|\mathcal{F}_m\right] &= \sum_{m\ge 0}\alpha_m^2\mathbb{E}\left[\|\mathbb{M}_{m+1}\|^2|\mathcal{F}_m\right] \nonumber\\
&\le K(1+\|\left(\theta_{m}, \omega_m\right)^\top\|^2) \text{ a.s.} \quad\text{(from Lemma \ref{lemma:relcasemartingale})} \nonumber\\
&< \infty \quad\text{(Assumptions \ref{assumption:vardist1} and \ref{assumption:relaxedboundedweights}).}
\end{align}
It follows from the Martingale Convergence Theorem that $\zeta_t$ converges as $t\rightarrow\infty$. This implies that the tail sum of $\{\mathbb{M}_{t}\}$, weighted by the learning rate, dies out, i.e., $\sum_{t>m} \alpha_t \mathbb{M}_{t+1} \rightarrow 0$ as $m \rightarrow \infty$.

Next, we show that the probability that the tail sum of the conditional bias, weighted by the learning rate, vanishes is $1$. Recall
\begin{equation}
    b_{t+1} = -\mathbb{E}[g_t \mid \mathcal{F}_t] + \mathbb{E}_{s \sim q}[g_t], \nonumber
\end{equation}
where $g_t = \nabla \hat{F}(\theta_t, \omega_t)$.
We assume that the Markov Chain governing the underlying state transitions is geometrically ergodic (Assumption \ref{assumption:mdpboundedergodic}). According to Levine's Markov Chain Mixing Convergence Theorem \citep{levin2017markov}, by geometric ergodicity, the Markov Chain mixes exponentially fast, $\rho \in (0, 1)$
\begin{equation}
    \label{ineq:levinemarkovmix}
    \sup_x \|P(s_{t'} \in \cdot | s_t = x) - \mu(\cdot)\|_{TV} \le C\rho^{t'-t}, \quad\text{for } t' > t,
\end{equation}
where $\mu$ is the stationary distribution of the Markov chain, C is a constant, and $\|x_1 - x_2\|_{TV}$ is the total variation distance between two probability distributions $x_1$ and $x_2$ on $X$ and is defined as
\begin{equation}
    \|x_1 - x_2\|_{TV} = \max_{A \subseteq X}|x_1(A) - x_2(A)|. \nonumber
\end{equation}
This implies that the dependence between $A_t$ (measurable w.r.t. $\mathcal{F}_t$) and $A_{t'}$ (measurable w.r.t. $\mathcal{F}_{t'}$) decays as $\rho^{t'-t}$.\\
For $t' > t$, the cross-covariance between $b_{t+1}$ and $b_{t'+1}$ satisfies the bound
\begin{equation}
    |\mathbb{E}[b_{t'+1}^\top b_{t+1}]| \leq C \rho^{t' - t} \sqrt{\mathbb{E}[\|b_{t'+1}\|^2]} \sqrt{\mathbb{E}[\|b_{t+1}\|^2]}. \nonumber
\end{equation}
From Lemma \ref{lemma:relcasemartingale}, we know that the conditional expectation of the squared norm of the $\|g_t\|^2$ is bounded and under typical stability assumptions (such as boundedness of the parameters $\theta_t, \omega_t$), the bias term $b_{t+1}$ is uniformly bounded in squared norm
\begin{equation*}
    \mathbb{E}[\|b_{t+1}\|^2] \leq C'.
\end{equation*}
Define the tail sum of the bias as
\begin{equation*}
    X_k = \sum_{t \geq k} \alpha_t b_{t+1}.
\end{equation*}
We wish to show that the variance of $X_k$ is finite
\begin{align}
    \mathbb{E}[\|X_k\|^2] &= \sum_{t \geq k} \alpha_t^2 \mathbb{E}[\|b_{t+1}\|^2] + 2 \sum_{t \geq k} \sum_{t' > t} \alpha_t \alpha_{t'} \mathbb{E}[b_{t'+1}^\top b_{t+1}]. \nonumber
\end{align}

Using the exponential mixing decay from inequality \ref{ineq:levinemarkovmix}, we can bound the cross-term
\begin{align}
    \sum_{t \geq k} \sum_{t' > t} \alpha_t \alpha_{t'} C \rho^{t' - t} 
    &\leq C \sum_{t \geq k} \alpha_t \sum_{d = 1}^\infty \alpha_{t + d} \rho^d. \nonumber
\end{align}
Since $\sum_{d=1}^\infty \rho^d < \infty$ and $\sum_t \alpha_t^2 < \infty$ (Assumption \ref{eq:robbins_monro}), it follows that
\begin{equation}
    \label{ineq:sumTailSum2}
    \mathbb{E}[\|X_k\|^2] < \infty.
\end{equation}

Leveraging Markov's Inequality, we state that for a random variable $X_k$ and for any $\epsilon > 0$
\begin{equation}\label{ineq:MarkovIneq2}
    P\left(\|X_k\|\ge\epsilon\right) \le \frac{\mathbb{E}[\|X_k\|^2]}{\epsilon^2}
\end{equation}
Take a summation over the sequence of $X_k$'s on both sides of the inequality
\begin{equation}\label{ineq:SumMarkovIneq2}
    \sum_{k=1}^\infty P\left(\|X_k\|\ge\epsilon\right) \le \frac{1}{\epsilon^2}\sum_{k=1}^\infty\mathbb{E}[\|X_k\|^2]. \nonumber
\end{equation}
Since $\epsilon^2 \in (0, \infty)$, if $\sum_{k=1}^\infty \mathbb{E}[\|X_k\|^2] < \infty$, the the left hand side of the inequality is finite and applying the Borel-Cantelli lemma, defined below, we can claim that $X_k \rightarrow 0$ almost surely, i.e., the tail sum of $\{b_t\}$, weighted by the learning rate, decays to zero, almost surely.
\\ \\
\textbf{Borel-Cantelli Lemma: } If $\sum_{k=1}^\infty P\left(\|X_k\|\geq \epsilon\right) < \infty$, then
\begin{equation} \label{ineq:BCLemma2}
    P\left(\lim\sup_{k\rightarrow\infty} |X_k| \geq \epsilon \right) = 0 \quad \forall \epsilon>0
\end{equation}
\\
This implies
\begin{equation}
    P\left(\lim_{k\rightarrow\infty} X_k = 0 \right) = 1 \nonumber
\end{equation}
\\
From inequality \ref{ineq:sumTailSum2}, we can state that $\sum_{k=1}^\infty \mathbb{E}[\|X_k\|^2] < \infty$. Therefore, from the Borel-Cantelli Lemma (Equation \ref{ineq:BCLemma2})
\begin{equation}
    P\left(\lim_{k\rightarrow\infty} \sum_{t>k}\alpha_t b_{t+1} = 0 \right) = 1.
\end{equation}
This implies that the tail sum of $\{b_t\}$, weighted by the learning rate, decays to zero, almost surely.
\end{proof}

The following lemma gives the Lyapunov drift for the mean update.

\begin{lemma}[Lyapunov drift]
\label{lemma:limitpointanalysis}
Let $x^*$ be the minimizer in Lemma~\ref{lemma:Fstrconvex-revised}. For the mean drift $h(x)=-\nabla F(x)$, the function $D(x):=\tfrac12\|x-x^*\|^2$ satisfies
\[
\langle\nabla D(x),h(x)\rangle\le-2\mu_F D(x)
\qquad\text{for all }x\in\mathcal{K}\times\Omega.
\]
\end{lemma}
\begin{proof}
Since $\nabla D(x)=x-x^*$, the left-hand side is $-\langle x-x^*,\nabla F(x)\rangle$. The gradient inequality in Lemma~\ref{lemma:Fstrconvex-revised} bounds this by $-\mu_F\|x-x^*\|^2=-2\mu_F D(x)$.
\end{proof}

\begin{proof}[Proof of almost-sure convergence in Theorem~\ref{theorem:finitesamplerrorbound}]
The stationary mean-gradient condition identifies the mean drift in Equation~\ref{eq:SGD-update} with $-\nabla F$. Lemmas~\ref{lemma:relcasemartingale} and~\ref{lemma:relcasemartingaledies-relaxedbiasdies} give the martingale second-moment bound and the vanishing weighted noise and bias tails for the stochastic approximation argument. Lemma~\ref{lemma:Fstrconvex-revised} supplies the unique minimizer, and Lemma~\ref{lemma:limitpointanalysis} establishes Lyapunov descent toward it. With iterates confined to $\mathcal{K}\times\Omega$, the stochastic approximation argument yields $x_t\to x^*$ almost surely \citep{borkar2008stochastic}.
\end{proof}

The next two lemmas give the gradient smoothness and noise constants for the finite-sample bound.

\begin{lemma}
\label{lemma:FhasLlipcontgrad}
The gradient of $F$ is $L$-Lipschitz on $\mathcal{K}\times\Omega$ with $L=L_0+\lambda$, where $L_0$ is an upper bound on the operator norm of the unregularized Hessian $\nabla^2\ell$.
\end{lemma}
\begin{proof}
\label{proof:FhasLlipcontgrad}
Let $G_{\max}$ uniformly bound $\|\nabla\delta\|$ over the parameter domain and the observed transitions. The finite state space, the bounded rewards and features, and Assumption~\ref{assumption:smoothness} give a finite $G_{\max}$. Equation~\ref{eq:hess-decomp} and the bounds on $\delta$ and $\nabla^2\delta$ give $\|\nabla^2\ell\|\le G_{\max}^2+L_H$, so one may take $L_0=G_{\max}^2+L_H$. The ridge term adds $\lambda I$ to the Hessian. Integrating the Hessian on line segments in the convex domain gives the Lipschitz bound.
\end{proof}

\begin{lemma}
\label{lemma:boundedstochgradnoise}
Under conditional unbiasedness, the stochastic gradient noise satisfies $\mathbb{E}\|g_t-\nabla F(x_t)\|^2\le\sigma^2$ uniformly in $t$. One admissible choice is $\sigma^2=C(1+B_\theta^2+B_\omega^2)$, with $C$ from Equation~\ref{ineq:relgtub5} and the parameter bounds from Assumption~\ref{assumption:smoothness}.
\end{lemma}
\begin{proof}
\label{proof:boundedstochgradnoise}
Conditional unbiasedness gives the second-moment decomposition
\[
\mathbb{E}\|g_t\|^2
=\mathbb{E}\|\nabla F(x_t)\|^2+\mathbb{E}\|g_t-\nabla F(x_t)\|^2.
\]
Equation~\ref{ineq:relgtub5} bounds the left side by $C(1+B_\theta^2+B_\omega^2)$. This bounds the noise and supplies a uniform second-moment bound even when $\nabla F(x^*)\ne0$.
\end{proof}

\begin{proof}[Proof of the finite-sample bound in Theorem~\ref{theorem:finitesamplerrorbound}]
Let $\Delta_t:=\mathbb{E}\|x_t-x^*\|^2$. Expanding the SGD update gives
\begin{equation}
\label{eq:fsanoise}
\Delta_{t+1}
=\Delta_t-2\alpha_t\mathbb{E}\langle g_t,x_t-x^*\rangle+\alpha_t^2\mathbb{E}\|g_t\|^2.
\end{equation}
Since $x_t$ is $\mathcal{F}_t$-measurable, conditional unbiasedness and Lemma~\ref{lemma:Fstrconvex-revised} imply
\[
\mathbb{E}\langle g_t,x_t-x^*\rangle
=\mathbb{E}\langle\nabla F(x_t),x_t-x^*\rangle
\ge\mu_F\Delta_t.
\]
The gradient inequality also applies to a boundary minimizer. When $\nabla F(x^*)=0$, the Lipschitz bound in Lemma~\ref{lemma:FhasLlipcontgrad} gives $\|\nabla F(x_t)\|\le L\|x_t-x^*\|$. The noise decomposition in Lemma~\ref{lemma:boundedstochgradnoise} then yields the stated second-moment bound. For $\nabla F(x^*)\ne0$, the theorem requires that bound directly. Substituting it into Equation~\ref{eq:fsanoise} gives
\[
\Delta_{t+1}\le(1-2\mu_F\alpha_t+L^2\alpha_t^2)\Delta_t+\alpha_t^2\sigma^2.
\]
With $\alpha_t=1/(\mu_F t)$, write $c=L^2/\mu_F^2$ and $b=\sigma^2/\mu_F^2$. Then
\[
\Delta_{t+1}\le (1-2/t+c/t^2)\Delta_t+b/t^2.
\]
For sufficiently large $t$, the coefficient is nonnegative and
\[
(1-2/t+c/t^2)\frac{4b}{t}+\frac{b}{t^2}\le\frac{4b}{t+1}.
\]
Indeed, the right side minus the left side is at least $3b/t^2-4bc/t^3$. Consequently the positive part of $\Delta_t-4b/t$ is bounded by its value at a fixed sufficiently large starting index times the product of the coefficients $1-2/k+c/k^2$. Using $1+u\le e^u$, the harmonic sum, and $\sum_k k^{-2}<\infty$, that product is $\mathcal{O}(t^{-2})$. Hence
\begin{equation}
\label{ineq:relsgdconvgncerate}
\mathbb{E}\|x_t-x^*\|^2
\le\frac{4\sigma^2}{\mu_F^2t}+\mathcal{O}(t^{-2}).
\end{equation}
\end{proof}

Algorithm~\ref{alg:rrstd} in Section~\ref{subsec:rstd} gives the practical critic update. The proof uses the gradient conditions in Theorem~\ref{theorem:finitesamplerrorbound}.

\subsection{Entropy-Shaped TD and Critic Error}
\label{app:ce-vs-kl}

The derivation of the tractable Bellman equation (Equation~\ref{eq:tractable-bellman} of the main text)
replaces the Donsker--Varadhan (DV) penalty
$\mathcal{D}_{\mathrm{KL}}(q_\omega\|p)$ by the log-likelihood
$\log q(s_{t+1}\mid s_t;\omega)$ of the observed transition. This subsection
makes precise what that substitution changes. (i) The substituted penalty is a
\emph{forward} divergence plus a state-conditional entropy
(Lemma~\ref{lemma:ce-decomp}). (ii) The resulting population operator is an
\emph{affine standard} Bellman operator with an entropy-shaped reward, and is a
$\gamma$-contraction (Proposition~\ref{prop:shaped-reward}). (iii) The
residual gap to the operator of Equation~\ref{eq:bellman} of the main text is exactly a
Jensen gap plus an entropy, and the Jensen gap is itself a forward KL
divergence to the optimal tilt (Lemma~\ref{lemma:dv-gap}, proving Equation~\ref{eq:jensen-gap}
of the main text). (iv) These pieces assemble into a four-term value-space
error decomposition for the Relaxed-RSTD iterates
(Corollary~\ref{cor:error-decomposition}), which accounts for the modeling
mismatch that Theorem~\ref{theorem:finitesamplerrorbound} alone does not.

\paragraph{Notation.}
Throughout this subsection the policy $\pi$ is fixed. We abbreviate
$p(\cdot\mid s):=P(\cdot\mid s,\pi(s))$ and $r_\pi(s):=r(s,\pi(s))$, with
$|r_\pi(s)|\le R_{\max}$. The state space $\mathcal S$ is finite and we adopt
the convention $0\log 0=0$. By Assumption~\ref{assumption:vardist1},
$q_\omega(\cdot\mid s)$ is a softmax distribution and therefore has full
support: $q_\omega(s'\mid s)>0$ for all $s,s'$. We write
\[
H\big(p(\cdot\mid s)\big) := -\sum_{s'} p(s'\mid s)\log p(s'\mid s)
\in \big[0,\log|\mathcal S|\big]
\]
for the Shannon entropy of the transition distribution \citep[Ch.~2]{cover2006elements},
$\mathrm{sp}(W):=\max_s W(s)-\min_s W(s)$ for the span seminorm, and
$\|q-p\|_{TV}=\tfrac12\sum_{s'}|q(s')-p(s')|$ for the total variation
distance. All operators act on
$(\mathbb{R}^{|\mathcal S|},\|\cdot\|_\infty)$, which is a complete normed
space. We write $\widetilde T^\pi_\beta$ for the discount-outside operator in Equation~\ref{eq:bellman},
$(\widetilde T^\pi_\beta W)(s) = \beta r_\pi(s)
+ \gamma\log\mathbb{E}_p[e^{W(s')}\mid s]$
(denoted $T^\pi_\beta$ in the main text), and $\bar V^\pi_\beta$ for its unique fixed
point (Proposition~\ref{prop:nested-exact}). The tilde follows the notation of Section~\ref{app:exact-surrogate-operators}. The
DV-penalized operator at parameter $\omega$ is
\[
(\widetilde T^{\mathrm{DV}}_{\omega} W)(s)
:= \beta r_\pi(s) + \gamma\Big(\mathbb{E}_{q_\omega(\cdot\mid s)}[W(s')]
- \mathcal{D}_{\mathrm{KL}}\big(q_\omega(\cdot\mid s)\,\big\|\,p(\cdot\mid s)\big)\Big),
\]
which satisfies $\widetilde T^{\mathrm{DV}}_{\omega} W \le
\widetilde T^\pi_\beta W$ pointwise by the Donsker--Varadhan inequality
\citep{donsker1976asymptotic}. One boundary case deserves care: because the
softmax $q_\omega(\cdot\mid s)$ has full support, $q_\omega(\cdot\mid s)\ll
p(\cdot\mid s)$ holds if and only if $p(\cdot\mid s)$ has full support (which
Assumption~\ref{assumption:mdpboundedergodic} does \emph{not} guarantee row by
row). If $q_\omega(\cdot\mid s)\not\ll p(\cdot\mid s)$, then
$\mathcal{D}_{\mathrm{KL}}(q_\omega\|p)=+\infty$ and
$(\widetilde T^{\mathrm{DV}}_{\omega}W)(s)=-\infty$. The DV inequality then
holds trivially in the extended reals. Every other operator below remains
finite-valued: in particular $\mathbb{E}_p[\log q_\omega(s'\mid s)]$ is finite
for every $s$ because $q_\omega>0$ everywhere and $p(\cdot\mid s)$ has finite
support.

\begin{lemma}[Cross-entropy decomposition]
\label{lemma:ce-decomp}
Fix $s\in\mathcal S$ and let $q(\cdot\mid s)$ be any distribution with
$q(s'\mid s)>0$ whenever $p(s'\mid s)>0$. Then
\begin{equation}
\label{eq:ce-decomp}
\mathbb{E}_{s'\sim p(\cdot\mid s)}\big[\log q(s'\mid s)\big]
\;=\; -\,H\big(p(\cdot\mid s)\big)
\;-\;\mathcal{D}_{\mathrm{KL}}\big(p(\cdot\mid s)\,\big\|\,q(\cdot\mid s)\big).
\end{equation}
\end{lemma}

\begin{proof}
Writing $p=p(\cdot\mid s)$, $q=q(\cdot\mid s)$ and summing over the (finite)
support of $p$,
\[
\sum_{s'} p(s')\log q(s')
= \sum_{s'} p(s')\log p(s') - \sum_{s'} p(s')\log\frac{p(s')}{q(s')}
= -H(p) - \mathcal{D}_{\mathrm{KL}}(p\,\|\,q),
\]
where every term is finite because $p$ has finite support and $q>0$ on it
\citep[Ch.~2]{cover2006elements}.
\end{proof}

Lemma~\ref{lemma:ce-decomp} makes a direction mismatch explicit. The DV
certificate that Equation~\ref{eq:tractable-bellman} of the main text budget-charges is the
\emph{reverse} divergence $\mathcal{D}_{\mathrm{KL}}(q_\omega\|p)$, whereas
the log-likelihood term the algorithm actually pays delivers, in expectation
over the observed transition, the \emph{forward} divergence
$\mathcal{D}_{\mathrm{KL}}(p\|q_\omega)$ \emph{plus} the entropy
$H(p(\cdot\mid s))$. The two divergences share the zero set
$\{q_\omega=p\}$ but have different geometry (mode-seeking versus
mass-covering, as discussed by \citealp{minka2005divergence}), and the entropy term is
independent of $\omega$: it acts as a state-dependent reward offset, not as a
penalty on $q_\omega$. The next definition and proposition turn this
observation into an exact operator statement.

\begin{definition}[Implemented population operator]
\label{def:implemented-op}
For fixed $\omega$, define $\widehat T_\omega:
\mathbb{R}^{|\mathcal S|}\to\mathbb{R}^{|\mathcal S|}$ by
\begin{equation}
\label{eq:hatT}
(\widehat T_\omega W)(s)
:= \beta r_\pi(s)
+ \gamma\Big(\mathbb{E}_{\bar s\sim q_\omega(\cdot\mid s)}\big[W(\bar s)\big]
+ \mathbb{E}_{s'\sim p(\cdot\mid s)}\big[\log q_\omega(s'\mid s)\big]\Big).
\end{equation}
Equivalently, $(\widehat T_\omega W)(s)=
\mathbb{E}_{s_{t+1}\sim p(\cdot\mid s)}\big[y_t(\theta,\omega)\,\big|\,s_t=s\big]$
whenever $W=\Phi\theta$, where $y_t$ is the TD target defining the objective
$F$ in Equation~\ref{eq:Fthetaomega} of the main text: $\widehat T_\omega$ is the conditional
mean of the target the Relaxed-RSTD critic regresses on. (The listing of
Algorithm~\ref{alg:rrstd} evaluates the $\mathbb{E}_{q_\omega}$ term of
$y_t$ at the observed next state $s_{t+1}$. The conditional means
of the two targets agree exactly when
$\mathbb{E}_{q_\omega(\cdot\mid s)}[W]=\mathbb{E}_{p(\cdot\mid s)}[W]$, in
particular at $q_\omega=p$. All statements below concern the population
objective $F$ as displayed in the main text.)
\end{definition}

\begin{proposition}[Shaped-reward equivalence]
\label{prop:shaped-reward}
Let Assumption~\ref{assumption:vardist1} hold and fix $\omega$. Write
$Q_\omega$ for the row-stochastic matrix $(Q_\omega)_{s\bar s}=
q_\omega(\bar s\mid s)$ and
$\hat r_\omega(s):=\beta r_\pi(s)
+\gamma\,\mathbb{E}_{p}[\log q_\omega(s'\mid s)\mid s]$. Then:
\begin{enumerate}
    \item[(i)] \textbf{(Exact identity.)} For every $W$ and every $s$ with
    $q_\omega(\cdot\mid s)\ll p(\cdot\mid s)$ (equivalently, with
    $p(\cdot\mid s)$ of full support, as discussed above),
    \begin{equation}
    \label{eq:hatT-vs-dv}
    (\widehat T_\omega W)(s)
    = (\widetilde T^{\mathrm{DV}}_{\omega} W)(s)
    + \gamma\Big[\mathcal{D}_{\mathrm{KL}}\big(q_\omega\|p\big)
    - \mathcal{D}_{\mathrm{KL}}\big(p\|q_\omega\big)
    - H\big(p(\cdot\mid s)\big)\Big],
    \end{equation}
    where all divergences are between the conditional distributions given $s$.
    At any $s$ with $q_\omega(\cdot\mid s)\not\ll p(\cdot\mid s)$, the
    left-hand side of \eqref{eq:hatT-vs-dv} is finite while
    $(\widetilde T^{\mathrm{DV}}_\omega W)(s)=-\infty$ and the bracket is
    $+\infty$: the identity is void there, and only $\widehat T_\omega$ is
    used in the sequel.
    \item[(ii)] \textbf{(Contraction and unique fixed point.)}
    $\widehat T_\omega W = \hat r_\omega + \gamma Q_\omega W$ is affine and
    monotone, and it is a $\gamma$-contraction:
    $\|\widehat T_\omega W_1 - \widehat T_\omega W_2\|_\infty \le
    \gamma\|W_1-W_2\|_\infty$ for all $W_1,W_2$. It therefore has the unique
    fixed point
    \begin{equation}
    \label{eq:hatV-omega}
    \widehat V_\omega = (I-\gamma Q_\omega)^{-1}\hat r_\omega,
    \end{equation}
    where $(I-\gamma Q_\omega)^{-1}=\sum_{k\ge 0}\gamma^k Q_\omega^k$ is an
    entrywise-nonnegative matrix with all row sums equal to $1/(1-\gamma)$.
    \item[(iii)] \textbf{(Entropy-shaped standard TD at the MLE limit.)} At
    the forward-KL optimum $q_\omega = p$,
    \begin{equation}
    \label{eq:shaped-operator}
    (\widehat T_p W)(s)
    = \underbrace{\big[\beta r_\pi(s) - \gamma H\big(p(\cdot\mid s)\big)\big]}_{=:\,\hat r(s)}
    + \gamma\,\mathbb{E}_{p}\big[W(s')\mid s\big] ,
    \end{equation}
    the standard risk-neutral policy-evaluation operator
    \citep{puterman2014markov, Sutton1998} for the shaped reward
    $\hat r(s)=\beta r_\pi(s)-\gamma H(p(\cdot\mid s))$. Consequently,
    $\widehat V_p = (I-\gamma P_\pi)^{-1}\big(\beta r_\pi - \gamma H_p\big)$,
    with $(P_\pi)_{ss'}=p(s'\mid s)$ and $H_p(s)=H(p(\cdot\mid s))$. If the
    dynamics are deterministic, then $H_p\equiv 0$ and moreover
    $\widehat T_p = \widetilde T^\pi_\beta$: the cross-entropy implementation
    is exact.
\end{enumerate}
\end{proposition}

\begin{proof}
(i) Under the absolute-continuity hypothesis every term below is finite
($\mathcal S$ is finite, $q_\omega>0$, and
$\mathcal{D}_{\mathrm{KL}}(q_\omega\|p)<\infty$). Subtracting the
definitions, for each such $s$,
\[
(\widehat T_\omega W)(s) - (\widetilde T^{\mathrm{DV}}_{\omega} W)(s)
= \gamma\Big(\mathbb{E}_{p}\big[\log q_\omega(s'\mid s)\big]
+ \mathcal{D}_{\mathrm{KL}}(q_\omega\|p)\Big),
\]
because the reward terms and the $\mathbb{E}_{q_\omega}[W]$ terms cancel.
Substituting Lemma~\ref{lemma:ce-decomp} for
$\mathbb{E}_{p}[\log q_\omega(s'\mid s)]$ yields \eqref{eq:hatT-vs-dv}.

(ii) Affinity is immediate from \eqref{eq:hatT}: the penalty term
$\mathbb{E}_p[\log q_\omega(s'\mid s)]$ does not depend on $W$, so
$\widehat T_\omega W=\hat r_\omega+\gamma Q_\omega W$. The reward $\hat r_\omega$ is
finite because $q_\omega>0$ (Assumption~\ref{assumption:vardist1}) and
$\mathcal S$ is finite. Monotonicity holds because $\gamma Q_\omega$ has
nonnegative entries. For the contraction, for every $s$,
\[
\big|(\widehat T_\omega W_1)(s)-(\widehat T_\omega W_2)(s)\big|
= \gamma\Big|\sum_{\bar s} q_\omega(\bar s\mid s)\big(W_1(\bar s)-W_2(\bar s)\big)\Big|
\le \gamma \sum_{\bar s} q_\omega(\bar s\mid s)\,\|W_1-W_2\|_\infty
= \gamma\|W_1-W_2\|_\infty ,
\]
using that $q_\omega(\cdot\mid s)$ is a probability distribution. Existence
and uniqueness of the fixed point follow from the Banach fixed-point theorem
on the complete space $(\mathbb{R}^{|\mathcal S|},\|\cdot\|_\infty)$
\citep{puterman2014markov}. Since $\|\gamma Q_\omega\|_{\infty\to\infty}
=\gamma<1$, the Neumann series $\sum_{k\ge0}\gamma^kQ_\omega^k$ converges and
equals $(I-\gamma Q_\omega)^{-1}$. Each summand is entrywise nonnegative with
row sums $\gamma^k$ (powers of a stochastic matrix are stochastic), giving row
sums $\sum_{k\ge0}\gamma^k=1/(1-\gamma)$.

(iii) Set $q_\omega=p$ in \eqref{eq:hatT} and apply
Lemma~\ref{lemma:ce-decomp} with $q=p$, for which
$\mathcal{D}_{\mathrm{KL}}(p\|p)=0$, so
$\mathbb{E}_p[\log p(s'\mid s)]=-H(p(\cdot\mid s))$. The closed form for
$\widehat V_p$ is \eqref{eq:hatV-omega} with $Q_\omega=P_\pi$. If
$p(\cdot\mid s)=\delta_{f(s)}$ for a deterministic transition map $f$, then
$H(p(\cdot\mid s))=-1\log 1=0$ and
$\log\mathbb{E}_p[e^{W(s')}\mid s]=W(f(s))=\mathbb{E}_p[W(s')\mid s]$, so
\eqref{eq:shaped-operator} coincides with $\widetilde T^\pi_\beta$ at every
$W$.
\end{proof}

\begin{remark}[Interpretation of the shaping]
\label{remark:shaping}
By Proposition~\ref{prop:shaped-reward}(iii), in the realizable
maximum-likelihood limit the implemented critic is \emph{exactly} standard TD
on $\hat r(s)=\beta r_\pi(s)-\gamma H(p(\cdot\mid s))$. The offset
$\gamma H(p(\cdot\mid s))\in[0,\gamma\log|\mathcal S|]$ penalizes visiting
states whose transition dynamics are noisy, independently of the sign of
$\beta$: an \emph{ambiguity-averse} shaping that vanishes exactly for
deterministic dynamics, where the implementation is exact. Two honest caveats.
First, this shaping is \emph{not} potential-based in the sense of
\citet{ng1999policy}, so a policy greedy for $\widehat V_p$ need not be greedy
for $\bar V^\pi_\beta$. The discrepancy is quantified by the relaxation term
of Corollary~\ref{cor:error-decomposition}. Second, for continuous state
spaces $H$ becomes the differential entropy of the transition density, which
can be negative. The sign statement above is specific to finite
$\mathcal S$.
\end{remark}

\begin{lemma}[DV gap as a divergence to the optimal tilt (corrected Equation~\ref{eq:jensen-gap})]
\label{lemma:dv-gap}
Fix $s$ and a function $W$, and define the exponentially tilted distribution
\begin{equation}
\label{eq:tilt}
q^*_s(s') := \frac{e^{W(s')}\,p(s'\mid s)}{\mathbb{E}_p\big[e^{W(s')}\mid s\big]} .
\end{equation}
Then, for every $q(\cdot\mid s)\ll p(\cdot\mid s)$,
\begin{equation}
\label{eq:dv-gap-identity}
\log\mathbb{E}_p\big[e^{W(s')}\mid s\big]
- \Big(\mathbb{E}_q\big[W(s')\big]
- \mathcal{D}_{\mathrm{KL}}\big(q\,\|\,p\big)\Big)
= \mathcal{D}_{\mathrm{KL}}\big(q(\cdot\mid s)\,\big\|\,q^*_s\big)\;\ge\;0 ,
\end{equation}
so the Donsker--Varadhan bound is exact if and only if $q=q^*_s$. Taking
$q=p$,
\begin{equation}
\label{eq:jensen-gap-supp}
\delta^W_\pi(s) := \log\mathbb{E}_p\big[e^{W(s')}\mid s\big]
- \mathbb{E}_p\big[W(s')\mid s\big]
= \mathcal{D}_{\mathrm{KL}}\big(p(\cdot\mid s)\,\big\|\,q^*_s\big)\;\ge\;0 ,
\end{equation}
with equality if and only if $W(s')$ is constant $p(\cdot\mid s)$-almost
surely. Equation~\ref{eq:jensen-gap} of the main text is
\eqref{eq:jensen-gap-supp} with $W=\bar V^\pi_\beta$, in which case we write
$\delta_\pi(s):=\delta^{\bar V^\pi_\beta}_\pi(s)$.
\end{lemma}

\begin{proof}
Abbreviate $Z:=\mathbb{E}_p[e^{W(s')}\mid s]\in(0,\infty)$ (finite because
$W$ is bounded on the finite $\mathcal S$). Expanding the definition of
$q^*_s$,
\[
\mathcal{D}_{\mathrm{KL}}(q\,\|\,q^*_s)
= \sum_{s'} q(s')\log\frac{q(s')\,Z}{e^{W(s')}\,p(s'\mid s)}
= \mathcal{D}_{\mathrm{KL}}(q\,\|\,p) - \mathbb{E}_q[W(s')] + \log Z ,
\]
which is \eqref{eq:dv-gap-identity}. Nonnegativity and the equality case
$q=q^*_s$ are the standard properties of the KL divergence
\citep[Ch.~2]{cover2006elements}. For $q=p$,
$\mathcal{D}_{\mathrm{KL}}(p\|p)=0$ gives \eqref{eq:jensen-gap-supp}, and
$\mathcal{D}_{\mathrm{KL}}(p\|q^*_s)=0$ if and only if $p=q^*_s$, i.e., if and
only if $e^{W(s')}=Z$ for $p(\cdot\mid s)$-almost every $s'$. By strict
monotonicity of $\exp$ this holds if and only if $W(s')$ is
$p(\cdot\mid s)$-a.s.\ constant.
\end{proof}

Before the error decomposition we record a population identity for the
objective $F$ of Equation~\ref{eq:Fthetaomega} of the main text (restated above
Lemma~\ref{lemma:Fstrconvex-revised}). This identity isolates exactly how $F$ relates to the
operator $\widehat T_\omega$.

\begin{lemma}[Population form of $F$]
\label{lemma:F-pop}
Let $V_\theta:=\Phi\theta$ and let the expectation in $F$ be over
$s_t\sim\mu$, $s_{t+1}\sim p(\cdot\mid s_t)$
(Assumption~\ref{assumption:mdpboundedergodic}). Then
\begin{equation}
\label{eq:F-pop}
F(\theta,\omega)
= \frac12\,\mathbb{E}_{s\sim\mu}\Big[\big((\widehat T_\omega V_\theta)(s)
- V_\theta(s)\big)^2\Big]
+ \frac{\gamma^2}{2}\,\mathbb{E}_{s\sim\mu}\Big[
\mathrm{Var}_{s'\sim p(\cdot\mid s)}\big(\log q_\omega(s'\mid s)\big)\Big]
+ \frac{\lambda}{2}\big(\|\theta\|^2+\|\omega\|^2\big),
\end{equation}
and the middle term does not depend on $\theta$.
\end{lemma}

\begin{proof}
Conditionally on $s_t=s$, the TD target decomposes as
$y_t = A(s)+\gamma\log q_\omega(s_{t+1}\mid s)$ with
$A(s):=\beta r_\pi(s)+\gamma\,\mathbb{E}_{q_\omega}[\phi^\top(\bar s)\theta]$
deterministic given $s$, so
$\mathbb{E}_p[y_t - V_\theta(s)\mid s]
=(\widehat T_\omega V_\theta)(s)-V_\theta(s)$ by
Definition~\ref{def:implemented-op}, and the conditional variance of
$y_t-V_\theta(s)$ is $\gamma^2\,
\mathrm{Var}_{p}(\log q_\omega(s'\mid s))$, which contains no $\theta$.
The identity
$\mathbb{E}[X^2]=(\mathbb{E}X)^2+\mathrm{Var}(X)$ applied conditionally on
$s$, followed by $\mathbb{E}_{s\sim\mu}$, gives \eqref{eq:F-pop}.
\end{proof}

\begin{corollary}[Value-space error decomposition]
\label{cor:error-decomposition}
Let Assumptions \ref{assumption:vardist1}--\ref{eq:robbins_monro} and
\ref{assumption:smoothness} hold with $\lambda > L_H$
(Equation~\ref{eq:LHdef-supp}), write $\mu_F := \lambda - L_H$
(Lemma~\ref{lemma:Fstrconvex-revised}), let
$\{(\theta_t,\omega_t)\}$ be the Relaxed-RSTD iterates, let
$(\theta^*,\omega^*)$ be the unique minimizer of $F$
(Theorem~\ref{theorem:finitesamplerrorbound}), let $\widehat V_{\omega^*}$ and
$\widehat V_p$ be the fixed points of $\widehat T_{\omega^*}$ and
$\widehat T_p$ (Proposition~\ref{prop:shaped-reward}(ii)), and let
$\bar V^\pi_\beta$ be the fixed point of $\widetilde T^\pi_\beta$. Let
$B_\phi:=\max_{s}\|\phi(s)\|_2$ (finite by
Assumption~\ref{assumption:relaxedlinfuncapp}) and let $\delta_\pi$ be the
Jensen gap of Lemma~\ref{lemma:dv-gap} evaluated at $W=\bar V^\pi_\beta$, and
define the approximation error
$\varepsilon_{\mathrm{approx}}:=\|\Phi\theta^*-\widehat V_{\omega^*}\|_\infty$.
Then, pathwise for every $t$,
\begin{align}
\big\|\Phi\theta_t - \bar V^\pi_\beta\big\|_\infty
\;\le\;{}&
\underbrace{\vphantom{\sup_s}B_\phi\,\big\|\theta_t-\theta^*\big\|_2}_{\textnormal{(i) optimization}}
\;+\;\underbrace{\vphantom{\sup_s}\big\|\Phi\theta^*-\widehat V_{\omega^*}\big\|_\infty}_{\textnormal{(ii) approximation }\varepsilon_{\mathrm{approx}}}
\nonumber\\
&+\underbrace{\frac{\gamma}{1-\gamma}\sup_{s}\Big[\big\|q_{\omega^*}(\cdot\mid s)-p(\cdot\mid s)\big\|_{TV}\,
\mathrm{sp}\big(\widehat V_p\big)
+\mathcal{D}_{\mathrm{KL}}\big(p(\cdot\mid s)\,\big\|\,q_{\omega^*}(\cdot\mid s)\big)\Big]}_{\textnormal{(iii) variational model error}}
\nonumber\\
&+\underbrace{\frac{\gamma}{1-\gamma}\sup_{s}\Big[\delta_\pi(s)+H\big(p(\cdot\mid s)\big)\Big]}_{\textnormal{(iv) relaxation bias}} .
\label{eq:err-decomp}
\end{align}
Moreover:
\begin{enumerate}
    \item[(a)] Only term (i) is random. Under the conditions of
    Theorem~\ref{theorem:finitesamplerrorbound}, for sufficiently large $t$,
    \[
    \mathbb{E}\big[B_\phi\|\theta_t-\theta^*\|_2\big]
    \;\le\; \frac{2 B_\phi\,\sigma}{\mu_F\sqrt{t}}
    + \mathcal{O}\!\left(\frac1t\right),
    \]
    and for any $\epsilon>0$,
    $\mathbb{P}\big(B_\phi\|\theta_t-\theta^*\|_2\ge\epsilon\big)\le
    B_\phi^2\big(4\sigma^2/(\mu_F^2 t)+\mathcal{O}(t^{-2})\big)/\epsilon^2$
    by Markov's inequality. Terms (ii)--(iv) are deterministic.
    \item[(b)] Term (iv) admits the exact representation
    \begin{equation}
    \label{eq:relax-exact}
    \bar V^\pi_\beta - \widehat V_p
    = (I-\gamma P_\pi)^{-1}\,\gamma\big(\delta_\pi + H_p\big)\;\ge\;0
    \quad\text{componentwise},
    \end{equation}
    whence $\widehat V_p \le \bar V^\pi_\beta$ everywhere (the implemented
    critic systematically \emph{under}-estimates the surrogate log-value) and
    \[
    \gamma\,\sup_s\big[\delta_\pi(s)+H(p(\cdot\mid s))\big]
    \;\le\;\big\|\bar V^\pi_\beta - \widehat V_p\big\|_\infty
    \;\le\;\frac{\gamma}{1-\gamma}\,\sup_s\big[\delta_\pi(s)+H(p(\cdot\mid s))\big]:
    \]
    the upper bound in (iv) is tight up to a factor $1/(1-\gamma)$, and is an
    equality whenever $\delta_\pi(s)+H(p(\cdot\mid s))$ is constant in $s$.
    \item[(c)] Degenerate cases. Term (i) tends to $0$ as $t\to\infty$ (almost surely
    by Theorem~\ref{theorem:finitesamplerrorbound}, at rate $\mathcal O(1/\sqrt t)$ in
    expectation by (a)). Term (ii) vanishes in the unregularized tabular limit
    when the fixed points are feasible
    (Remark~\ref{remark:tabular-exact}). Term (iii) is zero if
    $q_{\omega^*}=p$. For $\gamma\in(0,1)$, term (iv) is zero if and only if the
    dynamics are deterministic, since $H(p(\cdot\mid s))=0$ forces
    $p(\cdot\mid s)$ to be a point mass, which in turn forces
    $\delta_\pi(s)=0$ (at $\gamma=0$ the prefactor $\gamma/(1-\gamma)$ makes
    (iv) vanish for any dynamics).
\end{enumerate}
\end{corollary}

\begin{proof}
\textbf{Decomposition.} By the triangle inequality in
$(\mathbb{R}^{|\mathcal S|},\|\cdot\|_\infty)$,
\[
\big\|\Phi\theta_t - \bar V^\pi_\beta\big\|_\infty
\le \big\|\Phi(\theta_t-\theta^*)\big\|_\infty
+ \big\|\Phi\theta^*-\widehat V_{\omega^*}\big\|_\infty
+ \big\|\widehat V_{\omega^*}-\widehat V_p\big\|_\infty
+ \big\|\widehat V_p-\bar V^\pi_\beta\big\|_\infty .
\]
It remains to bound the first, third, and fourth summands by (i), (iii), and
(iv), respectively. The second summand is $\varepsilon_{\mathrm{approx}}$ by
definition.

\textbf{Term (i).} For every $s$, the Cauchy--Schwarz inequality gives
$|\phi(s)^\top(\theta_t-\theta^*)|\le\|\phi(s)\|_2\|\theta_t-\theta^*\|_2\le
B_\phi\|\theta_t-\theta^*\|_2$. Taking the maximum over $s$ gives the bound. For (a), since
$\theta_t-\theta^*$ is a sub-vector of
$(\theta_t,\omega_t)^\top-(\theta^*,\omega^*)^\top$,
$\|\theta_t-\theta^*\|_2\le\|(\theta_t,\omega_t)^\top-(\theta^*,\omega^*)^\top\|_2$,
and by Jensen's inequality (concavity of $x\mapsto\sqrt x$) and
Theorem~\ref{theorem:finitesamplerrorbound},
\[
\mathbb{E}\big[\|\theta_t-\theta^*\|_2\big]
\le\Big(\mathbb{E}\big[\|(\theta_t,\omega_t)^\top-(\theta^*,\omega^*)^\top\|_2^2\big]\Big)^{1/2}
\le\left(\frac{4\sigma^2}{\mu_F^2 t}+\frac{C'}{t^2}\right)^{1/2}
\le\frac{2\sigma}{\mu_F\sqrt t}+\frac{\sqrt{C'}}{t},
\]
using $\sqrt{a+b}\le\sqrt a+\sqrt b$ for $a,b\ge0$, where $C'$ is the constant
in Theorem~\ref{theorem:finitesamplerrorbound}'s $\mathcal O(t^{-2})$ term.
The Markov bound is immediate.

\textbf{Term (iii).} Both $\widehat T_{\omega^*}$ and $\widehat T_p$ are
$\gamma$-contractions (Proposition~\ref{prop:shaped-reward}(ii)). Using their
fixed-point property and the triangle inequality,
\begin{align*}
\big\|\widehat V_{\omega^*}-\widehat V_p\big\|_\infty
&= \big\|\widehat T_{\omega^*}\widehat V_{\omega^*}
- \widehat T_p\widehat V_p\big\|_\infty \\
&\le \big\|\widehat T_{\omega^*}\widehat V_{\omega^*}
- \widehat T_{\omega^*}\widehat V_p\big\|_\infty
+ \big\|(\widehat T_{\omega^*}-\widehat T_p)\widehat V_p\big\|_\infty \\
&\le \gamma\big\|\widehat V_{\omega^*}-\widehat V_p\big\|_\infty
+ \big\|(\widehat T_{\omega^*}-\widehat T_p)\widehat V_p\big\|_\infty ,
\end{align*}
and rearranging (legitimate since $1-\gamma>0$),
$\|\widehat V_{\omega^*}-\widehat V_p\|_\infty\le
(1-\gamma)^{-1}\|(\widehat T_{\omega^*}-\widehat T_p)\widehat V_p\|_\infty$.
From \eqref{eq:hatT}, for any $W$ and $s$, writing $q:=q_{\omega^*}(\cdot\mid s)$,
\begin{align*}
\big(\widehat T_{\omega^*}W-\widehat T_p W\big)(s)
&= \gamma\Big[\sum_{s'}\big(q(s')-p(s'\mid s)\big)W(s')
+ \mathbb{E}_p\big[\log q(s') - \log p(s'\mid s)\big]\Big] \\
&= \gamma\Big[\sum_{s'}\big(q(s')-p(s'\mid s)\big)W(s')
- \mathcal{D}_{\mathrm{KL}}\big(p(\cdot\mid s)\|q\big)\Big].
\end{align*}
For the first bracketed term, since $q-p(\cdot\mid s)$ has total mass zero,
for any constant $c$,
$\sum_{s'}(q-p)(s')W(s')=\sum_{s'}(q-p)(s')(W(s')-c)$. Choosing
$c=\tfrac12(\max_{s'}W(s')+\min_{s'}W(s'))$ so that
$\|W-c\|_\infty=\tfrac12\mathrm{sp}(W)$, H\"older's inequality gives
\[
\Big|\sum_{s'}(q-p)(s')W(s')\Big|
\le\Big(\sum_{s'}\big|q(s')-p(s'\mid s)\big|\Big)\cdot\frac{\mathrm{sp}(W)}{2}
=\big\|q-p(\cdot\mid s)\big\|_{TV}\,\mathrm{sp}(W).
\]
Applying the triangle inequality to the two bracketed terms (the divergence is
nonnegative) at $W=\widehat V_p$ and taking $\sup_s$ yields (iii).

\textbf{Term (iv) and claim (b).} By
Proposition~\ref{prop:shaped-reward}(iii) and the definition of
$\delta_\pi$, for every $s$,
\[
\big(\widetilde T^\pi_\beta \bar V^\pi_\beta\big)(s)
- \big(\widehat T_p \bar V^\pi_\beta\big)(s)
= \gamma\Big[\log\mathbb{E}_p\big[e^{\bar V^\pi_\beta(s')}\mid s\big]
- \mathbb{E}_p\big[\bar V^\pi_\beta(s')\mid s\big]
+ H\big(p(\cdot\mid s)\big)\Big]
= \gamma\big[\delta_\pi(s)+H_p(s)\big].
\]
Using the two fixed-point identities
$\bar V^\pi_\beta=\widetilde T^\pi_\beta\bar V^\pi_\beta$ and
$\widehat V_p=\hat r+\gamma P_\pi\widehat V_p$, the difference
$D:=\bar V^\pi_\beta-\widehat V_p$ satisfies
$D = \gamma P_\pi D + \gamma(\delta_\pi+H_p)$, i.e.,
\eqref{eq:relax-exact}, where $(I-\gamma P_\pi)^{-1}=\sum_{k\ge0}\gamma^k
P_\pi^k$ exists by Proposition~\ref{prop:shaped-reward}(ii). Since every
summand is entrywise nonnegative and the $k=0$ summand is $I$, we get the
componentwise sandwich
$\gamma(\delta_\pi+H_p)\le D\le
\|(I-\gamma P_\pi)^{-1}\|_{\infty\to\infty}\,
\gamma\|\delta_\pi+H_p\|_\infty\mathbf 1
=\tfrac{\gamma}{1-\gamma}\|\delta_\pi+H_p\|_\infty\mathbf 1$,
using that the row sums of $(I-\gamma P_\pi)^{-1}$ equal $1/(1-\gamma)$. This
proves $D\ge0$ (one-sidedness), the two-sided bound in (b), and term (iv). If
$\delta_\pi+H_p=c\mathbf 1$ then
$D=c\gamma(I-\gamma P_\pi)^{-1}\mathbf 1=\tfrac{c\gamma}{1-\gamma}\mathbf 1$,
the claimed equality case.

\textbf{Claim (c).} Theorem~\ref{theorem:finitesamplerrorbound} gives the limit in (i). Remark~\ref{remark:tabular-exact} establishes (ii). For (iii), both the TV and
KL terms vanish at $q_{\omega^*}=p$. For (iv), if $p(\cdot\mid s)$ is a point mass
for every $s$ then $H_p\equiv0$ and $\delta_\pi\equiv0$
(Lemma~\ref{lemma:dv-gap}, constant-on-support case). Conversely, for
$\gamma>0$, $H(p(\cdot\mid s))=0$ on a finite space forces $p(\cdot\mid s)$ to
be a point mass, and $\delta_\pi(s)\ge0$, so (iv) vanishes only in that case
(at $\gamma=0$ the factor $\gamma$ annihilates (iv) for any dynamics).
\end{proof}

Each term of \eqref{eq:err-decomp} has a distinct owner: (i) is the
statistical and optimization error of SGD, the only stochastic term, decaying
at rate $\mathcal O(1/\sqrt t)$ in expectation. Term (ii) is the price of linear
function approximation and ridge regularization. Term (iii) is the modeling
mismatch of the learned variational distribution at the optimum, the term
Theorem~\ref{theorem:finitesamplerrorbound} alone cannot see. Term (iv) is the
irreducible bias of relaxing the log-partition term by the cross-entropy
surrogate, which no amount of data, features, or model capacity removes, but
which is explicitly computable and vanishes for deterministic dynamics.

\begin{samepage}
\begin{remark}[Tabular approximation error]
Take $\Phi=I$, let $\mu$ have full support, and suppose $\mathcal K$ contains $\widehat V_\omega$ for each $\omega\in\Omega$.\label{remark:tabular-exact} For fixed $\omega$, the unregularized fit minimizes the squared residual of $\widehat T_\omega$ (Lemma~\ref{lemma:F-pop}). Its Hessian is $(I-\gamma Q_\omega)^\top\operatorname{diag}(\mu)(I-\gamma Q_\omega)$, which is positive definite by Proposition~\ref{prop:shaped-reward}(ii) and full support of $\mu$. Thus the fixed point $\widehat V_\omega$, where the residual vanishes, is the unique minimizer. The smallest eigenvalue of the Hessian is bounded away from zero by continuity and compactness of $\Omega$. Comparing the ridge objective at its minimizer with its value at the bounded fixed point shows uniform convergence to $\widehat V_\omega$ as $\lambda\downarrow0$. This applies at $\omega^*(\lambda)$, so the approximation term in Corollary~\ref{cor:error-decomposition} vanishes in this limit.
\end{remark}
\end{samepage}

\begin{remark}[No double sampling]
\label{remark:no-double-sampling}
Classical Bellman-residual gradients require two independent next-state draws from the same state \citep{baird1995residual, Sutton1998}. In finite state spaces, the continuation expectation in Equation~\ref{eq:Fthetaomega} is a finite sum under $q_\omega$. One observed transition supplies the log-likelihood term and suffices to form a sample gradient of $F$. The finite-sample bound in Theorem~\ref{theorem:finitesamplerrorbound} still assumes conditional unbiasedness given $\mathcal F_t$.

For fixed $\omega$, Equation~\ref{eq:F-pop} expresses $F$ as the squared residual of $\widehat T_\omega$ plus the ridge and a variance term independent of $\theta$. This variance depends on $\omega$, so joint residual minimization can yield a different transition model from the regularized likelihood update in Algorithm~\ref{alg:rrstd}. Corollary~\ref{cor:error-decomposition} bounds the error due to the discrepancy between $q_{\omega^*}$ and $p$.
\end{remark}

\subsection[Two-Sided Actor Convergence]{Proof of Theorem~\ref{thm:twosided_convergence}}
\label{app:proof-twosided}

The critic convergence result in Section~\ref{app:critic-convergence} supplies a finite-sample parameter bound for the inner runs. We now state the assumptions for the two-sided actor update, bound its finite-difference bias, perturbation noise, and critic error, and prove convergence and the local mean-square rate for the objective it evaluates.

\subsubsection{Assumptions.}
\label{app:ts-assumptions}

\begin{assumption}
\label{ass:ts-J}
The exponential-utility objective
$J(\psi):=\frac{1}{\beta}\log\mathbb{E}_{\tau\sim\pi_\psi}[e^{\beta R(\tau)}]$ for $\beta\ne0$ satisfies the following conditions.
\begin{enumerate}
    \item $|r(s,a)|\leq R_{\max}<\infty$ for all $(s,a)$.
    Hence $|R(\tau)|\leq\bar{R}:=R_{\max}/(1-\gamma)$ and
    $J_{\max}:=\sup_{\psi}|J(\psi)|\leq\bar{R}<\infty$.
  \item $J\in C^3(\mathbb{R}^d)$, so its derivatives through order three are bounded on compact sets.
  \item The unprojected iterates of Equation~\ref{eq:twosided-update} satisfy $\sup_{t\geq0}\|\psi_t\|<\infty$ almost surely.
\end{enumerate}
\end{assumption}

\begin{assumption}
\label{ass:ts-step}
The sequences $\{a_t\}_{t\geq 0}$ and $\{c_t\}_{t\geq 0}$ are
positive, deterministic, and satisfy:
\begin{enumerate}
  \item $a_t\to 0$ and $c_t\to 0$ monotonically.
  \item $\sum_{t=0}^{\infty}a_t=\infty$.
  \item $\sum_{t=0}^{\infty}\dfrac{a_t^2}{c_t^2}<\infty$.
  \item $\sum_{t=0}^{\infty}a_t c_t^2<\infty$.
\end{enumerate}
For power laws $a_t=a_0t^{-\alpha}$ and $c_t=c_0t^{-\eta}$ with $a_0,c_0>0$ and $t\ge1$, these conditions hold for
\[
\tfrac23<\alpha<1,\qquad \tfrac12(1-\alpha)<\eta<\alpha-\tfrac12.
\]
The upper bound on $\eta$ gives $2(\alpha-\eta)>1$, and the lower bound gives $\alpha+2\eta>1$.
\end{assumption}

\begin{assumption}
\label{ass:ts-rad}
The components $\{\Delta_{ti}\}_{i\in[d],\,t\geq 0}$ are mutually
independent with $\Delta_{ti}\sim\mathrm{Uniform}(\{-1,+1\})$,
independent of $\mathcal{F}_t$ for all $t$, and satisfy $\mathbb{E}[\Delta_{ti}]=0$, $|\Delta_{ti}^{-1}|=|\Delta_{ti}|=1$,
and $\Delta_{ti}^2=1$.
These are conditions (C3) of \citet[Theorem~1]{spall1992multivariate}.
The key moment identities are:
\begin{equation}
  \mathbb{E}[\Delta_{ti}^k]
  = \begin{cases} 0 & k\;\text{odd,} \\ 1 & k\;\text{even,} \end{cases}
  \qquad
  \mathbb{E}[\Delta_{ti}^a\Delta_{tj}^b]
  = \mathbb{E}[\Delta_{ti}^a]\,\mathbb{E}[\Delta_{tj}^b]
  \quad(i\neq j).
  \label{eq:ts-rad-moments}
\end{equation}
\end{assumption}

For $\beta\neq0$, the value fed to the actor and the critic-induced
surrogate objective are (restated from the main text)
\begin{equation}
  \widehat{J}_{\psi_t^{\pm}}
  \;:=\;
  \frac{1}{\beta}\,\phi(s_0^{\pm})^{\top}\theta_{N_t}^{\pm},
  \label{eq:actor-ce-estimate-appendix}
\end{equation}
\begin{equation}
  J_c(\psi)
  \;:=\;
  \frac{1}{\beta}\,\phi(s_0)^{\top}\theta^{*}(\psi),
  \label{eq:actor-surrogate}
\end{equation}
where $\theta_{N_t}^{\pm}$ are the inner-loop critic iterates after $N_t$
updates, $\theta^{*}(\psi)$ is the $\theta$-component of the unique
minimizer of $F$ under $\pi_\psi$, and $s_0$ is the fixed initial state.
When $\beta=0$, both read off the risk-neutral critic without the $1/\beta$
rescaling.

\begin{assumption}
\label{ass:ts-critic}
The Relaxed-RSTD critic run for $N_t$ inner iterations at the perturbed
policies $\psi_t^{\pm}$ produces, via \eqref{eq:actor-ce-estimate-appendix},
estimates $\widehat{J}_{\psi_t^{\pm}}$ whose errors relative to the
critic-induced surrogate \eqref{eq:actor-surrogate},
\[
  \varepsilon_t^{\pm}
  \;:=\;
  \widehat{J}_{\psi_t^{\pm}} - J_c(\psi_t^{\pm}),
\]
satisfy:
\begin{enumerate}
  \item[(i)] $\bigl|\mathbb{E}[\varepsilon_t^{\pm}\mid\mathcal{F}_t]\bigr|
        = O(\rho^{N_t}) \to 0$ for some $\rho\in(0,1)$
        (geometric burn-in of the conditional mean).
  \item[(ii)] $\mathbb{E}\bigl[(\varepsilon_t^{\pm})^2\bigr]
        \;\le\;
        \frac{4 B_\phi^2 \sigma^2}{\beta^2 \mu_F^2 N_t}
        + O\!\left(\frac{1}{N_t^2}\right)
        \;=:\;
        \frac{C_\varepsilon}{N_t} + O\!\left(\frac{1}{N_t^2}\right),
        \qquad
        C_\varepsilon := \frac{4 B_\phi^2 \sigma^2}{\beta^2\mu_F^2}
        < \infty$,
\end{enumerate}
where $B_\phi := \sup_{s\in\mathcal{S}}\|\phi(s)\|$
(Assumption~\ref{assumption:relaxedlinfuncapp}) and
$\mu_F = \lambda - L_H$ is the strong-convexity modulus of
Lemma~\ref{lemma:Fstrconvex-revised}. For the $\beta=0$
branch of \eqref{eq:actor-ce-estimate-appendix} the constant is
$C_\varepsilon = 4B_\phi^2\sigma^2/\mu_F^2$.
\end{assumption}

\begin{remark}[The critic error is controlled relative to $J_c$, not $J_\beta$]
\label{remark:actor-critic-assumption}
Part~(ii) follows when the finite-sample conditions of Theorem~\ref{theorem:finitesamplerrorbound} hold uniformly for the inner critic runs. By
\eqref{eq:actor-ce-estimate-appendix} and the Cauchy--Schwarz inequality,
\[
  (\varepsilon_t^{\pm})^2
  \;=\;
  \frac{1}{\beta^2}
  \bigl(\phi(s_0^{\pm})^{\top}
    (\theta_{N_t}^{\pm} - \theta^{*}(\psi_t^{\pm}))\bigr)^2
  \;\le\;
  \frac{B_\phi^2}{\beta^2}\,
  \bigl\|\theta_{N_t}^{\pm} - \theta^{*}(\psi_t^{\pm})\bigr\|^2
  \;\le\;
  \frac{B_\phi^2}{\beta^2}\,
  \bigl\|(\theta_{N_t}^{\pm},\omega_{N_t}^{\pm})^{\top}
    - (\theta^{*},\omega^{*})^{\top}\bigr\|^2,
\]
where $(\theta_{N_t}^{\pm},\omega_{N_t}^{\pm})$ are the two inner-loop
iterates (one critic run per perturbed policy), and
Theorem~\ref{theorem:finitesamplerrorbound} bounds the expected squared parameter distance by
$4\sigma^2/(\mu_F^2 N_t) + O(N_t^{-2})$, uniformly over the (random)
perturbed policies: $\sigma$ and $\mu_F$ depend only on the feature,
reward, weight, and variational-smoothness bounds --- not on $\psi$ --- and the
initialization-dependent $O(N_t^{-2})$ term is uniform over warm starts
in the compact $\mathcal{K}$.  The $\beta^{-2}$ factor is
the price of certainty-equivalent scaling.  Crucially,
Theorem~\ref{theorem:finitesamplerrorbound} controls the distance to
$\theta^{*}(\psi)$ --- hence to the \emph{surrogate} $J_c$ --- and not
to $J_\beta$.  Stating the assumption relative to $J_\beta$, as in the
originally submitted version, would silently absorb the approximation,
relaxation, and nested-vs-static gaps into an unverifiable hypothesis.
Those gaps are instead accounted for exactly once, in
Proposition~\ref{prop:actor-surrogate-gap}.  Part~(i) remains a genuine
assumption on the burn-in of the conditional mean:
Theorem~\ref{theorem:finitesamplerrorbound} together with conditional
Jensen only yields the polynomial bound
$\mathbb{E}\bigl[\bigl(\mathbb{E}[\varepsilon_t^{\pm}\mid
\mathcal{F}_t]\bigr)^2\bigr] \le C_\varepsilon/N_t + O(N_t^{-2})$.
\end{remark}

\begin{assumption}
\label{ass:ts-inner}
The number $N_t$ of inner updates satisfies $\sum_t a_t/(c_t\sqrt{N_t})<\infty$. For power-law step sizes as in Assumption~\ref{ass:ts-step}, it suffices that $N_t\ge t^\nu$ for $t\ge1$ with $\alpha-\eta+\nu/2>1$.
\end{assumption}

\begin{assumption}
\label{ass:ts-stable}
There exists a strict local maximum $\psi^\star\in\Psi$ with
$\nabla J(\psi^\star)=0$, $\nabla^2 J(\psi^\star)\prec 0$, and
$\mu:=-\lambda_{\max}(\nabla^2 J(\psi^\star))/2>0$, such that in
some neighbourhood $\mathcal{U}\ni\psi^\star$:
\begin{equation}
  \langle\nabla J(\psi),\,\psi-\psi^\star\rangle
  \;\leq\; -\mu\|\psi-\psi^\star\|^2
  \quad\forall\,\psi\in\mathcal{U}.
  \label{eq:ts-lyap-cond}
\end{equation}
\end{assumption}

\subsubsection{Decomposition of the Two-Sided Update.}
\label{app:ts-decomp}

We write the observed critic values as
$\widehat{J}_{\psi_t^\pm}=J(\psi_t^\pm)+\varepsilon_t^\pm$,
where $\varepsilon_t^\pm$ are the errors relative to the objective $J$ in Theorem~\ref{thm:twosided_convergence}. For the critic-induced objective $J_c$, Assumption~\ref{ass:ts-critic}(ii) supplies their mean-square bounds. Substituting into the actor update gives
\begin{align}
  \widehat{g}_t
  &=
  \underbrace{\frac{J(\psi_t^+)-J(\psi_t^-)}{2c_t}\Delta_t^{-1}}_{%
    =:\;\widehat{g}_t^{\,\circ}\;\text{(oracle estimator)}}
  +
  \underbrace{\frac{\varepsilon_t^+-\varepsilon_t^-}{2c_t}\Delta_t^{-1}}_{%
    =:\;\phi_t\;\text{(two-sample critic drift)}}.
  \label{eq:ts-g-decomp}
\end{align}
Decomposing the oracle estimator as
\begin{equation}
  \widehat{g}_t^{\,\circ}
  =
  \underbrace{\mathbb{E}[\widehat{g}_t^{\,\circ}\mid\mathcal{F}_t]}_{%
    =\,\nabla J(\psi_t)+b_t}
  +
  \underbrace{\widehat{g}_t^{\,\circ}-\mathbb{E}[\widehat{g}_t^{\,\circ}\mid\mathcal{F}_t]}_{%
    =:\,e_t\;\text{(zero-mean noise)}},
  \label{eq:ts-oracle-decomp}
\end{equation}
where $b_t:=\mathbb{E}[\widehat{g}_t^{\,\circ}\mid\mathcal{F}_t]-\nabla J(\psi_t)$
is the \emph{finite-difference bias}.
The full actor update is therefore
\begin{equation}
  \psi_{t+1}
  = \psi_t
  + a_t\bigl[\nabla J(\psi_t) + b_t + e_t + \phi_t\bigr],
  \label{eq:ts-full-decomp}
\end{equation}
Lemma~\ref{lem:ts-bias} bounds $b_t$ by $O(c_t^2)$ on bounded parameter sets. Lemma~\ref{lem:ts-noise} controls the martingale difference $e_t$, and Lemma~\ref{lem:ts-drift} gives $\sum_t a_t\|\phi_t\|<\infty$ almost surely.

\subsubsection{Bias of the Symmetric Estimator.}
\label{app:ts-biasA}

\begin{lemma}
\label{lem:ts-bias}
Under Assumptions~\ref{ass:ts-J}(ii) and~\ref{ass:ts-rad}, let $K_3$ bound $\|\nabla^{(3)}J\|$ on the perturbation segments through the iterates under consideration. Then $\|b_t\|\le (K_3d^2/6)c_t^2$.
\end{lemma}

\begin{proof}
Taylor expansion of $J(\psi_t\pm c_t\Delta_t)$ about $\psi_t$ cancels the constant and quadratic terms in the difference. Each third-order remainder has magnitude at most $K_3 c_t^3\|\Delta_t\|^3/6$. Dividing the difference by $2c_t$ and multiplying by $\Delta_t^{-1}=\Delta_t$ gives
\[
\widehat g_t^{\,\circ}=\Delta_t\Delta_t^\top\nabla J(\psi_t)+u_t,
\qquad
\|u_t\|\le (K_3/6)c_t^2\|\Delta_t\|^4.
\]
The perturbations are independent of $\mathcal F_t$ and satisfy $\mathbb E[\Delta_t\Delta_t^\top]=I$ and $\|\Delta_t\|^2=d$. Taking conditional expectations yields the bound.
\end{proof}

\subsubsection{Noise is a Bounded Martingale Difference Sequence.}
\label{app:ts-noiseB}

\begin{lemma}
\label{lem:ts-noise}
Under Assumptions~\ref{ass:ts-J} and~\ref{ass:ts-rad},
$\{e_t\}_{t\geq 0}$ is a martingale difference sequence adapted to
$\{\mathcal{F}_{t+1}\}_{t\geq 0}$ satisfying:
\begin{equation}
  \mathbb{E}[e_t\mid\mathcal{F}_t]=0,
  \qquad
  \mathbb{E}[\|e_t\|^2\mid\mathcal{F}_t]
  \;\leq\;
  \frac{4J_{\max}^2\,d}{4c_t^2}
  \;=\;
  \frac{J_{\max}^2\,d}{c_t^2}
  \;=:\;
  K_e\,c_t^{-2}.
  \label{eq:ts-noise-moments}
\end{equation}
Under Assumption~\ref{ass:ts-step}(iii), the weighted tail sums satisfy
\begin{equation}
  \sum_{t>m}a_t e_t
  \;\xrightarrow{\;\mathrm{a.s.}\;}
  0
  \quad\text{as }m\to\infty.
  \label{eq:ts-noise-tails}
\end{equation}
\end{lemma}

\begin{proof}
By definition~\eqref{eq:ts-oracle-decomp},
$e_t=\widehat{g}_t^{\,\circ}-\mathbb{E}[\widehat{g}_t^{\,\circ}\mid\mathcal{F}_t]$,
so $\mathbb{E}[e_t\mid\mathcal{F}_t]=0$ identically.
The oracle estimator depends on $\psi_t$ and $\Delta_t$, which are $\mathcal F_{t+1}$-measurable. Thus the sequence
$\{e_t,\mathcal{F}_{t+1}\}_{t\geq 0}$ is a martingale difference sequence.

\medskip
Since $|J(\psi_t^\pm)|\leq J_{\max}$ by Assumption~\ref{ass:ts-J}(i), the squared difference of the exact objective values is at most $4J_{\max}^2$.
Using $|\Delta_{ti}^{-1}|=1$ and $\|\Delta_t^{-1}\|^2=d$:
\begin{equation}
  \|\widehat{g}_t^{\,\circ}\|^2
  = \sum_{i=1}^d
    \left(\frac{J(\psi_t^+)-J(\psi_t^-)}{2c_t}\right)^{\!2}\!\!\Delta_{ti}^{-2}
  \;\leq\;
  \frac{4J_{\max}^2}{4c_t^2}\cdot d
  = \frac{J_{\max}^2\,d}{c_t^2}.
  \label{eq:ts-ghat-sq}
\end{equation}
By the variance decomposition
$\mathbb{E}[\|e_t\|^2\mid\mathcal{F}_t]
=\mathbb{E}[\|\widehat{g}_t^{\,\circ}\|^2\mid\mathcal{F}_t]
-\|\mathbb{E}[\widehat{g}_t^{\,\circ}\mid\mathcal{F}_t]\|^2
\leq\mathbb{E}[\|\widehat{g}_t^{\,\circ}\|^2\mid\mathcal{F}_t]$:
\begin{equation}
  \mathbb{E}[\|e_t\|^2\mid\mathcal{F}_t]
  \;\leq\;
  \frac{J_{\max}^2\,d}{c_t^2}
  \;=:\;
  K_e\,c_t^{-2},
  \quad K_e := J_{\max}^2\,d.
  \label{eq:ts-noise-var}
\end{equation}

\medskip
\noindent Using the summability of the quadratic variation, we have
\begin{equation}
  \sum_{t=0}^{\infty}\mathbb{E}[\|a_t e_t\|^2\mid\mathcal{F}_t]
  \;\leq\;
  K_e\sum_{t=0}^{\infty}\frac{a_t^2}{c_t^2}
  \;<\;\infty,
  \label{eq:ts-quad-var}
\end{equation}
by Assumption~\ref{ass:ts-step}(iii).

\medskip
Let $M_T:=\sum_{t=0}^{T-1}a_t e_t$.
By~\eqref{eq:ts-quad-var}, $\sup_T\mathbb{E}[\|M_T\|^2]\leq K_e\sum_t a_t^2/c_t^2<\infty$,
so $\{M_T,\mathcal{F}_T\}$ is an $L^2$-bounded martingale.
The $L^2$-Martingale Convergence Theorem \citep[Theorem B.1]{borkar2008stochastic} gives $M_T\xrightarrow{\mathrm{a.s.}}M_\infty\in\mathbb{R}^d$,
so $\sum_{t>m}a_t e_t=M_\infty-M_m\xrightarrow{\mathrm{a.s.}}0$.
\end{proof}

\subsubsection{Summability of the Critic Error.}
\label{app:ts-driftC}

\begin{lemma}
\label{lem:ts-drift}
If $\mathbb E[(\varepsilon_t^\pm)^2]=O(N_t^{-1})$ and Assumption~\ref{ass:ts-inner} holds, then $\sum_t a_t\|\phi_t\|<\infty$ almost surely.
\end{lemma}

\begin{proof}
With $M$ a uniform constant such that $\mathbb E[(\varepsilon_t^\pm)^2]\le M/N_t$, the definition of $\phi_t$ and $\|\Delta_t^{-1}\|^2=d$ give
\begin{equation}
\mathbb E\|\phi_t\|^2
\le \frac{d}{2c_t^2}\bigl(\mathbb E[(\varepsilon_t^+)^2]+\mathbb E[(\varepsilon_t^-)^2]\bigr)
\le \frac{dM}{c_t^2N_t}.
\label{eq:ts-phi-bound}
\end{equation}
Cauchy--Schwarz and Assumption~\ref{ass:ts-inner} give $\mathbb E\sum_t a_t\|\phi_t\|\le \sqrt{dM}\sum_t a_t/(c_t\sqrt{N_t})<\infty$. Thus the nonnegative sum is finite almost surely.
\end{proof}

\begin{proof}[Proof of Theorem~\ref{thm:twosided_convergence}]
Lemma~\ref{lem:ts-bias} gives the oracle bias. By almost-sure boundedness, each sample path of the iterates and its perturbation segments lies in a compact ball. On such balls the first and second derivatives of $J$ are bounded, and the mean-value theorem gives $\|\widehat g_t^{\,\circ}\|\le d\sup\|\nabla J\|$. In particular, the conditional second moment of $e_t$ is bounded up to the first exit from any fixed ball. The sum $\sum_t a_t^2$ is finite by Assumption~\ref{ass:ts-step}. The martingale $\sum_{t<n}a_t\langle\nabla J(\psi_t),e_t\rangle$, stopped at the first exit from that ball, therefore converges almost surely. Taking a countable sequence of increasing balls and using boundedness gives convergence of the unstopped sum almost surely.

Lemma~\ref{lem:ts-drift} gives $\sum_t a_t\|\phi_t\|<\infty$ almost surely, and Lemma~\ref{lem:ts-bias} with Assumption~\ref{ass:ts-step} gives $\sum_t a_t\|b_t\|<\infty$ almost surely. Also $\|\psi_{t+1}-\psi_t\|\le a_t(\|\widehat g_t^{\,\circ}\|+\|\phi_t\|)$, so $\sum_t\|\psi_{t+1}-\psi_t\|^2<\infty$ almost surely. Taylor expansion of $J(\psi_{t+1})$ in Equation~\ref{eq:ts-full-decomp} now gives
\[
J(\psi_{t+1})-J(\psi_t)
= a_t\|\nabla J(\psi_t)\|^2+a_t\langle\nabla J(\psi_t),e_t\rangle+q_t,
\qquad \sum_t|q_t|<\infty
\]
almost surely. The last sum includes the bias, critic error, and second-order remainder. Since $J$ is bounded on the iterates, $\sum_t a_t\|\nabla J(\psi_t)\|^2<\infty$ almost surely.

To show $\|\nabla J(\psi_t)\|\to0$, fix $\epsilon>0$ and consider excursions from gradient norm at least $\epsilon$ to norm below $\epsilon/2$. Every visit above $\epsilon$ must eventually be followed by a drop below $\epsilon/2$ because $\sum_t a_t=\infty$. On a bounded path, the gradient is Lipschitz and the oracle increments are bounded by a constant times $a_t$. The tail of $\sum_t a_t\|\phi_t\|$ vanishes. Thus each sufficiently late excursion requires a fixed positive sum of step sizes before dropping below $\epsilon/2$. Infinitely many such excursions would contradict $\sum_t a_t\|\nabla J(\psi_t)\|^2<\infty$. Continuity of the gradient then makes every limit point critical.

For the local result, choose the closed ball about $\psi^\star$ so that the curvature inequality in Assumption~\ref{ass:ts-stable} holds throughout. It makes $\psi^\star$ the only critical point in that ball, which gives $\psi_t\to\psi^\star$ almost surely if the iterates eventually stay there. For the mean-square bound, let $V_t=\mathbb E\|\psi_t-\psi^\star\|^2$ and work from the deterministic confinement time onward. Expanding the squared update, the inner product with $e_t$ has zero expectation and the inner product with $\nabla J(\psi_t)$ contributes at most $-2\mu a_t V_t$. Young's inequality bounds the contribution of $b_t+\phi_t$ by $\mu a_t V_t$ plus a constant times $a_t(c_t^4+(c_t^2N_t)^{-1})$. The local bound on $\widehat g_t^{\,\circ}$ and Equation~\ref{eq:ts-phi-bound} bound the squared update term by $O(a_t^2)$ when $c_t^4+(c_t^2N_t)^{-1}=O(a_t)$. Hence, for a constant $K$,
\[
V_{t+1}\le(1-\mu a_t)V_t+Ka_t^2.
\]
For $a_t=t^{-\alpha}$ with $\alpha<1$, $a_t-a_{t+1}=o(a_t^2)$. For sufficiently large $A$ and $t$, $(1-\mu a_t)Aa_t+Ka_t^2\le Aa_{t+1}$. Induction gives $V_t=O(t^{-\alpha})$. The schedule in the theorem satisfies Assumptions~\ref{ass:ts-step} and~\ref{ass:ts-inner} and makes both terms in $c_t^4+(c_t^2N_t)^{-1}$ at most $t^{-\alpha}$.
\end{proof}

Algorithm~\ref{alg:rrsac} in Section~\ref{sec:actor} gives the coupled actor and critic updates.

\begin{corollary}[Critic-in-the-loop SPSA converges on the surrogate]
\label{cor:actor-surrogate-spsa}
Suppose $J_c$ satisfies the smoothness and bounded-iterate conditions of Theorem~\ref{thm:twosided_convergence}. Under Assumptions~\ref{ass:ts-step}, \ref{ass:ts-rad}, \ref{ass:ts-critic}(ii), and \ref{ass:ts-inner}, every limit point of the critic-based actor is almost surely a critical point of $J_c$. The local convergence and mean-square rate of Theorem~\ref{thm:twosided_convergence} apply under its local conditions for $J_c$.
\end{corollary}

\begin{proof}
Assumption~\ref{ass:ts-critic}(ii) gives $\mathbb E|\widehat J_{\psi_t^\pm}-J_c(\psi_t^\pm)|^2=O(N_t^{-1})$, so Theorem~\ref{thm:twosided_convergence} applies with $J=J_c$.
\end{proof}

\begin{remark}[Smoothness of the surrogate]
\label{remark:actor-smoothness}
A sufficient condition for $J_c \in C^3(\Psi)$ is: (a)~the Hessian
$\nabla^2_{(\theta,\omega)}F \succeq \mu_F I \succ 0$ on
$\mathcal{K}\times\Omega$, with $\mu_F = \lambda - L_H > 0$ the
strong-convexity modulus of Lemma~\ref{lemma:Fstrconvex-revised} under
$\lambda > L_H$, and
(b)~the gradient map
$(\theta,\omega,\psi) \mapsto \nabla_{(\theta,\omega)}F(\theta,\omega;\psi)$
is $C^3$ \emph{jointly} in all of its arguments (equivalently, $F$ is
$C^4$ in $(\theta,\omega)$ with all such partial derivatives $C^3$ in
$\psi$).  Joint smoothness in $(\theta,\omega)$ reduces to smoothness of
$\omega \mapsto \log q(\cdot\mid\cdot\,;\omega)$, because $F$
(Equation~\ref{eq:Fthetaomega} of the main text) is a quadratic polynomial in $\theta$ (softmax
$q$ is $C^\infty$ in $\omega$). The $C^3$ dependence on $\psi$ holds
whenever the policy parameterization --- and hence the state--action
occupancy under Assumption~\ref{assumption:mdpboundedergodic} entering
the expectation defining $F$ --- is $C^3$ in $\psi$ (e.g.\
softmax or Gaussian policies).  The implicit function theorem
\citep[Thm.~9.28]{rudin1976principles} applied to the stationarity
identity $\nabla F(\theta^{*}(\psi),\omega^{*}(\psi);\psi)=0$, with
higher-order smoothness obtained by repeatedly differentiating that
identity, then yields $\theta^{*}\in C^3(\Psi)$, hence
$J_c\in C^3(\Psi)$ with
$K_3^{c} \le B_\phi \sup_{\psi\in\Psi}\|\nabla^{(3)}\theta^{*}(\psi)\| /
|\beta| < \infty$ on compact $\Psi$.  Absent such structure, $C^3$
smoothness of $J_c$ is an assumption, exactly parallel to the smoothness
previously assumed for $J$.
\end{remark}

\begin{remark}[Random initial states]
\label{remark:actor-init}
If the initial states $s_0^{\pm}$ used to query the critic are drawn
i.i.d.\ from a distribution $\xi$, independently of the critic's
training trajectories (by taking a fresh draw from $\xi$ at query time), define
$J_c(\psi) := \beta^{-1}\,\mathbb{E}_{s_0\sim\xi}
[\phi(s_0)^{\top}\theta^{*}(\psi)]$. Without this independence, $\theta_{N_t}^{\pm}$ and $s_0^{\pm}$ are correlated and the fluctuation below is no longer conditionally centered. With independent draws, two additional effects arise,
both explicitly bounded.
(1)~The sampling fluctuation
$\upsilon_t^{\pm} := \beta^{-1}\bigl(\phi(s_0^{\pm}) -
\mathbb{E}_\xi[\phi(s_0)]\bigr)^{\top}\theta_{N_t}^{\pm}$
(written $\upsilon$ because $\zeta_t$ already denotes the critic's
martingale tail sequence in Section~\ref{app:critic-convergence}) is conditionally
mean-zero given $\mathcal{F}_t \vee \sigma(\Delta_t)$ and satisfies
$|\upsilon_t^{\pm}| \le 2B_\phi D_{\mathcal{K}}/|\beta|$ almost surely. The induced
term $(\upsilon_t^{+}-\upsilon_t^{-})\Delta_t^{-1}/(2c_t)$ is a martingale
difference and is absorbed into Lemma~\ref{lem:ts-noise} by enlarging
the constant to
$K_e' \le 2d\,(J_{\max}^{c})^2 + 8d\,B_\phi^2 D_{\mathcal{K}}^2/\beta^2$
(via $(a+b)^2 \le 2a^2+2b^2$), leaving
Assumption~\ref{ass:ts-step}(iii) unchanged.
(2)~The population objective changes: with random $s_0$ the true
objective is
$J^{\xi}_\beta(\psi) = \beta^{-1}\log\mathbb{E}_{s_0\sim\xi}
\bigl[e^{\bar V^{\pi_\psi}_\beta(s_0)}\bigr]$,
whereas $J_c$ averages the log-value \emph{linearly}.  By Jensen's
inequality and Hoeffding's lemma
\citep[\S 2.6]{boucheron2013concentration} applied to the bounded
random variable $\bar V^{\pi_\psi}_\beta(s_0)$, $s_0\sim\xi$,
\[
  0
  \;\le\;
  \log\mathbb{E}_{\xi}\bigl[e^{\bar V^{\pi_\psi}_\beta}\bigr]
  - \mathbb{E}_{\xi}\bigl[\bar V^{\pi_\psi}_\beta\bigr]
  \;\le\;
  \tfrac{1}{8}\,\mathrm{sp}\bigl(\bar V^{\pi_\psi}_\beta\bigr)^2,
  \qquad
  \mathrm{sp}(W) := \sup W - \inf W ,
\]
where the supremum and infimum defining the span are taken over the
support of $\xi$ (so the gap vanishes for a Dirac $\xi$).
Hence an additional initial-state mixing term
$\varepsilon_{\mathrm{init}} := \tfrac{1}{8|\beta|}\sup_{\psi\in\Psi}
\mathrm{sp}\bigl(\bar V^{\pi_\psi}_\beta\bigr)^2$
must be added inside $\varepsilon_{\mathrm{tot}}$ in
Proposition~\ref{prop:actor-surrogate-gap}.  Since
$|\bar V^{\pi}_\beta| \le |\beta|\bar{R}$ with
$\bar{R} = R_{\max}/(1-\gamma)$
(Assumption~\ref{ass:ts-J}(i)), one has
$\mathrm{sp}(\bar V^{\pi}_\beta) \le 2|\beta|\bar{R}$ and so
$\varepsilon_{\mathrm{init}} \le |\beta|\bar{R}^2/2 \to 0$ as
$\beta\to 0$: the extra term does \emph{not} blow up in the weakly
risk-sensitive regime, and it vanishes identically for a Dirac $\xi$.
\end{remark}

\begin{lemma}[Bias of the empirical-log-mean estimator]
\label{lem:mc-logmean-bias}
Fix $\psi$ and $\beta\neq 0$, let $R_1,\dots,R_m$ be i.i.d.\ trajectory
returns under $\pi_\psi$ with $|R_i| \le \bar{R} = R_{\max}/(1-\gamma)$
(Assumption~\ref{ass:ts-J}(i)), and consider the empirical-log-mean
estimator
\[
  \widehat{J}^{\mathrm{MC}}_m
  \;:=\;
  \frac{1}{\beta}\,
  \log\!\Bigl(\frac{1}{m}\sum_{i=1}^{m} e^{\beta R_i}\Bigr).
\]
Write $X := e^{\beta R}$,
$x_{\min} := e^{-|\beta|\bar{R}} \le X \le e^{|\beta|\bar{R}} =:
x_{\max}$, $m_X := \mathbb{E}[X] \in [x_{\min},x_{\max}]$,
$\sigma_X^2 := \mathrm{Var}(X)$, and
$M_j := \mathbb{E}[(X-m_X)^j]$ for $j=3,4$.
Then for every $m\ge 1$:
\begin{enumerate}
  \item[(a)] (exact sign, all $m$) if $\sigma_X^2>0$ then
    \[
      \operatorname{sign}\bigl(\mathbb{E}[\widehat{J}^{\mathrm{MC}}_m]
        - J_\beta(\psi)\bigr) = -\operatorname{sign}(\beta):
    \]
    the estimator is pessimistic for $\beta>0$ and \emph{optimistic}
    --- it over-estimates the certainty equivalent --- for $\beta<0$.
    If $\sigma_X^2=0$ (degenerate return distribution) the estimator is
    exactly unbiased.
  \item[(b)] (delta-method expansion with explicit remainder)
    \begin{equation}
      \mathbb{E}\bigl[\widehat{J}^{\mathrm{MC}}_m\bigr]
      - J_\beta(\psi)
      \;=\;
      -\,\frac{\sigma_X^2}{2\beta m_X^2 m}
      \;+\; r_m,
      \qquad
      |r_m| \;\le\; \frac{C_{\mathrm{MC}}}{m^2},
      \label{eq:mc-bias}
    \end{equation}
    with the explicit constant
    \[
      C_{\mathrm{MC}}
      \;:=\;
      \frac{1}{|\beta|}
      \left(
        \frac{|M_3|}{3m_X^3}
        + \frac{3\sigma_X^4 + M_4}{4\,x_{\min}^4}
      \right)
      \;<\;\infty .
    \]
\end{enumerate}
\end{lemma}

\begin{proof}
Throughout, $\bar{X}_m := m^{-1}\sum_{i=1}^m X_i \in
[x_{\min},x_{\max}]$ almost surely, so $\log\bar{X}_m$ is well defined
and bounded, and $J_\beta(\psi) = \beta^{-1}\log m_X$.

(a) $\bar{X}_m$ is a positive random variable with
$\mathbb{E}[\bar{X}_m]=m_X$. If $\sigma_X^2>0$, then $\bar{X}_m$ is
non-degenerate for every $m\ge 1$ (indeed
$\mathrm{Var}(\bar{X}_m)=\sigma_X^2/m>0$), and strict Jensen for the
strictly concave logarithm gives
$\mathbb{E}[\log\bar{X}_m] < \log m_X$.  Dividing by $\beta$ preserves the
strict inequality for $\beta>0$ and reverses it for $\beta<0$, which is
the claim.  If $\sigma_X^2=0$ then $\bar{X}_m = m_X$ a.s.\ and the bias is
zero.

(b) Let $\delta := \bar{X}_m - m_X \in [x_{\min}-m_X,\,x_{\max}-m_X]$.  The
fourth-order Taylor expansion of $\log$ at $m_X$ with Lagrange remainder
provides, for each realization, a point $\chi_\delta$ between $m_X$ and
$m_X+\delta$ (so $\chi_\delta \ge x_{\min}$). We use $\chi$ to distinguish it from
both the initial-state distribution $\xi$ of
Remark~\ref{remark:actor-init} and the proof-local noise aggregate
$\xi_t$ of \S\ref{app:proof-twosided}. The expansion is
\[
  \log(m_X+\delta)
  = \log m_X + \frac{\delta}{m_X} - \frac{\delta^2}{2m_X^2}
  + \frac{\delta^3}{3m_X^3} - \frac{\delta^4}{4\chi_\delta^4}.
\]
Although $\chi_\delta$ need not be chosen measurably, the remainder
$\log(m_X+\delta)$ minus the fourth-order polynomial \emph{is} measurable
and is sandwiched pathwise between $-\delta^4/(4x_{\min}^4)$ and
$-\delta^4/(4x_{\max}^4)$, which is all the expectation step below uses.
Taking expectations and using the exact i.i.d.\ moment identities
$\mathbb{E}[\delta]=0$,
$\mathbb{E}[\delta^2]=\sigma_X^2/m$,
$\mathbb{E}[\delta^3]=M_3/m^2$, and
$\mathbb{E}[\delta^4] = \bigl(3(m-1)\sigma_X^4+M_4\bigr)/m^3 \le
(3\sigma_X^4+M_4)/m^2$ for $m\ge 1$,
\[
  \mathbb{E}[\log\bar{X}_m] - \log m_X
  \;=\;
  -\frac{\sigma_X^2}{2m_X^2 m}
  + \frac{M_3}{3m_X^3 m^2}
  - \mathbb{E}\!\left[\frac{\delta^4}{4\chi_\delta^4}\right],
  \qquad
  0 \le \mathbb{E}\!\left[\frac{\delta^4}{4\chi_\delta^4}\right]
  \le \frac{3\sigma_X^4+M_4}{4\,x_{\min}^4\,m^2}.
\]
Dividing by $\beta$ and collecting the two $O(m^{-2})$ terms into $r_m$
via the triangle inequality yields \eqref{eq:mc-bias} with the stated
$C_{\mathrm{MC}}$.
\end{proof}

\begin{remark}[Monte-Carlo-return actor variants]
\label{remark:variants}
An alternative actor evaluates two \emph{independent} batches of $m$
rollouts under $\pi_{\psi_t^{+}}$ and $\pi_{\psi_t^{-}}$ and feeds
$\widehat{J}^{\mathrm{MC},\pm}_m$ to the two-sided SPSA update of the
main text (\S\ref{sec:actor}).
Lemma~\ref{lem:mc-logmean-bias} shows this variant is \emph{not}
unbiased at any finite $m$: its value bias is
$-\sigma_X^2(\psi)/(2\beta m_X(\psi)^2 m) + O(m^{-2})$ with sign
$-\operatorname{sign}(\beta)$ exactly. In the risk-averse regime
$\beta<0$ the bias is \emph{positive}, i.e.\ systematically optimistic
about the certainty equivalent --- the unfavourable direction for
safety-critical use.  If
$\psi \mapsto B(\psi) := -\sigma_X^2(\psi)/(2\beta m_X(\psi)^2)$ is
$C^3$ on $\Psi$ (it is a smooth function of the first two tilted moments
$\mathbb{E}_{\pi_\psi}[e^{\beta R}]$ and
$\mathbb{E}_{\pi_\psi}[e^{2\beta R}]$, both bounded away from $0$), the
antisymmetric cancellation of Lemma~\ref{lem:ts-bias} applied to the
perturbed objective $\psi \mapsto J_\beta(\psi) + B(\psi)/m$ shows the
induced SPSA gradient bias is
$O(1/m_t) + O(c_t^2) + O\!\bigl(1/(c_t m_t^2)\bigr)$ --- not the naive
$O(1/(c_t m_t))$ obtained by treating $B/m$ as an unstructured error.
The third term stems from the $O(m^{-2})$ remainder $r_m$ of
\eqref{eq:mc-bias}, whose $\psi$-derivatives are \emph{not} controlled
by Lemma~\ref{lem:mc-logmean-bias}. This term is dominated by the $O(1/m_t)$
term as soon as $m_t \ge c_t^{-1}$.  The extra estimator variance
contributes $O(1/(c_t^2 m_t))$ to the martingale term.  Almost-sure
consistency is therefore guaranteed under batch sizes $m_t \to \infty$ with
$\sum_t a_t/m_t < \infty$ and (unless $m_t \ge c_t^{-1}$ eventually)
$\sum_t a_t/(c_t m_t^2) < \infty$ --- bias summability, paralleling
$\sum_t a_t c_t^2 < \infty$ in Assumption~\ref{ass:ts-step} --- together
with $\sum_t a_t^2/(c_t^2 m_t) < \infty$ for variance summability.  Neither
variant is uniformly preferable: the Monte-Carlo variant trades the
surrogate gap $\varepsilon_{\mathrm{tot}}$ of
Proposition~\ref{prop:actor-surrogate-gap} for an $O(1/m_t)$ bias and an
$O(1/(c_t^2 m_t))$ variance inflation per step. The critic-based
estimate \eqref{eq:actor-ce-estimate-appendix} is the variant analyzed
(Corollary~\ref{cor:actor-surrogate-spsa}) and implemented.  See
\citet[Ch.~3]{vaart1998asymptotic} for the general delta method and
\citet{spall1992multivariate} for the underlying SPSA framework.
\end{remark}

\subsection{Tightness of the Surrogate-Optimality Transfer}
\label{app:actor-tightness}

Section~\ref{app:ce-vs-kl} bounds the critic error relative to the scaled nested value, and Section~\ref{app:static-nested} compares the nested and static criteria. The following proposition gives the global and local transfer guarantees in Section~\ref{sec:actor} when the total surrogate error is bounded.

\begin{proposition}[Surrogate-optimality transfer]
\label{prop:actor-surrogate-gap}
Let $\Psi\subseteq\mathbb{R}^d$ be nonempty, $\beta\neq0$, and suppose
$\sup_{\psi\in\Psi}|J_c(\psi)-J_\beta(\psi)|\le\varepsilon_{\mathrm{tot}}$
for a finite constant $\varepsilon_{\mathrm{tot}}\ge0$ (the
certainty-equivalent-scaled sum of the critic optimization floor, LFA
approximation, variational model, Donsker--Varadhan relaxation, and
nested-vs-static errors).  Then (i) every
$\psi_c\in\arg\max_{\psi\in\Psi}J_c(\psi)$ satisfies
$J_\beta(\psi_c)\ge\sup_{\psi\in\Psi}J_\beta(\psi)
-2\varepsilon_{\mathrm{tot}}$, and if $J_c(\psi')\ge\sup_\Psi J_c-\delta$
then $J_\beta(\psi')\ge\sup_\Psi J_\beta-2\varepsilon_{\mathrm{tot}}-\delta$.
(ii) The same holds over any $U\subseteq\Psi$ on which $\psi^\star$
maximizes $J_c$.
\end{proposition}

\begin{proof}
If $J_c(\psi')\ge\sup_\Psi J_c-\delta$, the uniform error bound gives
\[
J_\beta(\psi')\ge J_c(\psi')-\varepsilon_{\mathrm{tot}}
\ge\sup_\Psi J_c-\delta-\varepsilon_{\mathrm{tot}}
\ge\sup_\Psi J_\beta-\delta-2\varepsilon_{\mathrm{tot}}.
\]
A global maximizer satisfies this inequality with $\delta=0$. Applying the same argument with both suprema taken over $U$ proves the local claim.
\end{proof}

\begin{remark}
\label{remark:actor-local}
Part (ii) requires $\psi^{\star}$ to \emph{maximize} $J_c$ over the
comparison region.  At a general critical point of $J_c$ no analogous
guarantee is possible without further curvature assumptions: two
functions at sup-distance $\varepsilon_{\mathrm{tot}}$ can have critical
points whose $J_\beta$-value falls short of $\sup J_\beta$ by the full
oscillation of $J_\beta$ (take $J_\beta$ two-peaked with peak gap much
larger than $\varepsilon_{\mathrm{tot}}$ and $J_c$ possessing a critical
point near the lower peak).  Under the strict-local-maximum condition of
\S\ref{app:proof-twosided} imposed on $J_c$, the iterates converge almost surely to such a
maximizer $\psi^{\star}$, and (ii) applies with $U$ the corresponding
basin.  Note also that sup-norm closeness controls values, not
gradients: $\nabla J_c(\psi^{\star})=0$ does not imply
$\|\nabla J_\beta(\psi^{\star})\|$ is small.
\end{remark}

\begin{remark}[The factor $2$ in Proposition~\ref{prop:actor-surrogate-gap} is sharp]
\label{remark:actor-tightness}
Let $\varepsilon_{\mathrm{tot}}>0$,
$\varsigma\in(0,\varepsilon_{\mathrm{tot}})$, and take
$\Psi=\{\psi_1,\psi_2\}$ with
\[
  J_\beta(\psi_1) = 2\varepsilon_{\mathrm{tot}}-\varsigma,
  \quad
  J_\beta(\psi_2) = 0,
  \qquad
  J_c(\psi_1) = \varepsilon_{\mathrm{tot}}-\varsigma,
  \quad
  J_c(\psi_2) = \varepsilon_{\mathrm{tot}} .
\]
Then $|J_c(\psi_i)-J_\beta(\psi_i)| = \varepsilon_{\mathrm{tot}}$ for
$i=1,2$, the unique maximizer of $J_c$ is $\psi_2$, and
$J_\beta(\psi_2) = \sup_{\Psi}J_\beta -
(2\varepsilon_{\mathrm{tot}}-\varsigma)$.  Letting $\varsigma\downarrow
0$ shows that no constant smaller than $2$ can replace the one in
Proposition~\ref{prop:actor-surrogate-gap}(i), and the same two-point
configuration embedded in a subregion $U$ shows sharpness of part~(ii).
Such sign-alternating surrogate errors are realizable, e.g.\ by a
one-parameter softmax policy interpolating between two deterministic
policies whose feature-approximation residuals have opposite signs.
\end{remark}

\section{More Related Work}
\label{appendix:related-work}
This appendix relates our method to Relative Entropy Policy Search (REPS) and control as inference.

\subsection{Connection with Relative Entropy Policy Search}

Relative Entropy Policy Search maximizes expected return subject to a KL-divergence constraint on the updated state-action distribution relative to a reference distribution \citep{peters2010relative}. Subsequent work has developed sample-based formulations and guarantees for relative entropy policy search \citep{van2017non, pacchiano2021near}. Relaxed-RSTD uses the Donsker--Varadhan representation of the next-state log-partition term in the nested entropic Bellman equation (Equation~\ref{eq:bellman}). Its variational distribution $q_\omega$ models next states under a fixed policy. The parameter $\beta$ sets the risk preference in the nested criterion.

\subsection{Connection with Control-as-Inference}

Control-as-inference methods formulate policy optimization as probabilistic inference. Variational formulations include bounds on trajectory objectives and risk-sensitive extensions \citep{chow2020variational, ito2024risk, granados2025risk}. In Relaxed-RSTD, the variational bound is applied to the next-state expectation for a fixed policy. When $q_\omega=p$, the log-likelihood relaxation adds a transition-entropy term to the critic reward (Section~\ref{app:ce-vs-kl}). The actor updates the policy using differences between the estimated certainty equivalents of perturbed policies (Section~\ref{sec:actor}).

\section{Experiments}
\label{appendix:experiments}
\suppressfloats[t]
Section~\ref{appendix:implementation-details} gives the experimental setup. Section~\ref{sec:gridworlds-windy} presents Windy Gridworld results, Section~\ref{sec:portfolio-experiments} expands on the portfolio optimization analysis, and Section~\ref{appendix:anm} presents the active network management results. Section~\ref{appendix:feature_dim_ablation} examines how critic feature dimension affects risk-sensitive learning on the MuJoCo and continuous Mountain Car benchmarks.

\subsection{Implementation Details}
\label{appendix:implementation-details}

The variational distribution $q(\cdot\mid s;\omega)$ in Relaxed-RS-AC is parameterized according to the nature of the action space: we utilize a learned discrete embedding for environments with discrete actions, and a Gaussian distribution for continuous control tasks. In the continuous regime, we apply spectral normalization \citep{miyato2018spectral} to the network weights to ensure Lipschitz continuity. The policy representations are uniformly implemented using single-layer neural networks across all settings. For discrete action spaces, the network maps the state space to categorical logits, from which actions are sampled via a softmax distribution. For continuous action spaces, the network outputs the mean and log-standard deviation of a diagonal Gaussian distribution used for action sampling. Both the linear function approximation parameters and the variational distribution weights are optimized using Adam \citep{kingma2014adam}. To ensure a rigorous and fair empirical evaluation, all baseline methods are adapted directly from their official, author-provided implementations with strict architectural parity enforced. Specifically, we replace all baseline critics with linear function approximators utilizing the identical state-space feature representation $\Phi$ as Relaxed-RS-AC. Furthermore, we standardize the actor architectures to match our single-layer implementations and explicitly restrict any natively off-policy baselines to operate in a strictly on-policy regime. Table~\ref{tab:comprehensive_hyperparams_1} lists the Relaxed-RS-AC hyperparameters for CliffWalking and Windy Gridworld (shared under Gridworlds), portfolio optimization, glycemic control, and active network management. Table~\ref{tab:continuous_hyperparams} lists the hyperparameters for the continuous control benchmarks in the feature-dimension ablation.

All learning experiments were conducted on machines equipped with an NVIDIA GeForce RTX 3050 OEM GPU (8 GB VRAM), 64 GB of system memory, and a 32-core CPU running PyTorch 2.9.0.

\begin{table}[htbp]
\footnotesize
\centering
\setlength{\tabcolsep}{4pt}
\begin{tabular}{lcccc}
\toprule
\textbf{Hyperparameter} & \textbf{Gridworlds} & \shortstack{\textbf{Portfolio}\\\textbf{optimization}} & \shortstack{\textbf{Glycemic}\\\textbf{control}} & \shortstack{\textbf{Active network}\\\textbf{management}} \\ 
\midrule
$\theta$ learning rate & 0.0005 & 0.0005 & 0.001 & 0.001 \\ 
$\gamma$ (discount factor) & 0.99 & 0.99 & 0.99 & 0.99 \\ 
$\lambda$ & 0.01 & 0.01 & 0.01 & 0.01 \\ 
$k$ (feature dimension) & 20 & 20 & 3 & 10 \\ 
$\mathcal{D}_{KL}$ scaling factor & 1.0 & 1.0 & 0.01 & 1.0 \\ 
\midrule
Actor update steps & 10 & 500 & 500 & 50 \\ 
Actor initial learning rate & 0.03 & 0.005 & 0.005 & 0.005 \\ 
Actor learning rate decay rate & 0.5 & 0.5 & 0.5 & 0.5 \\ 
Actor learning rate stability constant & 0.001 & 0.001 & 0.001 & 0.001 \\ 
Initial perturbation size & 0.2 & 0.2 & 0.2 & 0.2 \\ 
Perturbation stability constant & 0.161 & 0.161 & 0.161 & 0.161 \\ 
Perturbation decay rate & 0.001 & 0.001 & 0.001 & 0.001 \\ 
\midrule
Actor Optimizer & Adam & Adam & Adam & Adam \\ 
First moment decay rate & 0.9 & 0.9 & 0.9 & 0.9 \\ 
Second moment decay rate & 0.999 & 0.999 & 0.999 & 0.999 \\ 
Denominator stability constant & $1\times 10^{-7}$ & $1\times 10^{-7}$ & $1\times 10^{-7}$ & $1\times 10^{-7}$ \\ 
\midrule
Train episodes & 500 & 150 & 100 & 100 \\ 
Max episode steps & 200 & - & 200 & 200 \\ 
\bottomrule
\end{tabular}
\caption{\textbf{Hyperparameters for Relaxed-RS-AC across experimental domains.} Parameters shared universally across environments denote algorithmic defaults, while environment-specific parameters (such as feature dimension $k$ and training episodes) were tuned to accommodate the varying complexities of the respective state-action spaces.}
\label{tab:comprehensive_hyperparams_1}
\end{table}

\begin{table}[htbp]
\centering
\begin{tabular}{lcc}
\toprule
\textbf{Hyperparameter} & \textbf{MuJoCo} & \textbf{Mountain-Car-Continuous} \\ 
\midrule
$\theta$ learning rate & 0.001 & 0.001 \\ 
$\gamma$ (discount factor) & 0.99 & 0.99 \\ 
$\lambda$ & 0.01 & 0.01 \\ 
$k$ (feature dimension) & 32, 64, 128 & 32, 64, 128 \\ 
$\mathcal{D}_{KL}$ scaling factor & 1.0 & 1.0 \\ 
\midrule
Actor update steps & 500 & 500 \\ 
Actor initial learning rate & 0.005 & 0.005 \\ 
Actor learning rate decay rate & 0.5 & 0.5 \\ 
Actor learning rate stability constant & 0.001 & 0.001 \\ 
Initial perturbation size & 0.2 & 0.2 \\ 
Perturbation stability constant & 0.161 & 0.161 \\ 
Perturbation decay rate & 0.001 & 0.001 \\ 
\midrule
Actor Optimizer & Adam & Adam \\ 
First moment decay rate & 0.9 & 0.9 \\ 
Second moment decay rate & 0.999 & 0.999 \\ 
Denominator stability constant & $1\times 10^{-7}$ & $1\times 10^{-7}$ \\ 
\midrule
Train episodes & 500, 40 & 50 \\ 
Max episode steps & 1000 & 1000 \\ 
\bottomrule
\end{tabular}
\caption{\textbf{Hyperparameters for Relaxed-RS-AC in continuous control tasks.} To evaluate the effect of representation capacity, the feature dimension ($k$) was varied across 32, 64, and 128 for all continuous environments. Training episodes for MuJoCo were environment-dependent: 500 episodes for \texttt{Inverted-Double-Pendulum/v5} and 40 episodes for \texttt{Hopper/v5}.}
\label{tab:continuous_hyperparams}
\end{table}

\clearpage
\subsection{Windy Gridworld}
\label{sec:gridworlds-windy}
\begin{figure}[htbp]
    \centering
    \begin{subfigure}[b]{0.32\textwidth}
        \centering
        \includegraphics[width=\linewidth]{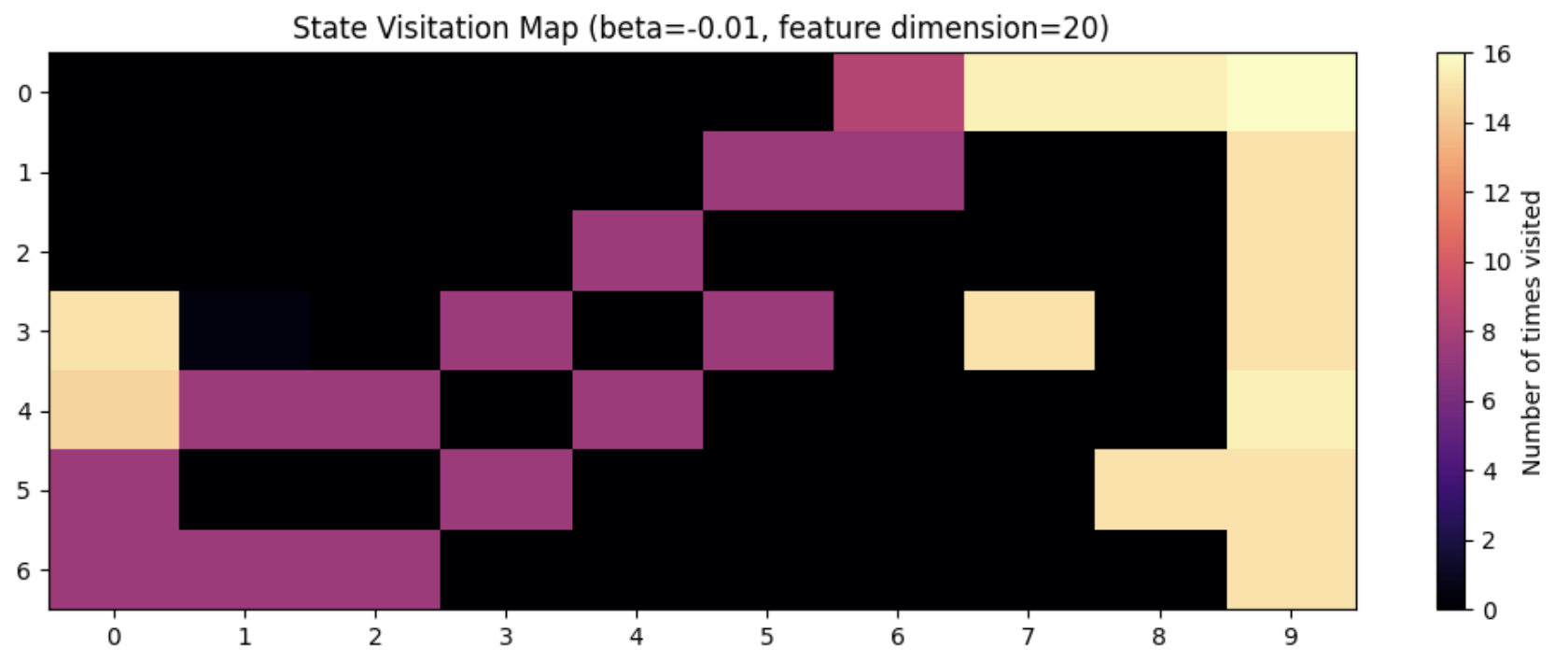}
        \caption{Relaxed-RS-AC $(\beta=-0.01)$}
        \label{fig:windy_averse}
    \end{subfigure}
    \hfill 
    \begin{subfigure}[b]{0.32\textwidth}
        \centering
        \includegraphics[width=\linewidth]{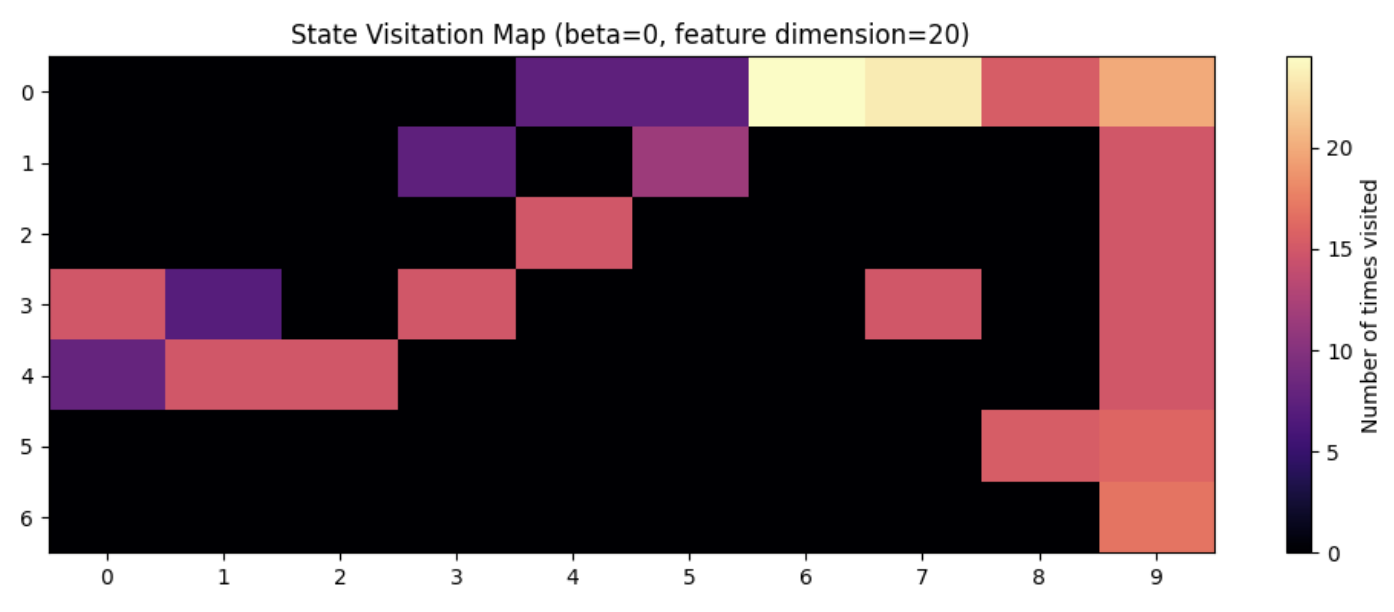}
        \caption{Relaxed-RS-AC $(\beta=0)$}
        \label{fig:windy_neutral}
    \end{subfigure}
    \hfill
    \begin{subfigure}[b]{0.32\textwidth}
        \centering
        \includegraphics[width=\linewidth]{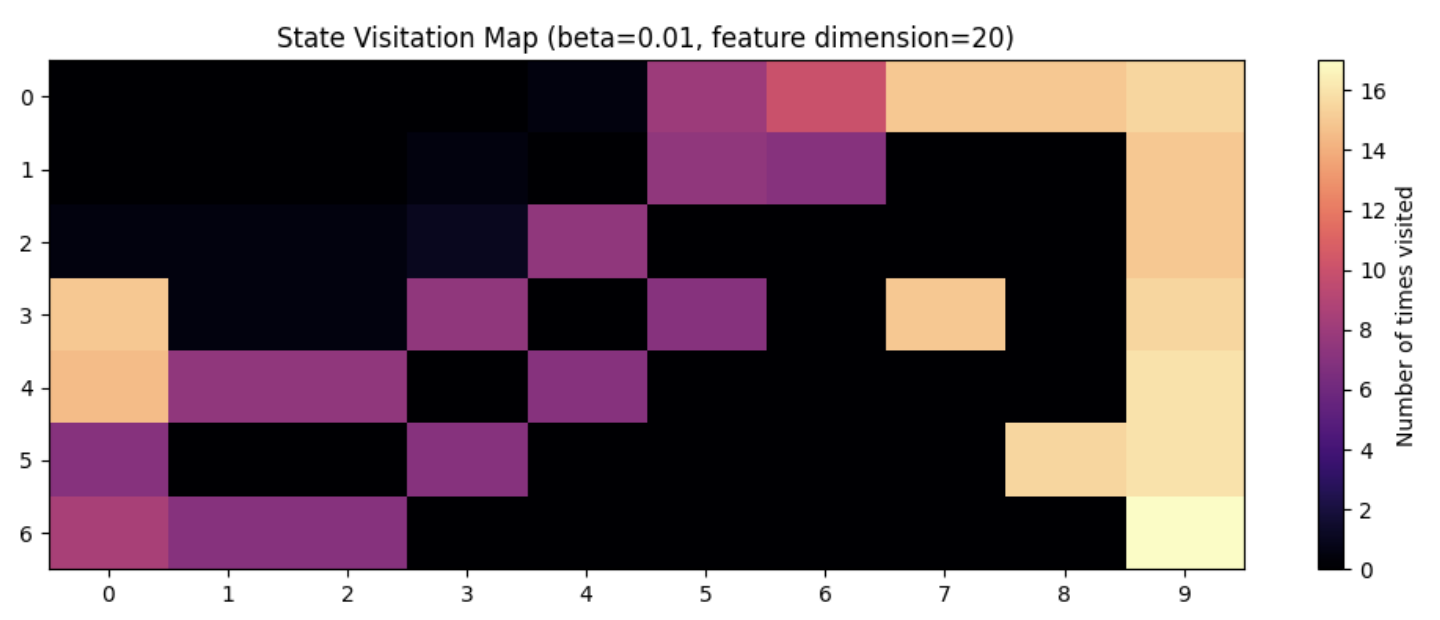}
        \caption{Relaxed-RS-AC $(\beta=0.01)$}
        \label{fig:windy_seeking}
    \end{subfigure}
    
    \par\medskip
    
    \begin{subfigure}[b]{0.32\textwidth}
        \centering
        \includegraphics[width=\linewidth]{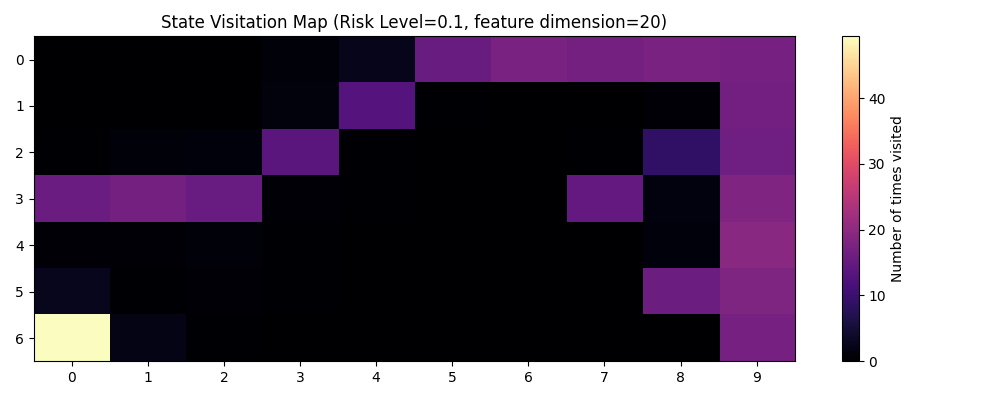}
        \caption{CVaR-PG}
        \label{fig:windy_cvarpg}
    \end{subfigure}
    \hfill
    \begin{subfigure}[b]{0.32\textwidth}
        \centering
        \includegraphics[width=\linewidth]{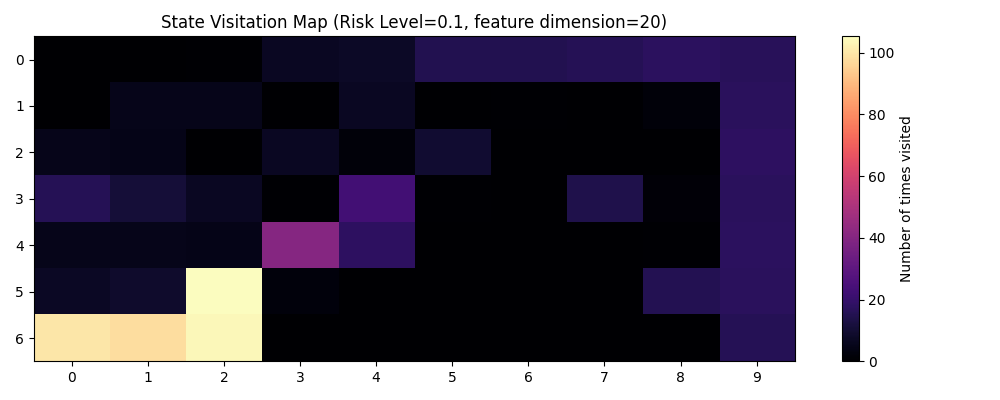}
        \caption{D4PG-CVaR}
        \label{fig:windy_d4pgcvar}
    \end{subfigure}
    \hfill
    \begin{subfigure}[b]{0.32\textwidth}
        \centering
        \includegraphics[width=\linewidth]{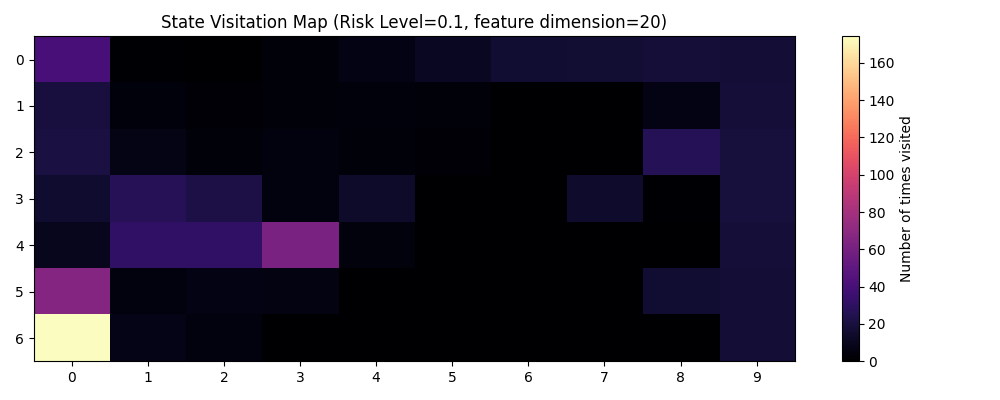}
        \caption{SDPG-CVaR}
        \label{fig:windy_sdpgcvar}
    \end{subfigure}
    
    \caption{\textbf{State visitation frequencies in the Windy Gridworld environment.} Heatmaps aggregate agent trajectories over evaluation episodes. Lighter colors indicate higher state visitation. \textbf{(a)--(c)} Relaxed-RS-AC demonstrates a clear behavioral spectrum: the risk-averse agent ($\beta=-0.01$) builds a spatial safety buffer by conservatively navigating the lower rows, while the risk-seeking agent ($\beta=0.01$) follows a wind-assisted route. \textbf{(d)--(f)} The CVaR baselines use risk level $0.1$ and show either wider detours or more scattered visitation.}
    \label{fig:windy_heatmaps}
\end{figure}

Windy Gridworld tests how Relaxed-RS-AC and the CVaR baselines respond to a stochastic crosswind. We use the same Relaxed-RS-AC settings ($\beta\in\{-0.01,0,0.01\}$), feature dimension ($20$), and CVaR risk level ($0.1$) as in CliffWalking (Section~\ref{sec:gridworlds}). Table~\ref{tab:windy_final_rewards} reports the final trajectory lengths, and Figure~\ref{fig:windy_heatmaps} shows the routes.

\begin{table}[htbp]
    \centering
    \footnotesize
    \setlength{\tabcolsep}{4pt}
    \begin{tabular}{lc}
        \toprule
        \textbf{Algorithm} & \textbf{Final Lengths $(\downarrow)$} \\
        \midrule
        Relaxed-RS-AC ($\beta = -0.01$) & $17.97 \pm 0.86$ \\
        Relaxed-RS-AC ($\beta = 0$)     & $18.12 \pm 1.70$ \\
        Relaxed-RS-AC ($\beta = 0.01$)  & $17.25 \pm 1.30$ \\
        CVaR-PG                          & $17.40 \pm 2.35$ \\
        D4PG-CVaR                        & $19.76 \pm 1.73$ \\
        SDPG-CVaR                        & $20.00 \pm 1.94$ \\
        \bottomrule
    \end{tabular}
    \caption{\textbf{Final trajectory lengths in Windy Gridworld} (mean $\pm$ std. dev. over 5 seeds).}
    \label{tab:windy_final_rewards}
\end{table}

The risk-averse agent follows lower rows, giving the smallest standard deviation in Table~\ref{tab:windy_final_rewards} at $17.97\pm0.86$ steps. The risk-seeking agent follows a wind-assisted route and has the shortest reported mean at $17.25\pm1.30$ steps. The risk-neutral agent takes an intermediate route and records $18.12\pm1.70$ steps. In Figure~\ref{fig:windy_heatmaps}, D4PG-CVaR and SDPG-CVaR make wider detours and have the longest mean trajectories. CVaR-PG has a competitive mean of $17.40$ steps and the highest reported standard deviation of $2.35$ steps.

\clearpage
\subsection{Portfolio Optimization}
\label{sec:portfolio-experiments}

We evaluate Relaxed Risk-Sensitive Actor-Critic (Relaxed-RS-AC) on a real-world equity portfolio optimization task spanning January 2013 to December 2023, a decade that contains two canonical
stress-event regimes: the COVID-19 market crash of March 2020 (a rapid tail-event drawdown and recovery, falling in our validation year) and the 2022 Federal Reserve tightening cycle (a sustained bear market in growth
equities, falling in our held-out test window). These two episodes stress-test the dual claims of exponential-utility optimization: (i) that risk-averse parameterizations
($\beta<0$) selectively suppress downside trajectories through the exponential penalty, while (ii) risk-seeking parameterizations ($\beta>0$)
amplify variance in the return distribution. We train models on data from January 2013 to December 2019 and test on data from January 2021 to December 2023. The data from January 2020 to December 2020 is considered the validation set and is used for model selection. \S\ref{sec:portfolio_optimization_main} summarizes a representative subset of configurations. All backtests reported in this section are therefore strictly out-of-sample: every figure spans the held-out test window (January 2021--December 2023), with cumulative returns indexed to $100\%$ at the start of the window. We compare against two distributional soft-actor-critic (DSAC) baselines, DSAC-Wang~\citep{bellemare2023distributional} and DSAC-CVaR, which represent the dominant risk-sensitive distributional RL approaches. We describe the experimental setup below.

\paragraph{Asset universe and state representation.}
The investable universe consists of ten large-cap US equities drawn
from the S\&P~500: Apple (AAPL), Microsoft (MSFT), Johnson \& Johnson
(JNJ), UnitedHealth Group (UNH), Visa (V), Home Depot (HD),
Walmart (WMT), Coca-Cola (KO), JPMorgan Chase (JPM), and
Procter \& Gamble (PG), together with a rest-of-market
proxy (Others).
These span five major GICS sectors—technology, healthcare, consumer
staples, consumer discretionary, and financials—providing both
high-$\beta$ growth assets and defensive dividend-paying assets,
thereby placing maximal stress on the risk-preference signal encoded
in $\beta$.

\paragraph{Action space and portfolio constraint.}
The policy outputs a vector $a_t\in\mathbb{R}^{d}$ of portfolio weights
with $\sum_i a_{ti}=1$.
Negative weights represent short positions. The policy is not
constrained to long-only, permitting leveraged strategies that emerge
naturally under risk-neutral and risk-seeking parameterizations
($\beta\ge0$, as discussed in \S\ref{sec:results-beta}).

\paragraph{Reward signal.}
At each daily timestep the reward is the portfolio's log-return
$r_t:=\log(1+\sum_i a_{ti}\tilde{r}_{ti})$,
where $\tilde{r}_{ti}$ is the adjusted closing-price return of asset $i$.

\paragraph{Hyperparameters.}
Full hyperparameter settings are reported in Table~\ref{tab:comprehensive_hyperparams_1}.

\paragraph{Evaluation metrics.}
We report historical compounded cumulative returns over the
test window, indexed to $100\%$, and historical uncompounded drawdown trajectories. The downside risk measures are the Ulcer Index (root-mean-squared drawdown), average and maximum drawdown, and the 95\% confidence-interval analogues of DaR, CDaR, EDaR, and RLDaR(0.3) \citep{chekhlov2005drawdown, z14}. We also report per-asset risk contribution relative to the equal-weight
benchmark, shown as a horizontal red line in the risk-contribution charts. The annualized Sharpe ratio is computed separately on the validation window (Jan 2020--Dec 2020) and the held-out test window (Jan 2021--Dec 2023). The validation-window Sharpe ratio is the
criterion used to select the reported $\beta$ for Relaxed-RS-AC and the reported configuration for each DSAC baseline in the main paper (\S\ref{sec:portfolio_optimization_main}).
Because the action space admits leveraged long--short positions, uncompounded drawdowns are not bounded below by $-100\%$ for highly levered policies.

For all of our portfolio-based plots, drawdown analysis and allocation summary, we employ RiskFolio \citep{riskfolio}.  
\subsubsection{Ablation Study: Effect of the Risk Parameter $\beta$}
\label{sec:results-beta}

\begin{figure*}[htbp]
    \centering
    \includegraphics[width=0.32\textwidth]{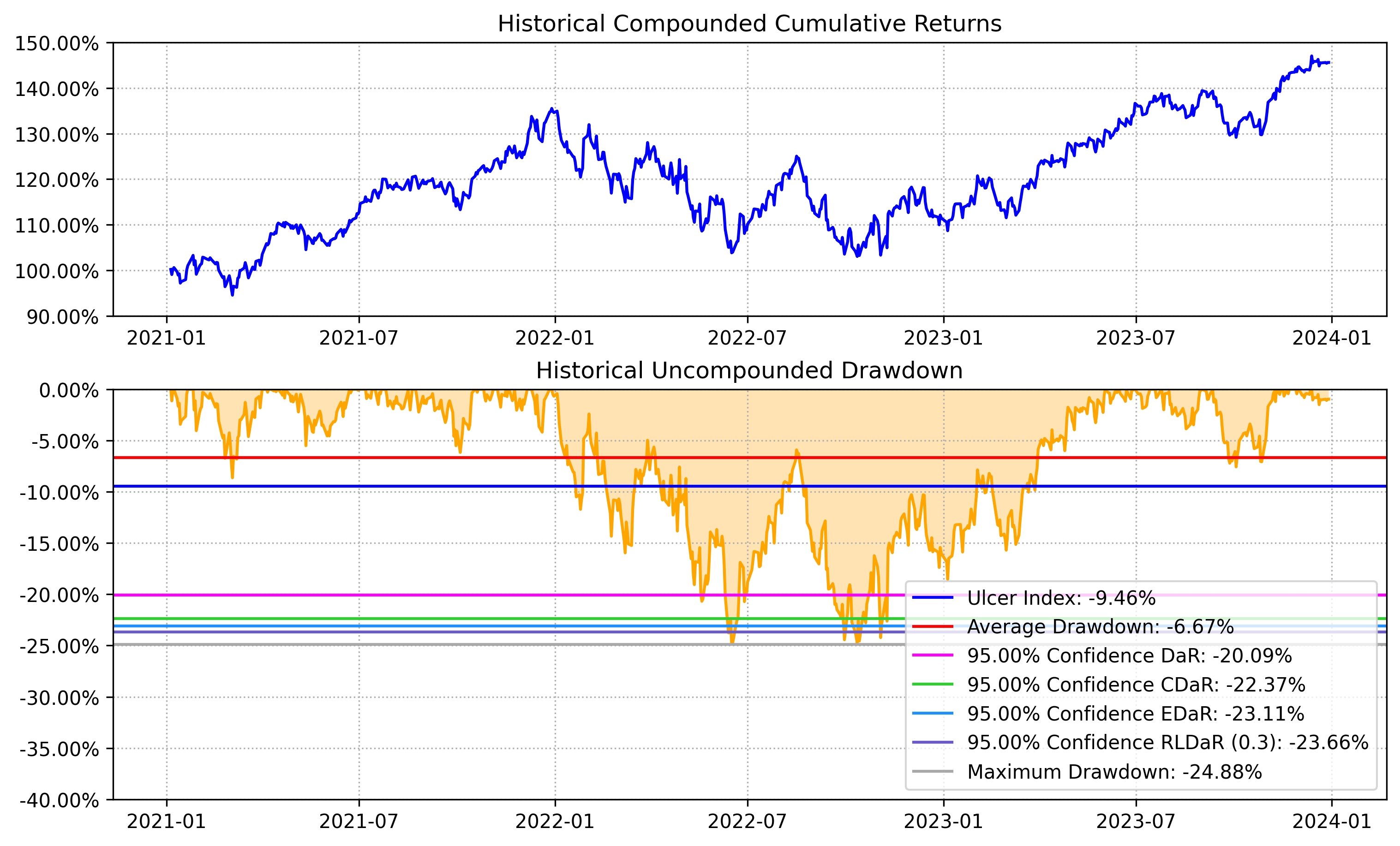}
    \hfill
    \includegraphics[width=0.32\textwidth]{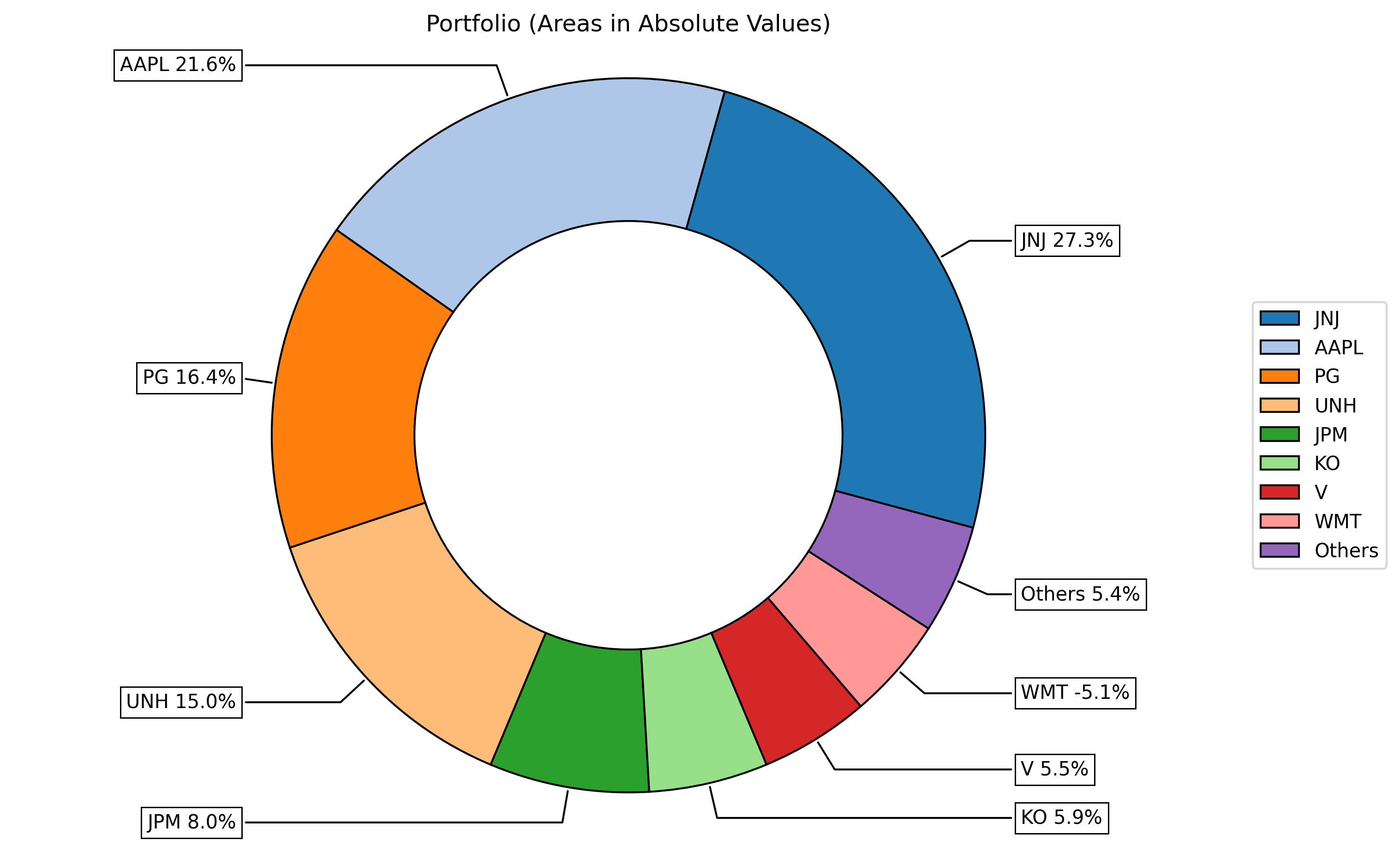}
    \hfill
    \includegraphics[width=0.32\textwidth]{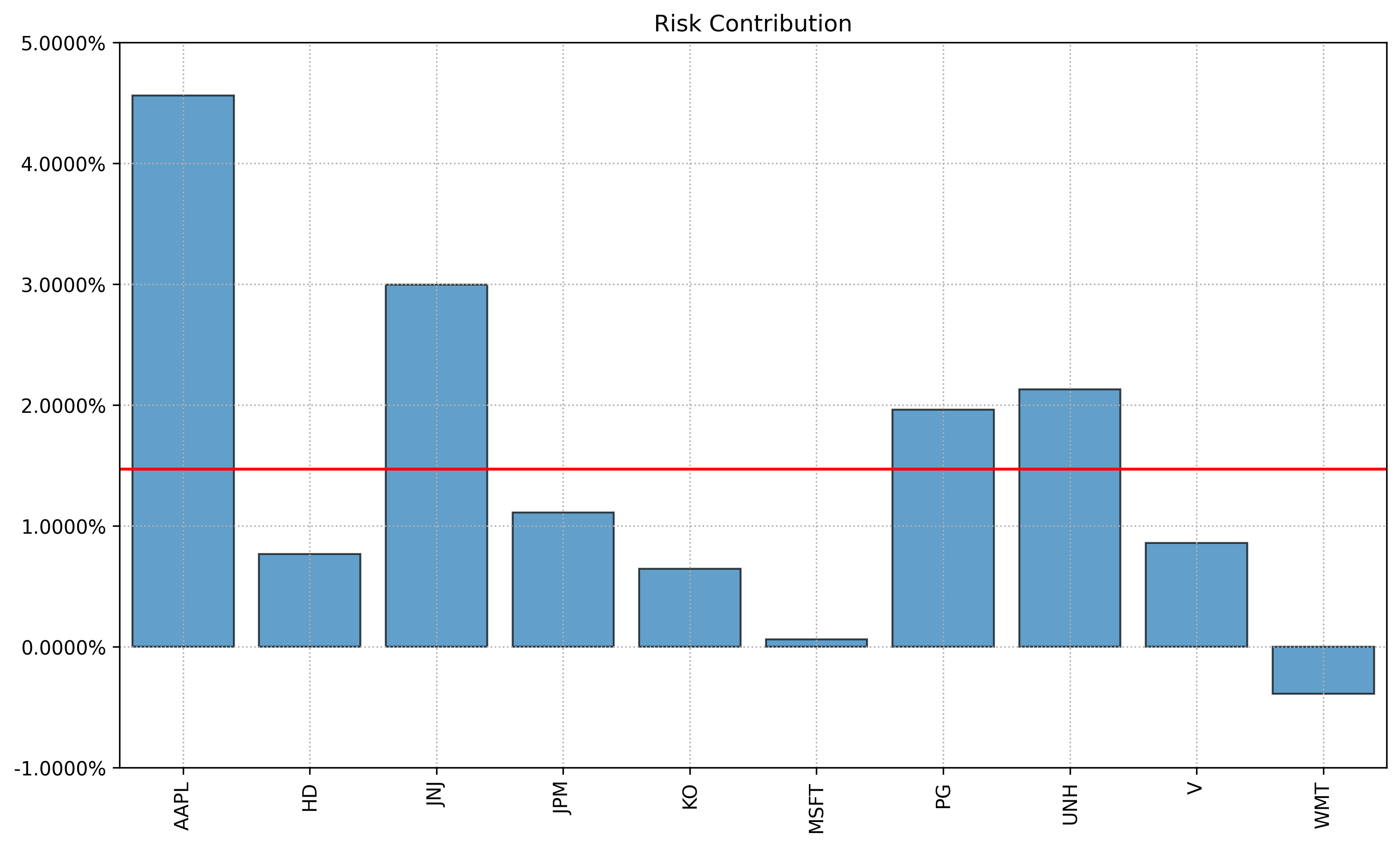}
    \caption{Historical cumulative-return and drawdown trajectories, mean asset allocation weights, and per-asset risk contribution profiles for the learned portfolio optimization policy for Relaxed-RS-AC ($\beta=-0.1$).}
    \label{fig:beta_minus_0_1}
\end{figure*}

\begin{figure*}[htbp]
    \centering
    \includegraphics[width=0.32\textwidth]{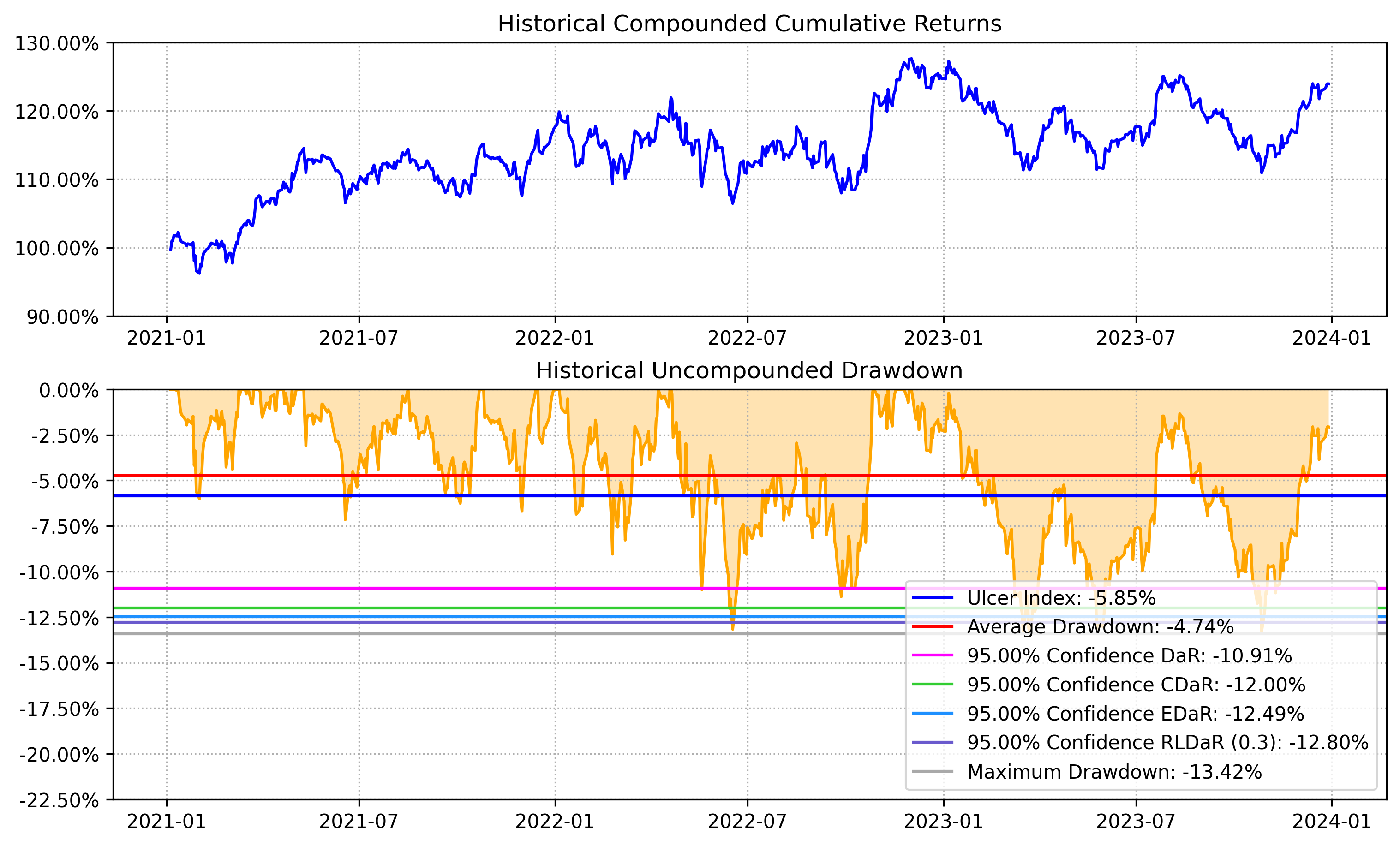}
    \hfill
    \includegraphics[width=0.32\textwidth]{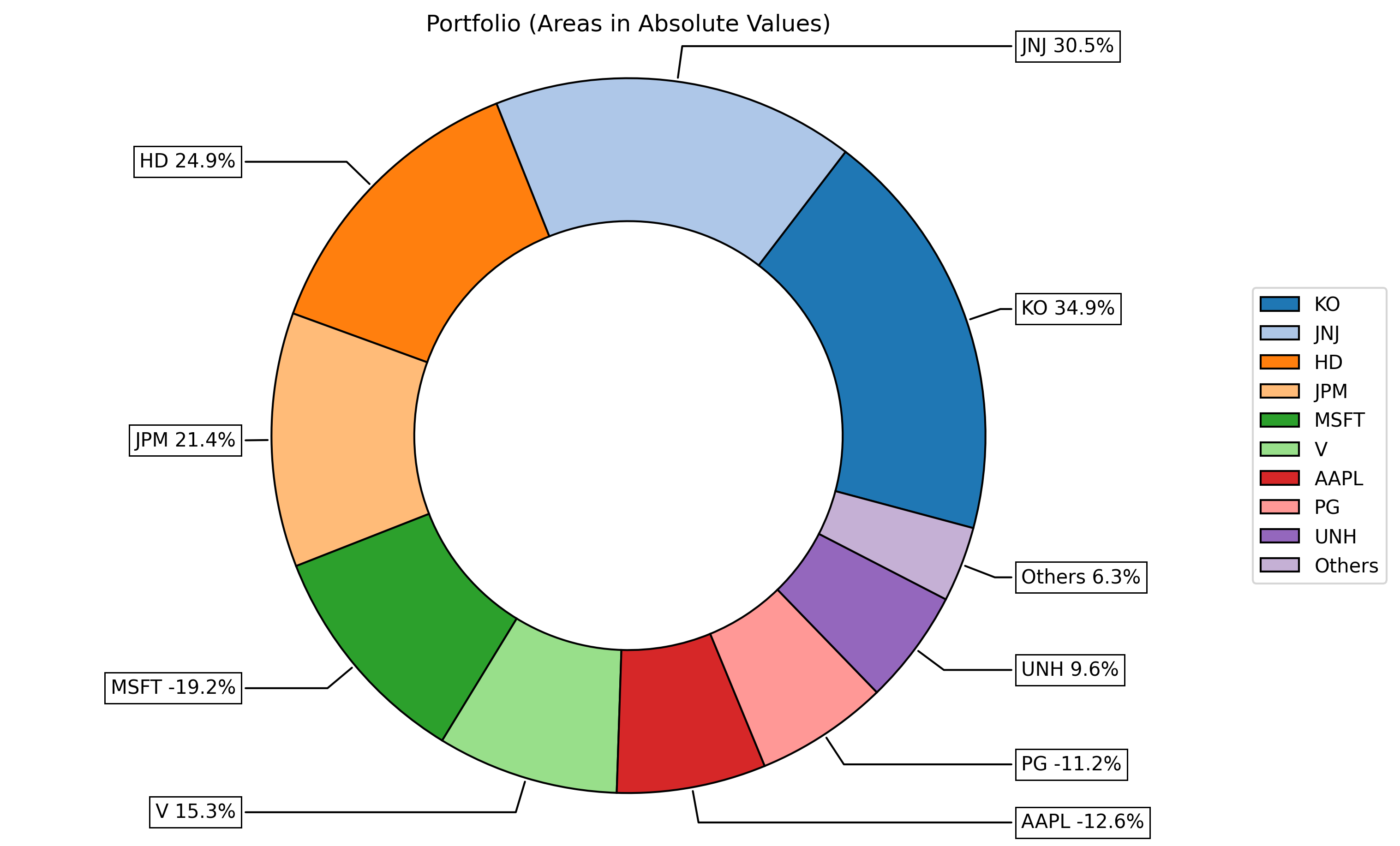}
    \hfill
    \includegraphics[width=0.32\textwidth]{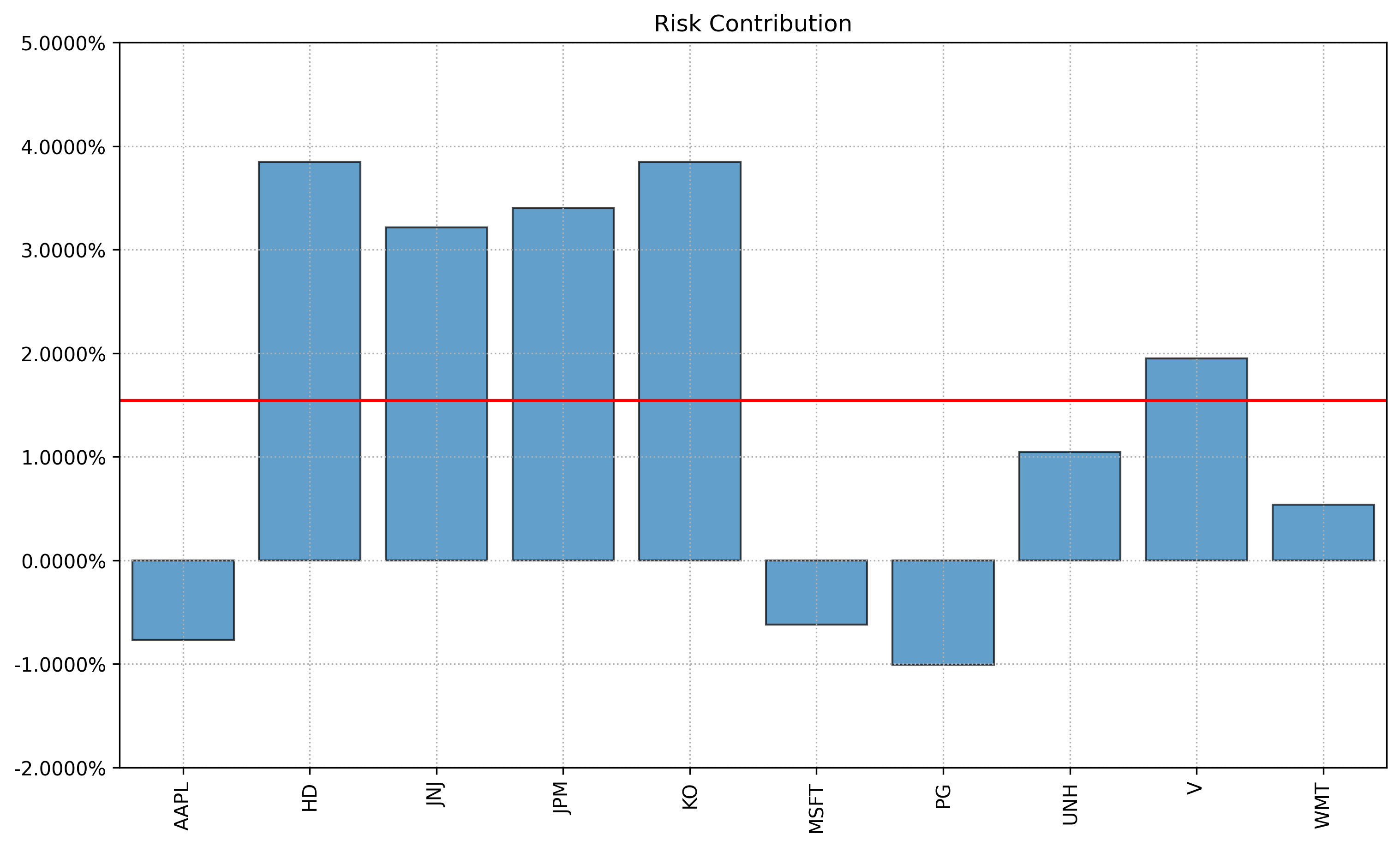}
    \caption{Historical cumulative-return and drawdown trajectories, mean asset allocation weights, and per-asset risk contribution profiles for the learned portfolio optimization policy for Relaxed-RS-AC ($\beta=-0.05$).}
    \label{fig:beta_minus_0_0_5}
\end{figure*}

\begin{figure*}[htbp]
    \centering
    \includegraphics[width=0.32\textwidth]{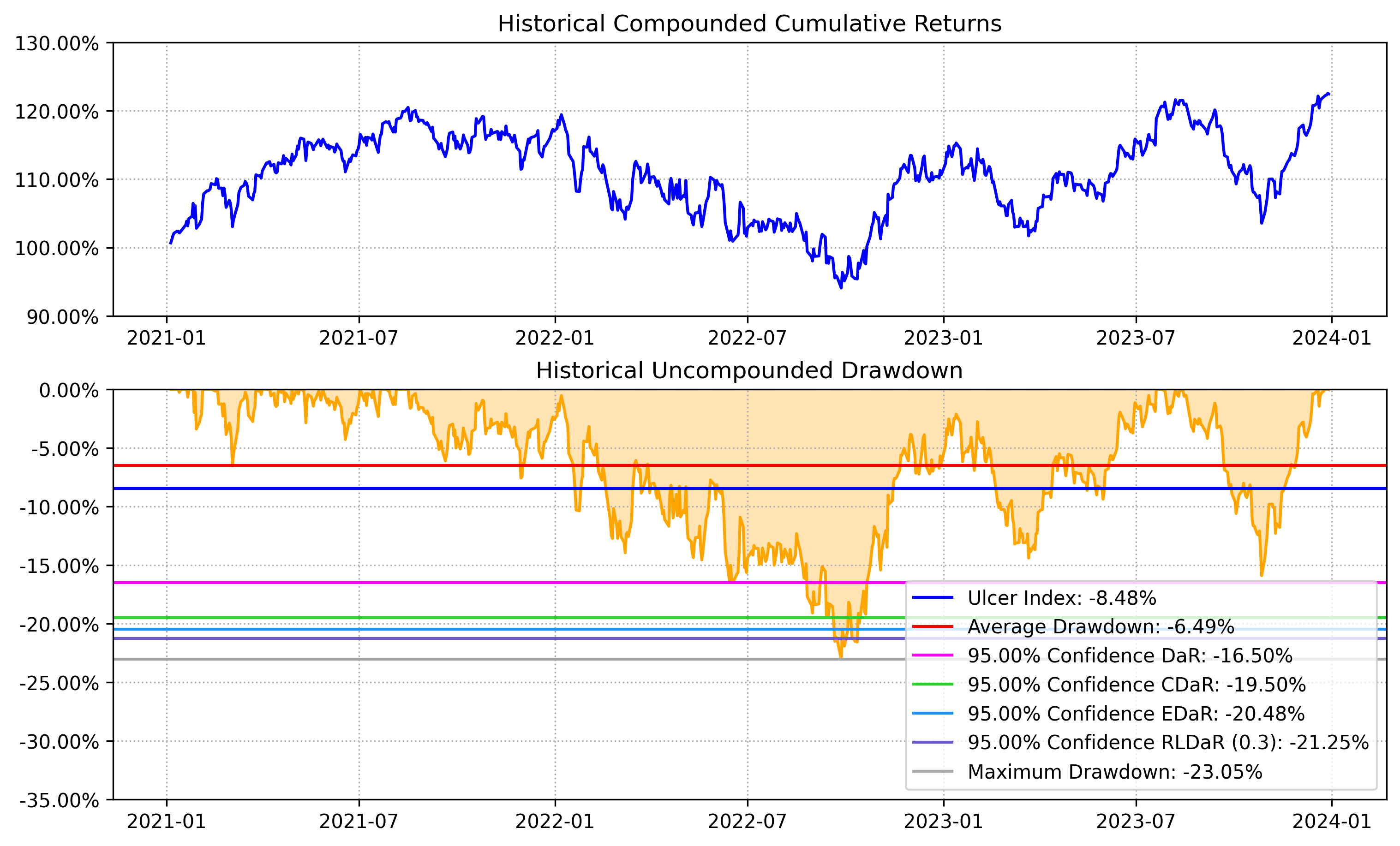}
    \hfill
    \includegraphics[width=0.32\textwidth]{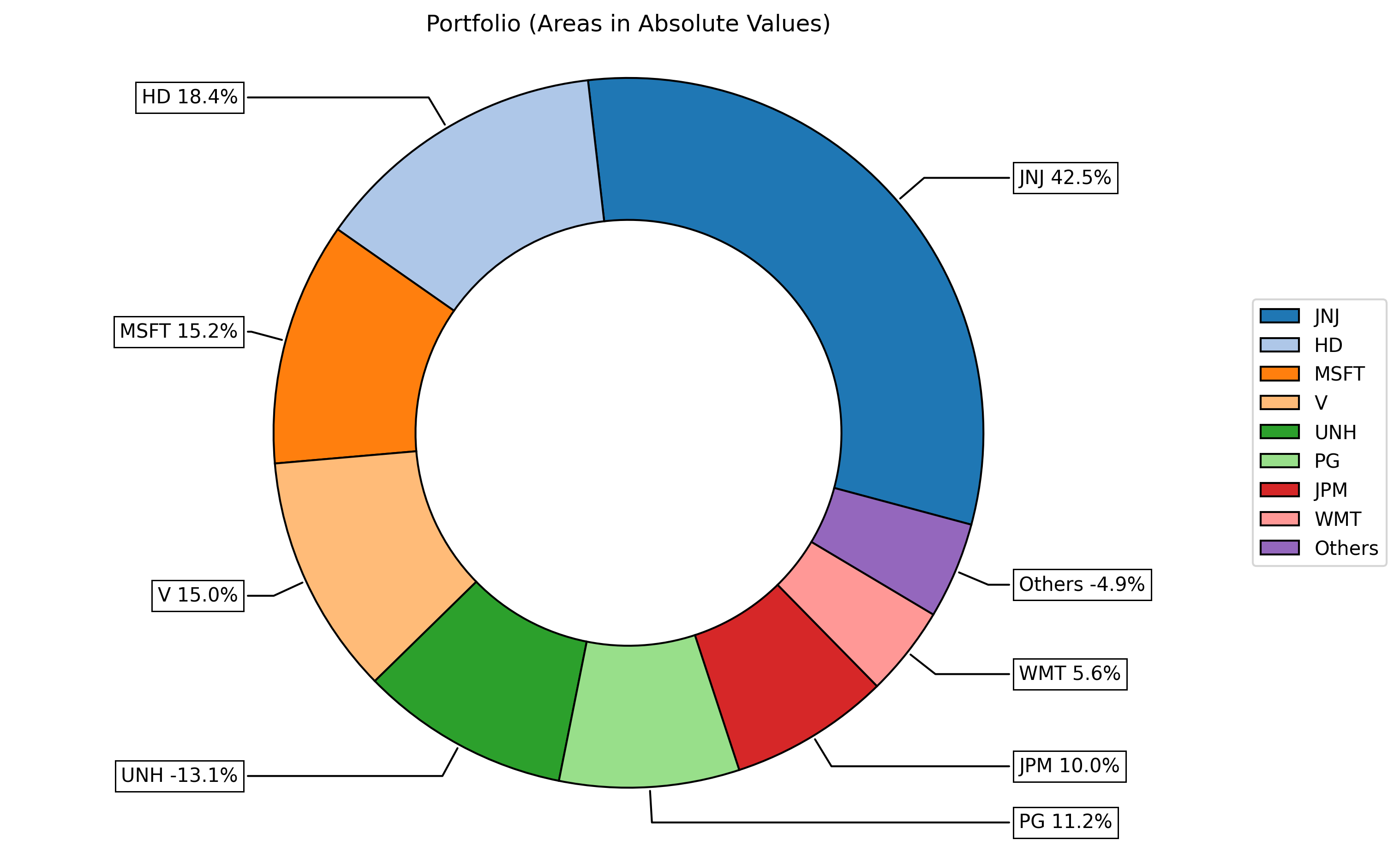}
    \hfill
    \includegraphics[width=0.32\textwidth]{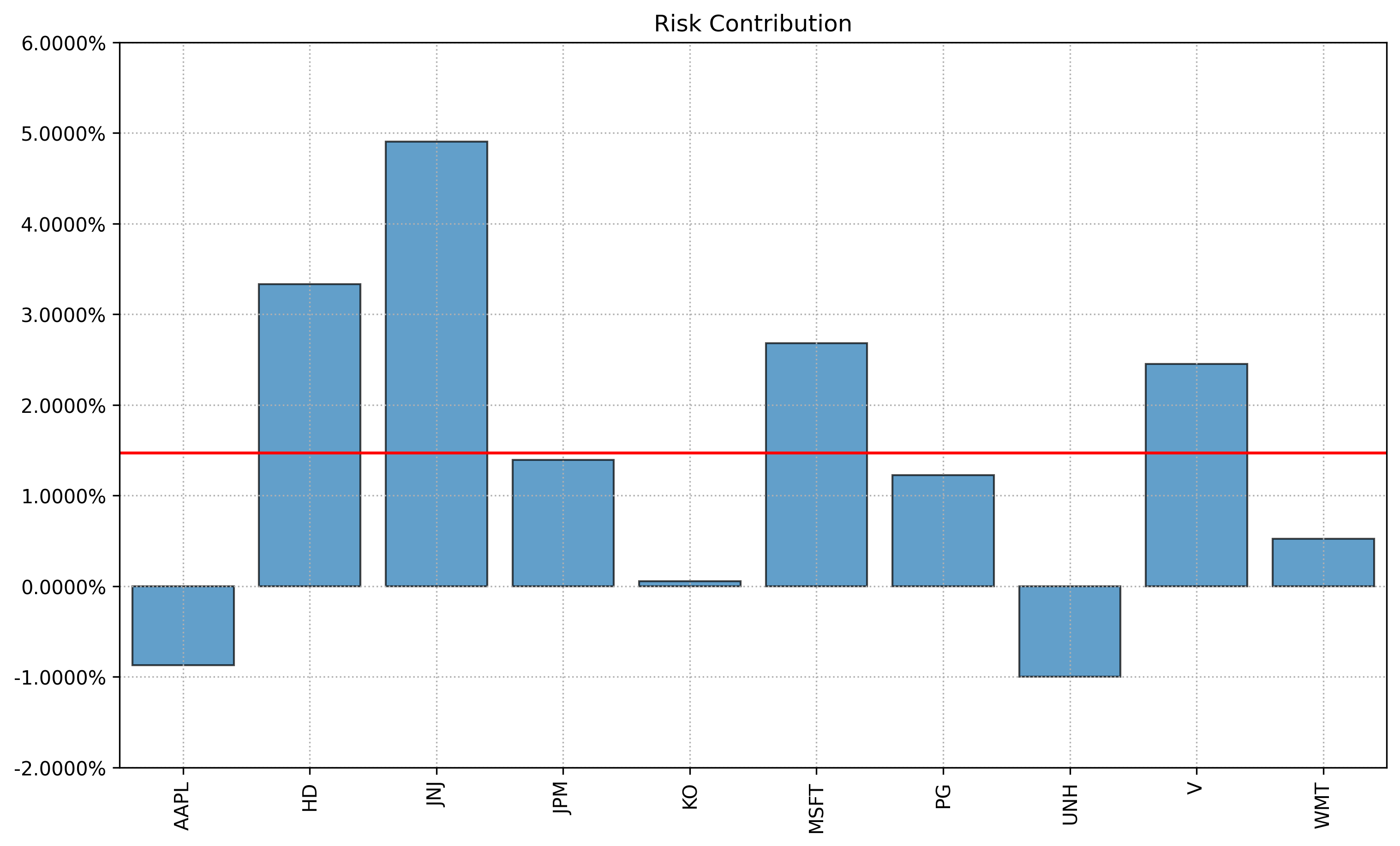}
    \caption{Historical cumulative-return and drawdown trajectories, mean asset allocation weights, and per-asset risk contribution profiles for the learned portfolio optimization policy for Relaxed-RS-AC ($\beta=-0.01$).}
    \label{fig:beta_minus_0_0_1}
\end{figure*}

\begin{figure*}[htbp]
    \centering
    \includegraphics[width=0.32\textwidth]{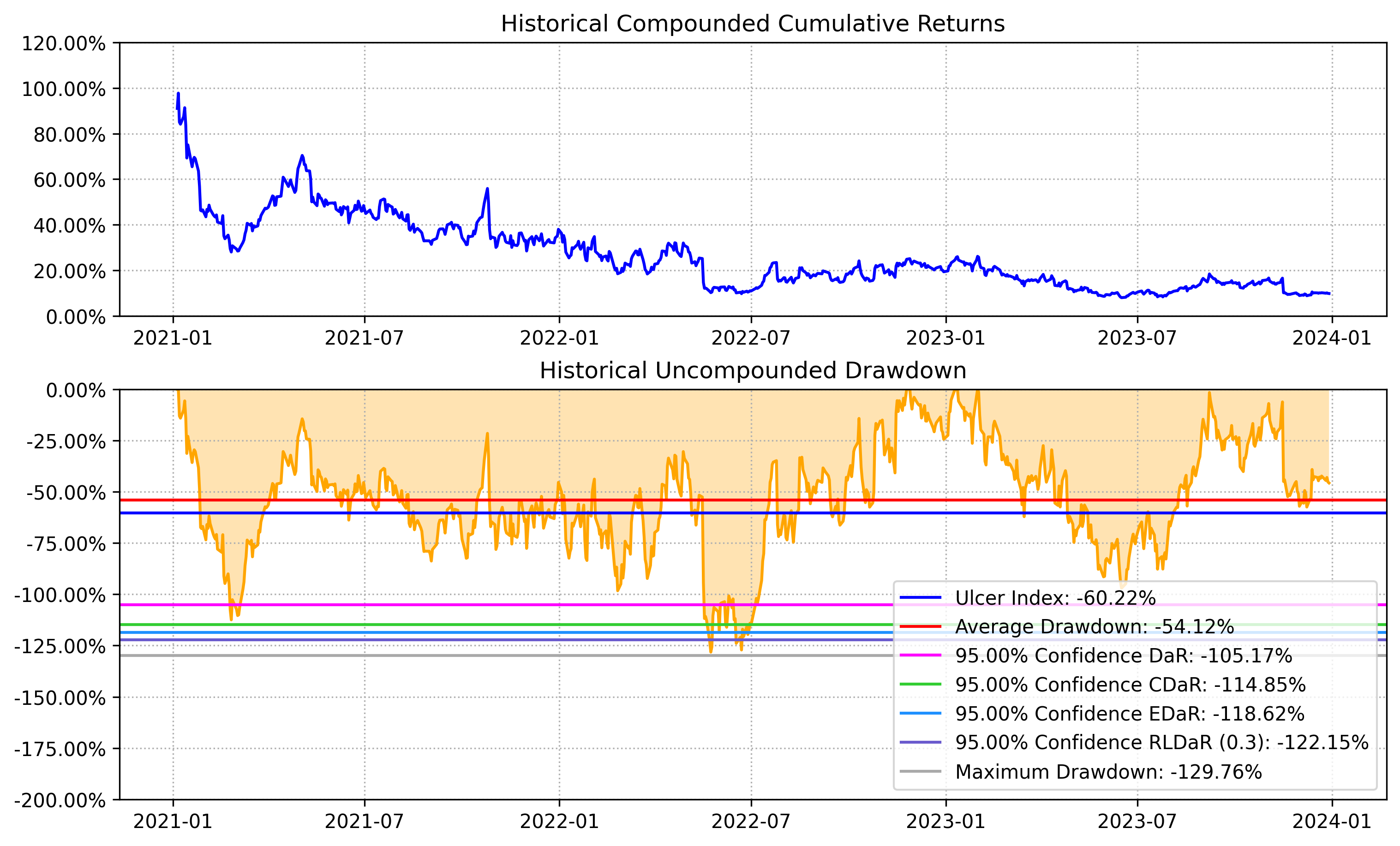}
    \hfill
    \includegraphics[width=0.32\textwidth]{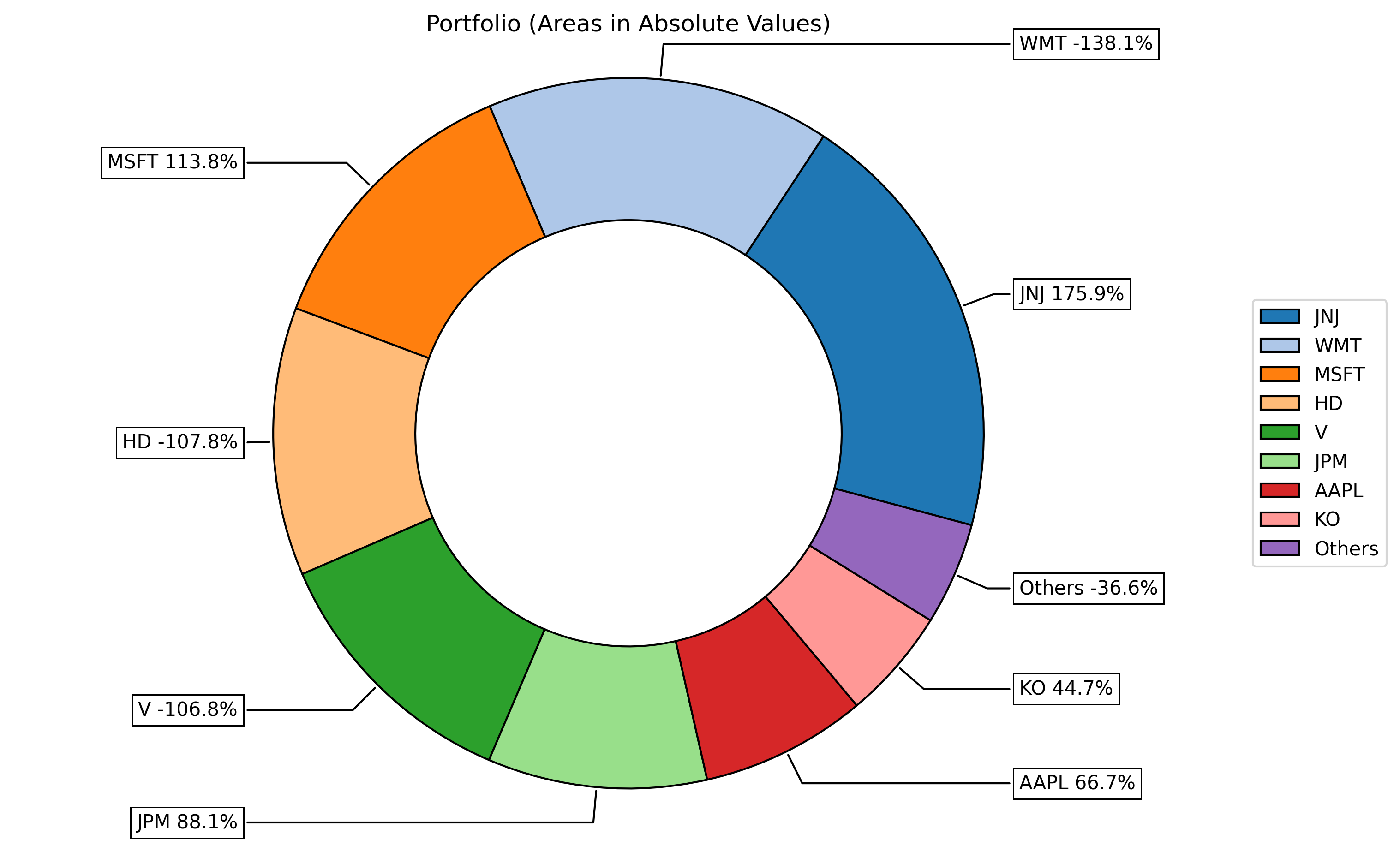}
    \hfill
    \includegraphics[width=0.32\textwidth]{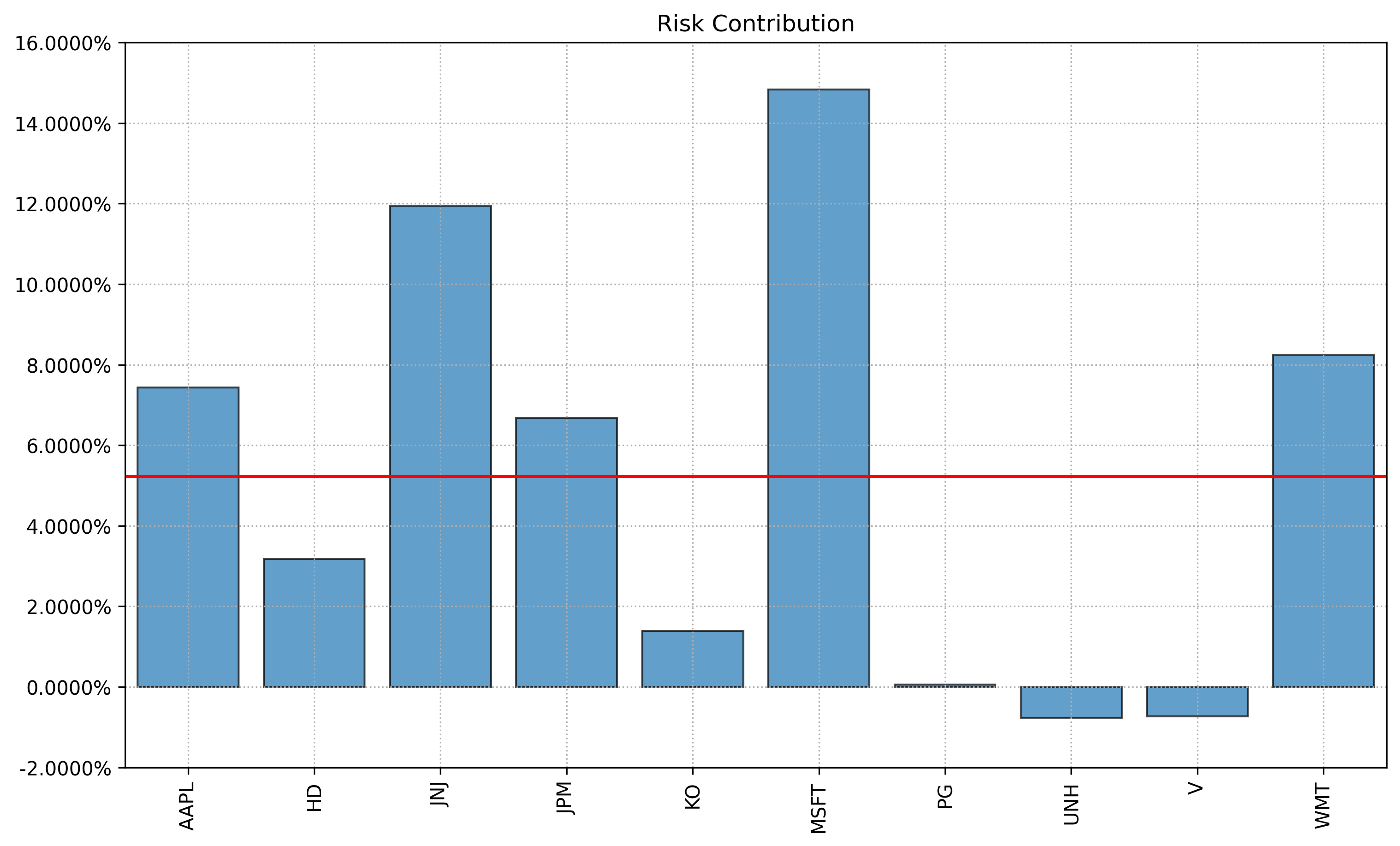}
    \caption{Historical cumulative-return and drawdown trajectories, mean asset allocation weights, and per-asset risk contribution profiles for the learned portfolio optimization policy for Relaxed-RS-AC ($\beta=0$).}
    \label{fig:beta_0}
\end{figure*}

\begin{figure*}[htbp]
    \centering
    \includegraphics[width=0.32\textwidth]{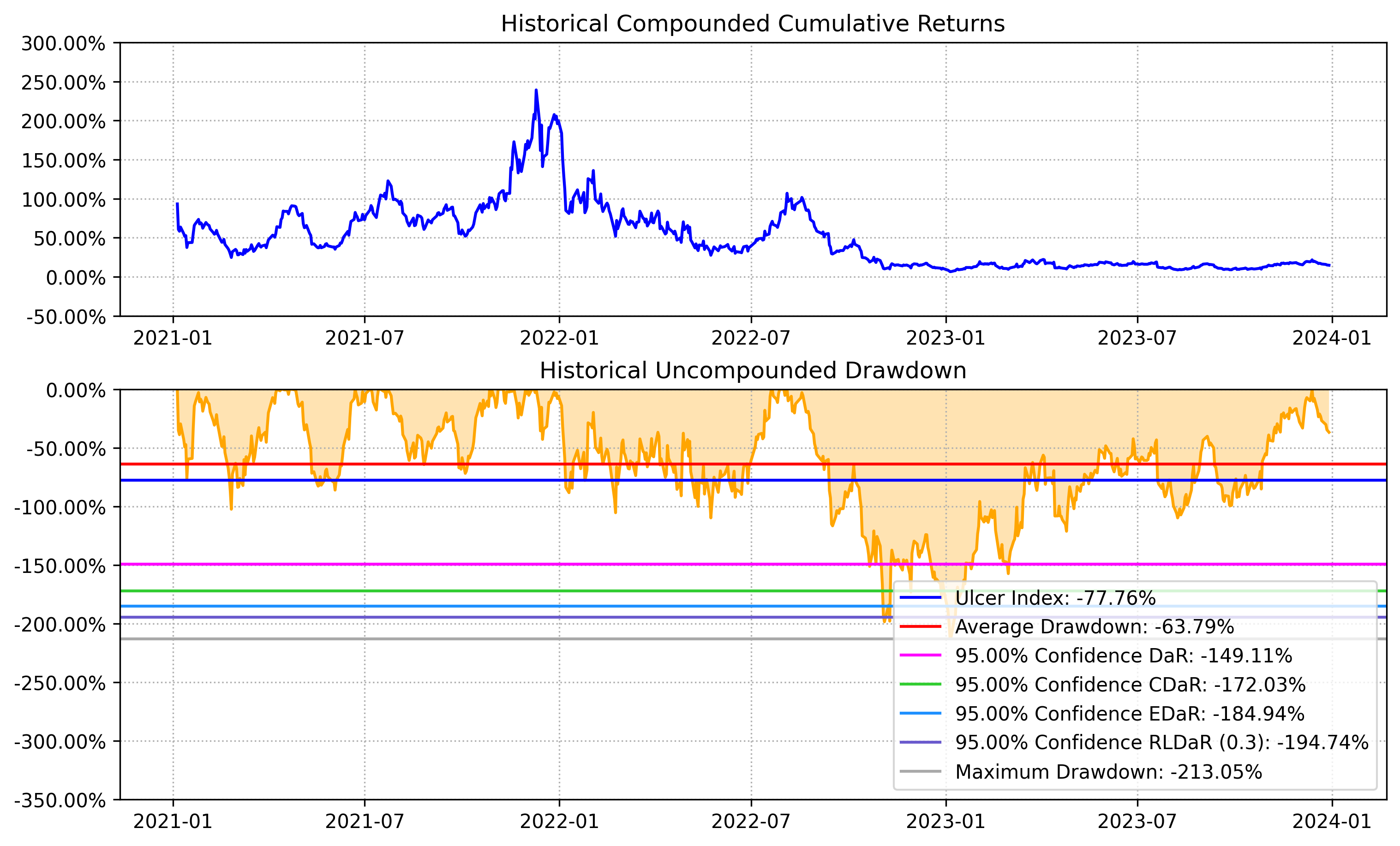}
    \hfill
    \includegraphics[width=0.32\textwidth]{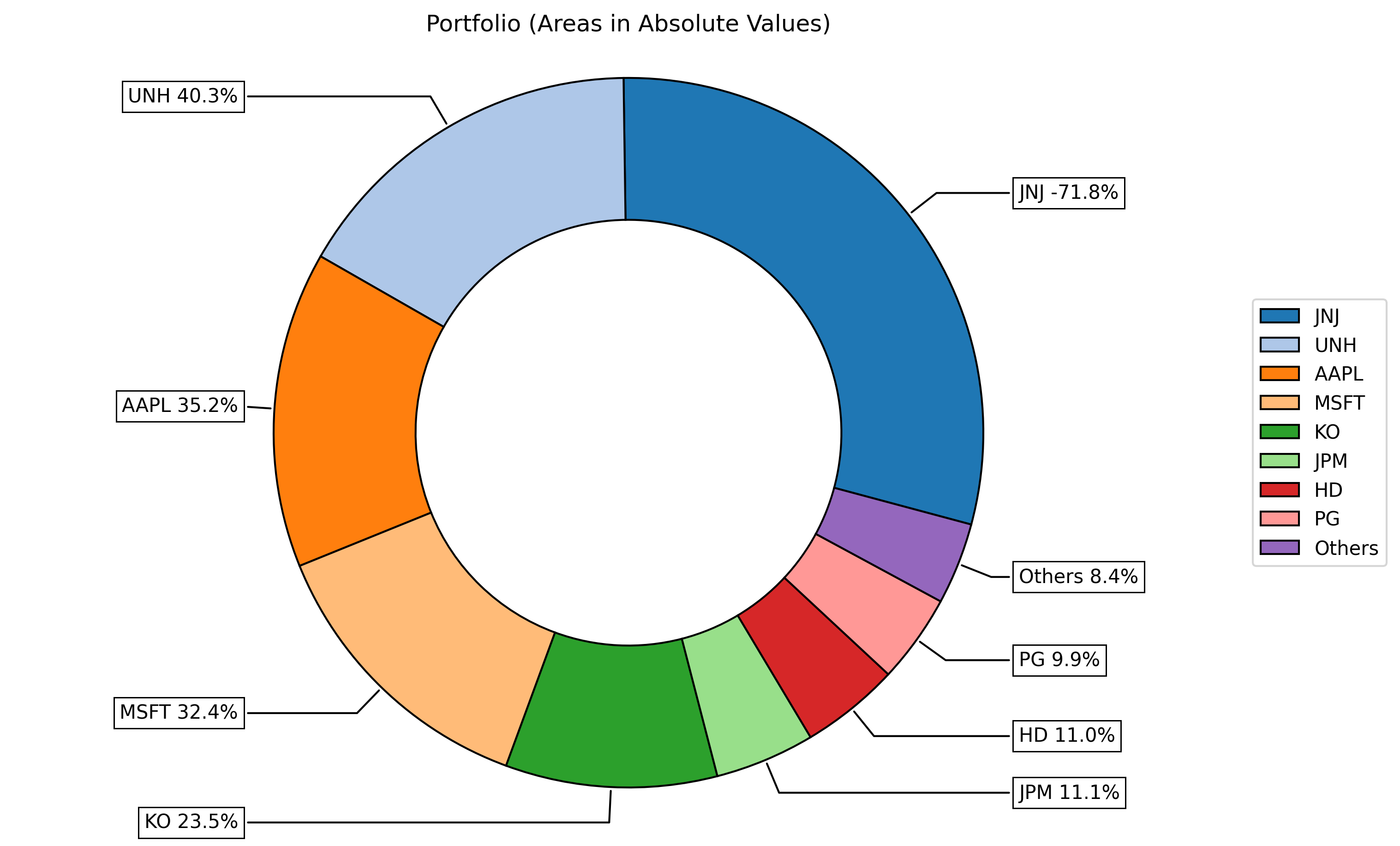}
    \hfill
    \includegraphics[width=0.32\textwidth]{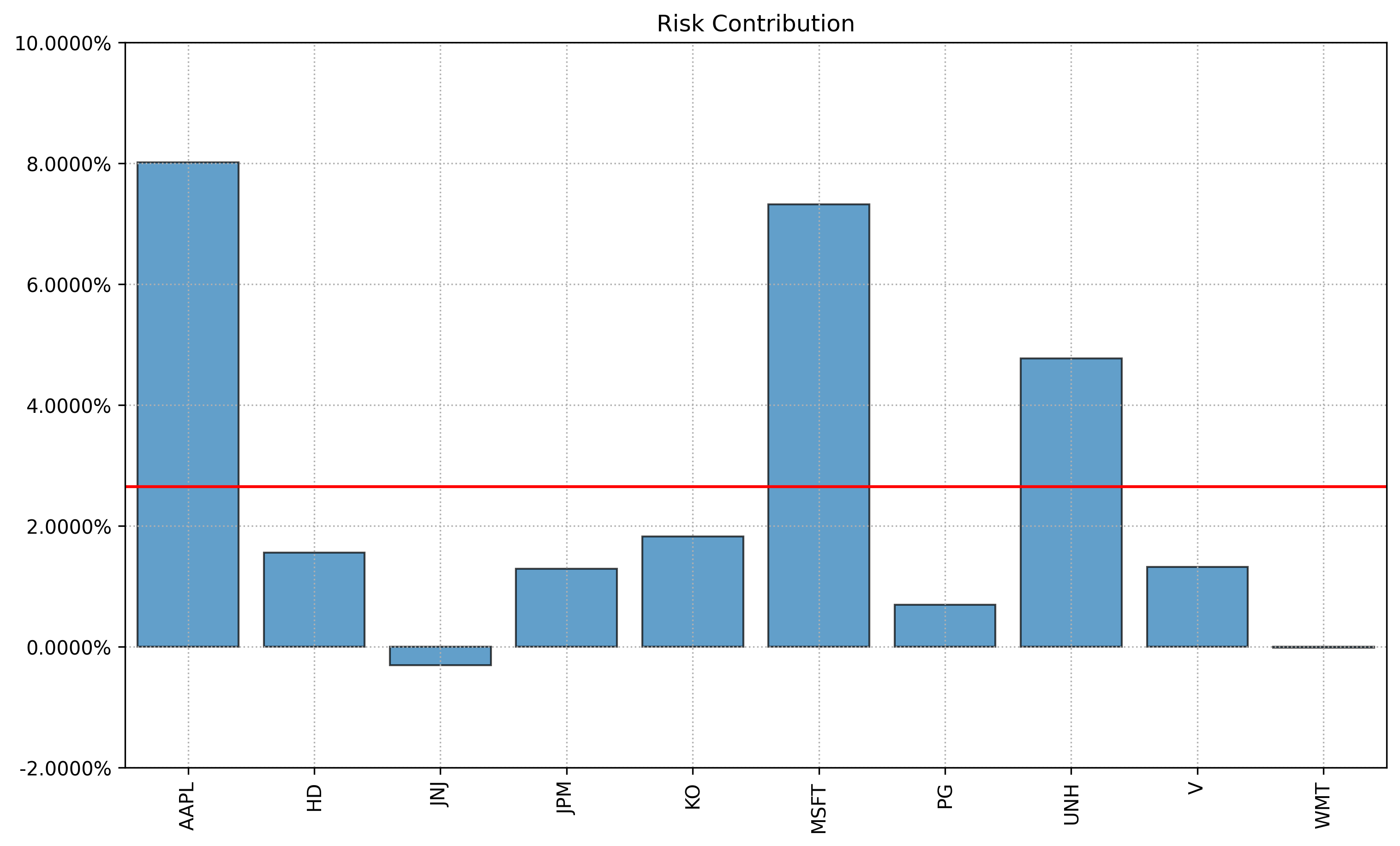}
    \caption{Historical cumulative-return and drawdown trajectories, mean asset allocation weights, and per-asset risk contribution profiles for the learned portfolio optimization policy for Relaxed-RS-AC ($\beta=0.01$).}
    \label{fig:beta_0_0_1}
\end{figure*}

\begin{figure*}[htbp]
    \centering
    \includegraphics[width=0.32\textwidth]{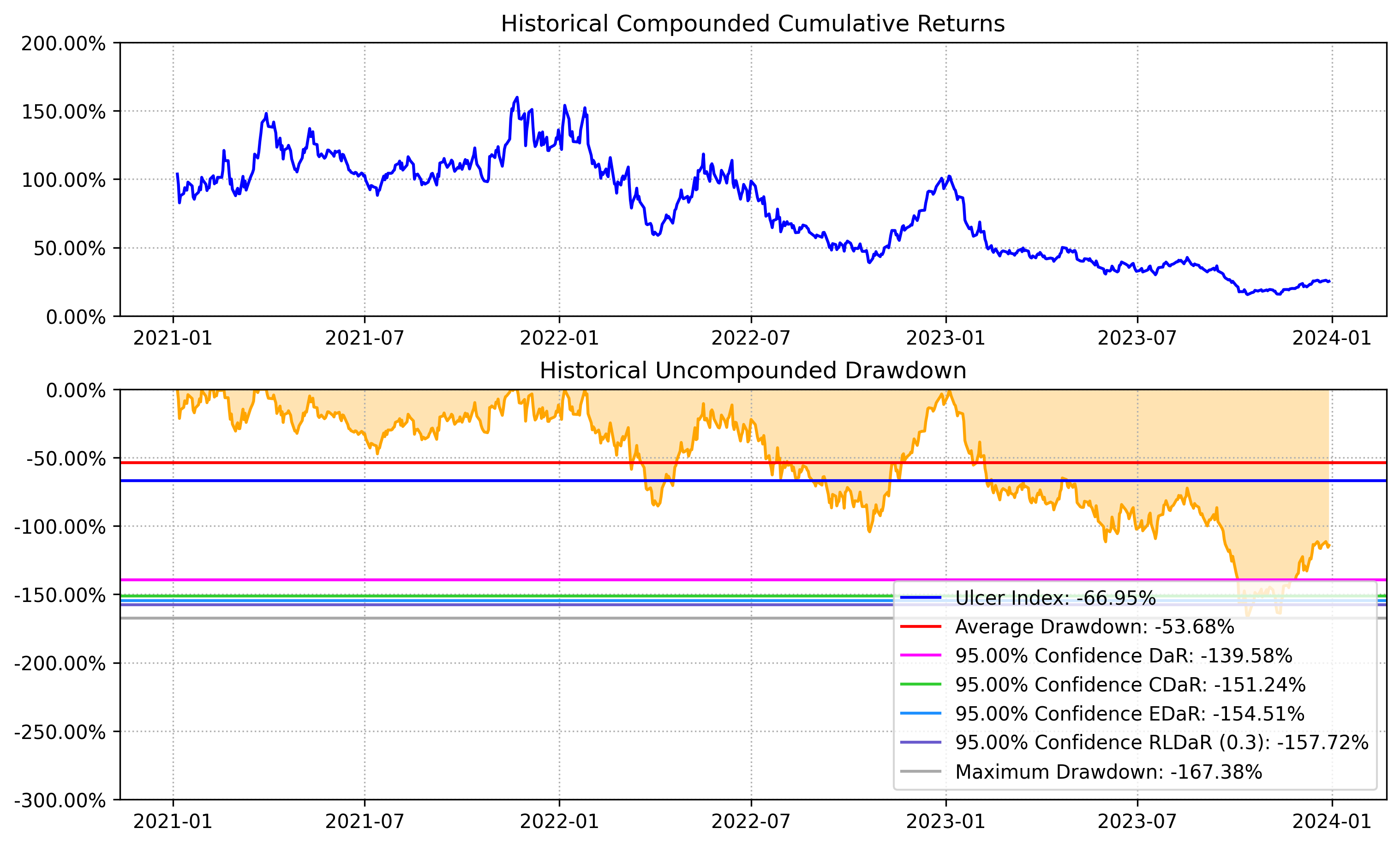}
    \hfill
    \includegraphics[width=0.32\textwidth]{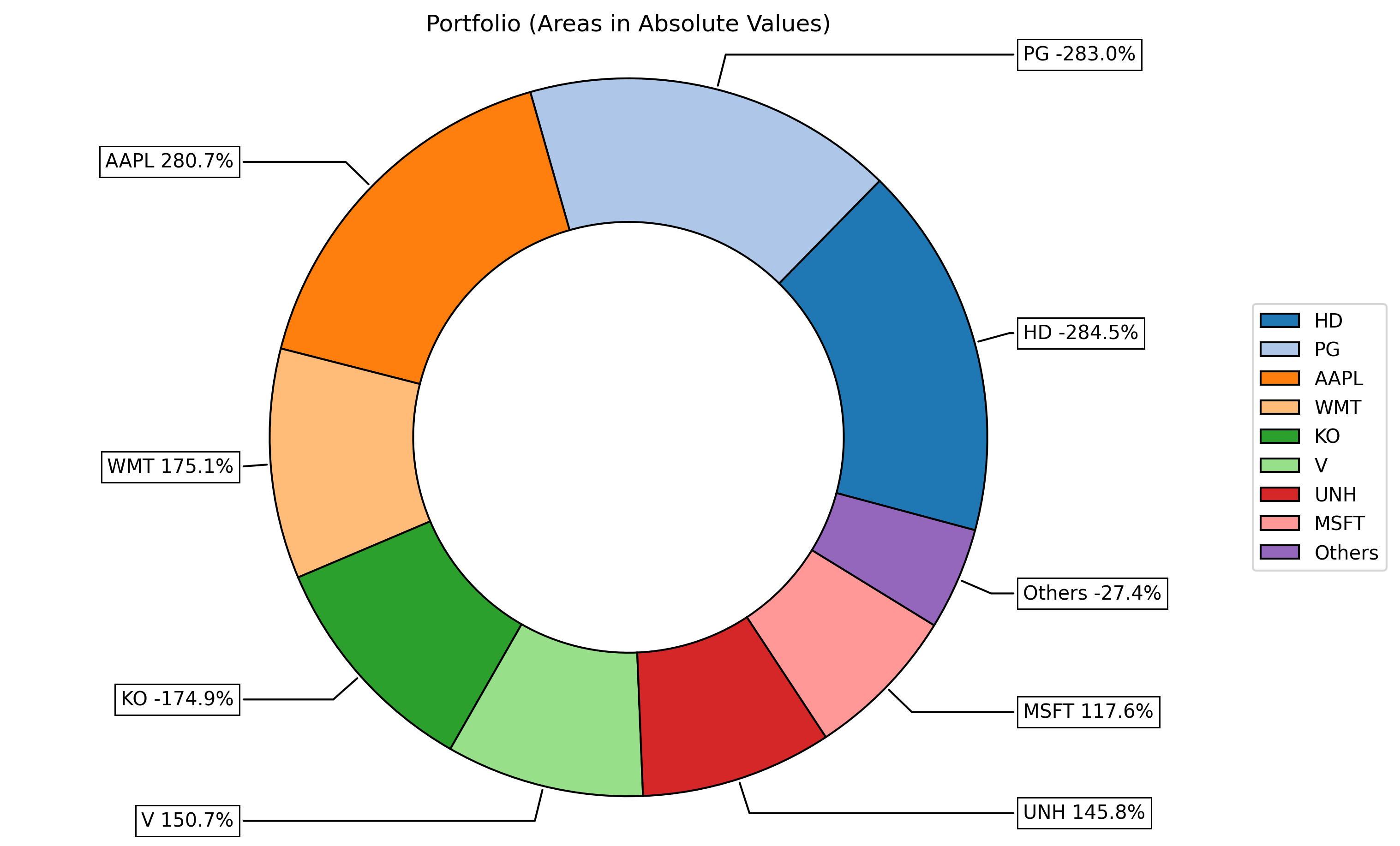}
    \hfill
    \includegraphics[width=0.32\textwidth]{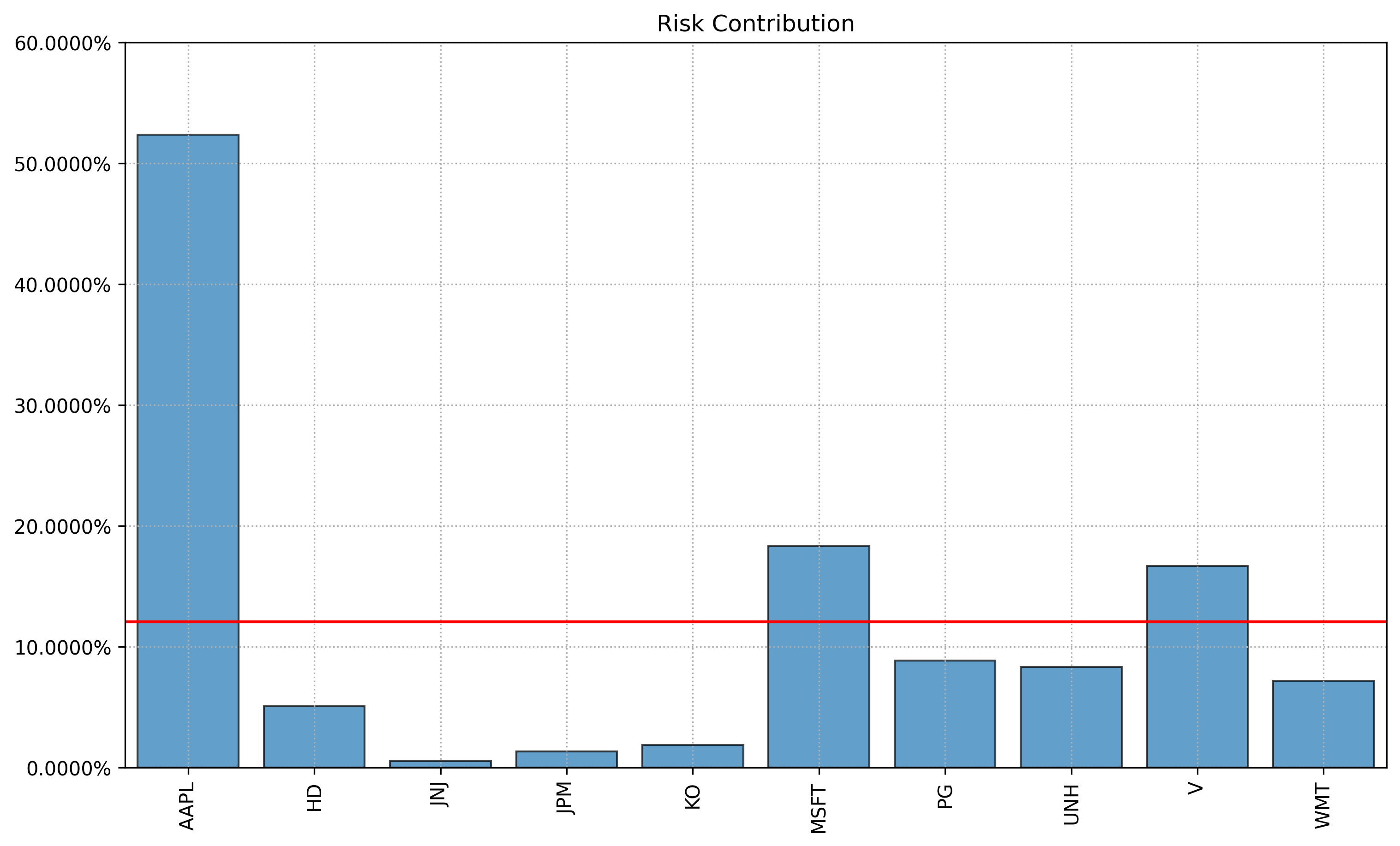}
    \caption{Historical cumulative-return and drawdown trajectories, mean asset allocation weights, and per-asset risk contribution profiles for the learned portfolio optimization policy for Relaxed-RS-AC ($\beta=0.05$).}
    \label{fig:beta_0_0_5}
\end{figure*}

\begin{figure*}[htbp]
    \centering
    \includegraphics[width=0.32\textwidth]{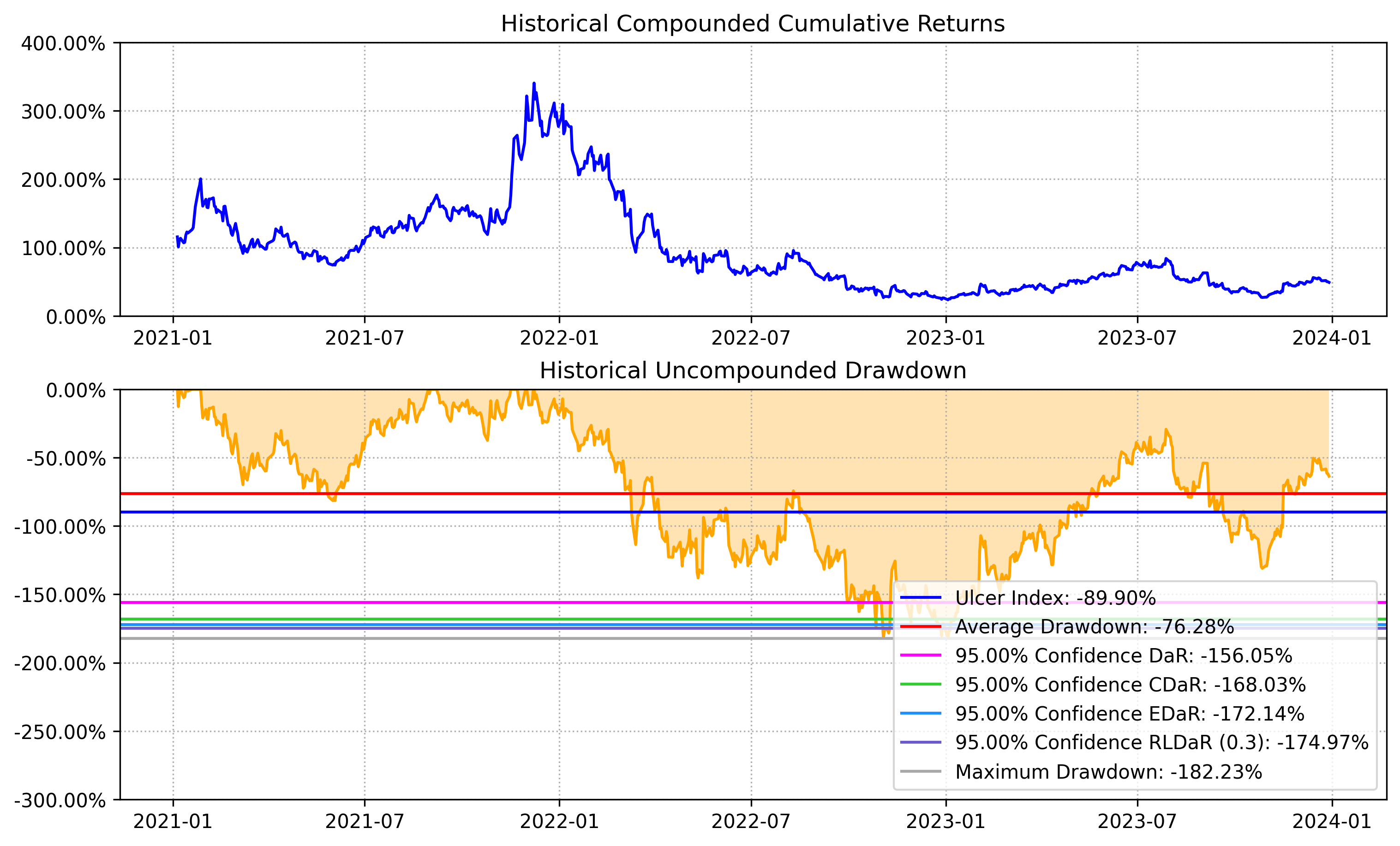}
    \hfill
    \includegraphics[width=0.32\textwidth]{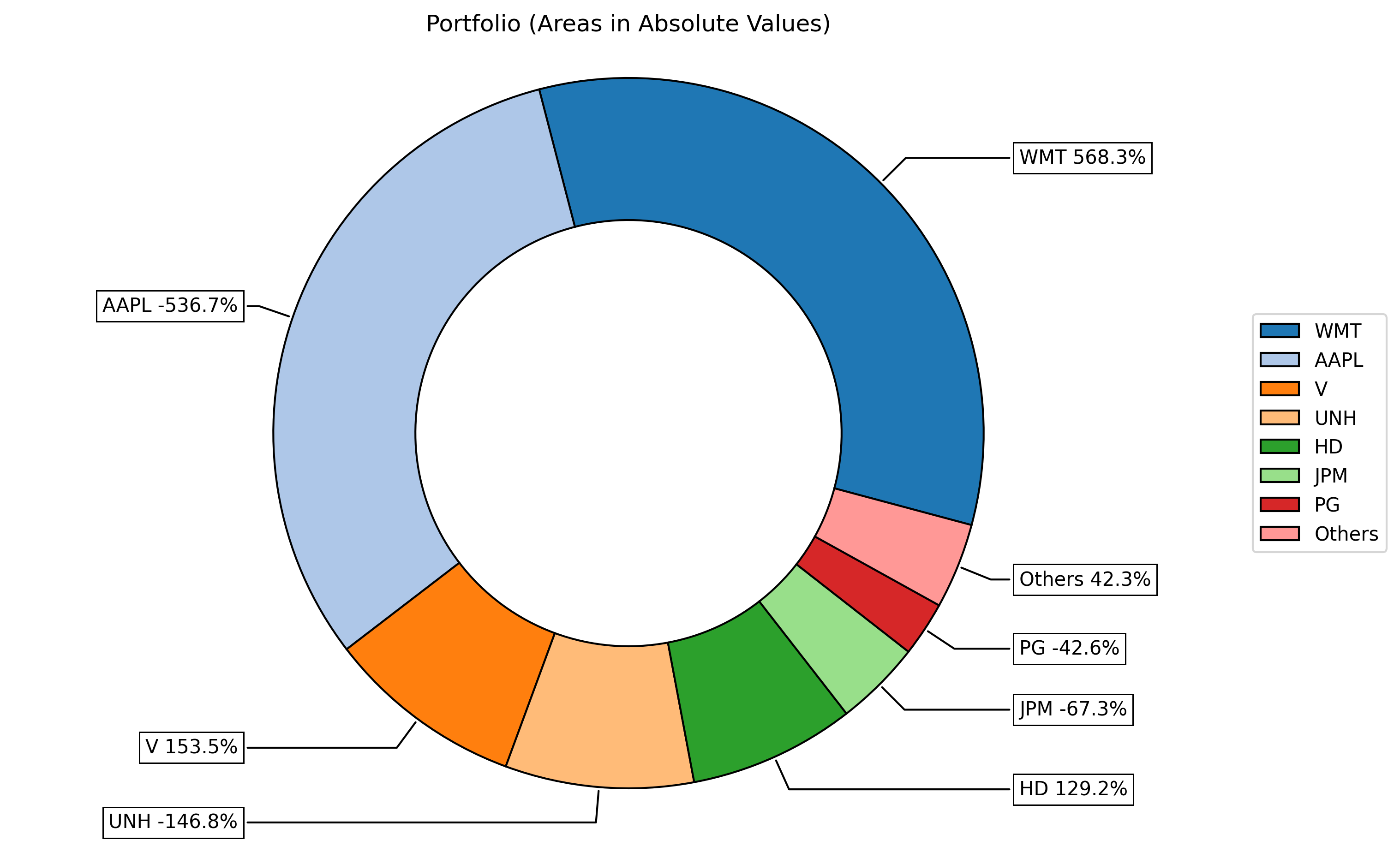}
    \hfill
    \includegraphics[width=0.32\textwidth]{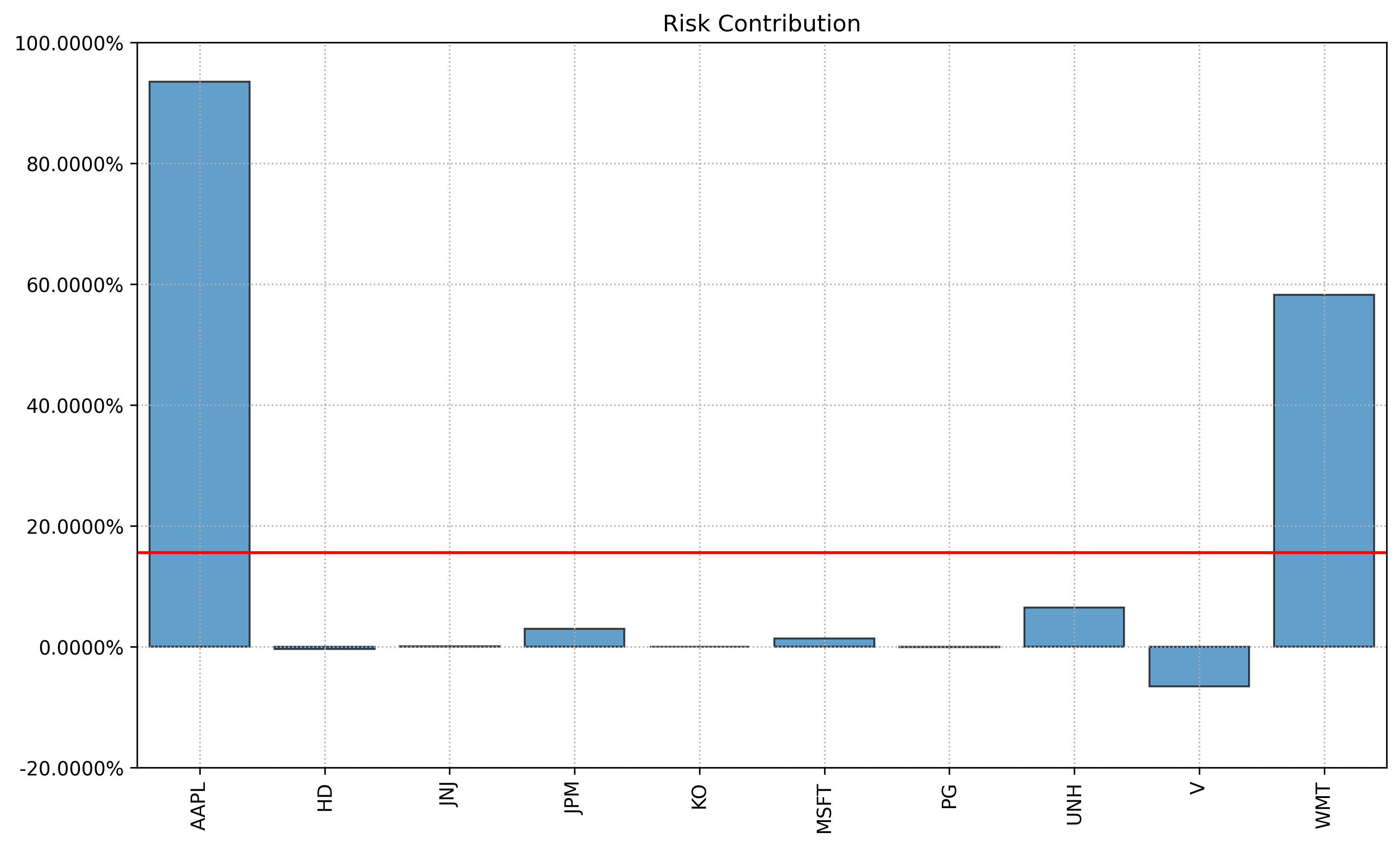}
    \caption{Historical cumulative-return and drawdown trajectories, mean asset allocation weights, and per-asset risk contribution profiles for the learned portfolio optimization policy for Relaxed-RS-AC ($\beta=0.1$).}
    \label{fig:beta_0_1}
\end{figure*}

We systematically sweep $\beta\in\{-0.1,\,-0.05,\,-0.01,\,0,\,0.01,\,0.05,\,0.1\}$,
covering the full spectrum from strongly risk-averse to strongly
risk-seeking.
Figures~\ref{fig:beta_minus_0_1}--\ref{fig:beta_0_1} display, for each
$\beta$, the cumulative-return trajectory, the uncompounded drawdown
profile with annotated risk statistics, the portfolio allocation
(absolute weight), and the per-asset risk contribution, all computed
on the held-out test window.

\paragraph{Risk-Averse Regime ($\beta < 0$).}
\label{sec:risk-averse}
For $\beta<0$, the exponential objective
$J(\psi)=(1/\beta)\log\mathbb{E}[e^{\beta R}]$
is a superlinear penalty on left-tail trajectories.
By the cumulant-generating function identity,
$J(\psi)=\mathbb{E}[R]-\tfrac{|\beta|}{2}\mathrm{Var}[R]+O(\beta^2)$,
small $|\beta|$ implements a mean-variance approximation,
penalising variance proportionally to $|\beta|/2$.
As $|\beta|$ increases, higher-order cumulants (skewness, kurtosis)
enter the penalty: the agent increasingly avoids trajectories that
are not merely high-variance but negatively skewed. This is precisely
the trajectory type generated by crash events.

Consider $\beta=-0.1$ (the most risk-averse configuration shown in Figure~\ref{fig:beta_minus_0_1}).
The learned allocation is an essentially unlevered long book
(gross exposure $\approx110\%$) anchored in defensive quality —
JNJ ($27.3\%$), PG ($16.4\%$), and UNH ($15.0\%$) — complemented by a
single concentrated growth position in AAPL ($21.6\%$) and small
long positions in JPM ($8.0\%$), KO ($5.9\%$), and V ($5.5\%$).
The only short is a modest WMT ($-5.1\%$) hedge.
The portfolio compounds to $\approx146\%$ of initial wealth
($\approx+46\%$ cumulative return, the highest of all Relaxed-RS-AC
configurations), with an Ulcer Index of $9.46\%$ and a maximum
drawdown of $-24.88\%$, incurred at the October 2022 trough of the
Federal Reserve tightening cycle.
The drawdown trajectory in Figure~\ref{fig:beta_minus_0_1} shows
shallow drawdowns (above $-7\%$) throughout 2021, a single prolonged
underwater period spanning the 2022 bear market, full recovery by
mid-2023, and a new high-water mark in December 2023.
The risk-contribution chart (right panel) confirms disciplined risk
budgeting: AAPL, the most volatile holding, is the largest single
contributor ($\approx4.6\%$ against an equal-weight benchmark of
$\approx1.5\%$), while the WMT short contributes negative risk,
acting as a partial hedge.

Consider $\beta=-0.05$ (intermediate risk aversion, shown in Figure~\ref{fig:beta_minus_0_0_5}).
The policy constructs a defensive value book — KO ($34.9\%$),
JNJ ($30.5\%$), HD ($24.9\%$), JPM ($21.4\%$), V ($15.3\%$), and
UNH ($9.6\%$) — hedged by explicit shorts in the three most volatile
large-cap names: MSFT ($-19.2\%$), AAPL ($-12.6\%$), and PG ($-11.2\%$).
This configuration achieves the best downside profile of the entire
study: an Ulcer Index of $5.85\%$, average drawdown of $-4.74\%$, and
a maximum drawdown of only $-13.42\%$.
The growth shorts function as an embedded volatility hedge: during
the 2022 tightening cycle, when growth equities sold off hardest,
the June--July 2022 trough is contained near $-13\%$
(Figure~\ref{fig:beta_minus_0_0_5}) and the portfolio sets a fresh
high-water mark by December 2022.
The cost is upside: the same shorts cap participation in the 2023
mega-cap rally, and the cumulative return of $\approx+24\%$ is roughly
half that of $\beta=-0.1$.
While $\beta=-0.1$ attains the higher terminal return and
$\beta=-0.05$ the stronger tail control, both sit on the risk--return
frontier of the sweep. The choice between them reflects the risk preference
that $\beta$ is designed to encode.

Consider $\beta=-0.01$ (near risk-neutral with a risk-averse tilt, shown in Figure~\ref{fig:beta_minus_0_0_1}).
With the penalty nearly switched off, the policy concentrates in
JNJ ($42.5\%$) and re-admits growth exposure (MSFT $15.2\%$)
alongside HD ($18.4\%$), V ($15.0\%$), PG ($11.2\%$), JPM ($10.0\%$),
and WMT ($5.6\%$), with a short in UNH ($-13.1\%$). Gross exposure
remains moderate at $\approx136\%$.
Performance degrades on both axes relative to $\beta=-0.05$:
cumulative return is $\approx+22\%$ with a maximum drawdown of
$-23.05\%$ and Ulcer Index of $8.48\%$ — nearly double the drawdown
for no additional return, a weakly Pareto-dominated outcome.
The drawdown trajectory in Figure~\ref{fig:beta_minus_0_0_1} shows a
prolonged 2022 underwater period bottoming at $\approx-23\%$ in
October 2022 and a secondary $\approx-16\%$ relapse in November 2023.
This confirms the theoretical prediction: as $\beta\to 0^-$,
$J(\psi)$ approaches pure expected return, the incentive to hedge
left-tail trajectories vanishes, and tail protection erodes before
any compensating growth materialises.

Table~\ref{tab:beta-summary} summarises the risk-return tradeoff.
Two regularities stand out.
First, every risk-averse parameterization is profitable out-of-sample
with bounded leverage (gross exposure between $\approx110\%$ and
$\approx186\%$), while every parameterization with $\beta\ge 0$
builds a heavily leveraged long--short book and suffers catastrophic
out-of-sample losses.
The exponential penalty thus acts as an implicit leverage and
concentration regulariser: penalising the variance (and higher
cumulants) of the return distribution directly penalises the gross
exposure that manufactures that variance.
Second, within the risk-averse regime the drawdown statistics respond
to $\beta$ as designed, with $\beta=-0.05$ attaining the best
downside metrics of the entire study.

\paragraph{Risk-Neutral Baseline ($\beta = 0$, shown in Figure~\ref{fig:beta_0}).}
\label{sec:risk-neutral}

The risk-neutral case ($\beta=0$) is the theoretically expected
degenerate case of the exponential framework and serves as an
internal ablation.
Optimizing pure expected discounted return in an unconstrained
long--short action space, the policy accumulates a gross exposure of
$\approx8.8\times$ initial wealth: leveraged longs in
JNJ ($175.9\%$), MSFT ($113.8\%$), JPM ($88.1\%$), AAPL ($66.7\%$),
and KO ($44.7\%$), financed by large shorts in WMT ($-138.1\%$),
HD ($-107.8\%$), and V ($-106.8\%$).
This book maximises the in-sample (2013--2019) return premium but is
catastrophically fragile out-of-sample: the equity curve in
Figure~\ref{fig:beta_0} collapses by $\approx-70\%$ within the first
two months of 2021, partially rebounds, then grinds down to
$\approx10\%$ of initial wealth by mid-2022, ending the test window
at a $\approx-90\%$ cumulative loss.
The drawdown panel records an uncompounded maximum drawdown of
$-129.76\%$ — leveraged books can lose more than initial wealth on
an uncompounded basis — with an Ulcer Index of $60.22\%$, and the
portfolio is underwater for essentially the entire test window.

This pathology is structurally expected.
Without the variance-penalising exponential term, the objective is
indifferent between a stable payoff and an extremely levered payoff
with the same in-sample mean, and the stochastic-approximation
iterates drift toward the leverage-maximising region of the
parameter space, overfitting the return premia estimated on the
2013--2019 training regime.
The exponential shaping at $\beta<0$ is therefore not a mere
refinement: it is the mechanism that keeps the learned policy inside
the economically meaningful, weakly-levered region of the action
space that the risk-averse variants occupy.

\paragraph{Risk-Seeking Regime ($\beta > 0$).}
\label{sec:risk-seeking} For $\beta>0$, the cumulant-expansion gives
$J(\psi)=\mathbb{E}[R]+(\beta/2)\mathrm{Var}[R]+O(\beta^2)$:
the agent explicitly rewards variance.
Since per-asset variance is bounded, the policy manufactures variance
through the action space instead: gross exposure rises to
$\approx2.4\times$ at $\beta=+0.01$ and to $\approx16$--$17\times$ at
$\beta\in\{+0.05,+0.1\}$, concentrated in offsetting long--short pair
trades.
Large realized returns receive a high exponential weight when $\beta>0$. Error in the perturbed value estimates can also be amplified by the factor $1/(2c_t)$ in Equation~\ref{eq:twosided-update}. The observed leveraged allocations produced large gains when they aligned with the market regime and sharp losses when the regime turned.

Consider $\beta=+0.01$ (weakly risk-seeking, shown in Figure~\ref{fig:beta_0_0_1}).
The policy shorts the most stable asset in the universe,
JNJ ($-71.8\%$), to lever up high-variance longs —
UNH ($40.3\%$), AAPL ($35.2\%$), MSFT ($32.4\%$), and KO ($23.5\%$) —
for a gross exposure of $\approx2.4\times$.
The equity curve rides the 2021 momentum regime to a peak of
$\approx240\%$ in December 2021, then collapses through 2022 to
$\approx8$--$15\%$ of initial wealth, ending at an $\approx-85\%$
cumulative loss.
The uncompounded maximum drawdown of $-213.05\%$ is the deepest of
all configurations, with an Ulcer Index of $77.76\%$: with no
downside penalty to protect the crash-prone book, and with unstable
gradient dynamics from exponential amplification, the policy is
unable to consolidate its transient 2021 gains.

Consider $\beta=+0.05$ (moderately risk-seeking, shown in Figure~\ref{fig:beta_0_0_5}).
Gross exposure explodes to $\approx16\times$: massive longs in
AAPL ($280.7\%$), WMT ($175.1\%$), V ($150.7\%$), UNH ($145.8\%$),
and MSFT ($117.6\%$) against equally massive shorts in
HD ($-284.5\%$), PG ($-283.0\%$), and KO ($-174.9\%$).
The risk-contribution chart shows more than half of total portfolio
risk loaded on AAPL alone.
After peaking near $160\%$ in late 2021, the book unwinds through
2022--2023 to end at an $\approx-75\%$ cumulative loss, with an
uncompounded maximum drawdown of $-167.38\%$ and an Ulcer Index of
$66.95\%$ — a particularly stark Pareto-dominated outcome relative to
the comparable risk-averse case $\beta=-0.05$
($\approx+24\%$ return at a $-13.42\%$ maximum drawdown).

Consider $\beta=+0.1$ (strongly risk-seeking, shown in Figure~\ref{fig:beta_0_1}).
The allocation degenerates into a single enormous pair trade —
long WMT ($568.3\%$) against short AAPL ($-536.7\%$), with secondary
positions in V ($153.5\%$) and HD ($129.2\%$) and shorts in
UNH ($-146.8\%$), JPM ($-67.3\%$), and PG ($-42.6\%$) — a gross
exposure of $\approx17\times$ whose risk-contribution profile is
almost entirely absorbed by the two legs of the pair.
The equity curve is correspondingly the most volatile of the sweep,
spiking to $\approx200\%$ in February 2021 and $\approx340\%$ in
December 2021 before collapsing below $30\%$ of initial wealth by the
end of 2022 and finishing at an $\approx-50\%$ cumulative loss.
The Ulcer Index of $89.90\%$ is the worst of the study, and the
uncompounded maximum drawdown reaches $-182.23\%$.
This is the pure expression of the variance-seeking objective: the
pair trade carries enormous idiosyncratic variance, which is exactly
what $J(\psi)$ rewards, independent of its (negative) out-of-sample
expected return.

\begin{table}[ht]
\centering
\small
\setlength{\tabcolsep}{4pt}
\begin{tabular}{lccccc}
\toprule
\textbf{Configuration} & \shortstack{\textbf{Cumulative}\\\textbf{return}} & \shortstack{\textbf{Ulcer}\\\textbf{index}} & \shortstack{\textbf{Average}\\\textbf{drawdown}} & \shortstack{\textbf{Maximum}\\\textbf{drawdown}} & \shortstack{\textbf{Sharpe}\\\textbf{(val. / test)}} \\
\midrule
$\beta=-0.10$          & $\approx +46\%$   & $-9.46\%$  & $-6.67\%$  & $-24.88\%$  & $0.3479\ /\ -0.8648$ \\
$\beta=-0.05$          & $\approx +24\%$   & $-5.85\%$  & $-4.74\%$  & $-13.42\%$  & $0.0229\ /\ -0.3999$ \\
$\beta=-0.01$          & $\approx +22\%$   & $-8.48\%$  & $-6.49\%$  & $-23.05\%$  & $\mathbf{0.8469}\ /\ \mathbf{0.0551}$ \\
$\beta=0$ (neutral)    & $\approx -90\%$   & $-60.22\%$ & $-54.12\%$ & $-129.76\%$ & $0.4228\ /\ -0.6826$ \\
$\beta=+0.01$          & $\approx -85\%$   & $-77.76\%$ & $-63.79\%$ & $-213.05\%$ & $-3.4806\ /\ -1.4752$ \\
$\beta=+0.05$          & $\approx -75\%$   & $-66.95\%$ & $-53.68\%$ & $-167.38\%$ & $-3.7894\ /\ -1.1963$ \\
$\beta=+0.10$          & $\approx -50\%$   & $-89.90\%$ & $-76.28\%$ & $-182.23\%$ & $-1.4214\ /\ -0.6392$ \\
\midrule
DSAC-Wang $\kappa=-0.75$ & $\approx +44\%$ & $-16.54\%$ & $-11.82\%$ & $-41.77\%$ & $1.7713\ /\ 0.7215$ \\
DSAC-Wang $\kappa=+0.75$ & $\approx +52\%$ & $-10.54\%$ & $-7.35\%$  & $-30.56\%$ & $1.3064\ /\ -0.0958$ \\
DSAC-CVaR $\alpha=0.9$   & $\approx +47\%$ & $-7.47\%$  & $-5.86\%$  & $-19.10\%$ & $\mathbf{1.8303}\ /\ 0.6605$ \\
\bottomrule
\end{tabular}
\caption{Risk-return summary across $\beta$ values and baselines on the held-out test window (January 2021--December 2023), together with annualized Sharpe ratio on the validation window (2020) and the
test window. Cumulative Return is the net compounded return over the test window. The Ulcer Index and drawdowns are negative-signed magnitudes (more negative = worse downside). Uncompounded drawdowns may exceed $-100\%$ for leveraged portfolios. Bolded Sharpe values mark the validation-window maximum within each method family. This
value is the selection criterion used for the main-paper configuration (\S\ref{sec:portfolio_optimization_main}).}
\label{tab:beta-summary}
\end{table}

\paragraph{Model selection via validation-set Sharpe ratio.}
Because $\beta$ is a free hyperparameter, we select the reported Relaxed-RS-AC configuration by validation-window Sharpe ratio rather than by test-window outcome, mirroring standard model-selection
practice and avoiding test-set leakage. Among the seven Relaxed-RS-AC configurations, $\beta=-0.01$ attains the highest validation Sharpe ratio ($0.8469$), driven by the COVID-19 recovery rally that dominates the 2020 validation year and rewards a lightly hedged, near-risk-neutral book. Encouragingly, $\beta=-0.01$ also attains the least-negative test-window Sharpe ratio among all seven
configurations ($0.0551$), so the validation-selected model remains the best Relaxed-RS-AC configuration by this metric out-of-sample as well, despite the test window's very different stress profile (the 2022 tightening cycle rather than a crash-and-recover pattern). The
same is not true of the baselines: DSAC-CVaR ($\alpha=0.9$) has the highest validation Sharpe ratio ($1.8303$) and is therefore the baseline selected for the main-paper comparison, but on the test window DSAC-Wang ($\kappa=-0.75$) edges it out ($0.7215$ versus $0.6605$). This reversal illustrates the same validation/test regime-shift sensitivity discussed throughout this section, and
motivates reporting both validation and test Sharpe ratios rather than test performance alone.


\subsubsection{Comparison with DSAC Baselines}
\label{sec:results-baselines}

\begin{figure*}[htbp]
    \centering
    \includegraphics[width=0.32\textwidth]{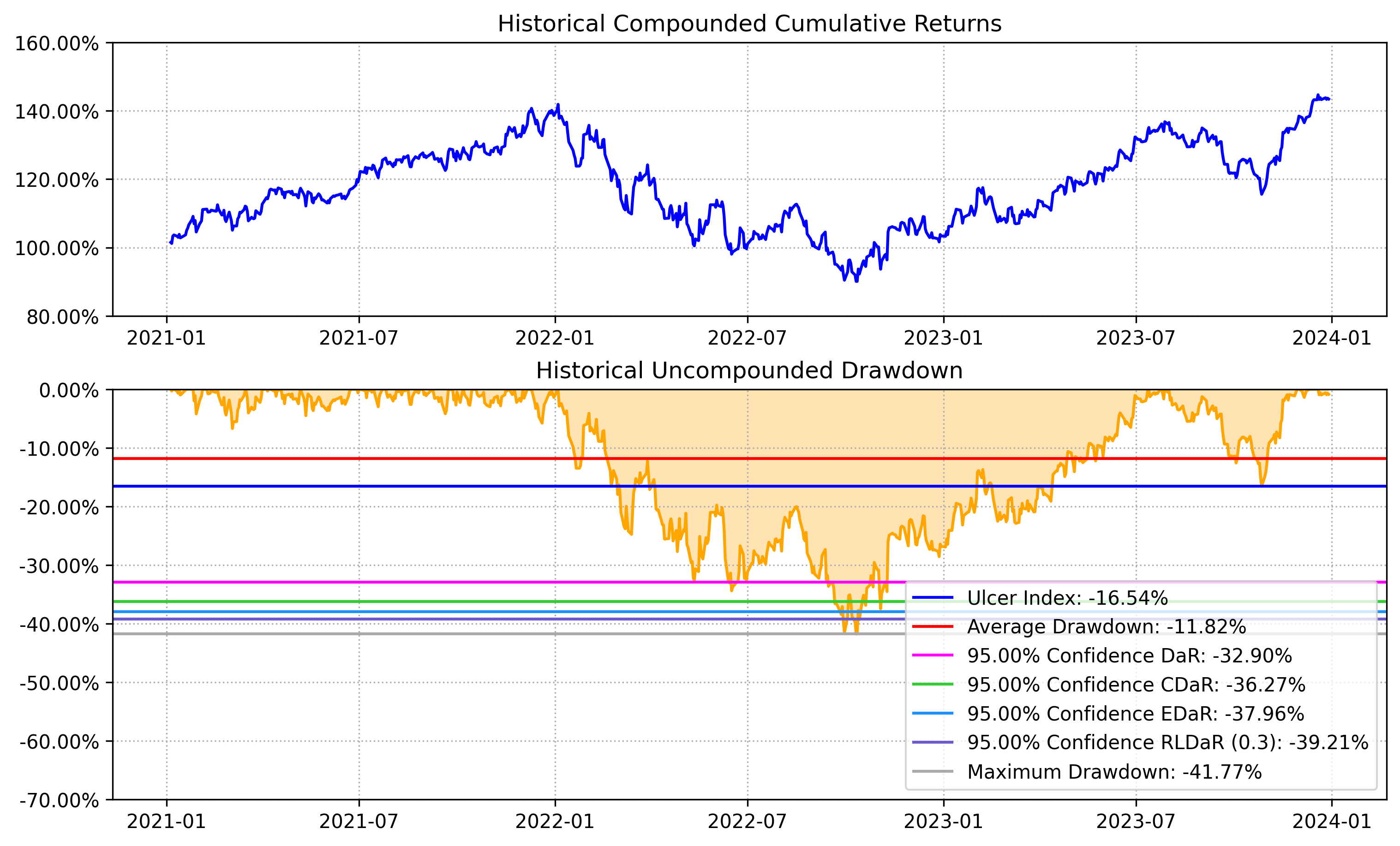}
    \hfill
    \includegraphics[width=0.32\textwidth]{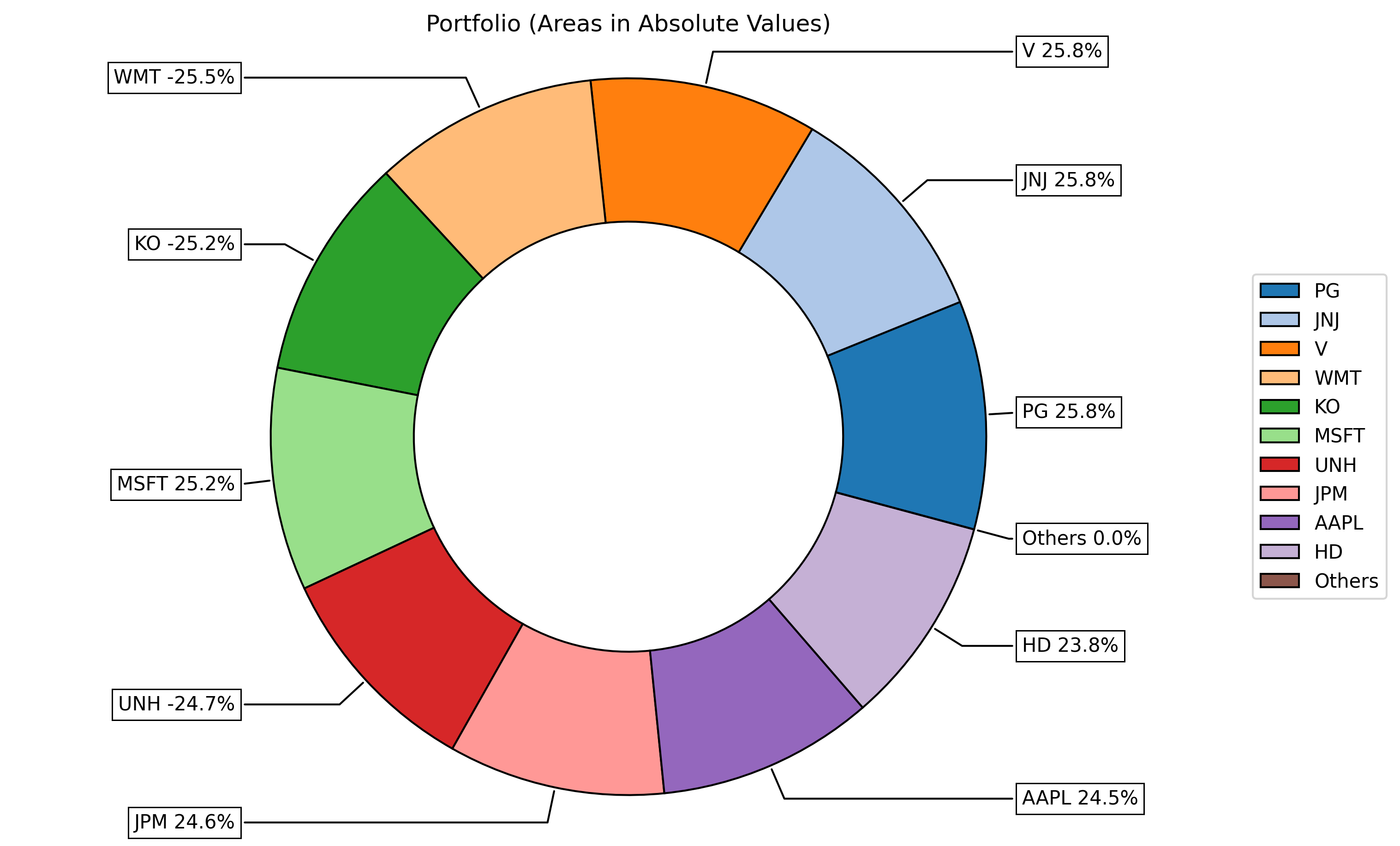}
    \hfill
    \includegraphics[width=0.32\textwidth]{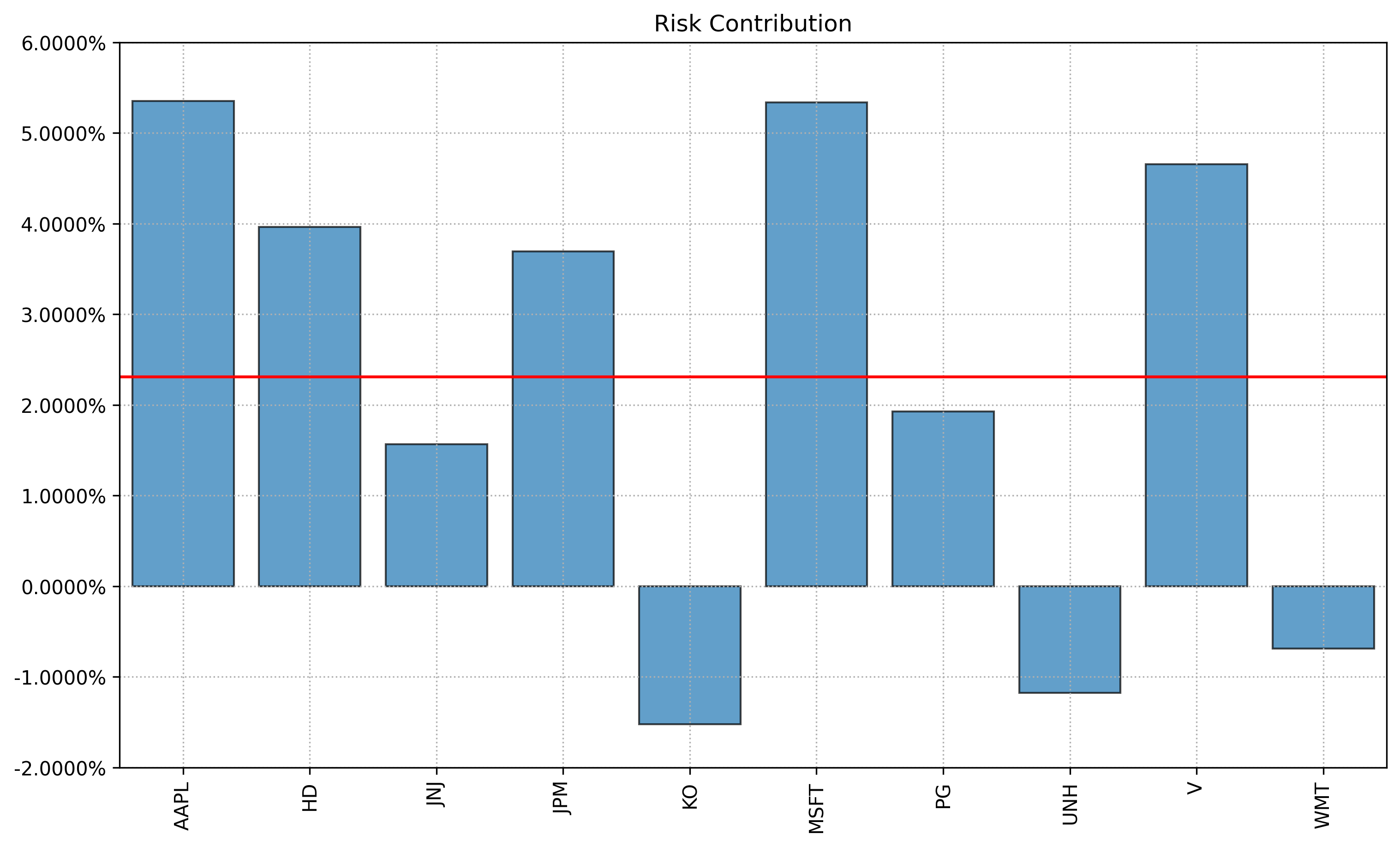}
    \caption{Historical cumulative-return and drawdown trajectories, mean asset allocation weights, and per-asset risk contribution profiles for the learned portfolio optimization policy for DSAC-Wang ($\kappa=-0.75$).}
    \label{fig:wang_-0.75}
\end{figure*}

Consider DSAC-Wang ($\kappa=-0.75$, shown in Figure~\ref{fig:wang_-0.75}).
The Wang distortion risk measure with $\kappa=-0.75$ imposes a
left-tail-pessimistic distortion on the return distribution,
concentrating probability mass on adverse outcomes.
The resulting portfolio exhibits a striking allocation signature:
near-uniform weight \emph{magnitudes} of $\approx24$--$26\%$ across
the entire universe, with the risk preference expressed only through
signs — longs in V, JNJ, PG, MSFT, JPM, AAPL, and HD (between
$23.8\%$ and $25.8\%$ each) against shorts in WMT ($-25.5\%$),
KO ($-25.2\%$), and UNH ($-24.7\%$).
Cumulative return is $\approx+44\%$, but the drawdown profile is the
worst of the profitable configurations: a maximum drawdown of
$-41.77\%$ at the October 2022 trough, an Ulcer Index of $16.54\%$,
and an underwater period stretching from January 2022 to mid-2023.
The three shorts fall precisely on the defensive names that held up
best during the 2022 tightening cycle, so the hedge leg amplified
rather than damped the bear-market drawdown: the distortion, fit to
the tail characteristics of the training distribution, mispriced
which tail would materialise out-of-sample.

\begin{figure*}[htbp]
    \centering
    \includegraphics[width=0.32\textwidth]{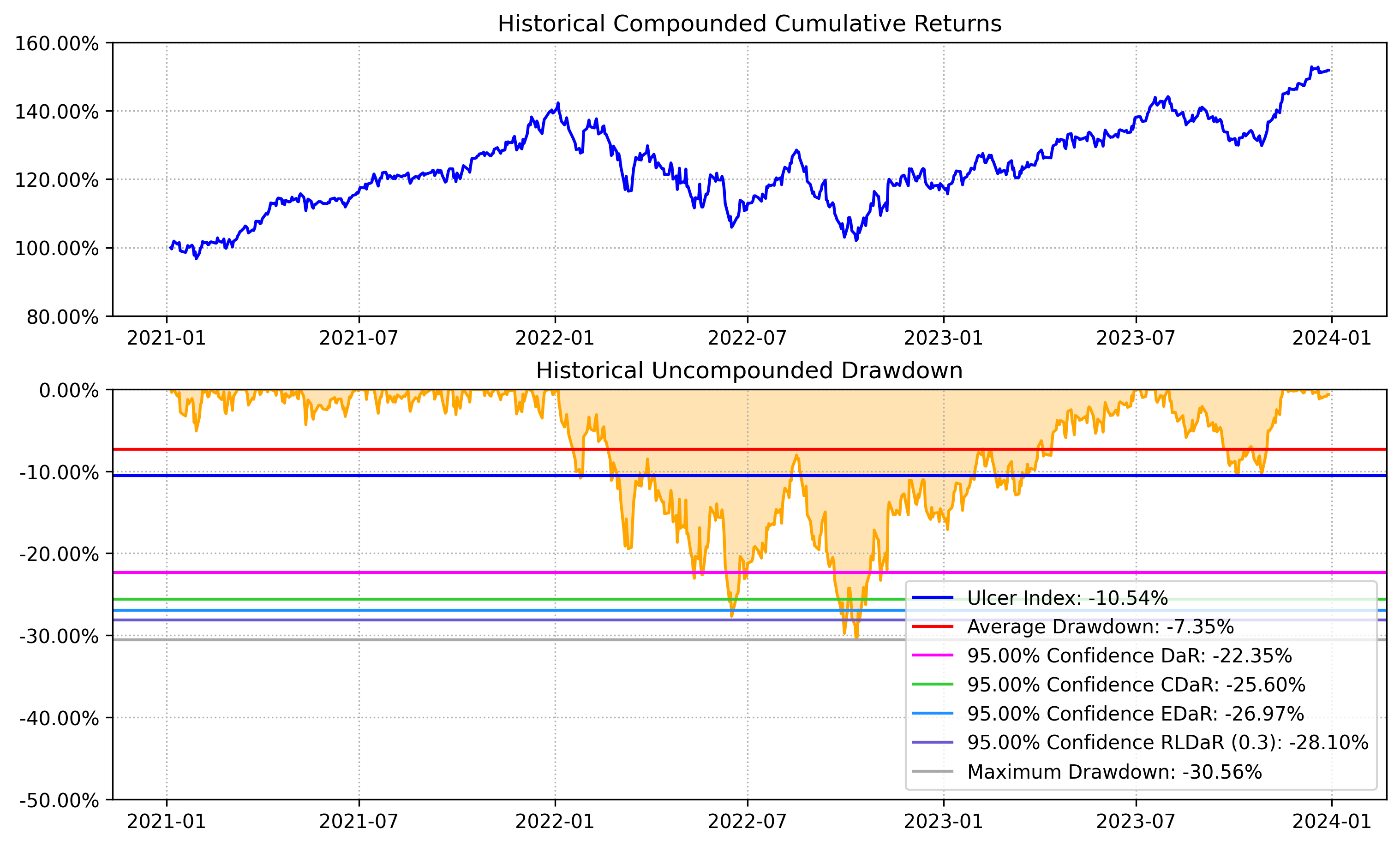}
    \hfill
    \includegraphics[width=0.32\textwidth]{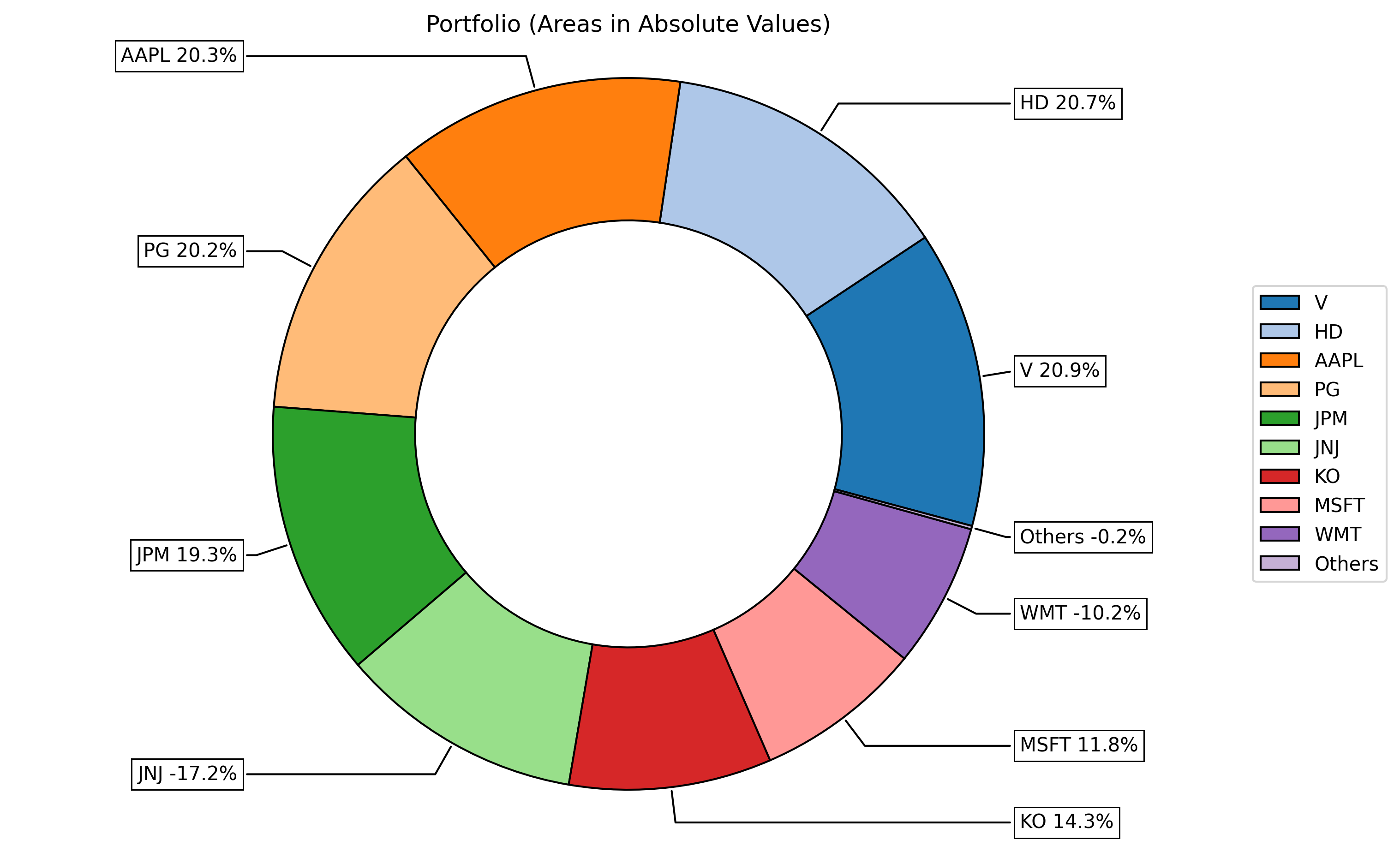}
    \hfill
    \includegraphics[width=0.32\textwidth]{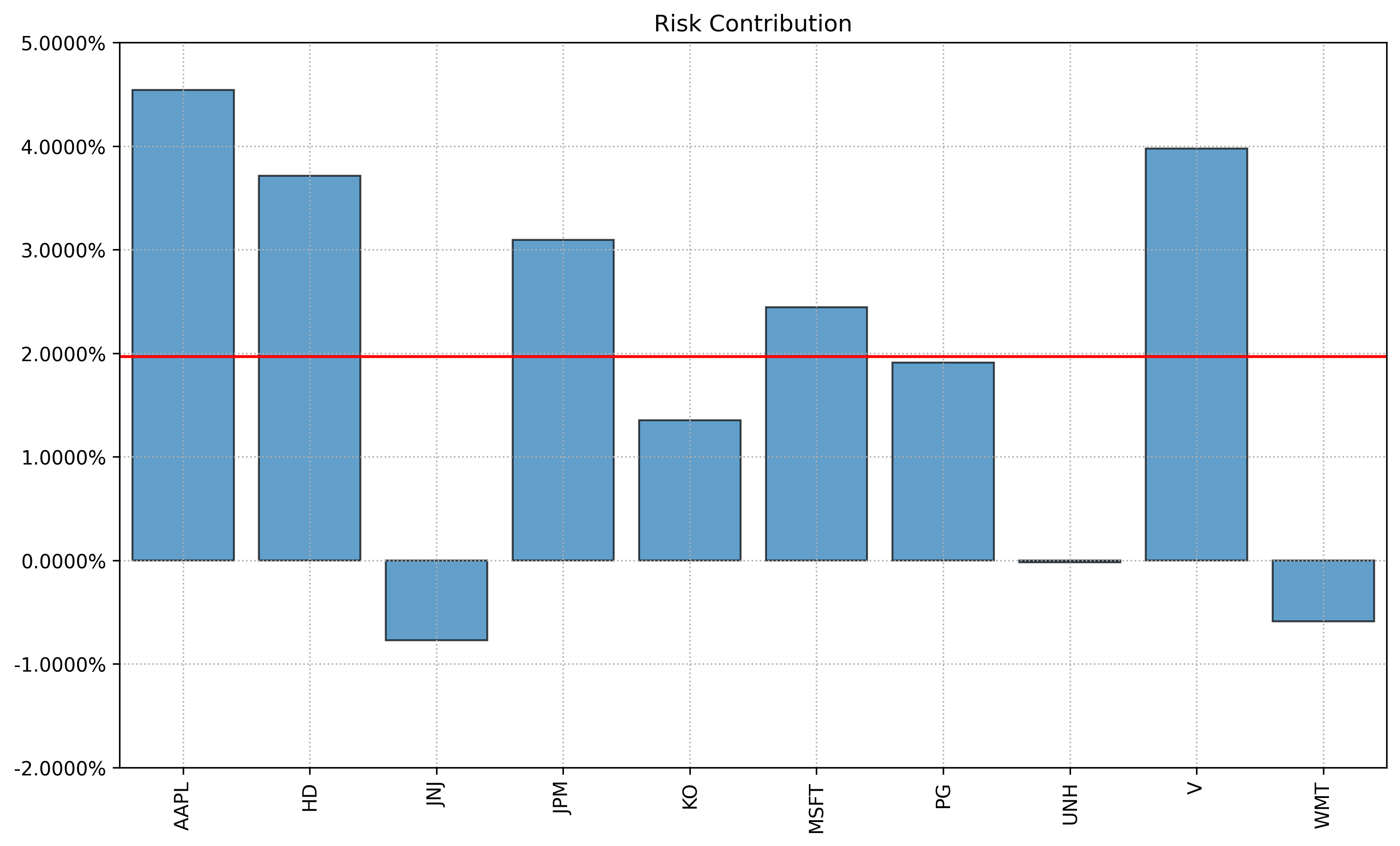}
    \caption{Historical cumulative-return and drawdown trajectories, mean asset allocation weights, and per-asset risk contribution profiles for the learned portfolio optimization policy for DSAC-Wang ($\kappa=0.75$).}
    \label{fig:wang_0.75}
\end{figure*}

Consider DSAC-Wang ($\kappa=+0.75$, shown in Figure~\ref{fig:wang_0.75}).
The optimistic Wang distortion yields the same quasi-uniform
signature at slightly lower magnitude: seven near-equal longs
between $\approx12\%$ and $21\%$ (V, HD, AAPL, PG, JPM, KO, MSFT)
with shorts confined to JNJ ($-17.2\%$) and WMT ($-10.2\%$), for a
gross exposure of $\approx1.6\times$ versus $\approx2.5\times$ for
$\kappa=-0.75$.
This lighter short book attains the highest terminal value of any
configuration ($\approx+52\%$ cumulative return) with a maximum
drawdown of $-30.56\%$ at the October 2022 trough and an Ulcer Index
of $10.54\%$, ending the window at or near its all-time high.
The structural similarity of the two Wang books — near-equal weight
magnitudes, differing mainly in a few sign flips — indicates that at
moderate magnitudes the distortion is comparatively insensitive to
the sign of $\kappa$: it reshapes tail quantiles of the critic's
return distribution but provides little gradient signal to
differentiate assets, a weakness of quantile-based risk functionals
in settings where the entire return distribution is time-varying.

\begin{figure*}[htbp]
    \centering
    \includegraphics[width=0.32\textwidth]{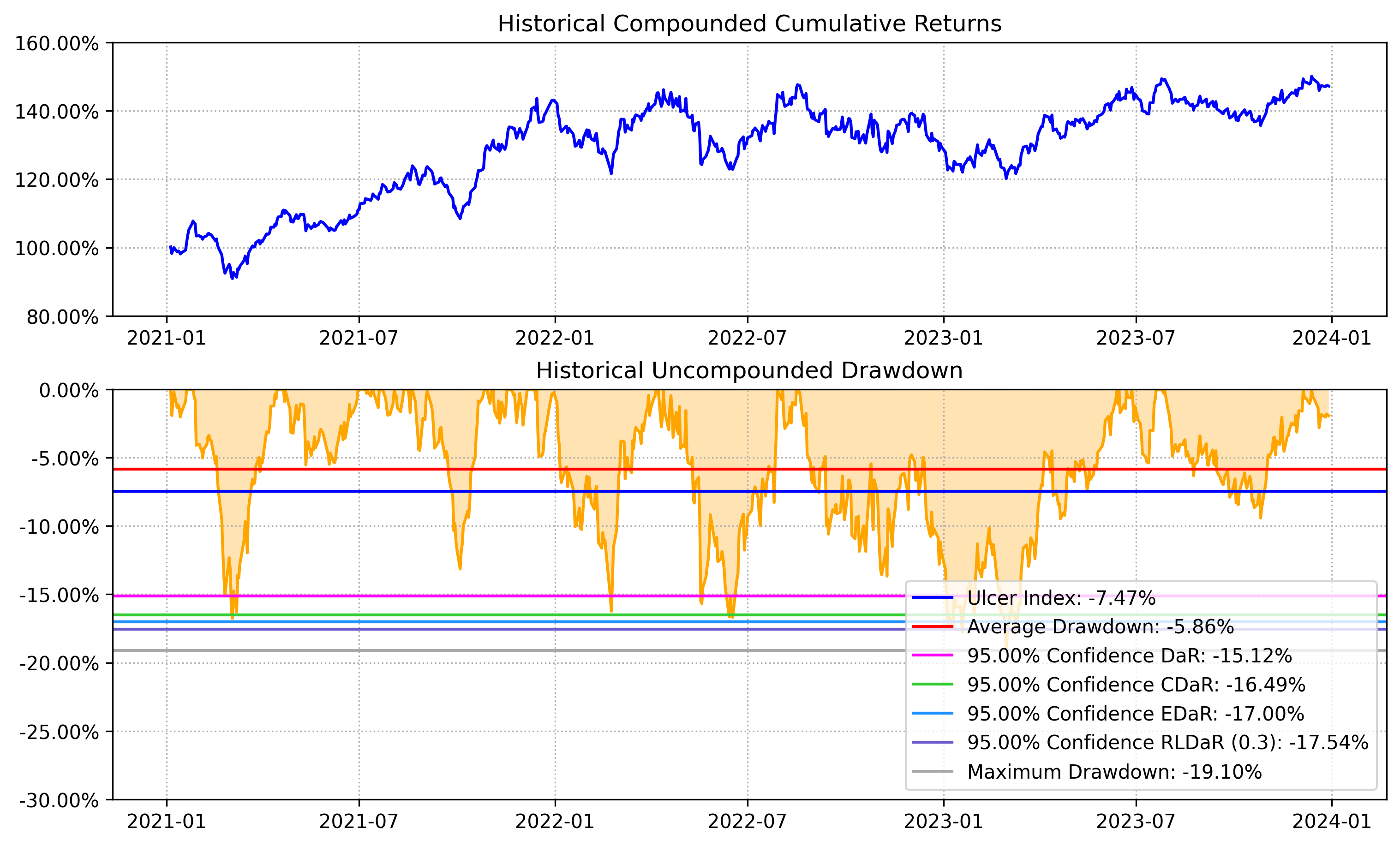}
    \hfill
    \includegraphics[width=0.32\textwidth]{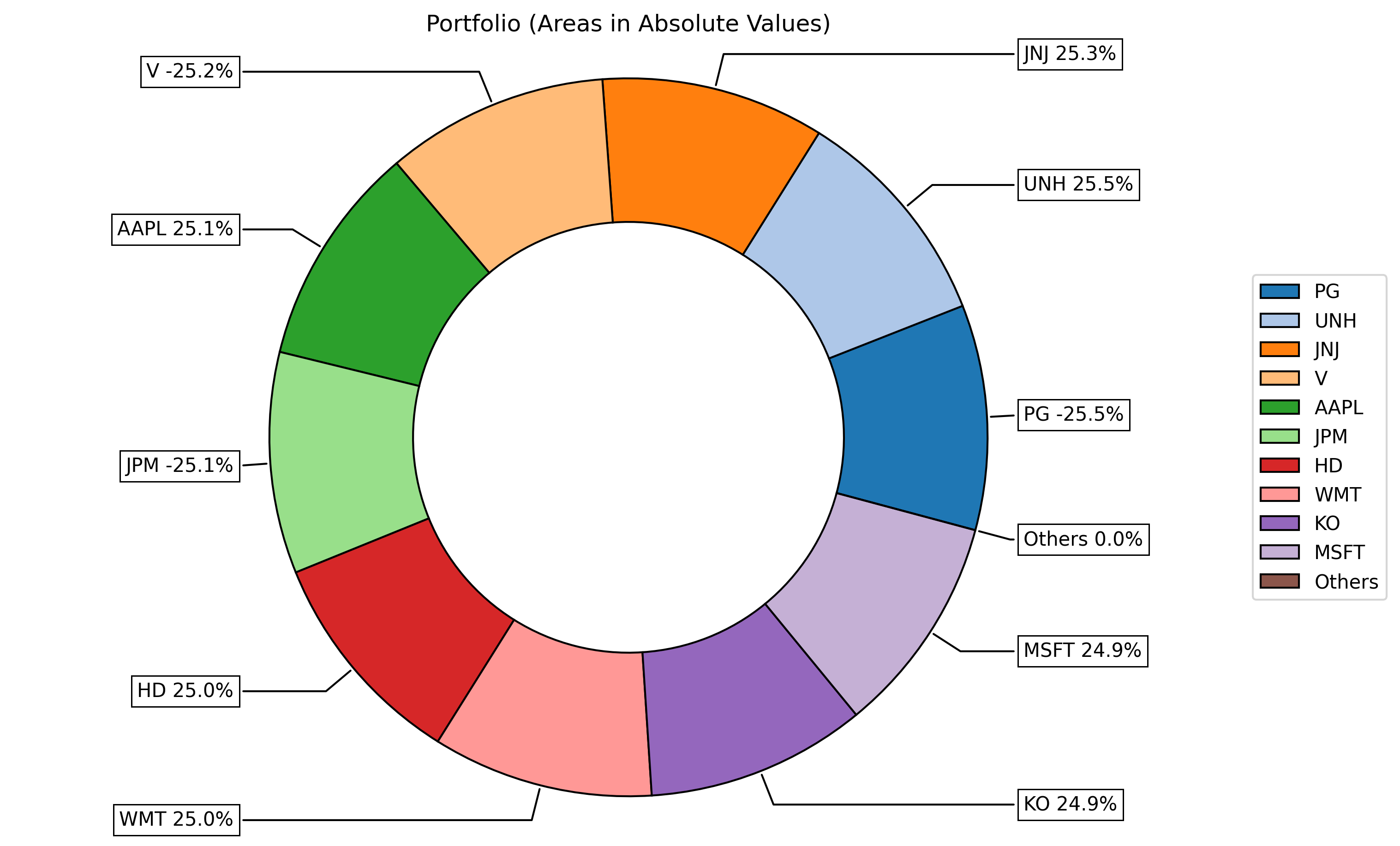}
    \hfill
    \includegraphics[width=0.32\textwidth]{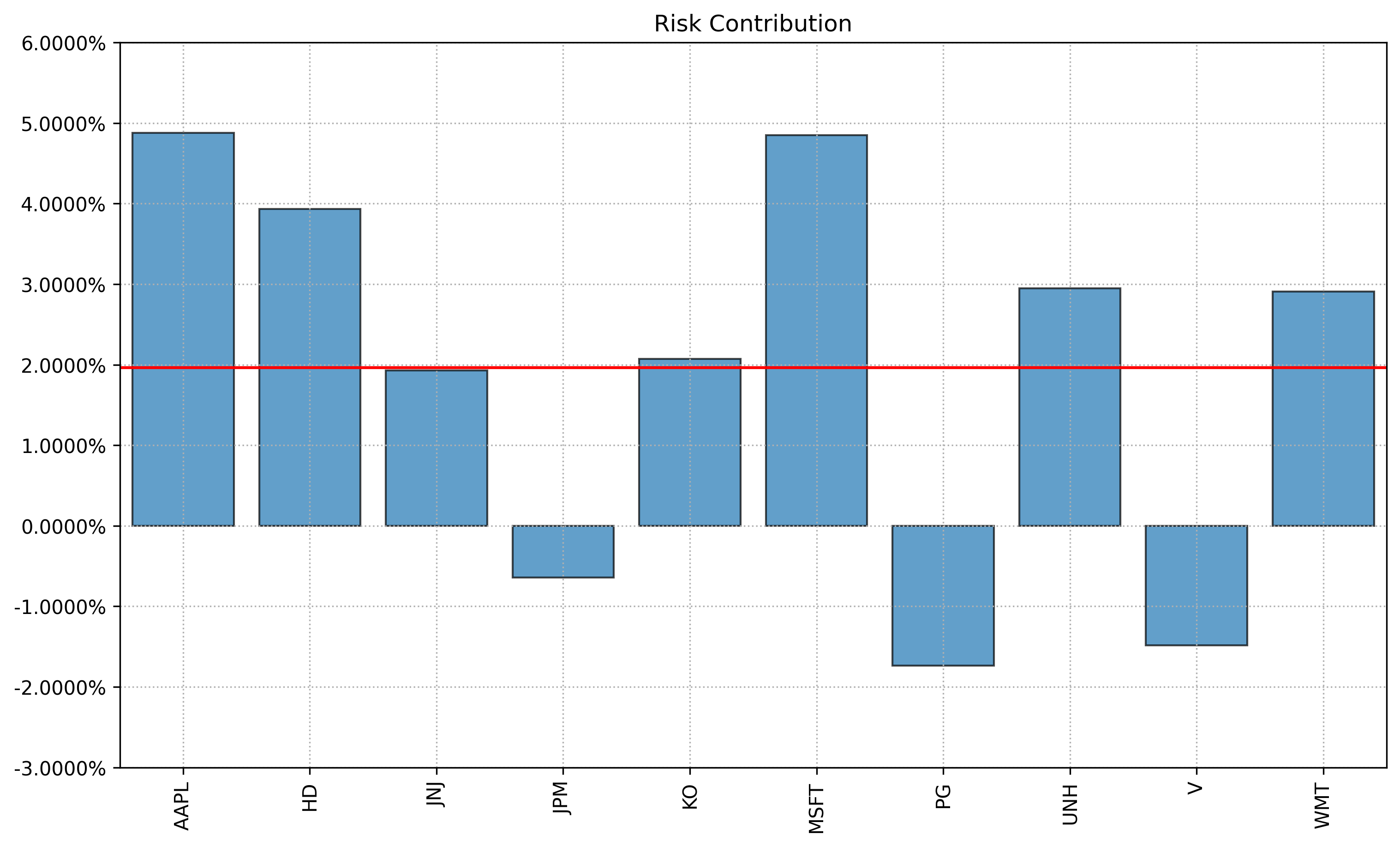}
    \caption{Historical cumulative-return and drawdown trajectories, mean asset allocation weights, and per-asset risk contribution profiles for the learned portfolio optimization policy for DSAC-CVaR ($\alpha=0.9$).}
    \label{fig:cvar_0.9}
\end{figure*}

Consider DSAC-CVaR ($\alpha=0.9$, shown in Figure~\ref{fig:cvar_0.9}).
CVaR at the $90\%$ level optimizes the expected return conditional on
being in the worst $10\%$ of trajectory outcomes, a sharp focus on
deep tail events, and produces the same equal-magnitude signature:
seven longs at $\approx25\%$ each (JNJ, UNH, MSFT, KO, WMT, HD, AAPL)
against shorts in PG ($-25.5\%$), V ($-25.2\%$), and JPM ($-25.1\%$).
It is the strongest baseline: $\approx+47\%$ cumulative return with
the best baseline drawdown profile (maximum drawdown $-19.10\%$,
Ulcer Index $7.47\%$).
Notably, its maximum drawdown falls in February--March 2023 rather
than at the October 2022 bear-market trough, indicating that the
deep-tail objective successfully damped the tightening-cycle
drawdown: the V and JPM shorts hedged exactly the payments and
financials exposure that sold off as consumer-spending and rate
expectations deteriorated through 2022.
However, the strict tail-optimization prevents the policy from
concentrating into the highest-Sharpe assets, capping terminal
wealth at roughly the level of the unlevered $\beta=-0.1$ book.

All three DSAC baselines cluster tightly in the $+44$ to $+52\%$
cumulative-return band, and all three converge to near-uniform weight magnitudes ($\approx24$--$26\%$ per asset for $\kappa=-0.75$
and CVaR, $\approx12$--$21\%$ for $\kappa=+0.75$) that differ only in their sign patterns.
This clustering reflects the inherent coarseness of distributional RL approaches based on fixed-quantile distortions: the distortion
imposes a static risk preference calibrated to the tail characteristics of the training distribution and provides a weak, undifferentiated gradient signal across assets, so the learned books
are structurally interchangeable. The exponential-utility family, by contrast, spans the entire
risk--return frontier as $\beta$ varies: $\beta=-0.1$ matches the baselines' returns ($\approx+46\%$) at an intermediate maximum
drawdown ($-24.88\%$), while $\beta=-0.05$ attains the best downside profile of the study (maximum drawdown $-13.42\%$, Ulcer Index $5.85\%$) at a reduced but positive return. The distinguishing capability is controllability: a single continuous parameter moves the learned policy smoothly along this frontier, whereas each distortion baseline delivers one fixed, weakly tunable operating point. By Sharpe ratio the baselines are similarly close on the validation window ($1.31$--$1.83$) but separate more on the test window ($-0.10$ to $0.72$), with DSAC-CVaR's validation-set advantage
narrowly reversed by DSAC-Wang ($\kappa=-0.75$) out-of-sample.

\subsubsection{Market Stress-Event Analysis}
\label{sec:stress}

We examine the principal stress episodes contained in the held-out
test window (January 2021--December 2023) and discuss how the
exponential-utility objective shapes policy behaviour in each.
Note that the COVID-19 crash of March 2020 lies in the validation
year, outside the plotted test window. The dominant out-of-sample
stress event is the 2022 Federal Reserve tightening cycle.

\paragraph{2022 Federal Reserve Tightening Cycle.}
The Federal Reserve raised the federal funds rate from $0.25\%$
to $4.5\%$ over 2022, driving the NASDAQ Composite down $-33\%$
and inflating the discount rate on long-duration growth assets.
This episode produces the deepest trough in nearly every drawdown
profile, with most configurations bottoming at the October 2022
market low.
The response depends critically on $\beta$:

\begin{itemize}
  \item \emph{\mbox{$\beta=-0.1$ (defensive-quality long book)}}:
    the drawdown deepens in two legs (June and October 2022) to the
    maximum of $-24.88\%$, but the unlevered book recovers fully by
    mid-2023 and closes the window at a new high-water mark
    (Figure~\ref{fig:beta_minus_0_1}).
  \item \emph{$\beta=-0.05$ (growth-hedged book)}:
    the embedded MSFT/AAPL shorts profit precisely as growth sells
    off. The 2022 trough is contained near $-13\%$ and the portfolio
    posts a fresh high-water mark by December 2022
    (Figure~\ref{fig:beta_minus_0_0_5}).
    This is the clearest demonstration that the exponential penalty
    pre-positions the policy for the left tail rather than reacting
    to it.
  \item \emph{$\beta\ge0$ (leveraged books)}:
    the levered long--short structures amplify the bear market
    catastrophically, with uncompounded drawdowns between
    $-129.76\%$ and $-213.05\%$
    (Figures~\ref{fig:beta_0}--\ref{fig:beta_0_1}). None recovers
    within the test window.
  \item \emph{Baselines}:
    DSAC-Wang $\kappa=-0.75$ suffers the deepest drawdown among the
    profitable configurations ($-41.77\%$) because its shorts sit on
    the defensive names that outperformed in 2022, while DSAC-CVaR's
    deep-tail objective contains the 2022 trough and defers its
    shallower maximum drawdown of $-19.10\%$ to February--March 2023
    (Figures~\ref{fig:wang_-0.75} and~\ref{fig:cvar_0.9}).
\end{itemize}

\paragraph{Early-2021 rotation and the late-2023 rate scare.}
Two smaller episodes bracket the window.
In February--March 2021, a sharp rotation out of the prior regime's
winners produced only shallow dips ($-5$ to $-7\%$) in the
risk-averse books, but instantly exposed the overfit leverage of the
$\beta\ge0$ policies: the risk-neutral book lost $\approx-70\%$
within the first two months of the test window
(Figure~\ref{fig:beta_0}).
In October--November 2023, a rates-driven correction produced a
final $-7$ to $-17\%$ dip across the profitable configurations,
followed by a sharp year-end rally in which the risk-averse
Relaxed-RS-AC books and the baselines closed at or near their
high-water marks.
The contrast across all three episodes is consistent: policies
trained with $\beta<0$ enter each shock with bounded leverage and
hedged or defensive books and therefore recover, whereas the
$\beta\ge0$ policies enter with fragile levered exposure and convert
each shock into a permanent capital loss.

\subsubsection{Portfolio Concentration, Risk Contribution,
            and the Emergence of Leverage}
\label{sec:concentration}

\paragraph{Moderate concentration as an emergent property of exponential utility.}
A recurring theme across the risk-averse ($\beta<0$) parameterizations
is concentrated but disciplined allocation: the top three positions
hold $\approx65$--$90\%$ of net exposure (JNJ/AAPL/PG at
$\beta=-0.1$, KO/JNJ/HD at $\beta=-0.05$, and JNJ/HD/MSFT at
$\beta=-0.01$), while gross exposure stays between $\approx110\%$
and $\approx186\%$.

This concentration is not a hyperparameter artifact but a
mathematically necessary consequence of the exponential
objective.
The exponential utility $U(R)=e^{\beta R}/\beta$ with $\beta<0$
has a uniquely determining property: for a portfolio of assets with
independent Gaussian returns, the optimal portfolio under $U$
is proportional to the mean-variance efficient portfolio
of \citet{markowitz1952modern} with risk-aversion coefficient
$|\beta|/2$.
For small $|\beta|$, this portfolio is nearly the maximum-Sharpe
portfolio, which is inherently concentrated in the assets with the
highest Sharpe ratios estimated over the 2013--2019 training window —
here the defensive quality cluster (JNJ, PG, UNH, KO) plus AAPL.
The exponential utility therefore acts as an implicit Sharpe-ratio
maximiser subject to an implicit leverage budget, automatically
identifying the high-Sharpe asset cluster without requiring explicit
Sharpe computation.

\paragraph{Leverage as the failure mode of $\beta\ge0$.}
The complementary observation is where leverage emerges.
Gross exposure jumps from at most $\approx1.9\times$ across all
risk-averse configurations to $\approx8.8\times$ at $\beta=0$,
$\approx2.4\times$ at $\beta=+0.01$, and $\approx16$--$17\times$ at
$\beta\in\{+0.05,+0.1\}$, culminating in the degenerate
WMT ($+568.3\%$) / AAPL ($-536.7\%$) pair trade at $\beta=+0.1$.
The mechanism is structural: because the reward is the log-return of
the weighted book, expected return grows at most linearly in gross
exposure while variance grows quadratically.
A variance-penalising objective ($\beta<0$) therefore admits an
interior optimum in leverage, a variance-indifferent objective
($\beta=0$) leaves leverage unregularised, and a variance-rewarding
objective ($\beta>0$) actively maximises it.
The uncompounded drawdowns beyond $-100\%$ in
Figures~\ref{fig:beta_0}--\ref{fig:beta_0_1} are the direct
empirical signature of this mechanism.

\paragraph{Risk contribution and the diversification structure.}
The risk-contribution charts (right panels of all figures) show the
per-asset contribution to total portfolio variance relative to
the equal-weight benchmark (red horizontal line).
For $\beta=-0.1$, AAPL — the largest-variance holding — contributes
$\approx4.6\%$ against an equal-weight benchmark of $\approx1.5\%$,
a $3\times$ concentration that the exponential penalty deems
worthwhile for the asset's compounding.
For $\beta=-0.05$, all three shorted assets (PG, AAPL, MSFT) carry
\emph{negative} risk contributions: the book is explicitly
variance-hedged rather than merely diversified.
For $\beta=+0.1$, virtually the entire risk budget collapses onto
the two legs of the WMT/AAPL pair.
The DSAC portfolios
(Figures~\ref{fig:wang_-0.75}--\ref{fig:cvar_0.9}) show
comparatively flat contribution profiles, consistent with their
near-uniform weights, with their short legs (KO/UNH/WMT for
$\kappa=-0.75$ and PG/V/JPM for CVaR) appearing as modest negative
contributors.

We conclude that the exponential-utility parameterization with
$\beta<0$ trades some diversification for compounding under an
implicit leverage budget, and that the location of the learned
policy on this spectrum — from the hedged defensive book at
$\beta=-0.05$ to the unlevered quality-growth book at $\beta=-0.1$ —
is directly and interpretably controlled by $\beta$.
Whether the additional terminal wealth of the $\beta=-0.1$
configuration justifies its deeper 2022 drawdown depends on the
investor's horizon and absolute drawdown tolerance, precisely the
decision the $\beta$ parameter is designed to encode.

\subsection{Active Network Management}
\label{appendix:anm}
In Active Network Management (ANM), the primary operational directive is not merely to maximize the integration of renewable energy, but to do so while strictly guaranteeing the physical safety of the power grid under highly stochastic conditions. Electricity distribution networks are constantly subjected to aleatoric shocks, such as sudden wind gusts, cloud cover over solar arrays, and unpredictable local load spikes. A traditional, risk-neutral reinforcement learning agent will frequently learn an aggressive control policy that pushes power flows exactly to their physical limits to maximize average efficiency. However, operating on this razor's edge leaves the system highly vulnerable to catastrophic failures. A sudden weather variance can easily push the network into severe constraint breaches. Therefore, risk-sensitive control is fundamentally necessary in this domain to explicitly penalize heavy-tailed failure states, mathematically forcing the agent to maintain a proactive safety buffer below the network capacities to absorb unexpected environmental shocks.

We utilize the \texttt{ANM6Easy-v0} environment of the Gym-ANM suite \citep{HENRY2021100092} that simulates the operation of electricity distribution networks. In this domain, the agent controls the active and reactive power setpoints of renewable energy generators and distributed storage devices to maximize the integration of stochastic renewable energy. To rigorously evaluate risk-sensitive behavior beyond scalar reward, we track four complementary metrics over a 200-step evaluation episode: Thermal Violation Rate, the fraction of timesteps where line current exceeds rated capacity, worst overload (in per-unit), the peak ratio of observed current to rated capacity across all lines and timesteps, CVaR at 5\%, the expected reward in the worst 5\% of timesteps, capturing tail-risk exposure, and reward standard deviation, measuring the consistency of the agent's control policy. Together, these metrics paint a picture not just of how well an agent  performs on average, but how safely and reliably it operates under the stochastic conditions inherent to renewable energy integration. We benchmark our proposed Relaxed-RS-AC across all three risk profiles against the risk-averse Distributional SAC, configured the Wang ($\kappa=-0.75$) distortion risk measure \citep{wang1998ordering}. The reward metrics are listed in Table \ref{tab:anm_reward} and the risk-specific metrics are listed in Table \ref{tab:anm_risk}.

\begin{table}[]
\centering
\small
\begin{tabular}{lccc}
\toprule
\textbf{Algorithm} & \textbf{Total Reward} $(\uparrow)$ & \textbf{Mean Reward} $(\uparrow)$ & \textbf{Reward Std} $(\downarrow)$ \\
\midrule
Relaxed-RS-AC ($\beta = -0.01$) & -182.46 & -0.91 & \textbf{1.00} \\
Relaxed-RS-AC ($\beta = 0$)     & \textbf{-169.55} & \textbf{-0.85} & 1.12 \\
Relaxed-RS-AC ($\beta = 0.01$)  & -242.15 & -1.21 & 1.33 \\
DSAC-Wang ($\beta = -0.75$)     & -491.63 & -2.46 & 3.11 \\
\bottomrule
\end{tabular}
\caption{\textbf{\texttt{ANM6Easy-v0}.} Reward metrics across varying risk profiles ($\beta$) and the 
DSAC-Wang baseline.}
\label{tab:anm_reward}
\end{table}

\begin{table}[]
\centering
\small
\setlength{\tabcolsep}{4pt}
\begin{tabular}{lcccc}
\toprule
\textbf{Algorithm} & \shortstack{\textbf{Voltage}\\\textbf{violations} $(\downarrow)$} & \shortstack{\textbf{Thermal}\\\textbf{violations} $(\downarrow)$} & \shortstack{\textbf{Overload}\\\textbf{(pu)} $(\downarrow)$} & \shortstack{\textbf{CVaR}\\\textbf{5\%} $(\uparrow)$} \\
\midrule
Relaxed-RS-AC ($\beta = -0.01$) & 0.00 & \textbf{0.54} & 1.51 & \textbf{-3.11} \\
Relaxed-RS-AC ($\beta = 0$)     & 0.00 & 0.61 & 1.56 & -4.05 \\
Relaxed-RS-AC ($\beta = 0.01$)  & 0.00 & 0.64 & 1.80 & -5.07 \\
DSAC ($\text{Wang} = -0.75$)     & 0.00 & 0.57 & \textbf{1.49} & -7.49 \\
\bottomrule
\end{tabular}
\caption{\textbf{\texttt{ANM6Easy-v0}.} Evaluation of reward and safety metrics across varying risk 
profiles ($\beta$) and the DSAC-Wang baseline. The risk-averse Relaxed-RS-AC ($\beta = -0.01$) 
achieves the best CVaR(5\%) and lowest thermal violation rate among all Relaxed-RS-AC 
configurations, demonstrating improved tail-risk performance as the risk-averse objective 
takes effect. While DSAC-Wang (-0.75) attains the lowest worst-case overload 
(1.49 pu), it does so at a significant cost to overall performance, exhibiting the highest 
worst CVaR(5\%) across all agents.}
\label{tab:anm_risk}
\end{table}

The results reveal a clear and interpretable risk-sensitivity spectrum. The risk-averse agent 
($\beta = -0.01$) achieves the best CVaR(5\%) of $-3.11$ and the lowest thermal violation rate of $0.54$, confirming that the risk-averse objective successfully forces the agent to maintain a proactive safety buffer, absorbing stochastic shocks at the cost of a modest reduction in mean reward ($-0.91$ vs. $-0.85$ for the risk-neutral agent). The risk-neutral 
agent ($\beta = 0$) attains the highest mean reward of $-0.85$ and total reward of $-169.55$, consistent with its objective of maximizing expected return without penalizing variance, but this comes at the expense of a higher thermal violation rate ($0.61$) and a significantly worse CVaR ($-4.05$), exposing the system to greater tail-risk. The risk-seeking agent ($\beta = 0.01$) performs worst across nearly all metrics, with the highest thermal violation rate ($0.64$), worst overload ($1.80$ pu), and worst CVaR ($-5.07$), confirming that actively 
seeking variance leads to reckless control behavior that frequently breaches network capacities. The DSAC-Wang baseline ($\beta = -0.75$) presents a particularly instructive 
failure mode: despite applying the strongest risk penalty of all agents, it achieves the lowest worst-case overload ($1.49$ pu) yet simultaneously exhibits the highest reward variance 
($3.11$) and the worst CVaR ($-7.49$) across all evaluated agents. This apparent contradiction suggests that the Wang distortion risk measure, while effective at suppressing the magnitude of individual overloads, fails to produce a globally stable policy, instead inducing erratic 
behavior that manifests as catastrophically poor outcomes in the tail of the reward distribution. Taken together, these results demonstrate that the Relaxed-RS-AC framework with 
a risk-averse objective achieves the most favorable safety-performance trade-off, successfully internalizing the risk-sensitive requirements of active network management.

\subsection{Ablation Study: Feature Dimension}
\label{appendix:feature_dim_ablation}

\begin{figure}[p]
\centering
\begin{subfigure}[b]{0.32\linewidth}
         \centering
         \includegraphics[width=\linewidth]{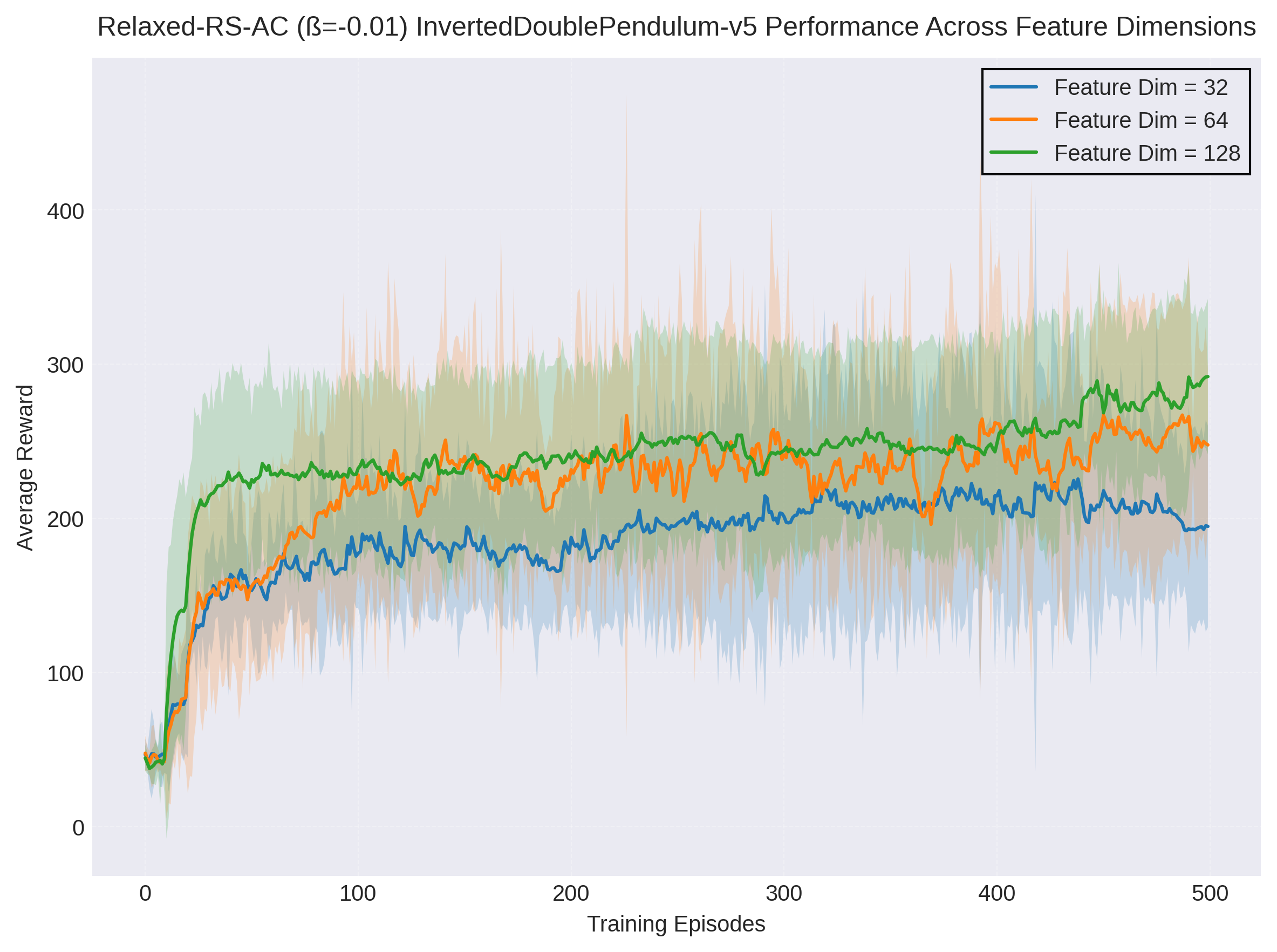}
         \caption{Risk-Averse ($\beta=-0.01$)}
         \label{fig:inverted-double-pendulum-averse}
     \end{subfigure}\hfill
     \begin{subfigure}[b]{0.32\linewidth}
         \centering
         \includegraphics[width=\linewidth]{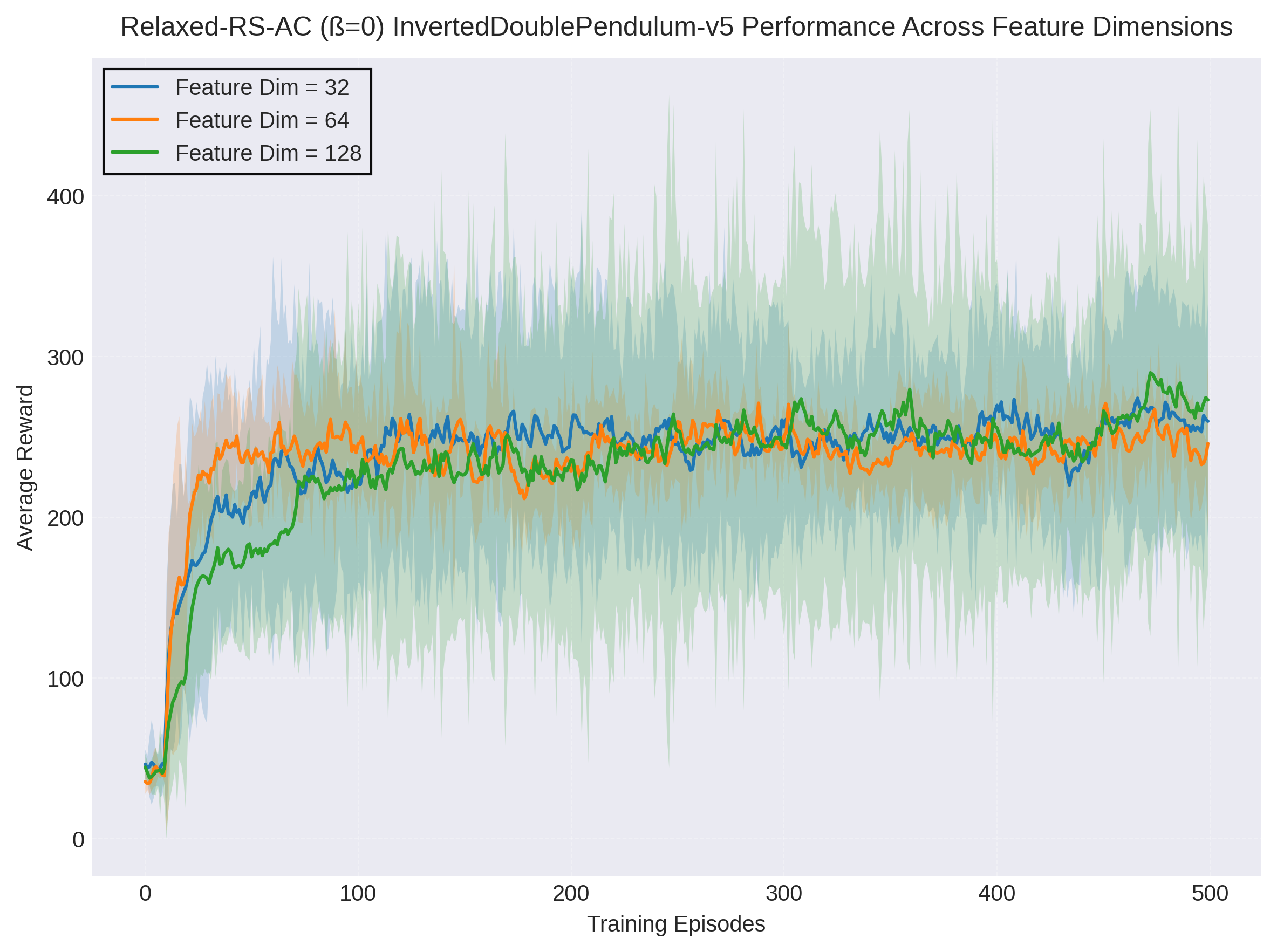}
         \caption{Risk-Neutral ($\beta=0$)}
         \label{fig:inverted-double-pendulum-neutral}
     \end{subfigure}\hfill
     \begin{subfigure}[b]{0.32\linewidth}
         \centering
         \includegraphics[width=\linewidth]{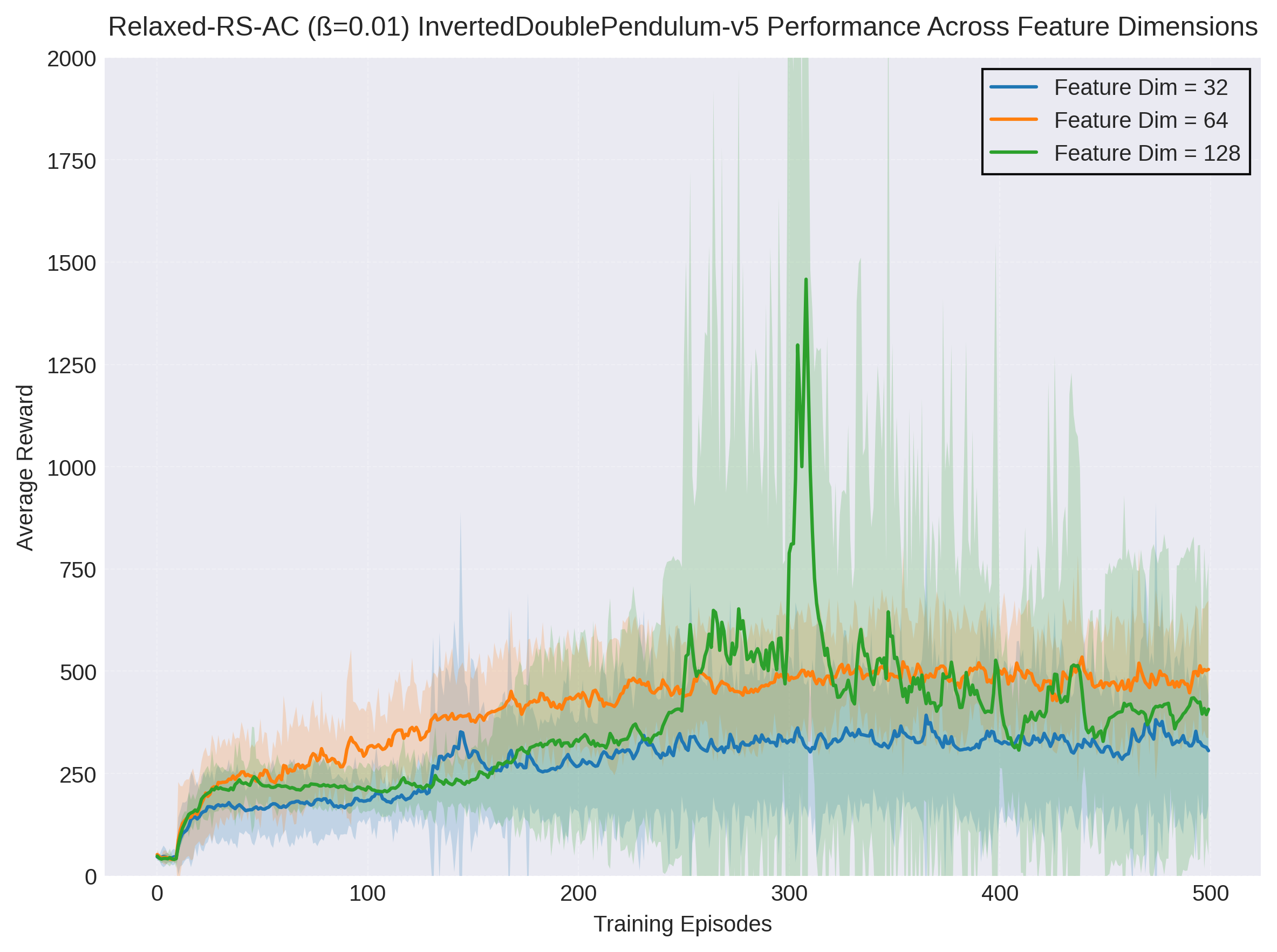}
         \caption{Risk-Seeking ($\beta=0.01$)}
         \label{fig:inverted-double-pendulum-seeking}
     \end{subfigure}
    \caption{\textbf{\texttt{Inverted-Double-Pendulum/v5}.} The learning curves show the effect of the linear feature dimension ($k$) on Relaxed-RS-AC in \texttt{Inverted-Double-Pendulum/v5}. Across 500 episodes, performance is compared under risk-neutral ($\beta = 0$), risk-seeking ($\beta = 0.01$), and risk-averse ($\beta = -0.01$) profiles. Shaded regions represent variance. The plots highlight that increasing feature capacity actively shapes the algorithm's tail-risk behavior, triggering severe upper-tail instability when risk-seeking, but providing crucial structural robustness when risk-averse. }
    \label{fig:inverted-double-pendulum}
\par\medskip
\begin{subfigure}[b]{0.32\linewidth}
         \centering
         \includegraphics[width=\linewidth]{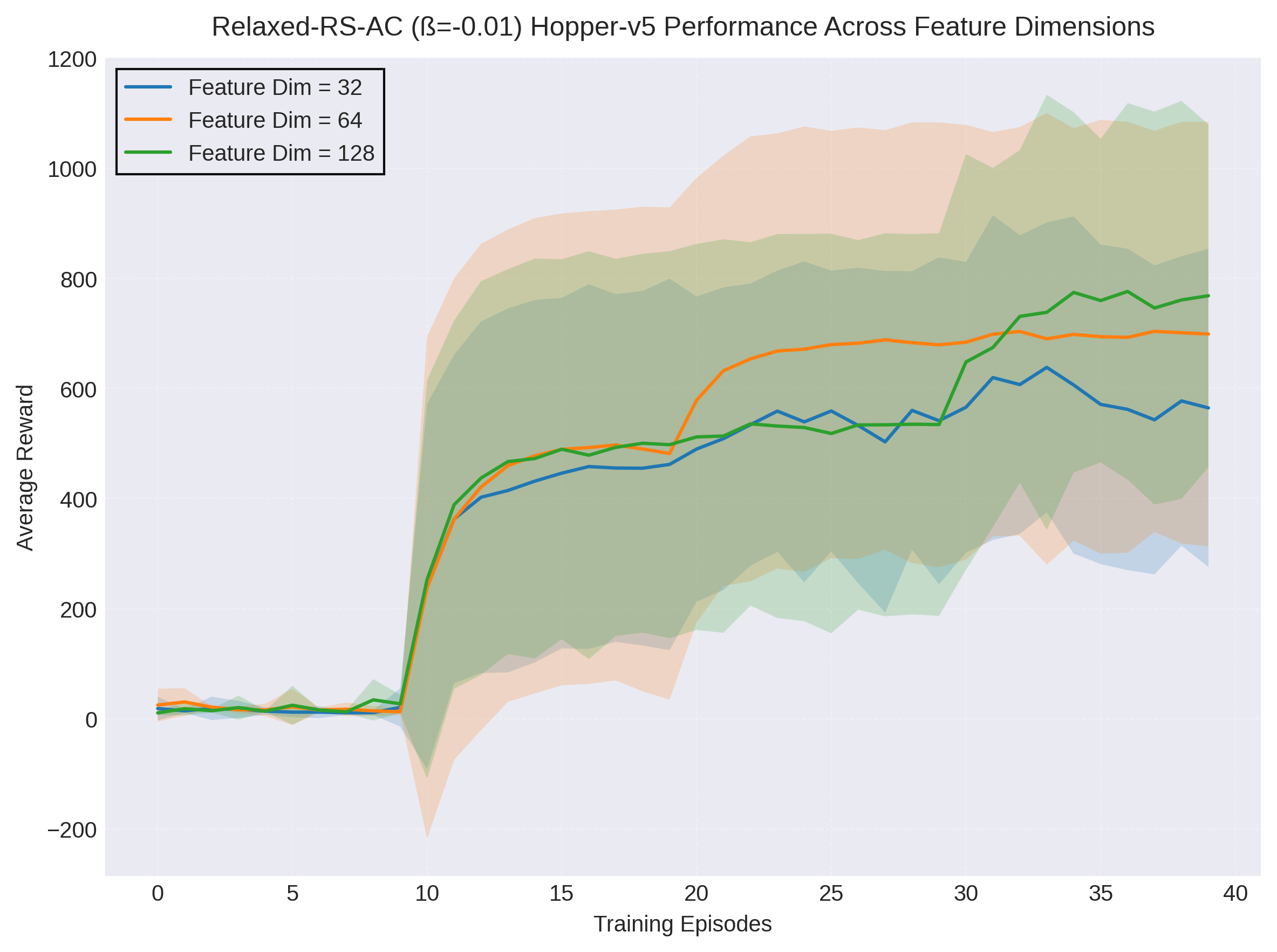}
         \caption{Risk-Averse ($\beta=-0.01$)}
         \label{fig:hopper-averse}
     \end{subfigure}\hfill
     \begin{subfigure}[b]{0.32\linewidth}
         \centering
         \includegraphics[width=\linewidth]{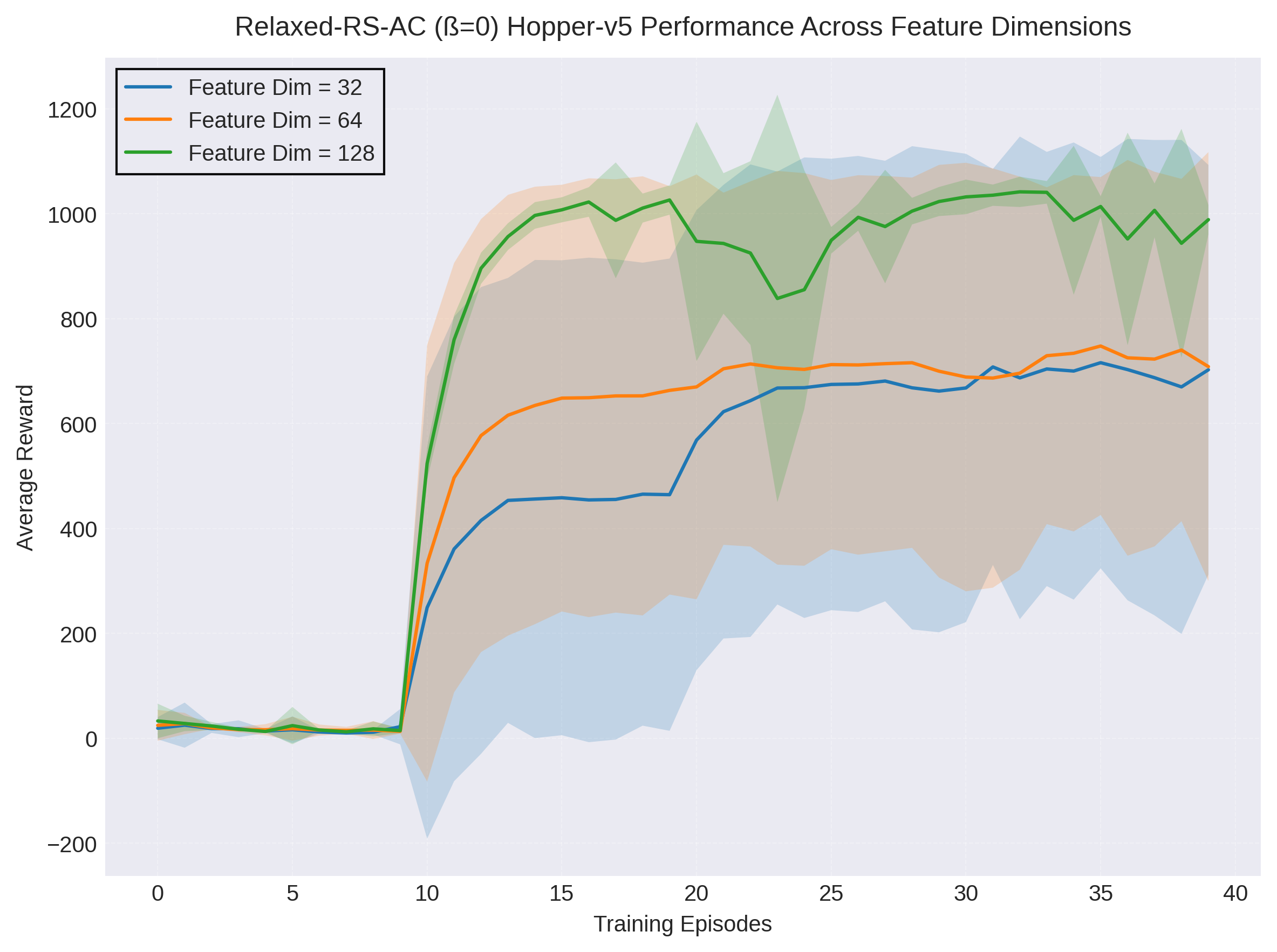}
         \caption{Risk-Neutral ($\beta=0$)}
         \label{fig:hopper-neutral}
     \end{subfigure}\hfill
     \begin{subfigure}[b]{0.32\linewidth}
         \centering
         \includegraphics[width=\linewidth]{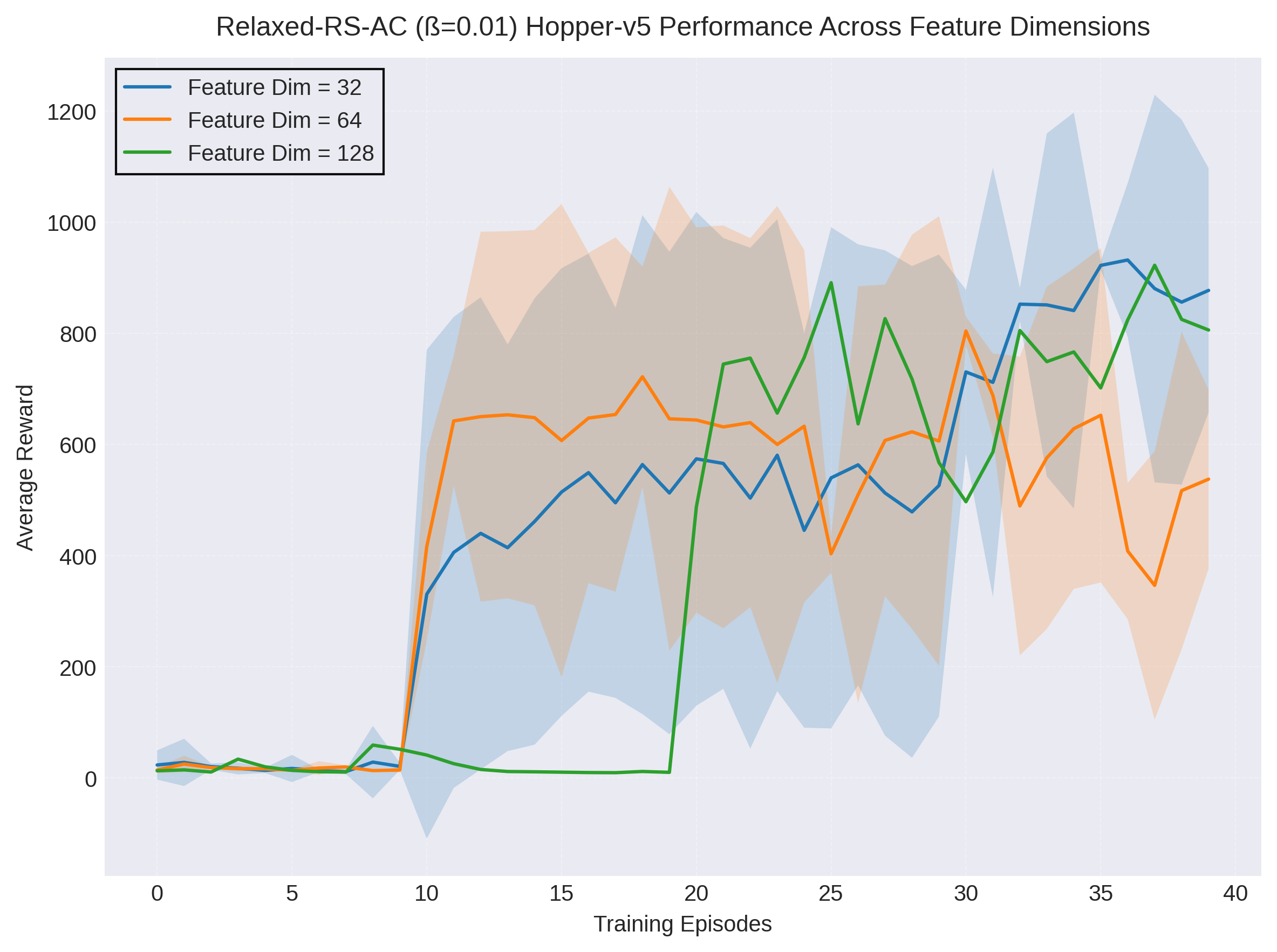}
         \caption{Risk-Seeking ($\beta=0.01$)}
         \label{fig:hopper-seeking}
     \end{subfigure}
    \caption{\textbf{\texttt{Hopper/v5}.} The learning curves show the effect of the linear feature dimension ($k$) across risk profiles ($\beta$). While expanded capacity ($k=128$) yields robust performance gains and tighter variance bounds in risk-neutral and risk-averse settings, it triggers pronounced instability and delayed learning under the risk-seeking objective ($\beta=0.01$) due to a fundamental conflict with the environment's strict survival-gated geometry. }
    \label{fig:hopper}
\par\medskip
\begin{subfigure}[b]{0.32\linewidth}
         \centering
         \includegraphics[width=\linewidth]{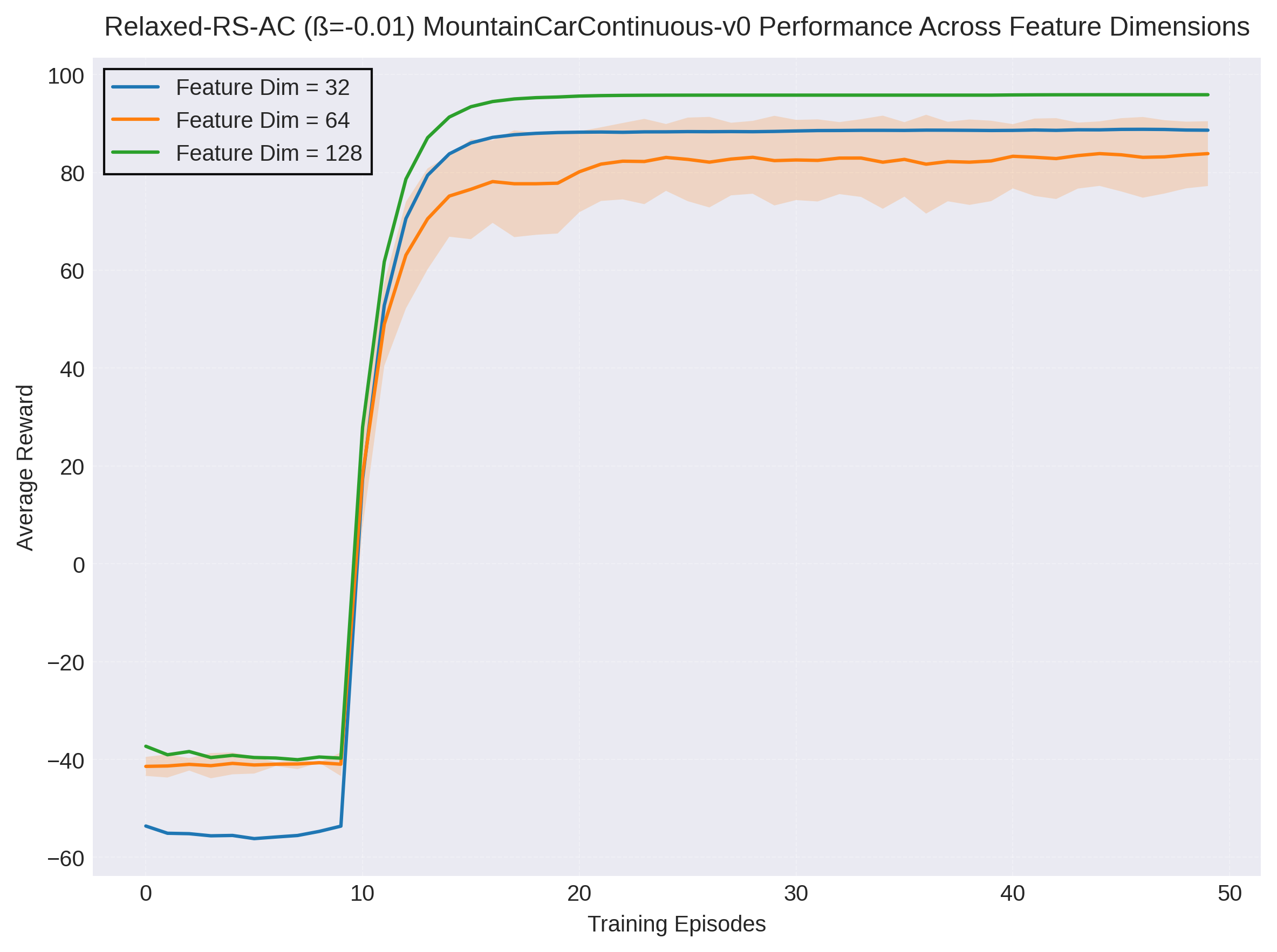}
         \caption{Risk-Averse ($\beta=-0.01$)}
         \label{fig:mountain-car-continuous-averse}
     \end{subfigure}\hfill
     \begin{subfigure}[b]{0.32\linewidth}
         \centering
         \includegraphics[width=\linewidth]{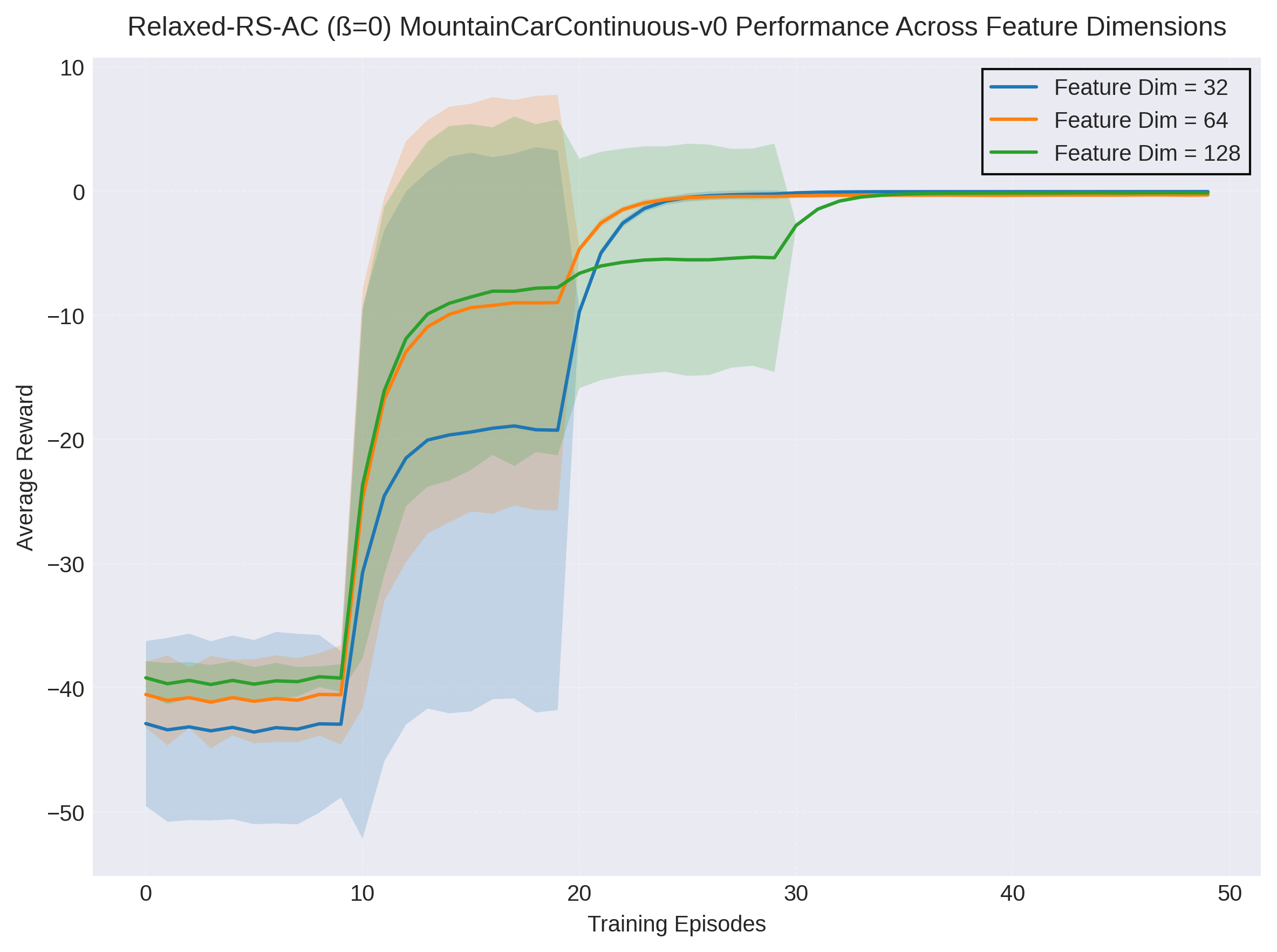}
         \caption{Risk-Neutral ($\beta=0$)}
         \label{fig:mountain-car-continuous-neutral}
     \end{subfigure}\hfill
     \begin{subfigure}[b]{0.32\linewidth}
         \centering
         \includegraphics[width=\linewidth]{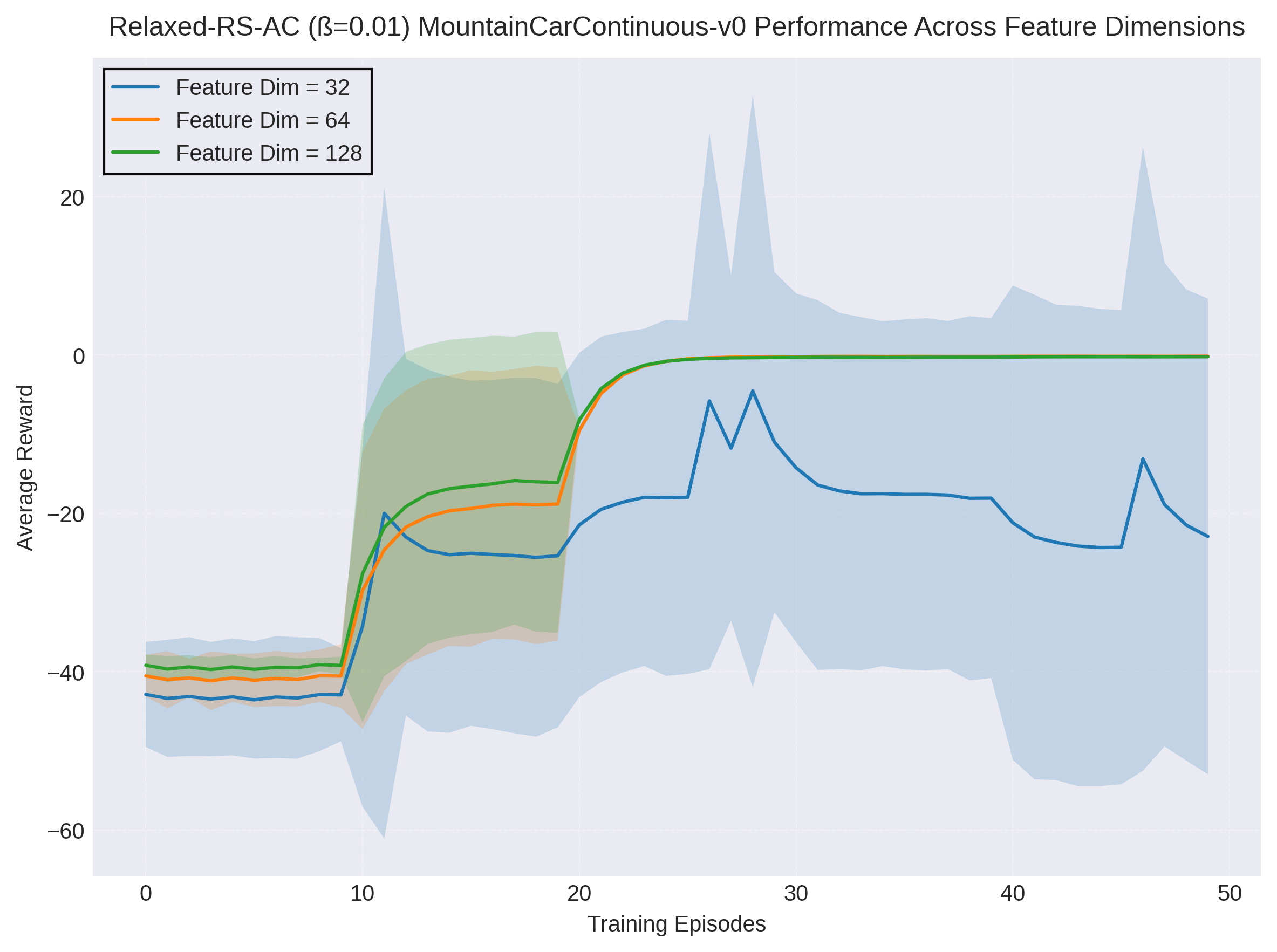}
         \caption{Risk-Seeking ($\beta=0.01$)}
         \label{fig:mountain-car-continuous-seeking}
     \end{subfigure}
    \caption{\textbf{\texttt{Mountain-Car-Continuous/v0}.} The learning curves show the effect of the linear feature dimension ($k$) across risk profiles ($\beta$). While risk-neutral ($\beta = 0$) and risk-seeking ($\beta = 0.01$) agents fail to escape a 0-reward local minimum, the risk-averse agent ($\beta = -0.01$) successfully captures the sparse terminal bonus. Under this risk-averse objective, increased capacity ($k=128$) yields a highly stable, near-zero variance policy that optimally solves the task.}
    \label{fig:fig:mountain-car-continuous}
\end{figure}

In this section, we present a comprehensive ablation study investigating the sensitivity of the Relaxed-RS-AC algorithm to the dimensionality of the feature space within its linear function approximator. To rigorously evaluate the algorithm's robustness, we conduct experiments across three canonical continuous control environments: \texttt{Inverted-Double-Pendulum/v5} and \texttt{Hopper/v5} from the MuJoCo suite \citep{MuJoCoTassa2012}, and \texttt{Mountain-Car-Continuous/v0} \citep{Moore-1990-13223}. These tasks were deliberately selected for their markedly distinct empirical return landscapes: heavy-tailed (frequent failures vs. rare balances), survival-gated (right-skewed penalty mode obscuring long-tail gradients), and sparse/bimodal (polarizing terminal rewards). Evaluating across these contrasting regimes allows us to systematically isolate the effect of feature dimension ($k\in \{32, 64, 128\}$) on learning stability, tail-risk exposure, and expected return. We train Relaxed-RS-AC across 5 seeds per configuration and report the resulting training curves in Figures \ref{fig:inverted-double-pendulum}, \ref{fig:hopper} and \ref{fig:fig:mountain-car-continuous}.

\textbf{\texttt{Inverted-Double-Pendulum/v5}.} Under a risk-neutral objective ($\beta=0$, Figure \ref{fig:inverted-double-pendulum-neutral}), the algorithm is relatively robust to feature dimension, with all capacities ($k \in \{32, 64, 128\}$) converging to a comparable asymptotic average reward. However, risk-sensitive profiles reveal a strict dependence on representational capacity. When risk-seeking ($\beta=0.01$, Figure \ref{fig:inverted-double-pendulum-seeking}), a narrow network ($k=32$) acts as a necessary regularizer against over-optimizing for rare, full-horizon balances. Conversely, a high-capacity network ($k=128$) aggressively exploits these high-variance trajectories, leading to a transient spike followed by catastrophic policy collapse. The inverse is true for the risk-averse profile ($\beta=-0.01$, Figure \ref{fig:inverted-double-pendulum-averse}). Minimizing downside risk (frequent tip-overs) proves fundamentally more complex than maximizing expected return. Here, $k=32$ is insufficient, yielding high variance. The expanded $k=128$ feature space is required to successfully map the intricate boundaries of failure states, yielding the tightest confidence intervals and highest reliable performance.

\textbf{\texttt{Hopper/v5}.} For standard expected-return optimization ($\beta=0$, Figure \ref{fig:hopper-neutral}), feature capacity monotonically improves performance, with $k=128$ distinctly outperforming narrower architectures. The risk-averse objective ($\beta=-0.01$, Figure \ref{fig:hopper-averse}) perfectly complements this survival-gated environment. Scaling to $k=128$ uniformly enhances this synergy, providing the expressivity needed to model subtle state-space boundaries, effectively preventing catastrophic falls and tightening lower-bound variance. Conversely, the risk-seeking profile ($\beta=0.01$, Figure \ref{fig:hopper-seeking}) exposes severe pathological behaviors when paired with high capacity. The $k=128$ configuration suffers from delayed learning and explosive variance, as the objective incentivizes risky maneuvers that trigger early termination. In this regime, the bottleneck of $k=32$ is uniquely required to regularize dangerous exploration.

\textbf{\texttt{Mountain-Car-Continuous/v0}}. In this sparse, bimodal geometry, both the risk-neutral and risk-seeking agents fail entirely to cross the exploration chasm, converging instead to a zero-reward local minimum of passive avoidance (Figures \ref{fig:mountain-car-continuous-neutral} and \ref{fig:mountain-car-continuous-seeking}). Strikingly, the risk-averse objective ($\beta=-0.01$, Figure \ref{fig:mountain-car-continuous-averse}) is the only profile capable of solving the environment. By escaping the valley and isolating the optimal trajectory, all feature dimensions successfully capture the sparse terminal bonus. Within this success, representational capacity remains critical: the $k=128$ configuration is unambiguously superior, constructing a highly confident policy that reliably achieves the terminal state with virtually zero variance.

The overarching conclusion from our feature dimension ablation is that iterates remain stable without divergence, provided the architecture matches the risk geometry of the task. Representational capacity is key: under a risk-averse profile, $k=128$ networks isolate optimal trajectories with near-zero variance, while restricted architectures yield high-variance policies. Importantly, linear function approximation enables our finite-sample $\mathcal{O}(1/t)$ MSE critic bound (Theorem \ref{theorem:finitesamplerrorbound}) via the strong convexity of $F(\theta, \omega)$. Deep networks destroy this strong convexity, downgrading guarantees to asymptotic stationary points. Closing this gap is our core theoretical contribution. Empirically, $q_\omega$ is implemented as a spectrally-normalized neural network, and the policy is a single-layer Gaussian network. While SPSA's $\mathcal{O}(d)$ variance is a universal property of zeroth-order estimators \citep{spall1992multivariate, mania2018simple}, it safely trades sample efficiency for convergence as established in Theorem \ref{thm:twosided_convergence}. Extending these methods to deep critics requires stationary-point analysis and trust-region stabilization. These remain principled avenues for future work, whereas our current focus resolves the missing finite-sample bounds for linear function approximation.

\end{document}